\documentclass{tlp}

\usepackage[dvips]{epsfig}
\usepackage{wrapfig}
\usepackage{subfigure}
\usepackage{latexsym}
\usepackage{xspace}
\usepackage{fancyvrb}
\usepackage{listings}
\newcommand{\prend}{\hfill \rule{2.3mm}{2.3mm} \hspace*{3mm} $ $}

\newcommand{\mrule}{constraint}
\newcommand{\mrules}{constraints}
\newcommand{\mRules}{Constraints}

\newtheorem{theorem}{Theorem}[section]
\newtheorem{lemma}[theorem]{Lemma}

\newtheorem{corollary}[theorem]{Corollary}

\begin{document}

\title{\bf Redundant Sudoku Rules
}
\author[Bart Demoen and Maria Garcia de la Banda]
{
BART DEMOEN \\
Department of Computer Science, KU Leuven, Belgium\\
bart.demoen@cs.kuleuven.be
\and
MARIA GARCIA DE LA BANDA \\
Faculty of Information Technology, Monash University, Australia \\
Maria.GarciaDeLaBanda@monash.edu
}
\date{}
\maketitle

\begin{abstract}
The rules of Sudoku are often specified using twenty seven
\texttt{all\_different} constraints, referred to as the {\em big}
\mrules. Using graphical proofs and exploratory logic programming, the
following main and new result is obtained: many subsets of six of
these big \mrules~are redundant (i.e., they are entailed by the
remaining twenty one \mrules), and six is maximal (i.e., removing more
than six \mrules~is not possible while maintaining equivalence). The
corresponding result for binary inequality constraints, referred to
as the {\em small} \mrules, is stated as a
conjecture.
\end{abstract}

\section{Introduction}

On the 18th of May 2008, the following question was posted on
\texttt{rec.puzzles}: ``what's the minimum amount of checking that
needs to be done to show that a completed 9x9 grid is valid?''. We
prove that the short answer is: ``21 all\_different constraints''. The
complete answer shown here is the result of a set of theorems
whose proofs are presented in an intuitive graphical representation, 
together with 
a set of Prolog programs\footnote{The relevant programs are
available at http://people.cs.kuleuven.be/bart.demoen/sudokutplp. We have used different Prolog
  systems, including SICStus Prolog, B-Prolog and hProlog. These
  programs run in other systems with little change.}
whose help was welcome for guiding our intuition and for dealing with
some of the combinatorial explosion resulting from the symmetries of
the Sudoku puzzle.

A very common formulation of the Sudoku~\cite{sudoku,sudokunarendra}
puzzle is as follows: each 3$\times$3 box, as well as each row and
each column, must contain all the numbers from 1 to 9. As a constraint
satisfaction problem (CSP), the Sudoku puzzle can be modeled using a
set of 81 variables $x_{ij}$, one per row $i \in [1..9]$ and column
$j\in [1..9]$, 81 \emph{domain} constraints indicating that the domain
of each $x_{ij}$ is [1..9], and 27 \emph{all\_different} constraints
with 9 variables each (9 constraints for the variables in each of the
rows, another 9 for each of the columns, and a final 9 for each of the
boxes). We refer to these 27 constraints as {\em the big \mrules} and
use the word {\em Sudoku} in italics to denote the associated CSP model, i.e., the one containing
all 27 big \mrules~ together with the 81 domain constraints.

An \emph{all\_different} constraint can also be formulated as the
pairwise binary inequality constraints of its input variables. For
example, \emph{all\_different}($\{y_1,y_2,y_3\}$) is logically
equivalent to the conjunction of the constraints $y_1 \neq y_2$, $y_1
\neq y_3$, and $y_2 \neq y_3$. We refer to these binary
$\neq$-constraints as the {\em small \mrules}. When Sudoku is
modeled using small \mrules, it is easy to see that each cell is
involved in 20 small \mrules: 8 in the same box, 6 in the same
row and 6 in the same column. Since there are 81 cells, and each
\mrule~is posted twice, there are in total 810 different small \mrules~
(as opposed to 27 big ones). Whenever a CSP model \emph{M} specified  using (big or small)
\mrules, together with the 81 domain constraints, is equivalent to
       {\em Sudoku} (i.e., it has the same set of solutions), we say \emph{M is Sudoku}.

It was always intuitively clear to us that some of the small \mrules~
must be {\em redundant}, i.e., entailed by the others. However, the
questions ``which and what is the size of the largest redundant set of
small \mrules?'' remained to be answered. The situation was even worse
for big \mrules: when we started this research, it was not even clear
to us whether any single big \mrule~is redundant. Both issues are
attacked here: we give a complete answer for the big \mrules~and a
partial answer for the small \mrules.

We begin by recalling some common Sudoku-related terminology in
Section \ref{terminology}.
Section~\ref{representation} introduces our graphical representation
of {\em Sudoku} modeled with big \mrules. This 
representation significantly simplifies the reasoning required for
showing that \emph{some} sets of big \mrules~with six or less elements
are redundant (Section~\ref{twolemmas}).
We then describe a Prolog
 program that
systematically applies these two positive lemmas to find {\em all}
sets of redundant big \mrules~with six or less elements
(Section~\ref{full}). While doing this, we discover seven {\em
  negative} lemmas (Section~\ref{negativelemmas}). The combination of
positive and negative lemmas results in a complete classification of
all sets of 21 (27 - 6) big \mrules~(Section~\ref{usingnegative}).
We then turn to the study of sets of seven big \mrules~and show that
none of them are redundant (Section~\ref{sevenisnotredundant}). As
before, our Prolog program discovers a new negative lemma, whose proof
is also presented graphically.
We then show that at least 20\% of the small \mrules~can be redundant
(Section~\ref{redundantsmall}), and conjecture that no more is
possible.
Finally, in Section~\ref{concl} we conclude and discuss related work
and possible extensions.

\section{Terminology}\label{terminology}

\setlength{\columnsep}{0.5cm}
\setlength{\intextsep}{-0.5cm}

\begin{wrapfigure}{r}{.35\textwidth}
\begin{center}\includegraphics[%
  width=0.28\textwidth,
  keepaspectratio]{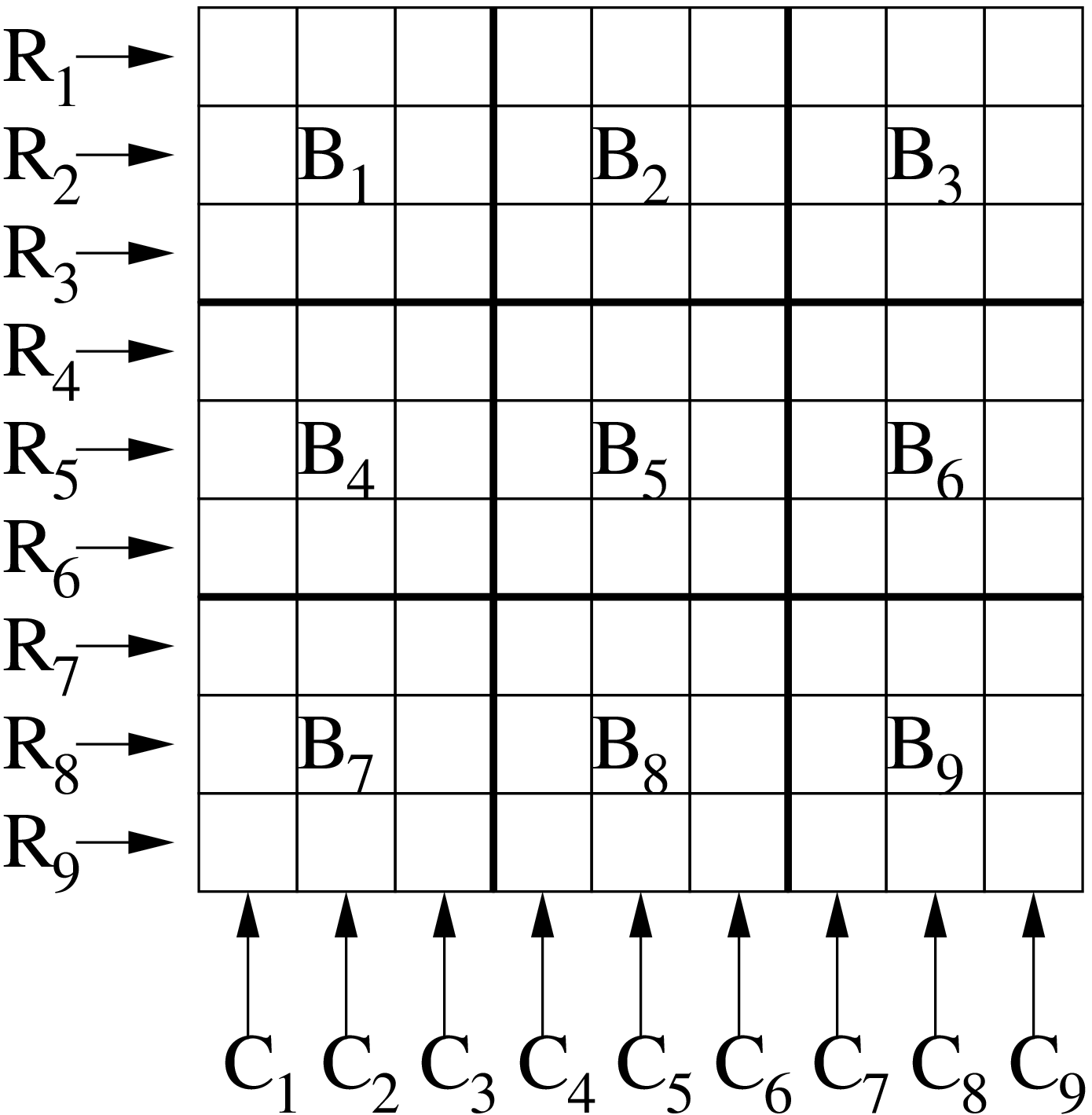}\end{center}
\end{wrapfigure}
The 27 big \mrules~in Sudoku correspond to the 27 regions in which
its board is usually divided (see the attached picture): the 9 rows, 9
columns and 9 boxes, whose big \mrules~will be denoted as
$R_1,\ldots,R_9$, as $C_1,\ldots,C_9$, and as $B_1,\ldots,B_9$,
respectively.  We use the word horizontal (vertical) {\em chute} to
refer to three horizontal (vertical) boxes. For instance, the boxes
associated to constraints $B_2$, $B_5$, and $B_8$ denote a vertical
chute.

As mentioned before, we are interested in exploring Sudoku models where some
big \mrules~are missing. In the following, we will use $\mathit{Missing}(n)$ to
denote the set of Sudoku models that have $27-n$ big~\mrules. For example,
every model in $\mathit{Missing}(5)$ has 22 big \mrules~(plus, of course, the
usual 81 domain constraints).

\section{A graphical Representation of Sets of Big \mrules}\label{representation}

\begin{center}
\begin{figure}[h]
\hspace{1cm}
\includegraphics[width=0.16\textwidth,keepaspectratio]{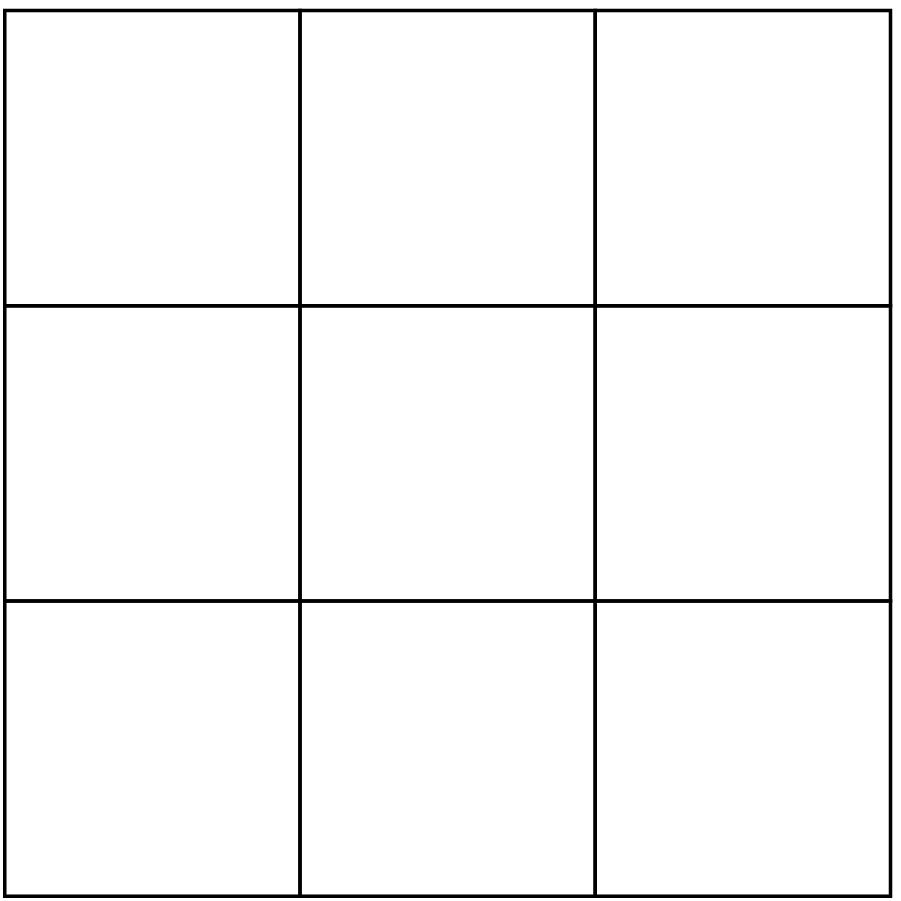}
\hspace{4cm}
\includegraphics[width=0.18\textwidth,keepaspectratio]{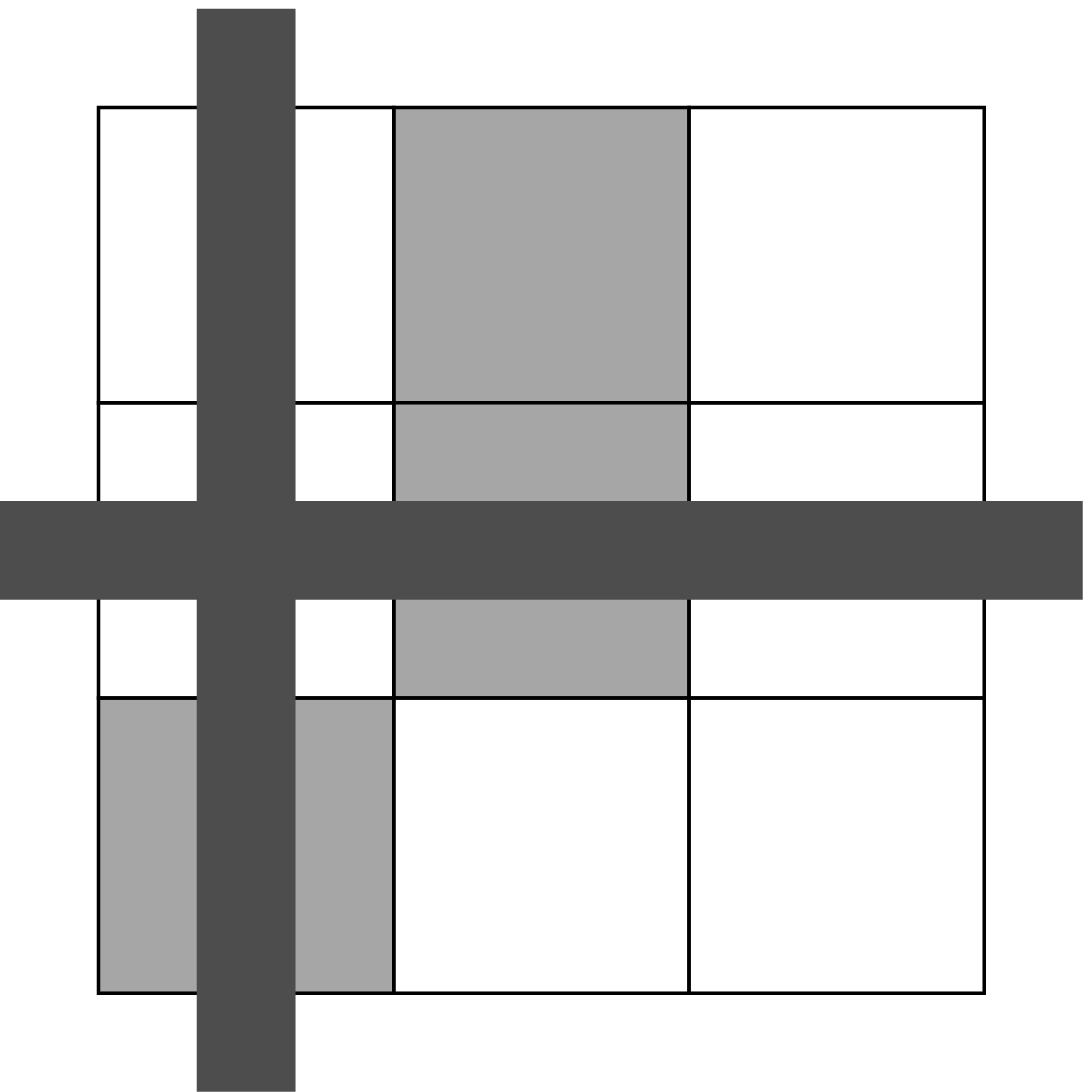}
\caption{The left-hand side of the figure shows {\em Sudoku}, i.e., a
  CSP model with all domain constraints and all big \mrules; the right-hand
  side shows a model with all domain constraints as usual, but with only 22 out of the 27
  big \mrules, since $C_2, R_5, B_2, B_5$, and $B_7$ are marked as
  missing.}
 \label{visual}
\end{figure}
\end{center}

The standard set notation is not visually clear once the number of elements
in the set is high. Since we will be dealing mostly with sets of
more than 20 big \mrules, we have developed a graphical representation of the
Sudoku model that we find more useful. This graphical representation always
shows the borders of the boxes of a Sudoku board and assumes all 81 domain
constraints are specified in the model. Further, all 27 big \mrules~are
also specified unless they are explicitly represented as missing in the
figure. A column, row or box \mrule~is represented as missing if it is
shaded. Figure~\ref{visual} shows an example.

\hspace{1cm}

The pictures provide a quick and intuitive view into which big
\mrules~are present in the model and which are not. Note that the absence of a big
\mrule~does not mean it is violated, simply that it has not been
specified in the associated model.

\begin{center}
\begin{figure}[h]
\includegraphics[height=0.04\textheight,keepaspectratio]{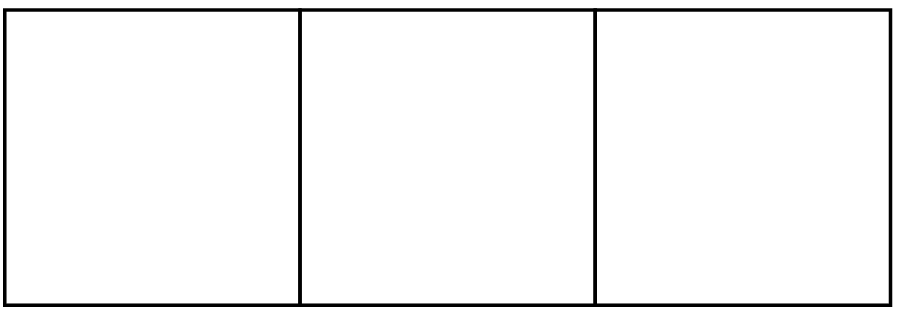}
\hspace{2cm}
\includegraphics[height=0.04\textheight,keepaspectratio]{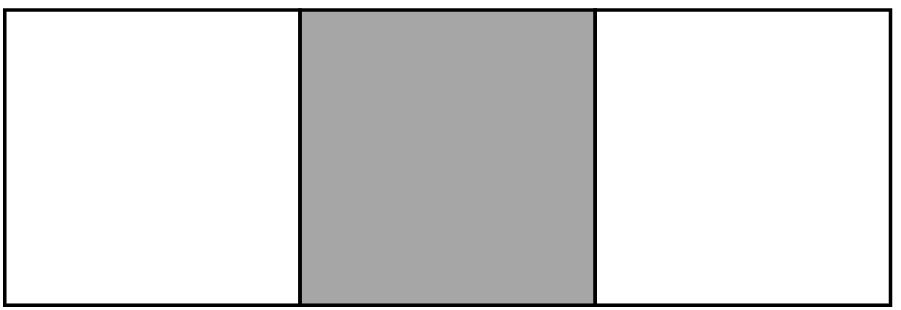}
\hspace{2cm}
\includegraphics[height=0.04\textheight,keepaspectratio]{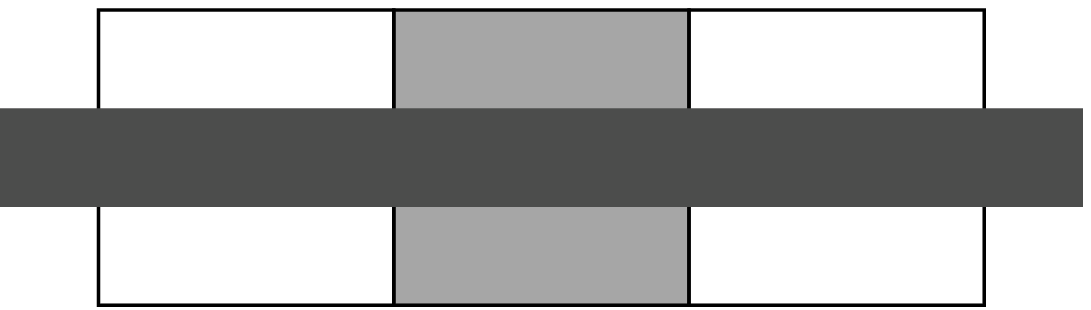}
\caption{The left hand-side represents a chute where all row ($R_1,
  R_2$, and $R_3$) and box constraints ($B_1,B_2$, and $B_3$) are
  present, while the middle one is missing $B_2$ and the right one is
  missing both $B_2$ and $R_2$. Note that, as before, the 27 domain constraints
  associated to the cells in the chute are assumed to be present in
  all three pictures.}\label{chute}
\end{figure}
\end{center}

Using the same idea, we can represent a set of big \mrules~applicable
only to a chute (any chute): this is illustrated in
Figure~\ref{chute}.

\hspace{2cm}

\section{Two Constructive Lemmas}\label{twolemmas}

Let us now prove two positive lemmas, i.e., how a subset of the big
\mrules~can be shown to entail another big \mrule.

\begin{lemma}\label{lemmaI}
The conjunction of the big \mrules~in $\{R_1,R_2,R_3,B_1,B_3\}$ entails
$B_2$. This is represented graphically by means of the following
picture:

\medskip
\hspace{3cm}\includegraphics[height=0.04\textheight,keepaspectratio]{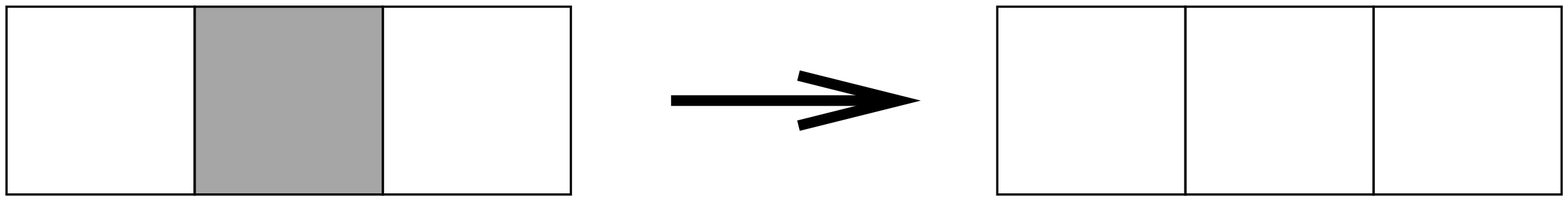}

\end{lemma}
\underline{Proof} Let us fill the chute with 27 numbers, so that the
constraints in
\includegraphics[height=0.02\textheight,keepaspectratio]{chute2.eps}
are satisfied. To do so, let us try to place any value $N \in
1..9$ in the chute. Since $R_1, R_2$ and $R_3$ are present, there must be exactly
one $N$ in each row, which means there must be three $N$s in the
chute. Since $B_1$ and $B_3$ are also present, exactly one of these
three $N$s must be in box 1 and exactly another one in box 3. This
leaves exactly one (the third) $N$ in box 2. Since this holds for any $N \in
1..9, B_2$ also
holds. \prend

\bigskip

The dual of Lemma \ref{lemmaI} is Lemma \ref{lemmaII}.

\begin{lemma}\label{lemmaII}
\hspace{3cm}\includegraphics[height=0.04\textheight,keepaspectratio]{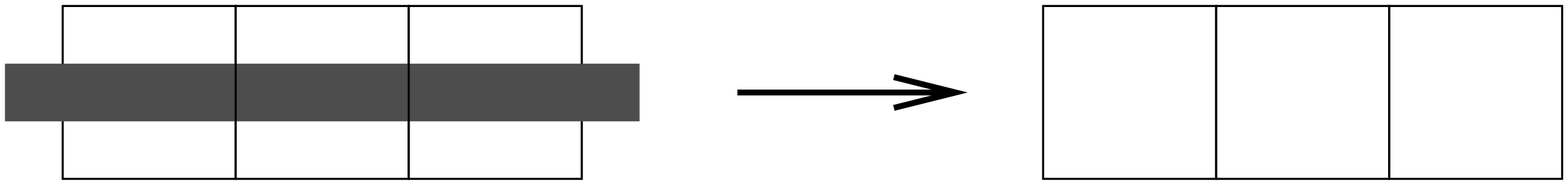}

\end{lemma}
\underline{Proof} Let us fill the chute with 27 numbers, so that the constraints in
\includegraphics[height=0.02\textheight,keepaspectratio]{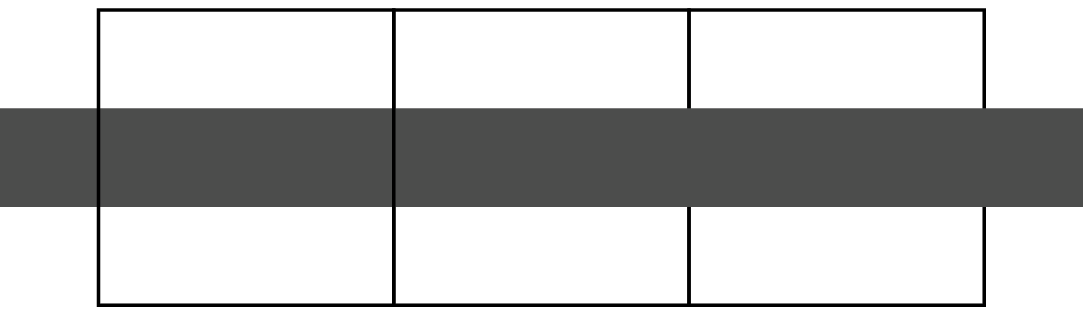}
are satisfied. To do so, let us again try to place any value $N \in
1..9$ in the chute. Since $B_1, B_2$ and $B_3$ are present, there must be exactly
one $N$ in each box, which means there must be three $N$s in the
chute. Since $R_1$ and $R_3$ are also present, exactly one of these
three $N$s must be in row 1 and exactly another one in row 3. This
leaves exactly one (the third) $N$ in row 2. As before, this means $R_2$ also
holds. \prend

\medskip
From now on we assume the graphical
representation is clear enough not to require accompanying text. 
Together with the trivial lemma
\includegraphics[height=0.02\textheight,keepaspectratio]{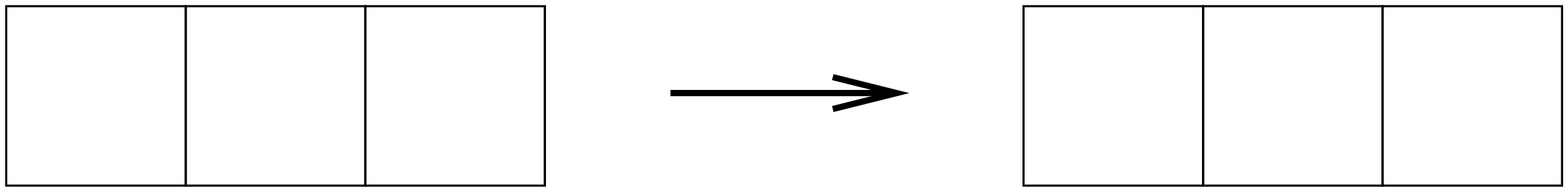}, the
above two lemmas form the building blocks of a corollary and of a whole set of
theorems: we simply glue several applications of these lemmas to form a new
one, as exemplified in the following picture:

\vspace{0.5cm}{\hspace{2cm}\includegraphics[height=0.07\textheight,keepaspectratio]{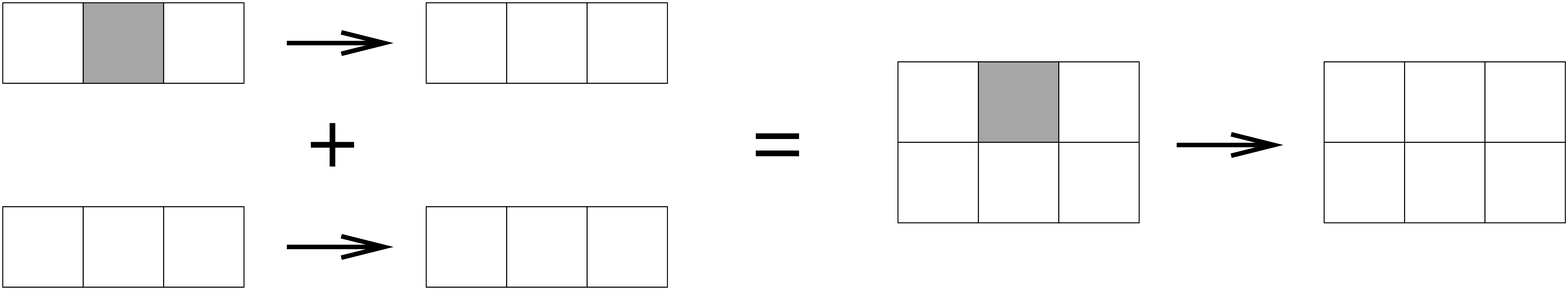}

\medskip
We are now ready for our corollary:

\begin{corollary}\label{cor1}

\vspace{1cm}{~~~~~~~~~~~~~~~~~~~~~~~~~~~~~~~~~~and~~~~~~~~~~~~~~~~are both {\em Sudoku}.}

\vspace{-1cm}{\hspace{2cm}\includegraphics[width=0.1\textwidth,keepaspectratio]{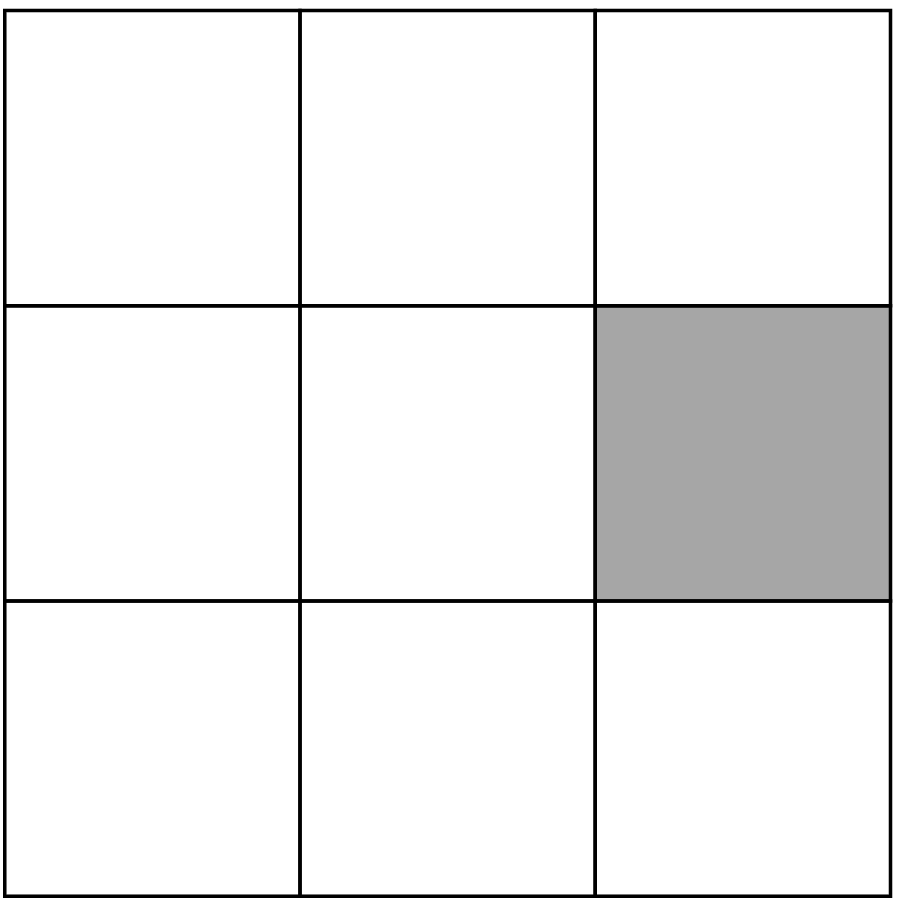}
\hspace{1cm}\includegraphics[width=0.1\textwidth,keepaspectratio]{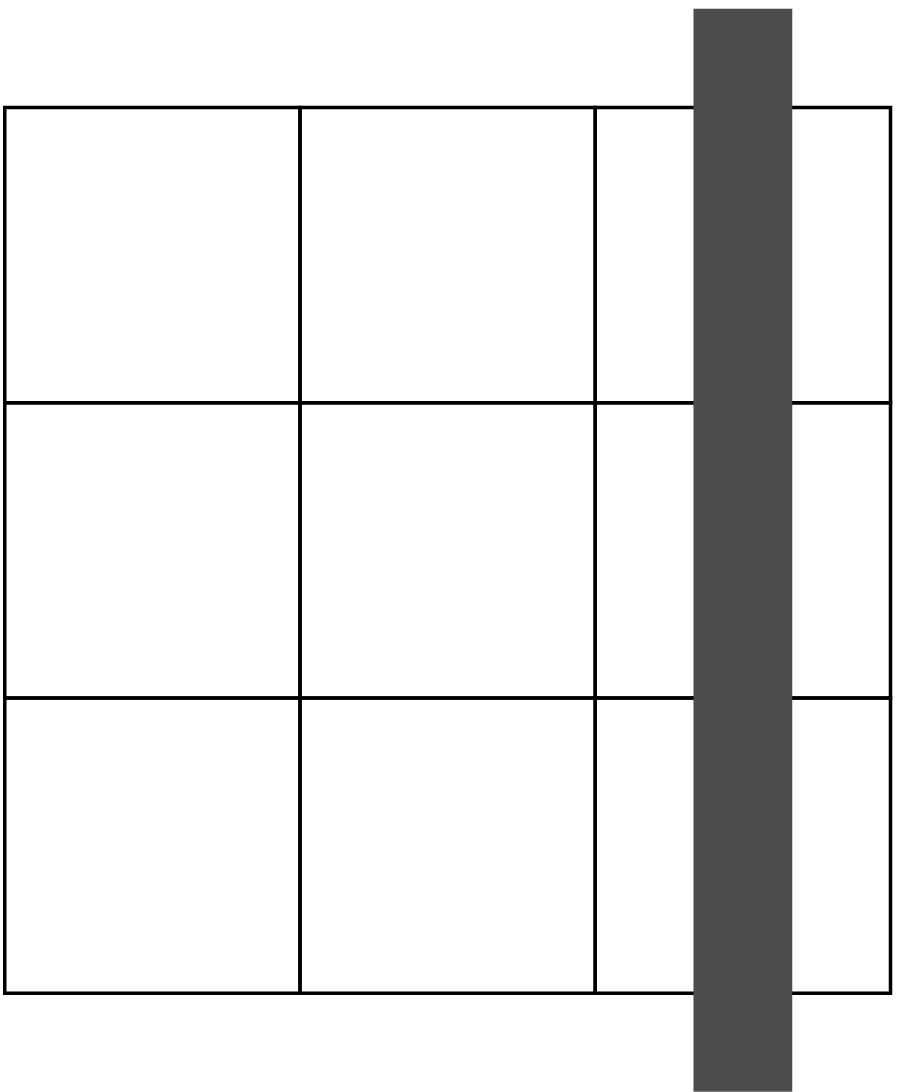}}
\end{corollary}

\underline{Proof}
Glue together twice the trivial lemma with Lemma \ref{lemmaI} and Lemma
\ref{lemmaII}, respectively, and obtain the result immediately. \prend

Taking into account the symmetries of the puzzle, it follows that
every single big \mrule~is (by itself) redundant, i.e., that
every model in $\mathit{Missing}(1)$ is {\em Sudoku}! We will see later that this is
not true for any other $\mathit{Missing}(n)$ with $n > 1$.

Note that the two lemmas really are {\em constructive}, i.e., they show how to
infer one new big \mrule~from a set of big \mrules. The following two
theorems exploit that constructive power to reason further about
redundancy.

\begin{theorem}\label{th1}
\hspace{2cm}\includegraphics[width=0.1\textwidth,keepaspectratio]{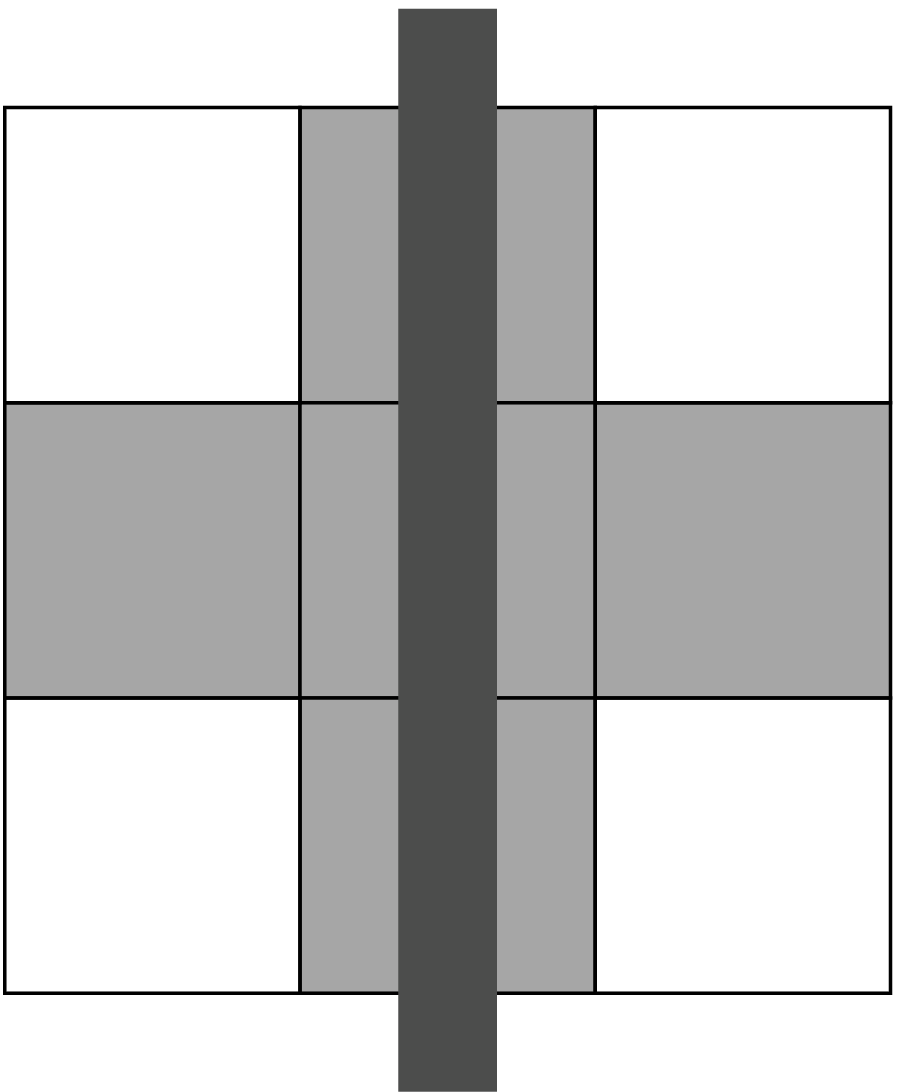}

\vspace{-1.1cm}{~~~~~~~~~~~~~~~~~~~~~~~~~~~~is {\em Sudoku}.}

\end{theorem}
\bigskip
\underline{Proof}
We prove this by repeatedly using Lemmas \ref{lemmaI}
and \ref{lemmaII} as follows: ~\\

\noindent\includegraphics[width=0.9\textwidth,keepaspectratio]{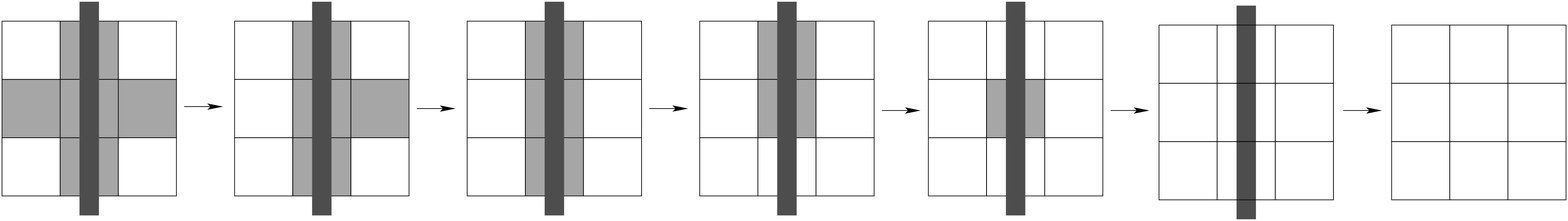}

\noindent
where the first five rewrites use Lemma \ref{lemmaI}, and the last step uses Lemma
\ref{lemmaII}.\prend

\begin{theorem}\label{th2}
\hspace{2cm}\includegraphics[width=0.1\textwidth,keepaspectratio]{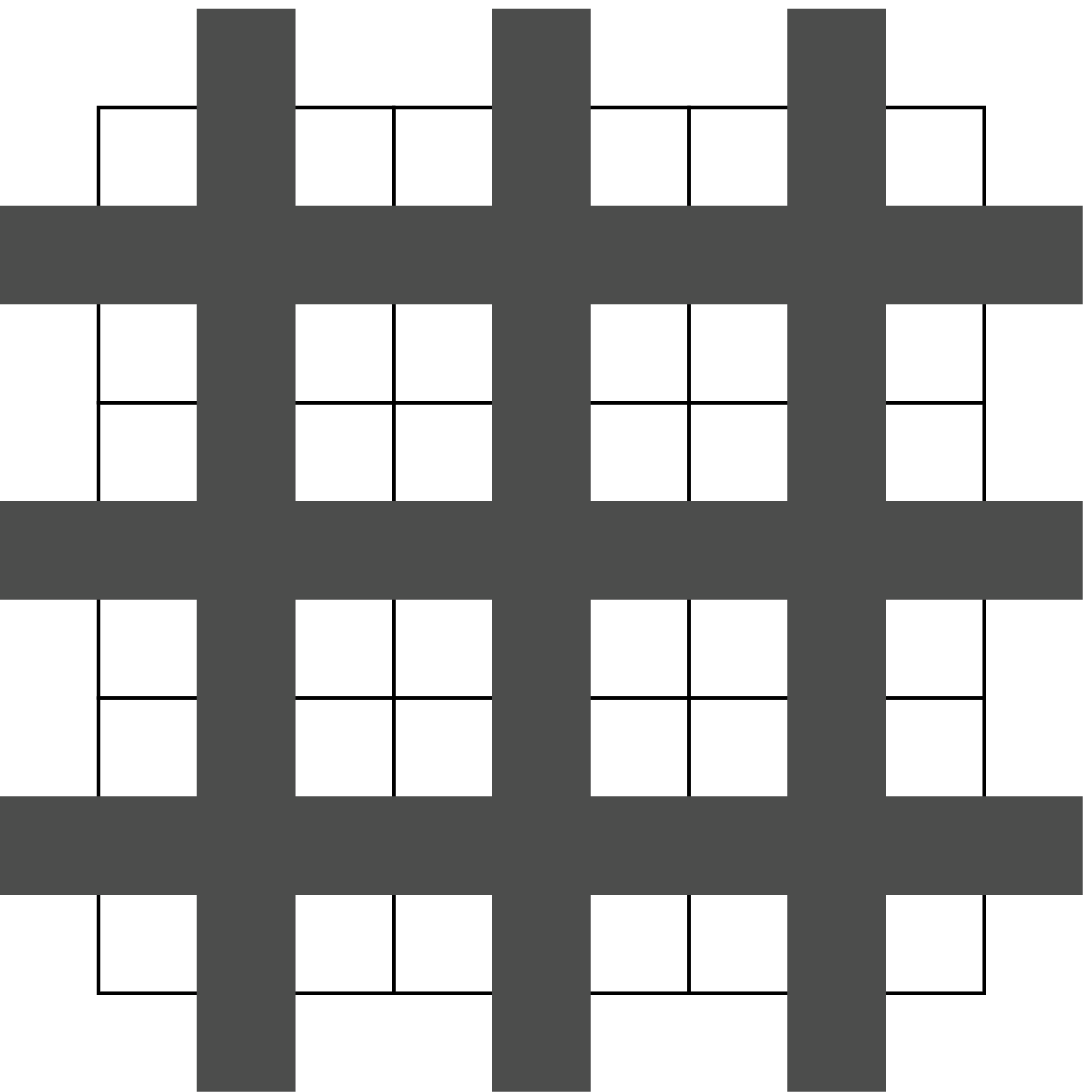}

\vspace{-1.1cm}{~~~~~~~~~~~~~~~~~~~~~~~~~~~~is {\em Sudoku}.}

\end{theorem}
\bigskip
\underline{Proof}
We prove this by repeatedly using Lemma \ref{lemmaII} as follows: ~\\

\noindent\includegraphics[width=0.9\textwidth,keepaspectratio]{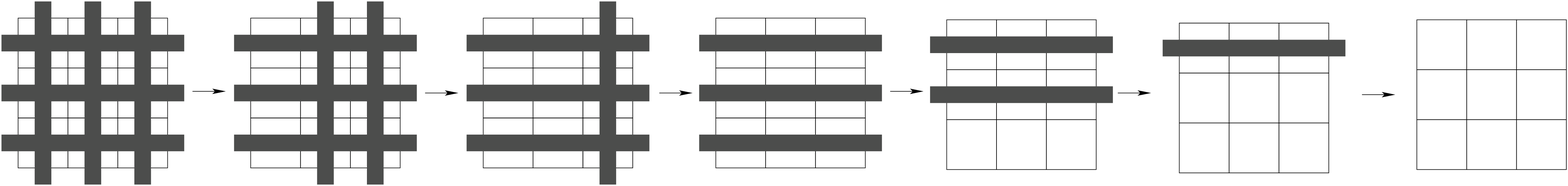}
\prend

Each of the two theorems above shows a model in $\mathit{Missing}(6)$ that is {\em
  Sudoku}. While there are many symmetric versions of these 
theorems, we have chosen those that are visually most pleasing to us. 
The next section fully classifies $\mathit{Missing}(6)$.

\section{A Full Classification of $\mathit{Missing}(6)$}\label{full}

Lemmas \ref{lemmaI} and \ref{lemmaII} allow us to add a new
big \mrule~to a set of big \mrules~while retaining
equivalence, as shown in the proof of Theorem \ref{th1}. We use
this to implement a Prolog program that attempts to classify all models in $\mathit{Missing}(6)$ as either {\em
  Sudoku} or not, and whose simplified form is shown Figure~\ref{alg1}. Intuitively, the program receives as input in
\texttt{MissingN} a list with all models in $\mathit{Missing}(n)$, for
some particular $n$. Then, for each model \texttt{Model} of
\texttt{MissingN}, it exhaustively applies lemmas~\ref{lemmaI}
and~\ref{lemmaII} computing the (possibly reduced) model in
\texttt{NewModel}. If \texttt{NewModel} contains the 27
big~\mrules~(and, thus, it is $Sudoku$) it adds \texttt{Model} to the \texttt{Reducible} list
and, otherwise, it adds
\texttt{NewModel} to the list \texttt{Stuck} of models with less than 27
big~\mrules~at which it got stuck. These latter models need special
attention.

\begin{figure}
\begin{center}
\begin{minipage}{0.7\textwidth}
\begin{Verbatim}[fontsize=\footnotesize,frame=single,xleftmargin=1em]
classify_each([], [], []).
classify_each([Model|MissingN], Stuck, Reducibles) :-
    exhaustively_apply_lemmas(Model, NewModel),
    ( has_all_bigs(NewModel) ->
         Reducibles = [Model|Tail],
         classify_each(MissingN, Stuck, Tail)
    ;
         Stuck = [NewModel|Tail],
         classify_each(MissingN, Tail, Reducibles)
    ).

exhaustively_apply_lemmas(Model, NewModel) :-
    ( apply_lemmaI(Model, ModelI) ->
         exhaustively_apply_lemmas(ModelI, NewModel)
    ; apply_lemmaII(Model, ModelII) ->
         exhaustively_apply_lemmas(ModelII, NewModel)
    ;    
         NewModel = Model
    ).
\end{Verbatim}
\end{minipage}
\end{center}
\caption{Program I}\label{alg1}
\end{figure}

While the number of models in $\mathit{Missing}(6)$ is relatively small
(296,010), we can further reduce it by eliminating the spatially
symmetric models. We have run\footnote{See file \texttt{classify.pl}
at the already mentioned website: the actual \mbox{Prolog} code has an extra argument collecting the models shown in Appendix II}  the complete Program I over the (reduced) set of $\mathit{Missing}(n)$ for $n=6$ (from which we can
also derive the results for $n=[2..5]$).  Surprisingly, the program
only failed to prove equivalence to {\em Sudoku} for the following
models:

\medskip
\includegraphics[width=0.1\textwidth,keepaspectratio]{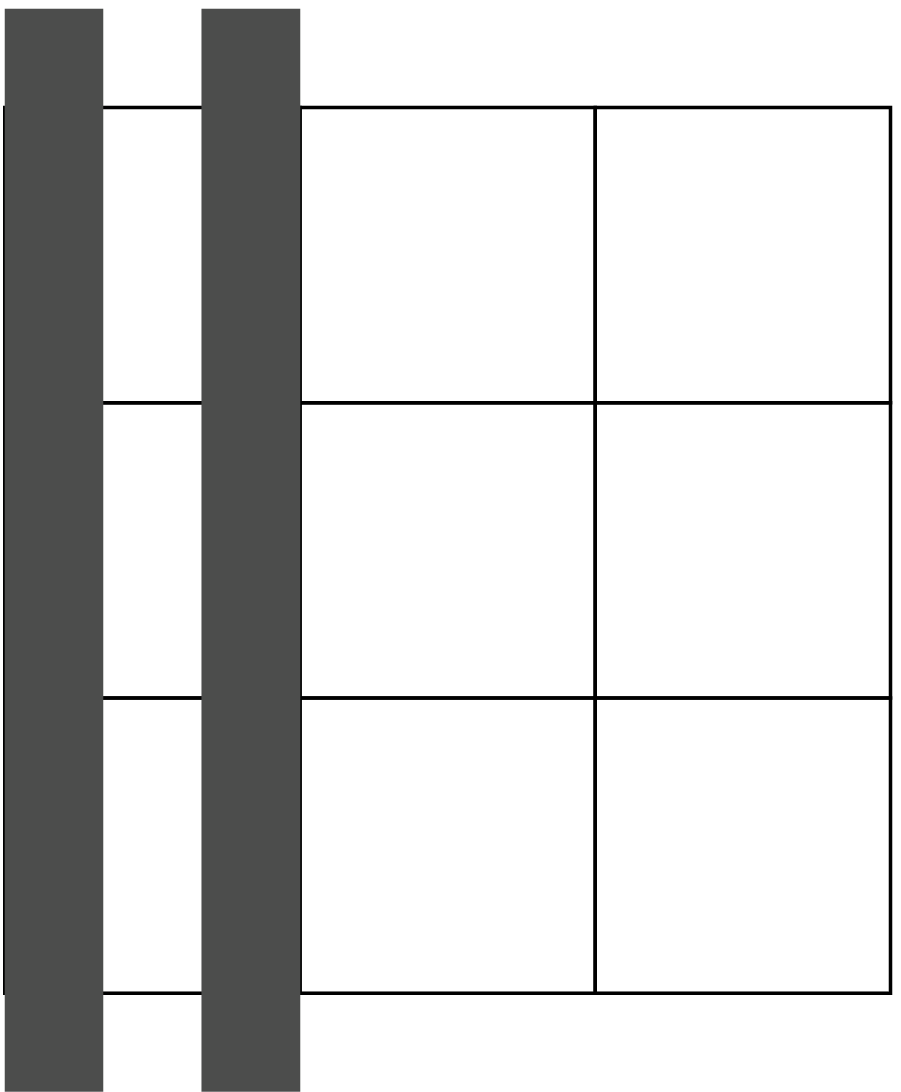}
\hspace{0.3cm}
\includegraphics[width=0.1\textwidth,keepaspectratio]{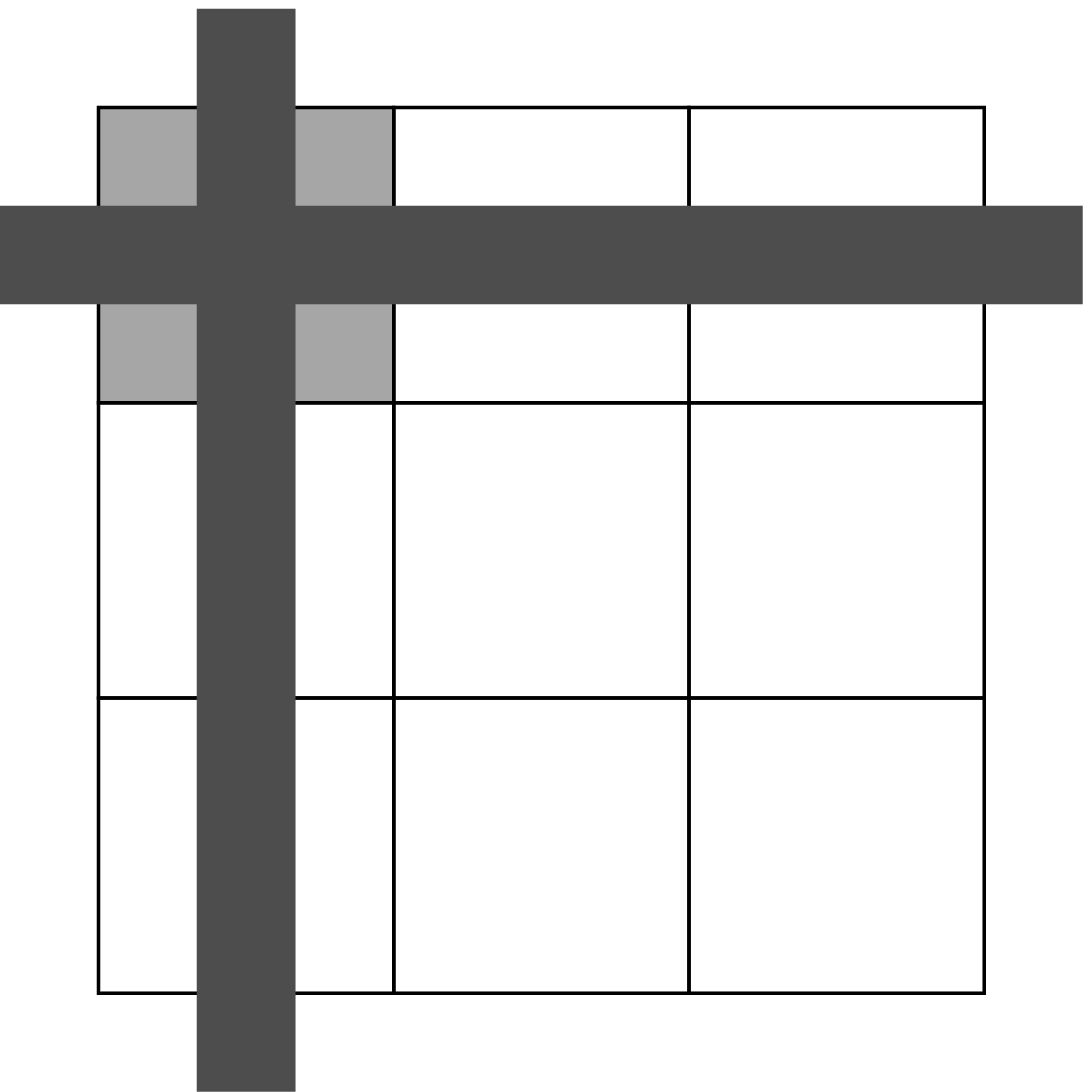}
\hspace{0.3cm}
\includegraphics[width=0.1\textwidth,keepaspectratio]{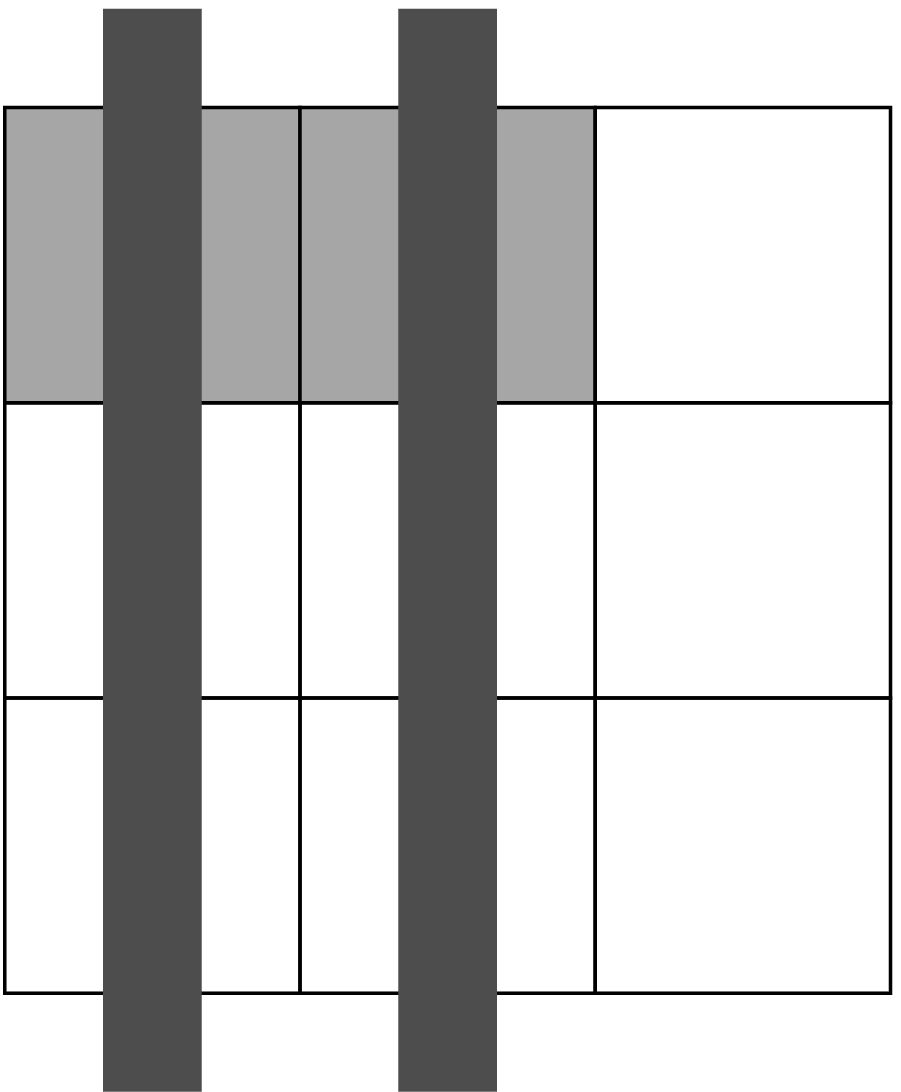}
\hspace{0.3cm}
\includegraphics[width=0.1\textwidth,keepaspectratio]{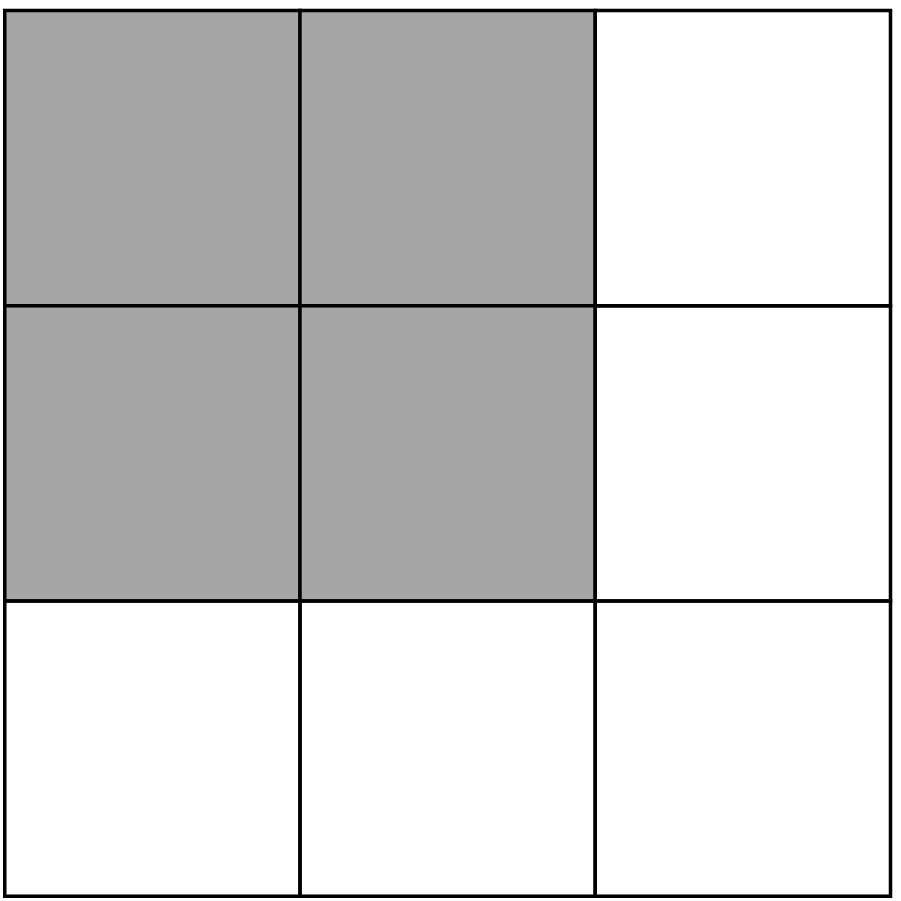}
\hspace{0.3cm}
\includegraphics[width=0.1\textwidth,keepaspectratio]{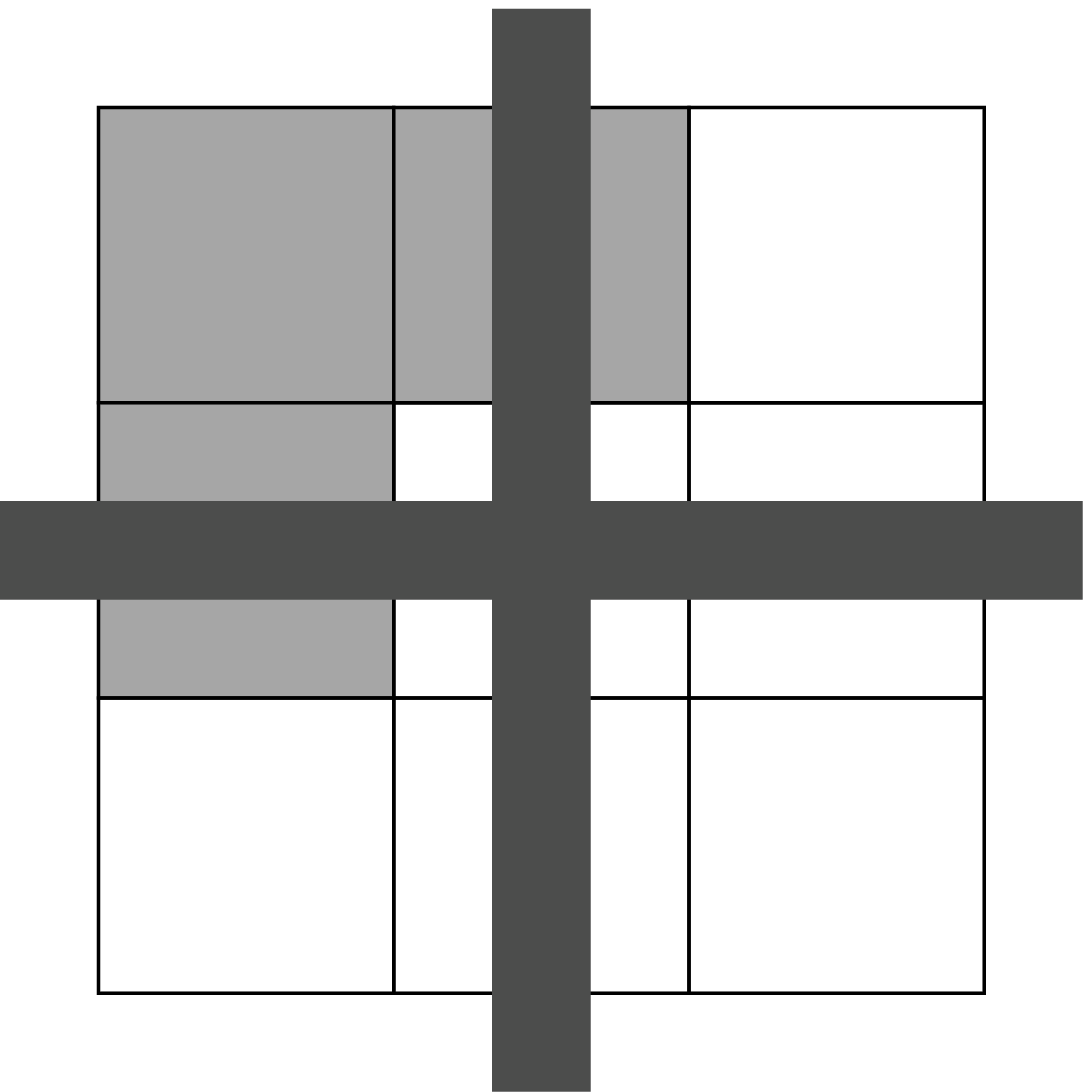}
\hspace{0.3cm}
\includegraphics[width=0.1\textwidth,keepaspectratio]{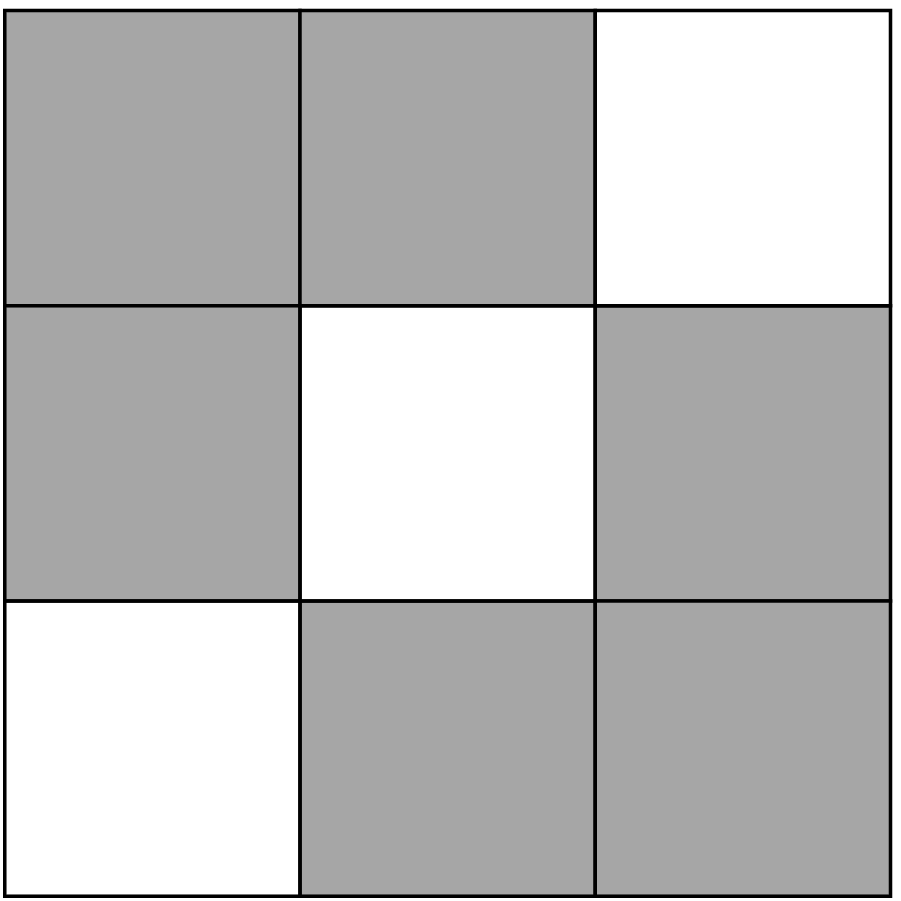}
\hspace{0.3cm}
\includegraphics[width=0.1\textwidth,keepaspectratio]{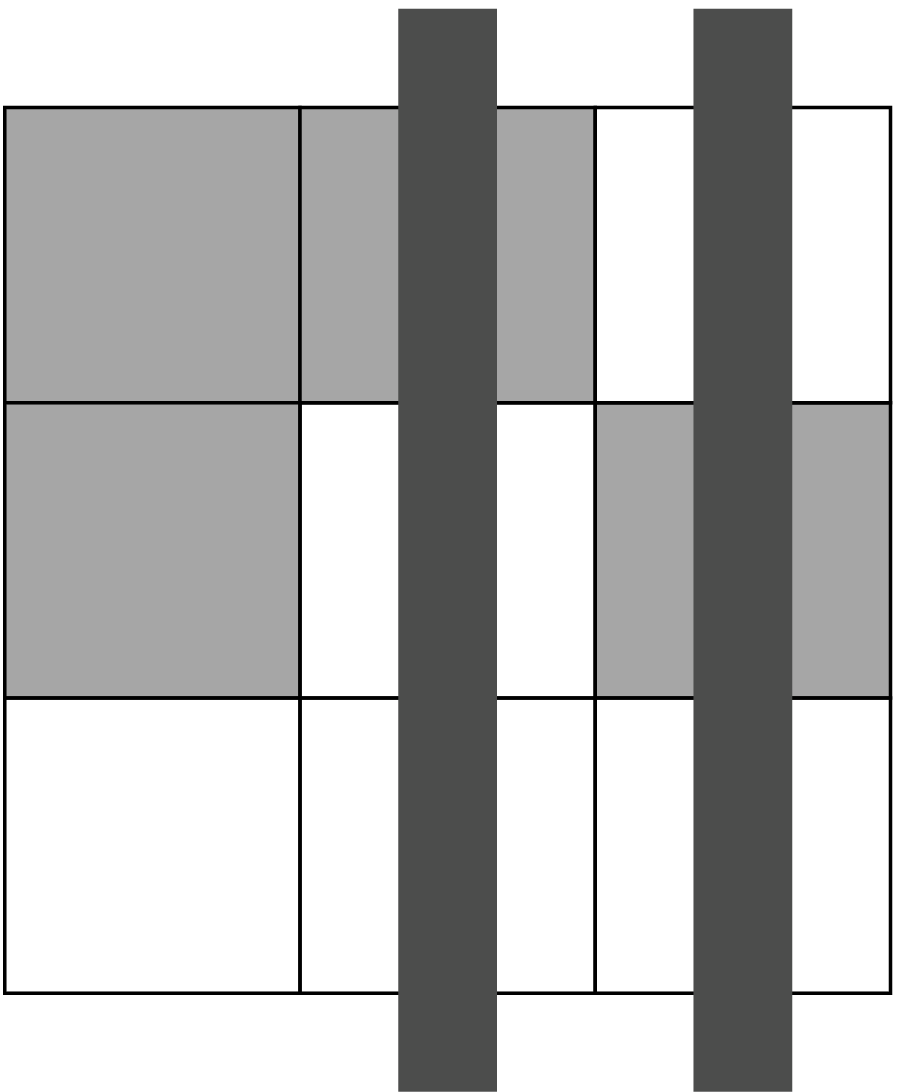}

\medskip
Note that while the last two are models in $\mathit{Missing}(6)$, the others
(from right to left) are models in $\mathit{Missing}(5), \mathit{Missing}(4),
\mathit{Missing}(3)$, and $\mathit{Missing}(2)$, which were obtained during the proving
process by applying Lemma \ref{lemmaI} or \ref{lemmaII} to some model in $\mathit{Missing}(6)$. As
we prove in the next section, none of these seven models is {\em Sudoku}
and, thus, none of the models in $\mathit{Missing}(6)$ whose proof got stuck is {\em Sudoku} either. This is because if a model
$M$ in $\mathit{Missing}(n)$ is not {\em Sudoku}, then any model $M'$ in
$\mathit{Missing}(n')$ where $n' > n$ and the constraints in $M'$ are a subset
of those in $M$, cannot be {\em Sudoku} either. 

\subsection{Seven {\em Negative} Lemmas}\label{negativelemmas}

We proceed by proving seven {\em negative} lemmas, stating that each of the
seven models shown above is not {\em Sudoku}. The proof to each
lemma consists of two pictures: the left picture represents a solution
to the Sudoku puzzle where the circled cells have the specified 4 or 5
value (note that there might be many solutions that satisfy this). For
example, in the first lemma, the left picture represents any solution where
cell $x_{11}$ has value 4 and cell $x_{13}$ has value 5. The right
picture in a proof represents the result of changing every circled 4 in the
left picture by a circled 5, and vice versa. In all cases the result is a
non-solution (to {\em Sudoku}) with the violated big \mrules~depicted
as shaded.
These violated constraints are exactly those that, if removed, the
lemma claims cannot yield {\em Sudoku}.
Since the picture proves that if the big \mrules~in question are removed the
non-solution is accepted as a solution, the lemma is proved.

\begin{lemma}\label{lemmatrivial}
\hspace{2cm}\includegraphics[width=0.1\textwidth,keepaspectratio]{2c.eps}

\vspace{-1cm}{~~~~~~~~~~~~~~~~~~~~~~~~~~~~is not {\em Sudoku}.}
\end{lemma}
\underline{Proof}

\hspace{2cm}
\includegraphics[width=0.35\linewidth,keepaspectratio]{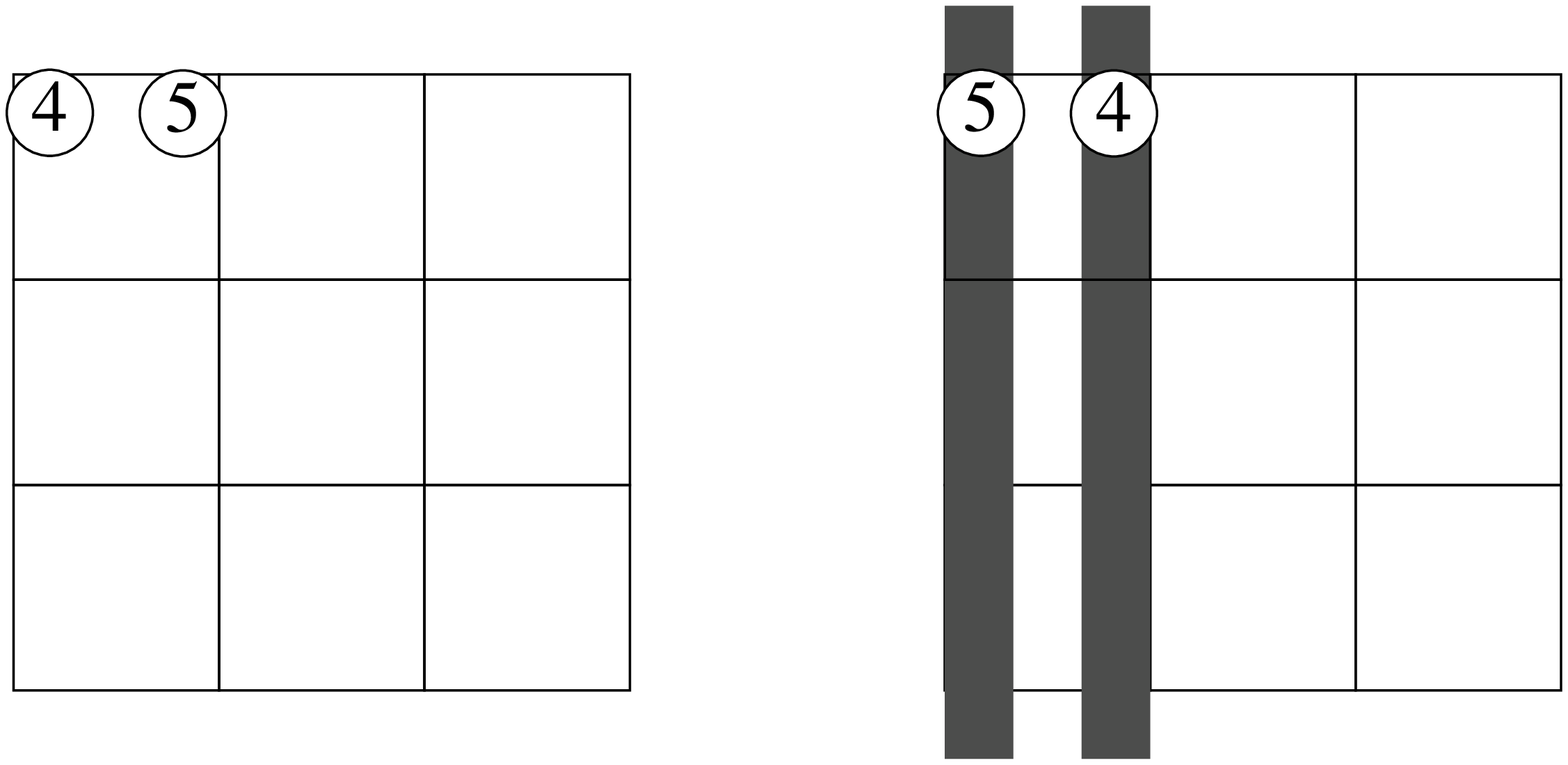}
\prend

The other six negative lemmas follow the same schema.  We expect 
readers to work out the details for them after convincing
themselves that such an initial solution exists for each proof
(some such solutions are provided in Appendix III). 

\begin{lemma}\label{lemmaM}
\hspace{2cm}
\includegraphics[width=0.1\linewidth,keepaspectratio]{M.eps}

\vspace{-1cm}{~~~~~~~~~~~~~~~~~~~~~~~~~~~~is not {\em Sudoku}.}
\end{lemma}
\underline{Proof}

\hspace{2cm}
\includegraphics[width=0.35\linewidth,keepaspectratio]{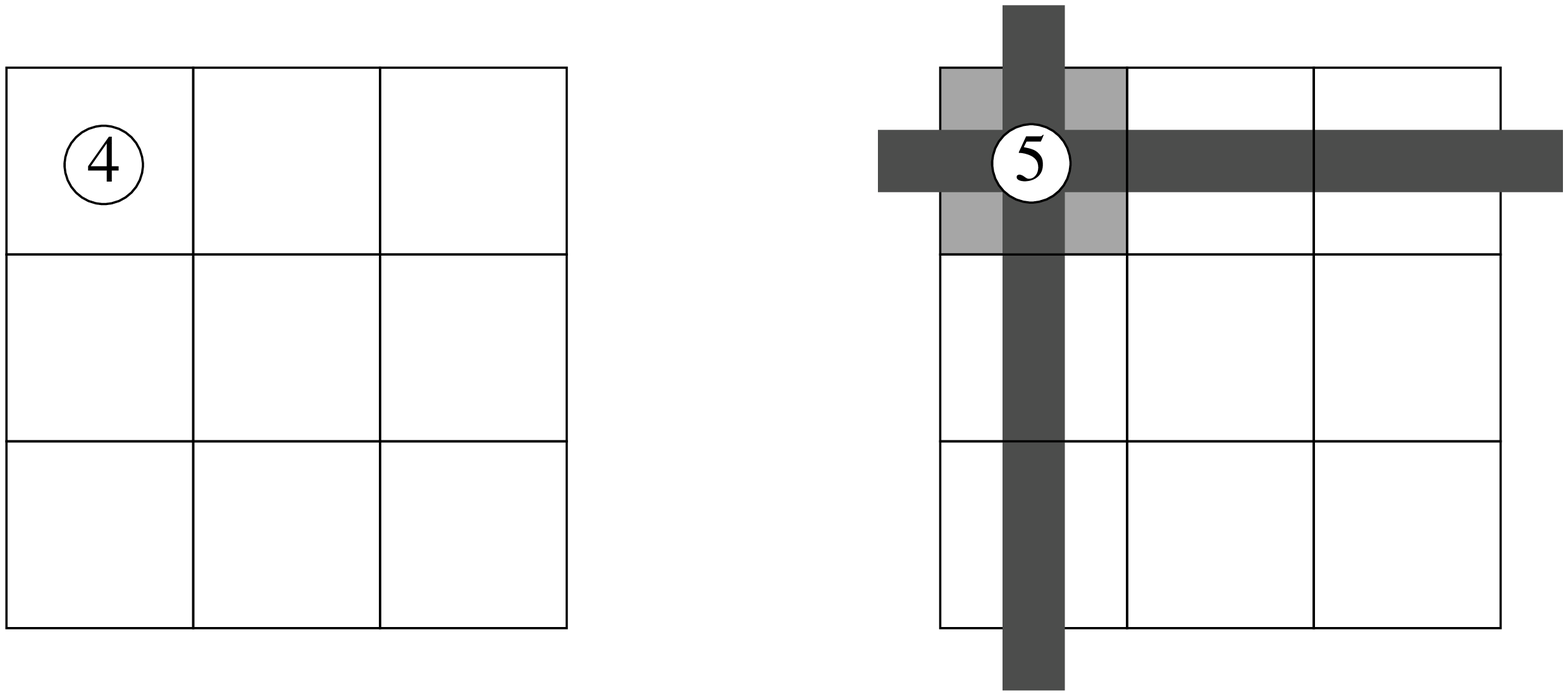}
\prend

\begin{lemma}\label{lemma2}
\hspace{2cm}
\includegraphics[width=0.1\linewidth,keepaspectratio]{2.eps}

\vspace{-1cm}{~~~~~~~~~~~~~~~~~~~~~~~~~~~~is not {\em Sudoku}.}
\end{lemma}
\underline{Proof}

\hspace{2cm}
\includegraphics[width=0.35\linewidth,keepaspectratio]{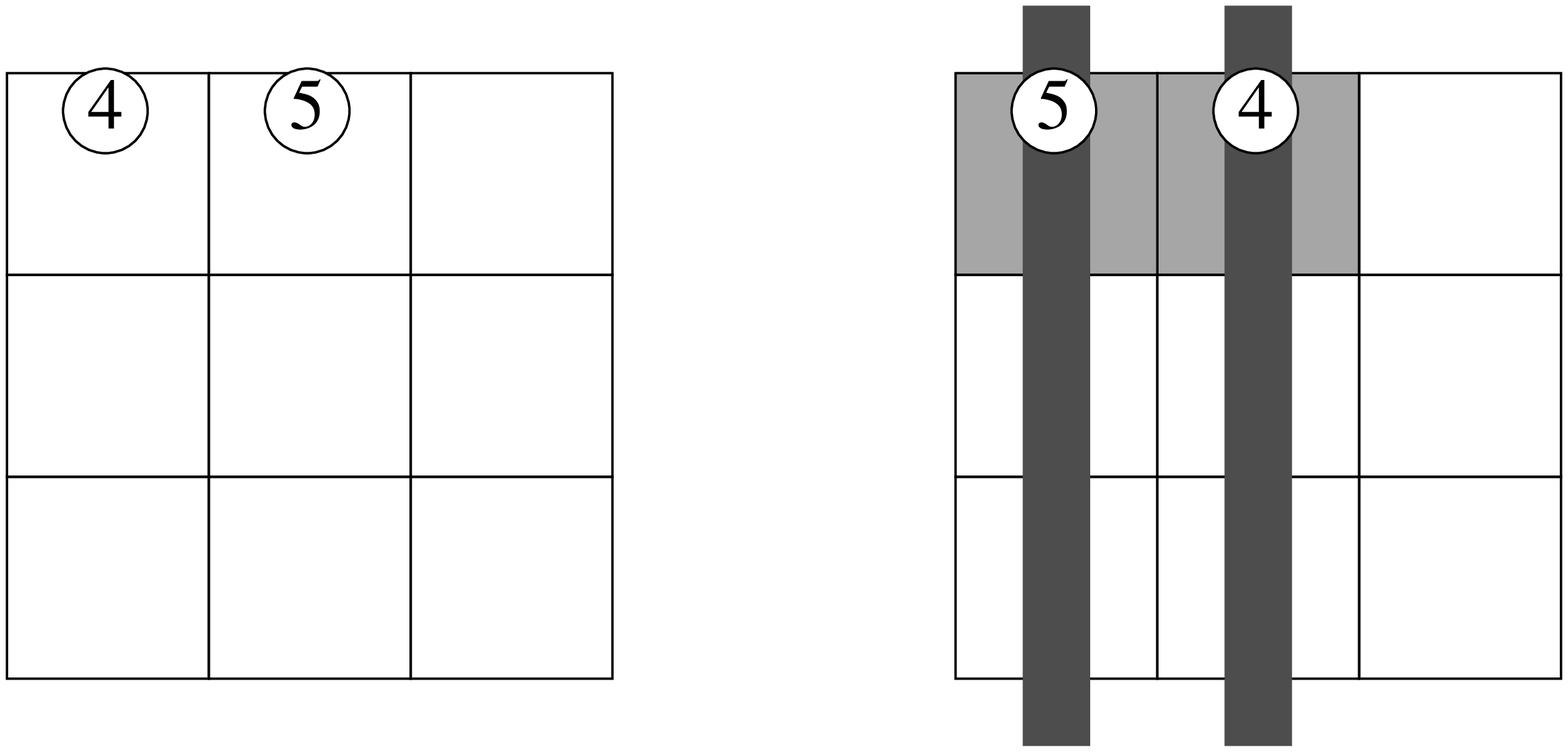}
\prend

\begin{lemma}\label{lemma4}
\hspace{2cm}
\includegraphics[width=0.1\linewidth,keepaspectratio]{4.eps}

\vspace{-1cm}{~~~~~~~~~~~~~~~~~~~~~~~~~~~~is not {\em Sudoku}.}
\end{lemma}
\underline{Proof}

\hspace{2cm}
\includegraphics[width=0.35\linewidth,keepaspectratio]{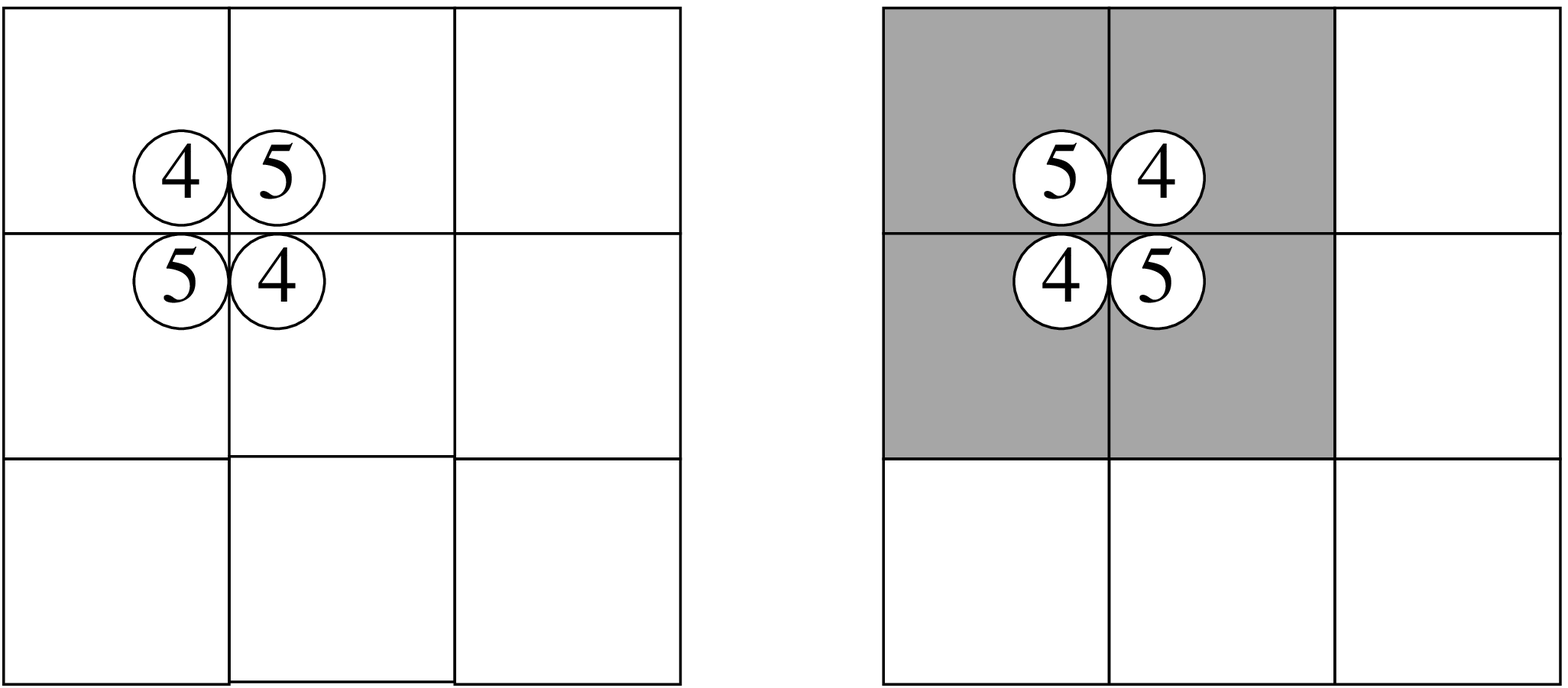}
\prend

\begin{lemma}\label{lemmaT}
\hspace{2cm}
\includegraphics[width=0.1\linewidth,keepaspectratio]{T.eps}

\vspace{-1cm}{~~~~~~~~~~~~~~~~~~~~~~~~~~~~is not {\em Sudoku}.}
\end{lemma}
\underline{Proof}

\hspace{2cm}
\includegraphics[width=0.35\linewidth,keepaspectratio]{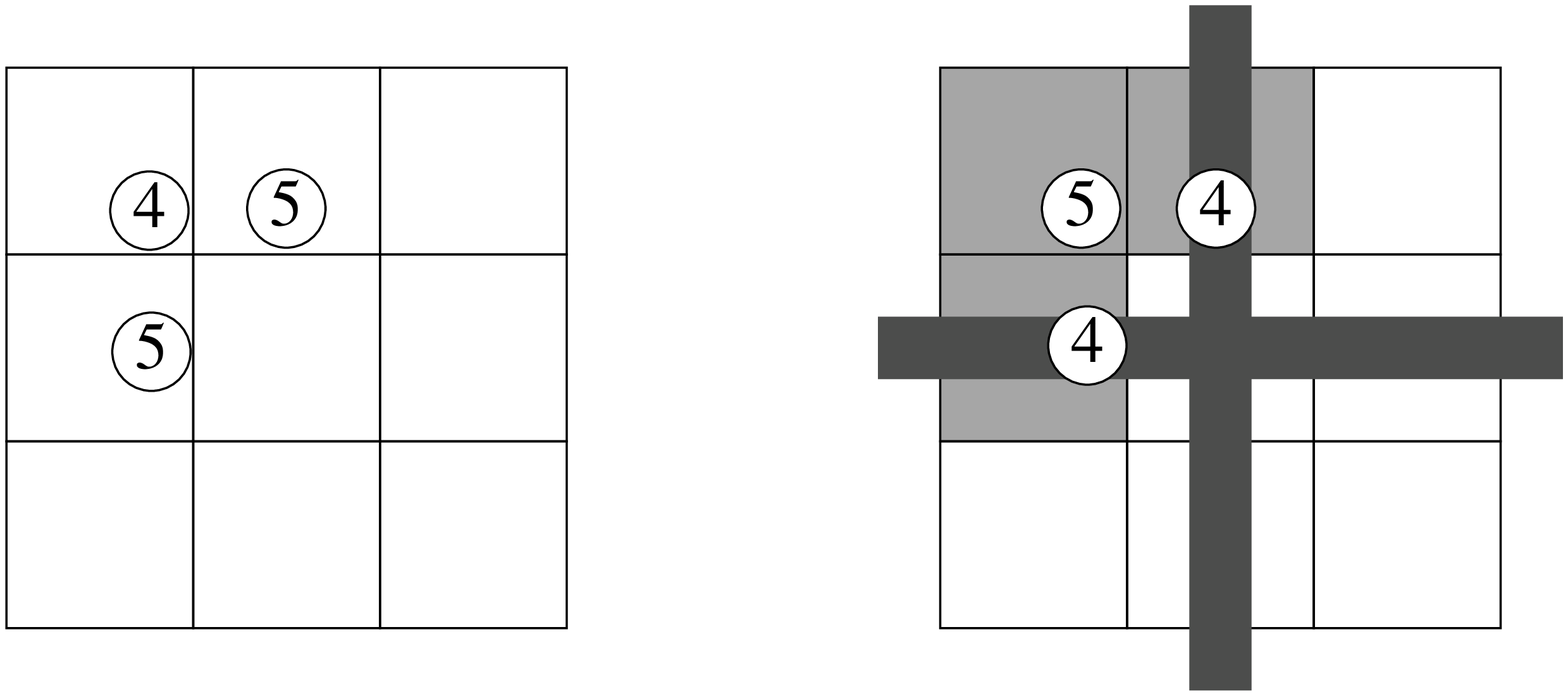}
\prend

\begin{lemma}\label{lemmaF}
\hspace{2cm}
\includegraphics[width=0.1\linewidth,keepaspectratio]{F.eps}

\vspace{-1cm}{~~~~~~~~~~~~~~~~~~~~~~~~~~~~is not {\em Sudoku}.}
\end{lemma}
\underline{Proof}

\hspace{2cm}
\includegraphics[width=0.35\linewidth,keepaspectratio]{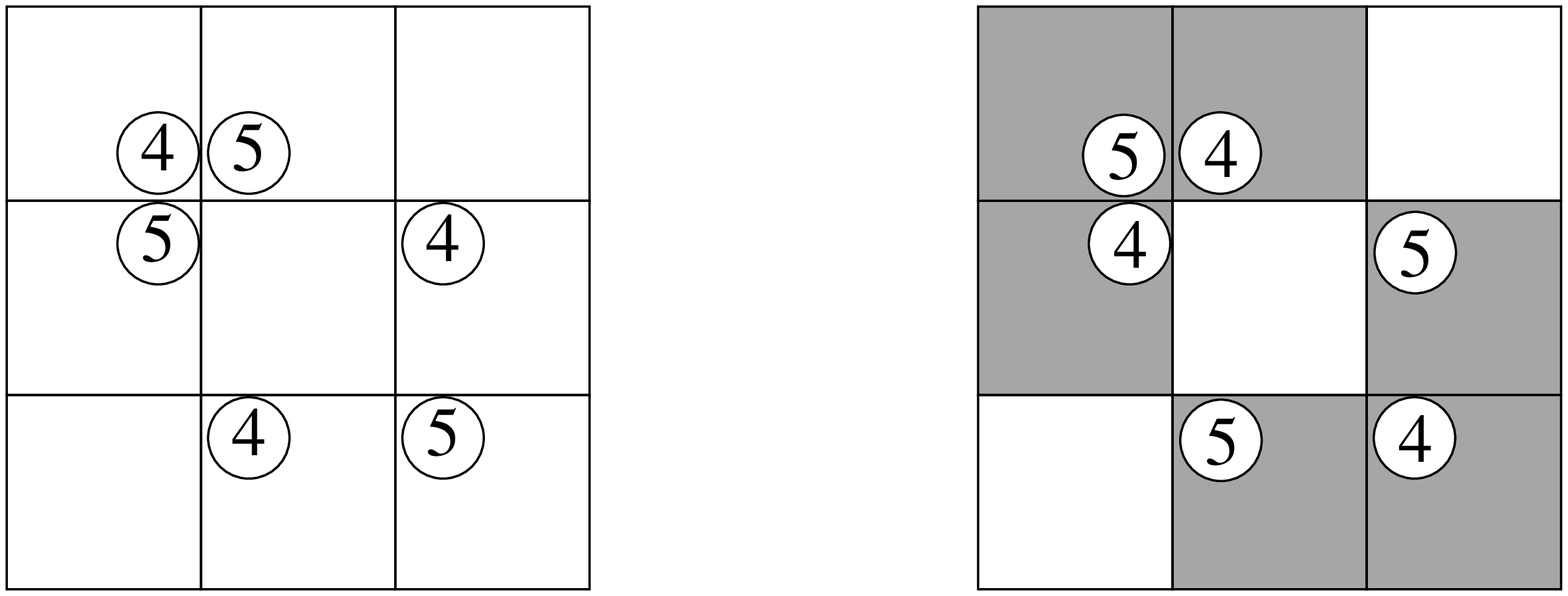}
\prend

\begin{lemma}\label{lemmaxxx}
\hspace{2cm}
\includegraphics[width=0.1\linewidth,keepaspectratio]{X.eps}

\vspace{-1cm}{~~~~~~~~~~~~~~~~~~~~~~~~~~~~is not {\em Sudoku}.}
\end{lemma}
\underline{Proof}

\hspace{2cm}
\includegraphics[width=0.35\linewidth,keepaspectratio]{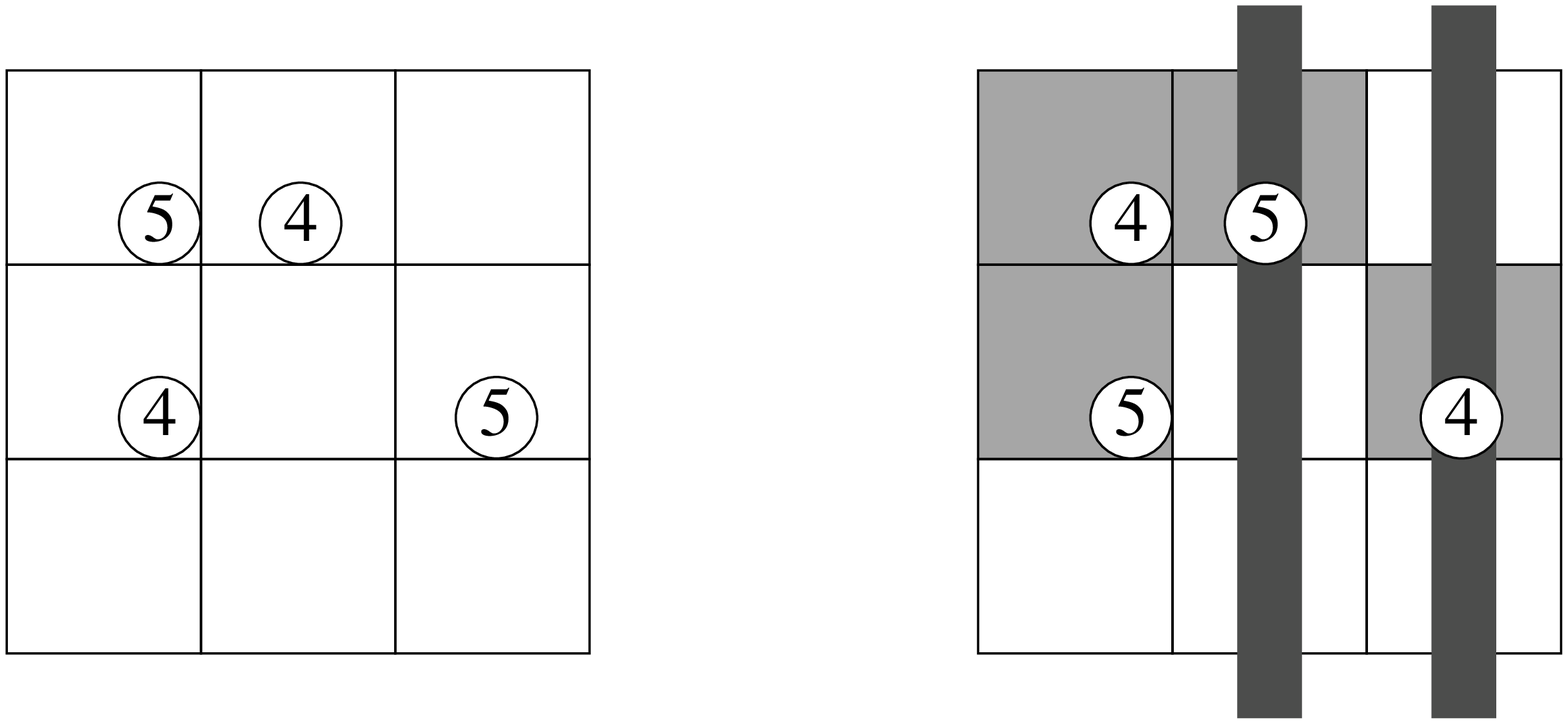}
\prend

\subsection{Making Use of the Negative Lemmas}\label{usingnegative}

The above positive and negative lemmas give us a complete method for
determining whether any model in $\mathit{Missing}(6)$ is {\em Sudoku} or
not: if the application of the constructive lemmas results in {\em
  Sudoku}, then the model is {\em Sudoku}; otherwise it will get
stuck in one of the seven negative models and, thus, is known not to
be {\em Sudoku}. In this sense, the two constructive lemmas are
complete (and also confluent). This can be used to render our first program more useful
by changing the \texttt{classify\_each/3} predicate to 
also check whether the models who do not have all
27 big constraints are one of the seven negative lemmas. If so, it ignores them,
otherwise, as before, it adds them to \texttt{Stuck}. Note that, for the
  case of $\mathit{Missing}(6)$, \texttt{Stuck} is then empty. We
refer to the modified version of Program I by
Program II, and we have further modified it to generate the pictures\footnote{See file \texttt{genfigs.pl}
at the website.} that can be found in
the appendices~I and~II: we run this modified program with $n = 6$ and, for each
model in (the reduced) $\mathit{Missing}(6)$ a picture is
output. Interestingly, there are 39 different models in
(the symmetry reduced) $\mathit{Missing}(6)$ that are {\em Sudoku}, and 70 that are not.

\subsection{No Model in $\mathit{Missing}(7)$ is {\em Sudoku}}\label{sevenisnotredundant}

When we run Program II with $n = 7$, every model gets stuck either in one of the previous seven lemmas, or in a model with a new set
of big \mrules. This model results in one more negative lemma, which is not
implied by any of the previous negative lemmas.

\begin{lemma}\label{lemmaon7}
\hspace{2cm}
\includegraphics[width=0.1\linewidth,keepaspectratio]{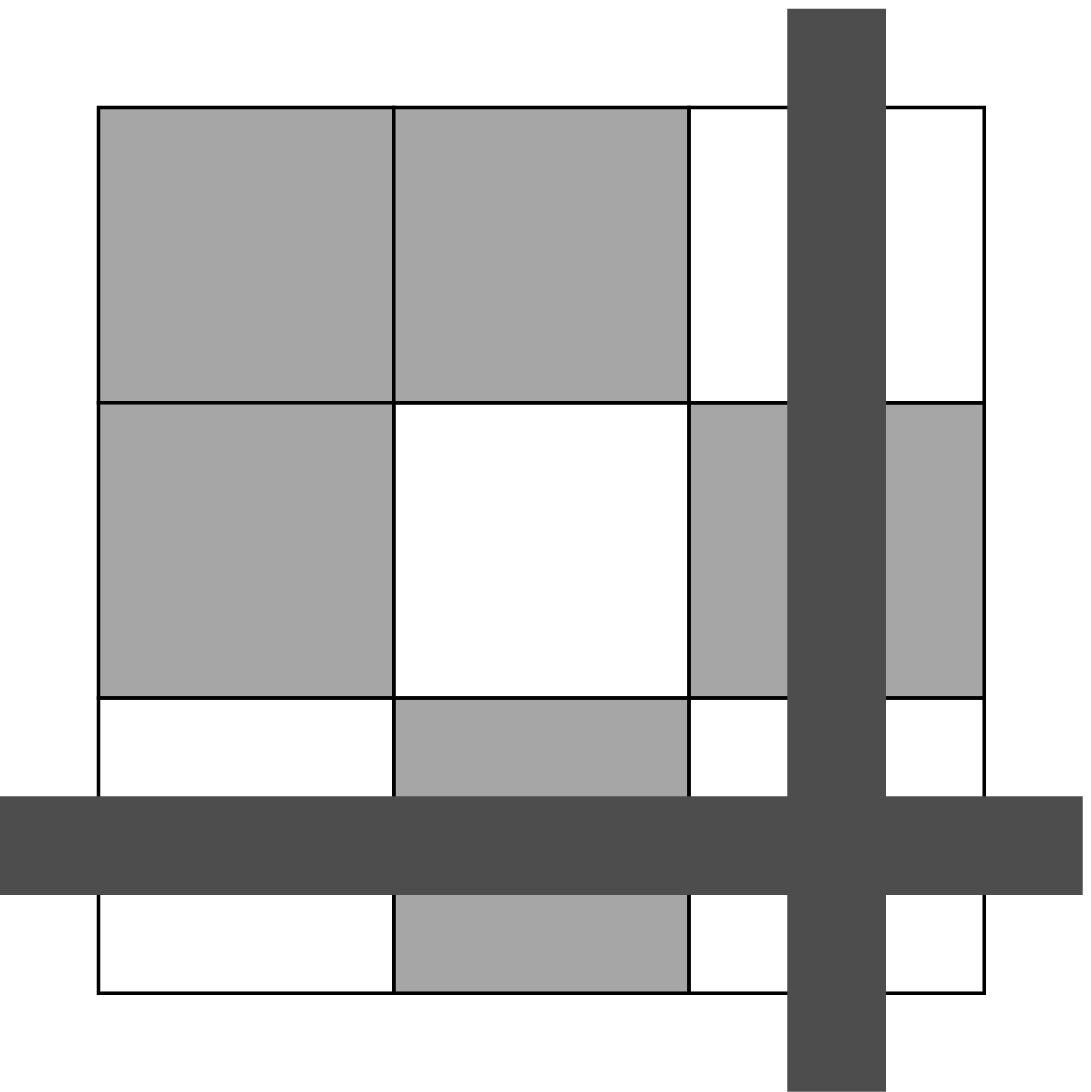}

\vspace{-1cm}{~~~~~~~~~~~~~~~~~~~~~~~~~~~~is not {\em Sudoku}.}
\end{lemma}
\underline{Proof}

\hspace{2cm}
\includegraphics[width=0.35\linewidth,keepaspectratio]{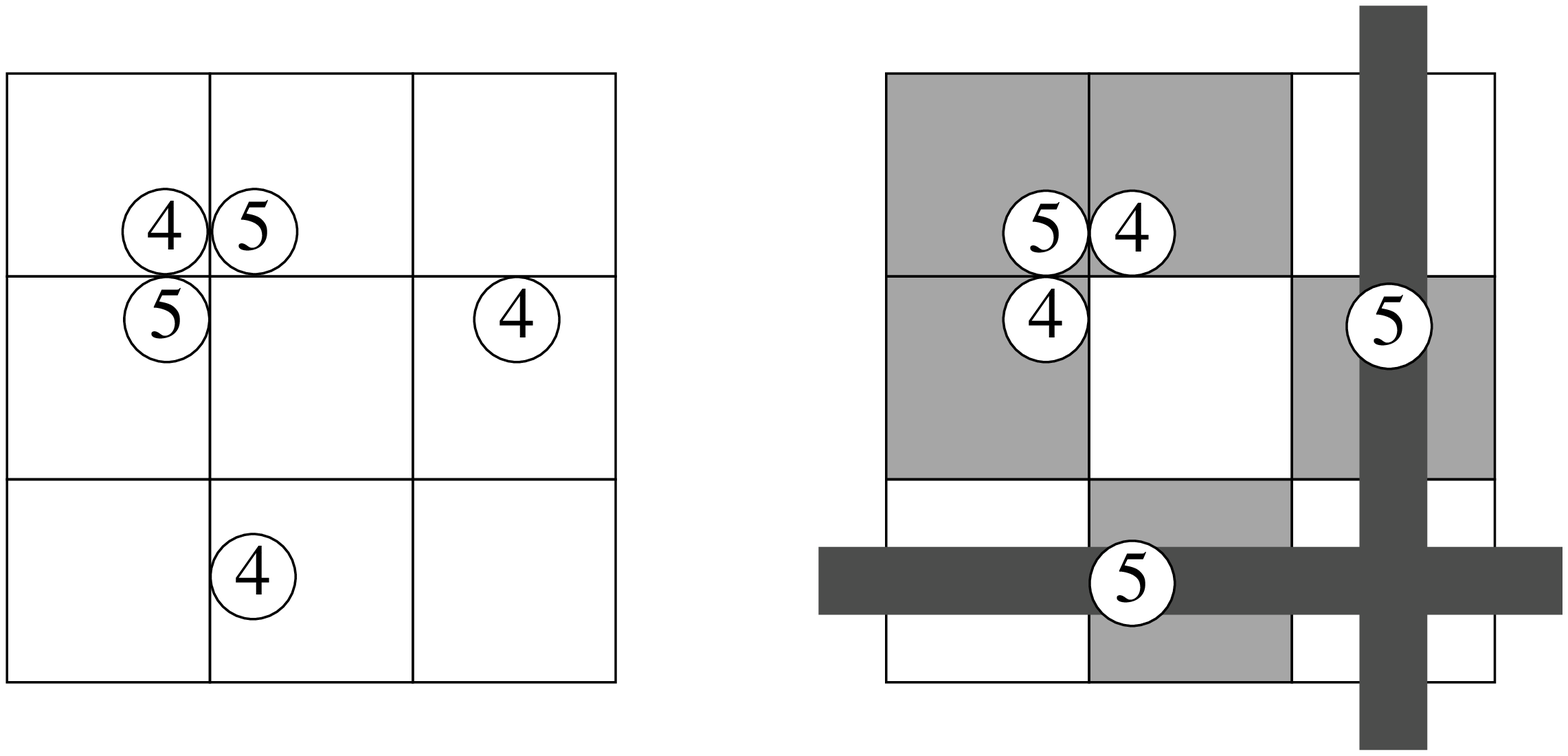}
\prend

\medskip
Readers can easily check that none of the models in
Lemma~\ref{lemmatrivial} up to Lemma~\ref{lemmaxxx} is contained in the
above model.
As a result, no model in
$\mathit{Missing}(7)$ is {\em Sudoku}. Or put otherwise: no redundant set of
big \mrules~has more than six elements. 

\subsection{Generalizing to Puzzles of Size N}\label{generalisation}

Our techniques can be readily applied to the investigation of
Sudoku puzzles of different sizes. Up to now, we have dealt with puzzles of
size 3, i.e., there are $3^4$ cells, in a $3^2$ by $3^2$ board, with $3^2$
rows, columns and boxes.  Clearly, Lemmas \ref{lemmaI} and \ref{lemmaII}
generalize easily to other sizes. For example, for size 4, one just needs
to add one non-shaded block constraint to the pictures to ensure the lemmas remain true.

This suggests that for size $n$, no model in $\mathit{Missing}(2*n+1)$ is {\em
  Sudoku}. Proving this is, however, outside the scope of the current paper. 

\section{Redundancy for the Small~\mRules}\label{redundantsmall}

For each of the models in $\mathit{Missing}(6)$ one can easily count the
number of different small \mrules~it represents: for the ones that are
{\em Sudoku}, the highest count is 690, and the lowest count is 648.
This lowest count occurs only for the set of Theorem \ref{th2},
and we denote the model with this set of small constraints by $Small_{\ref{th2}}$.

It seems worth trying to remove small \mrules~from $Small_{\ref{th2}}$
and check whether the resulting model is still {\em Sudoku}. To achieve
this, we have implemented a Prolog program\footnote{See file \texttt{sudoku648.pl}
at the website.} that selects every
small~\mrule~$x\neq y$ in $Small_{\ref{th2}}$, creates a new set
$Rest=Small_{\ref{th2}}\setminus \{x\neq y\}$, and then tries to prove
$Rest$ is not $Sudoku$ by posting all constraints in $Rest$ plus constraint
$x=y$ to a constraint solver and running the solver on a set of Sudoku
puzzles. If a solution is found, then $Rest$ cannot be $Sudoku$, since $x$
cannot be equal to $y$ in it. Note that this is similar to our manual
treatment of the set
of models classified as stuck by Program I, where each model
is proved not to be \emph{Sudoku} by finding a solution to the model that is
not a solution of \emph{Sudoku}. The (simplified) Prolog program is 
provided in Figure~\ref{alg3}.
The set of Sudoku puzzles we have used comes from Gordon Royle's website~\cite{gordonroyle} and consists of more than 50,000 \emph{minimal}
Sudoku puzzles each containing 17 given entries: their minimality was
proven recently in~\cite{no16cluesudoku}. We refer to this set as
$GR$.

\medskip
\begin{figure}
\begin{center}
\begin{minipage}{0.7\textwidth}
\begin{Verbatim}[fontsize=\footnotesize,frame=single,xleftmargin=1em]
try_each_inequality(Model):-
    remove(X#\=Y,Model,Rest),
    (gordonRoyle(Givens), solve([X#=Y|Rest],Givens) ->
        writeln(is_not_Sudoku(Rest))
    ; 
        writeln(maybe_Sudoku(Rest))
    ).
\end{Verbatim}
\end{minipage}
\end{center}
\caption{Program III}\label{alg3}
\end{figure}

Interestingly, the above program determines that every strict subset
$Rest$ of $Small_{\ref{th2}}$ is not {\em Sudoku}: for each $Rest$, there is
indeed a puzzle in $GR$ which has a solution that makes the two variables in the removed inequality equal. This proves that the set
$Small_{\ref{th2}}$ forms a locally minimal set of small \mrules~for
{\em Sudoku}. This was independently verified \cite{mikepersonal} by running a
CNF-encoding of that statement using the BEE-compiler described in
\cite{Metodi:tool:appendix}. Moreover, using
the same technology, we were jointly able to prove that each
of the 39 models $M$ of $\mathit{Missing}(6)$ that are {\em Sudoku} (see
Appendix I) has the following property:

\begin{itemize}
\item[]
$M$ has a subset of inequalities of size 648 that is {\em Sudoku} and
is also a locally minimal set of small \mrules
\end{itemize}

We were not able to reduce those $M$'s any further, i.e., beyond
648. Although these results do not allow us to conclude that 
Sudoku models with a smaller set of small \mrules~ are not possible, we dare to conjecture
the following:

\paragraph{{\bf Conjecture:}}
No model with less than 648 small \mrules~is {\em Sudoku}.

\section{Discussion and Conclusion}\label{concl}

The message in rec.puzzles mentioned in the introduction also refers
essentially to our Corollary~\ref{cor1}, i.e., that in every chute, one row (or column)
constraint needs no checking, if the other constraints in that chute
are validated.\footnote{At the time of that post, we had already
  completed our classification of the big \mrules.} Clearly, other
people have wondered about redundant big \mrules~in Sudoku, and our
main result -- many sets of six big \mrules~are
redundant -- often surprises people. It is all the more interesting that the popular
\cite{Ist06sudokuas} refers to the ``minimal encoding'' as one
containing \emph{all} big rules: our results clearly indicate that
such encoding is not minimal at all. Further, while redundant rules
can strengthen propagation and, thus, reduce the search space, it has
already been noted ~\cite{Kwonoptimizedcnf} that the classical
Conjunctive Normal Form encodings for Sudoku in SAT generate too
many redundant clauses, and compact encodings (which eliminate
redundant clauses) are more efficient. Our work can be used to inform
such encodings. 

Our conjecture that no model with less than 648 small \mrules~is {\em
  Sudoku} remains to be proven. While the combinatorial challenge is great,
we are currently investigating the use
of {\em unavoidable sets} as in~\cite{no16cluesudoku}. We have also
obtained a full classification of models that use small constraints for the
more restricted problem of Latin Squares~\cite{latsq}.

Apart from our novel results themselves, and the use of exploratory
(Constraint) Logic Programming, this paper also introduces a powerful
graphical representation of sets of constraints that
renders the proofs easy to understand, and that can be re-used for
larger Sudoku puzzles.

Exploratory programming was essential in this research: it helped us
discover potential theorems and lemmas which we subsequently turned
into hard general proofs.
Further, the use of Prolog has been critical: as it can be
seen from
the website, the
programs are small, fast, easy to read and modify. This would have
been very difficult without the combined power of backtracking (for
almost everything, particularly finding all solutions satisfying a set
of conditions), constraint solving (to easily define Sudoku and test
the satisfiability of many of its subsets) and logic variables (to
easily identify and access the variables in the model).

Redundant constraints are very often good for the performance of CP
systems, and indeed, all solvers we checked perform much slower (about
a factor 2000) with a minimal set of big constraints. So it
might seem counterproductive to try to find redundant constraints if
the aim is to remove them.
However, our work gives some insight into the {\em
  construction} of new (redundant) inequality constraints: while
deriving new equalities from a set of equalities is easy because
equality is transitive, this does not hold for inequalities. The
difficulty and possibility of deriving new inequalities depends
crucially on the domains of the variables. For instance, from a chain
of inequalities $x_1 \neq x_2 \neq \dots \neq x_n$ between {\em boolean}
variables, one may conclude that $x_1 \neq x_4$ (amongst others), but
if the domains have a larger cardinality, this is no longer
true. Since our work provides a complete set of rewrite rules on sets
of all\_different constraints (together with the domain constraints)
for a particular CSP, it forms a first step in the development of a
more general inequality inference framework.

Finally, note that our result on big \mrules~completes in some sense
the result in~\cite{no16cluesudoku}: 17 clues is necessary, and so are
21 big \mrules. It would be interesting to have the corresponding result for
small \mrules.

\paragraph{Acknowledgements}
The main results reported here were obtained while the first author
was on a research visit at Monash University in April 2008, and
enjoying the Stuckey hospitality in Apollo Bay and Elwood,
Australia. Many thanks for a most enjoyable stay. 
We are grateful to Michael Codish for his help with obtaining some of
the results related to the conjecture.
This research was partly sponsored by he Australian Research Council grant DP110102258, by the Brussels-Capital
Region through project ParAps, and by the Research Foundation Flanders
(FWO) through projects {\em WOG: Declarative Methods in Computer
Science} and {\em G.0221.07}.

\bibliographystyle{acmtrans}
\bibliography{sudokutplp}

\newpage

\section*{Appendix I: All {\em Sudoku} Models in $\mathit{Missing}(6)$ up to Symmetry} \label{appI}

\noindent
\includegraphics[width=0.12\linewidth,keepaspectratio]{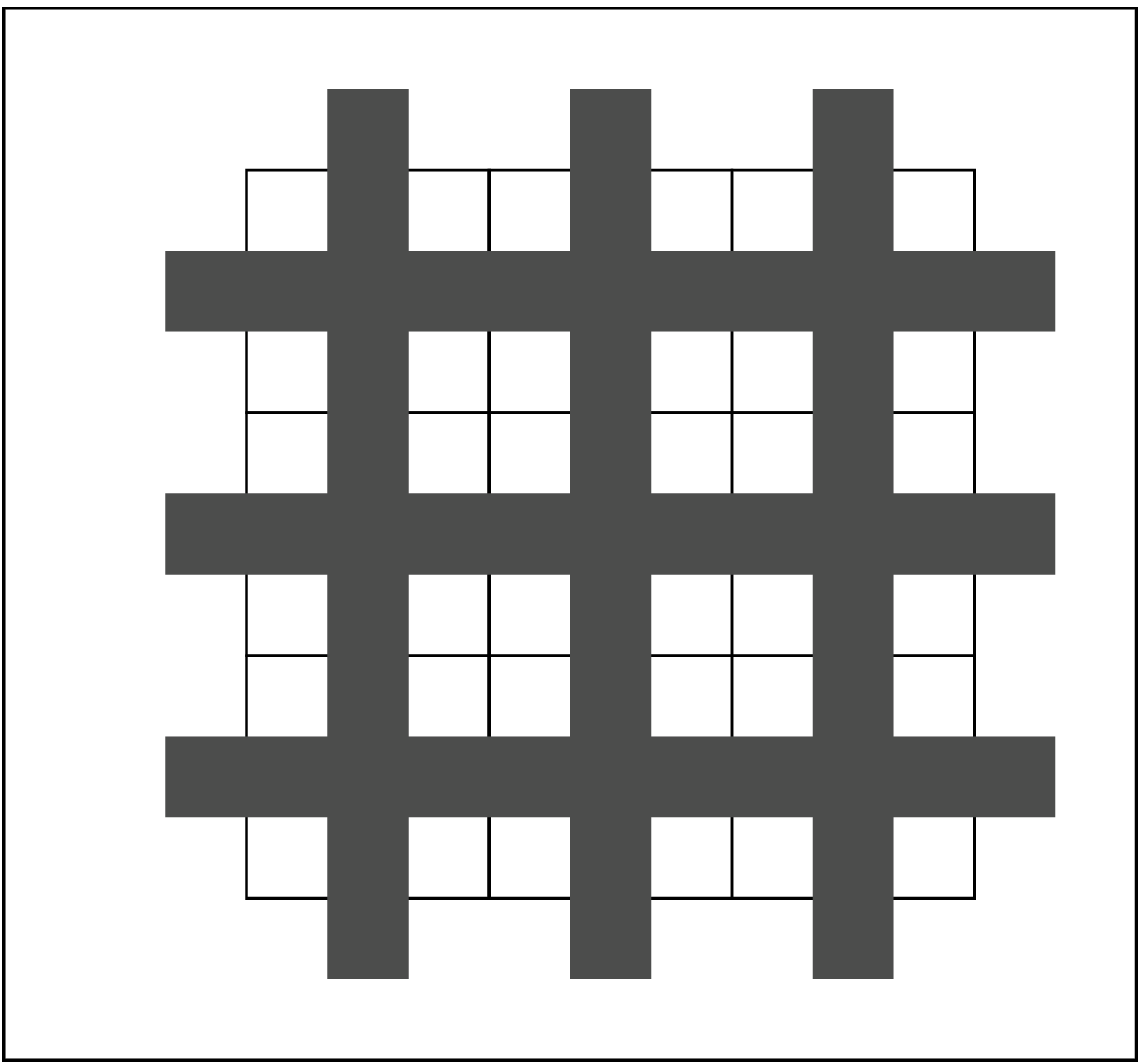}
\includegraphics[width=0.12\linewidth,keepaspectratio]{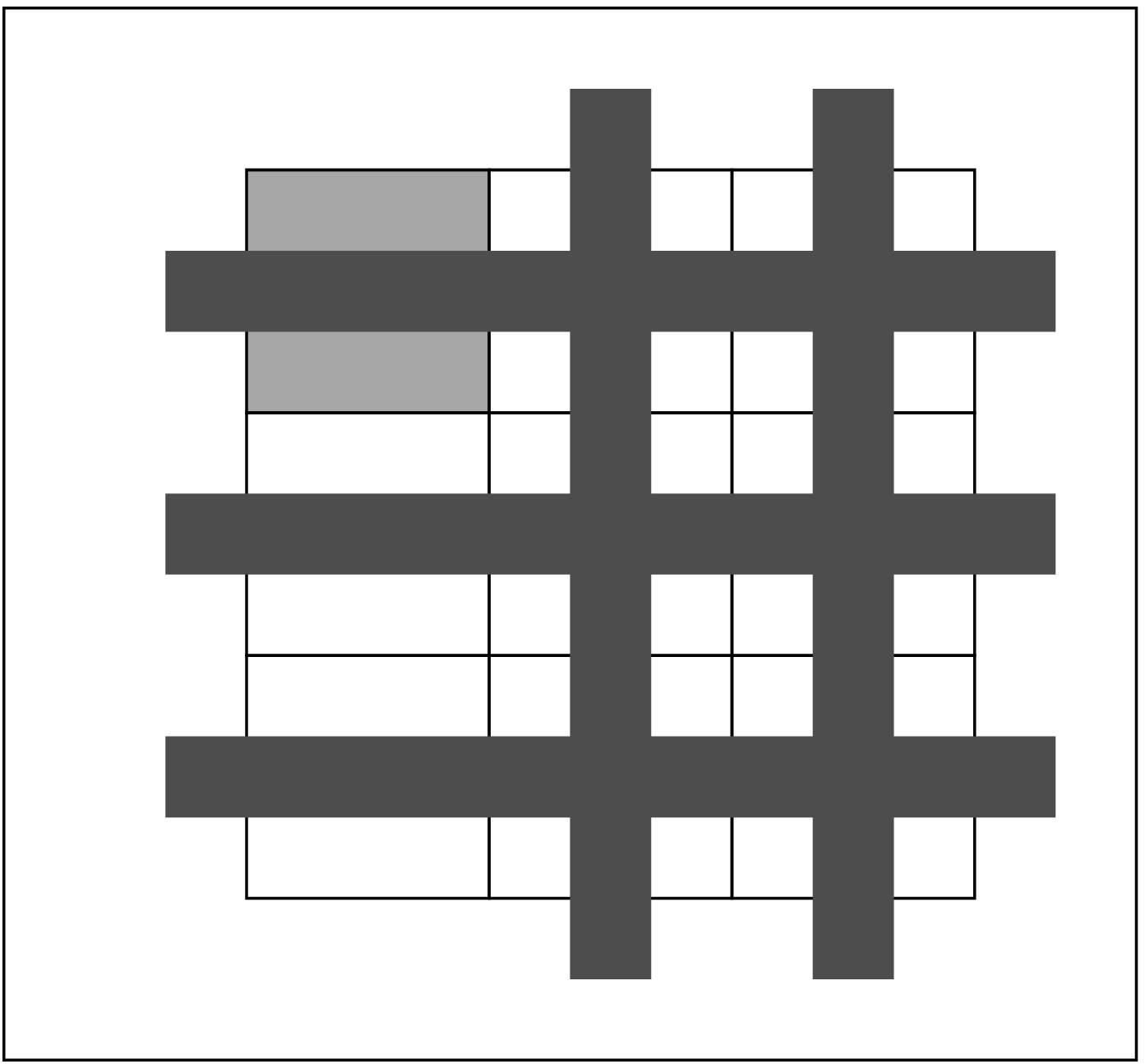}
\includegraphics[width=0.12\linewidth,keepaspectratio]{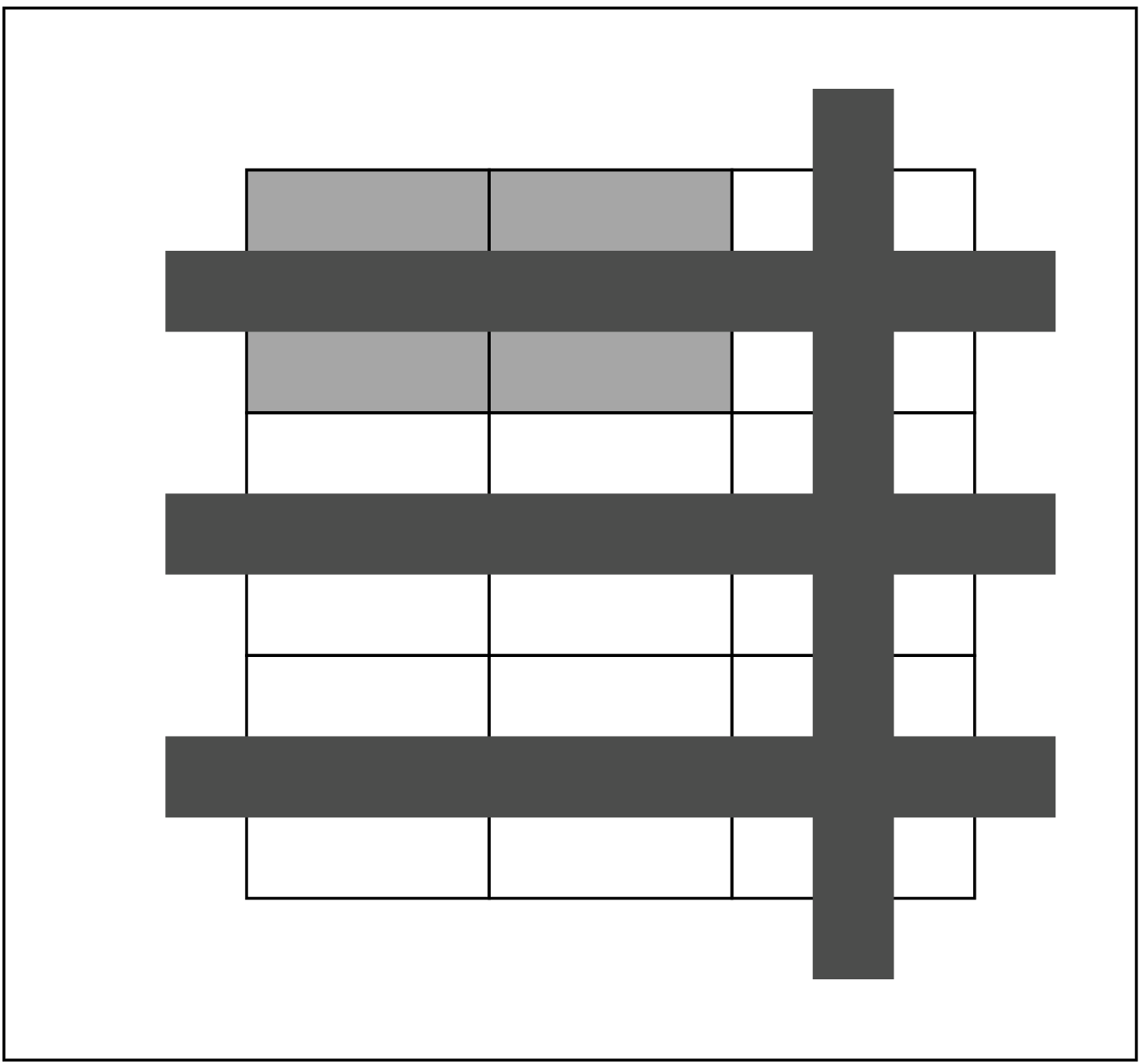}
\includegraphics[width=0.12\linewidth,keepaspectratio]{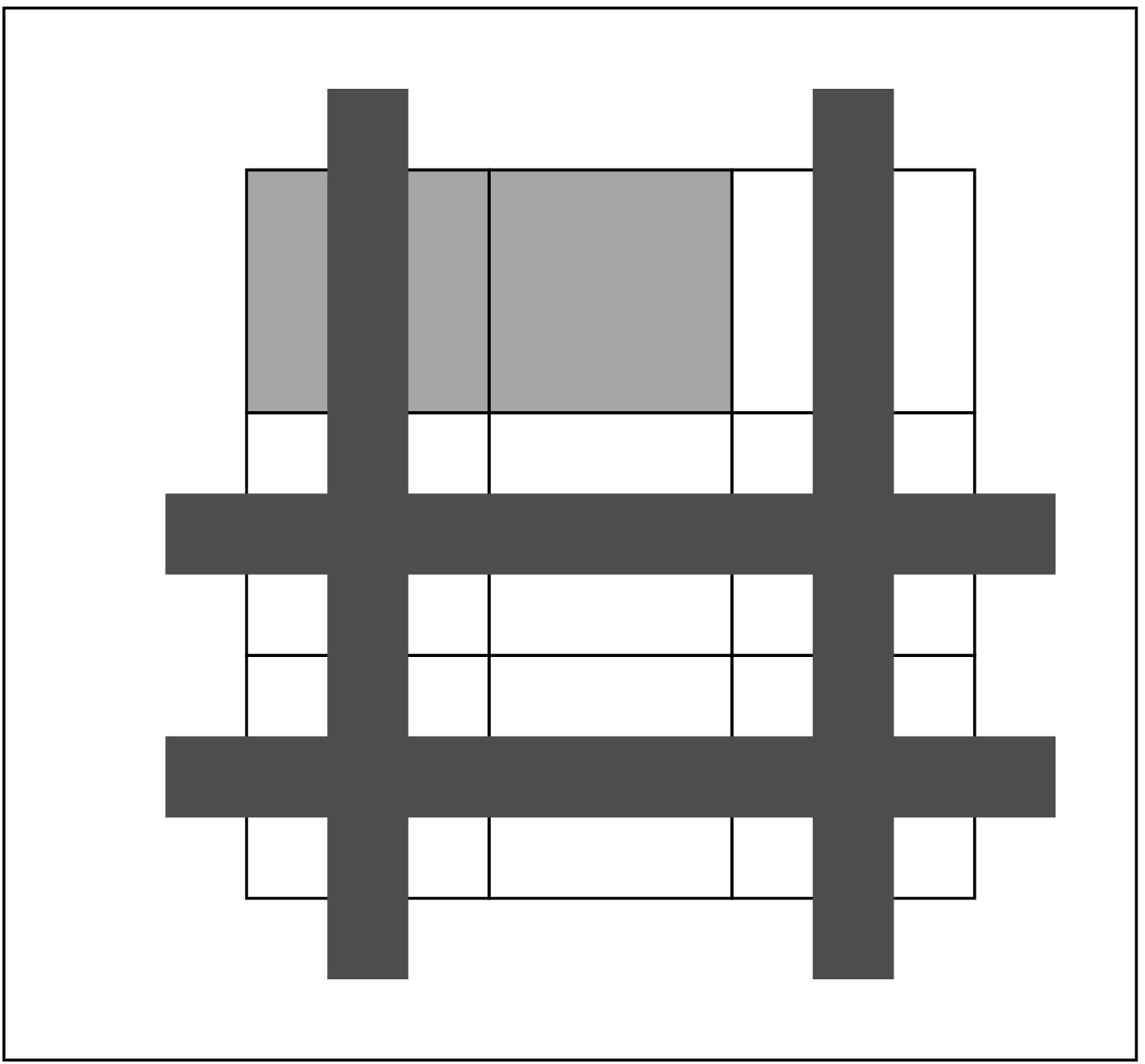}
\includegraphics[width=0.12\linewidth,keepaspectratio]{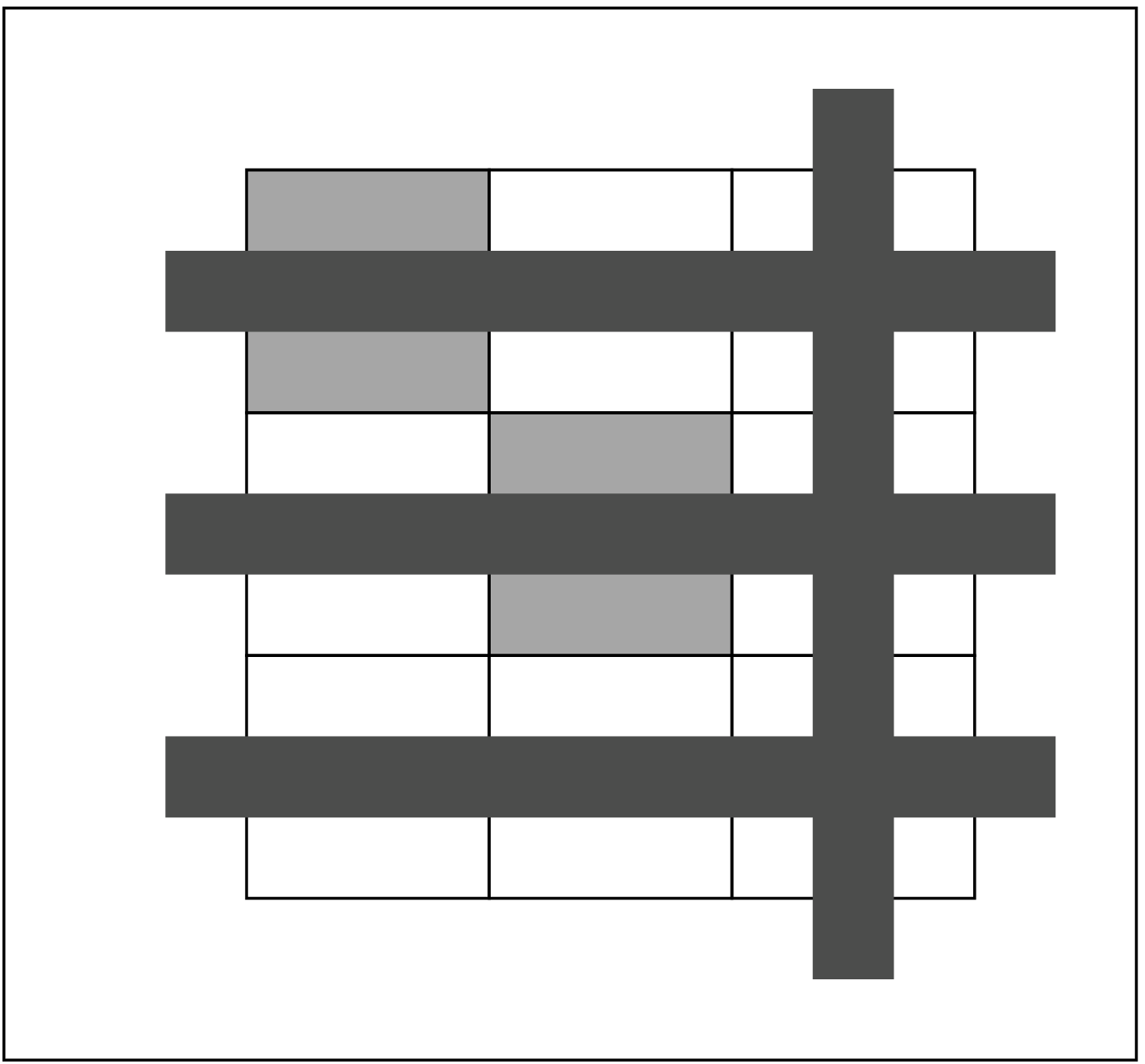}
\includegraphics[width=0.12\linewidth,keepaspectratio]{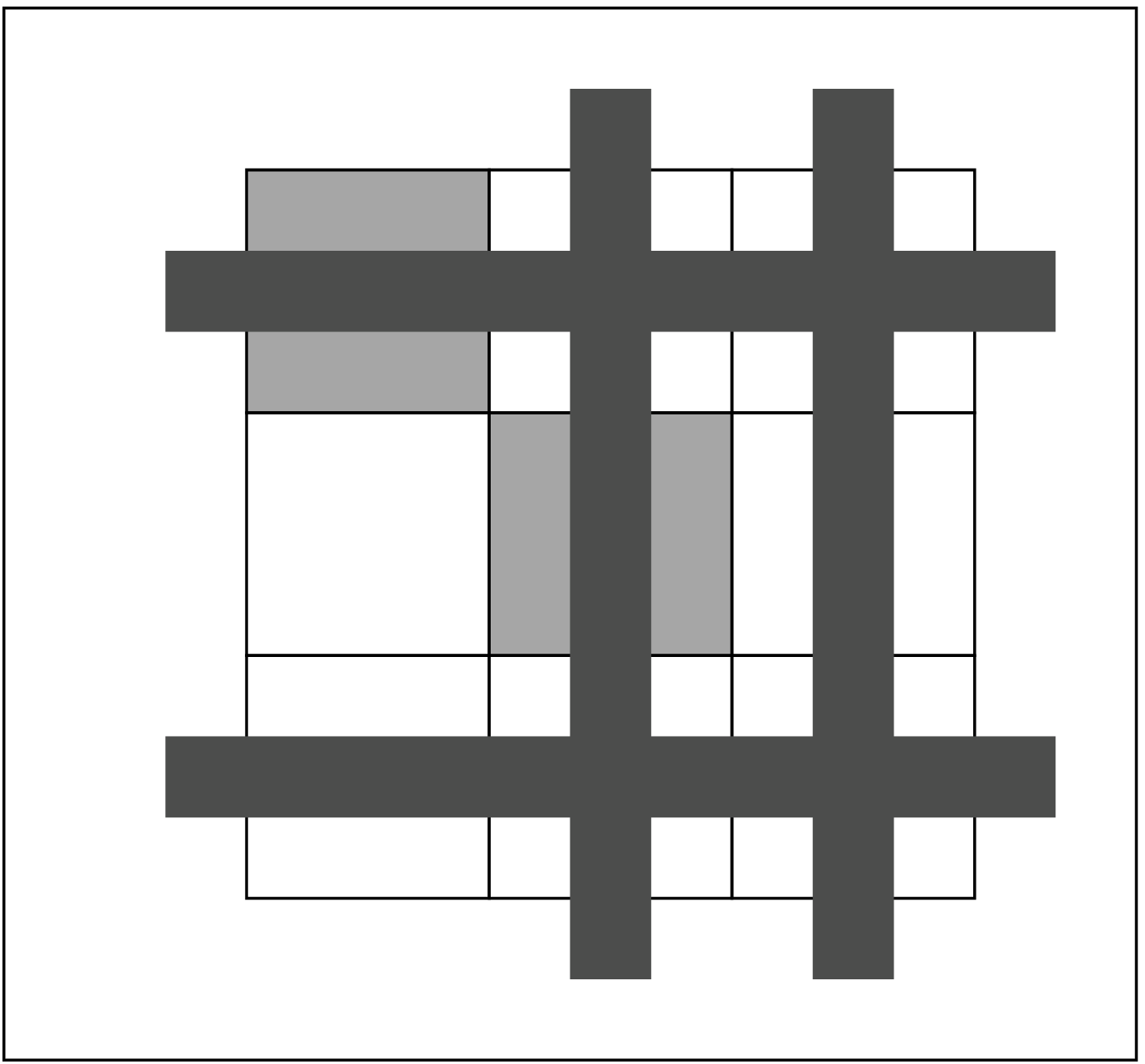}
\includegraphics[width=0.12\linewidth,keepaspectratio]{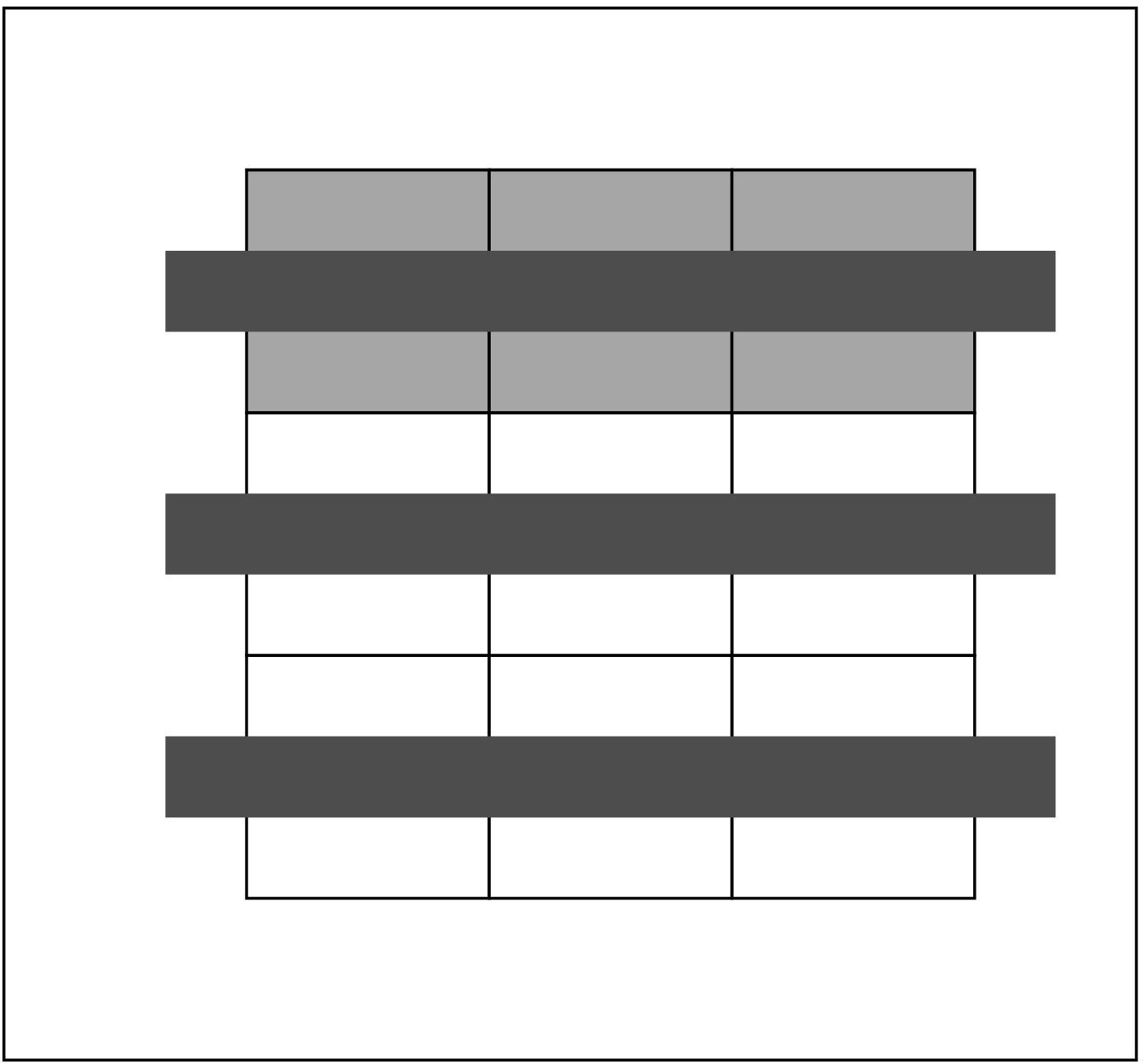}
\includegraphics[width=0.12\linewidth,keepaspectratio]{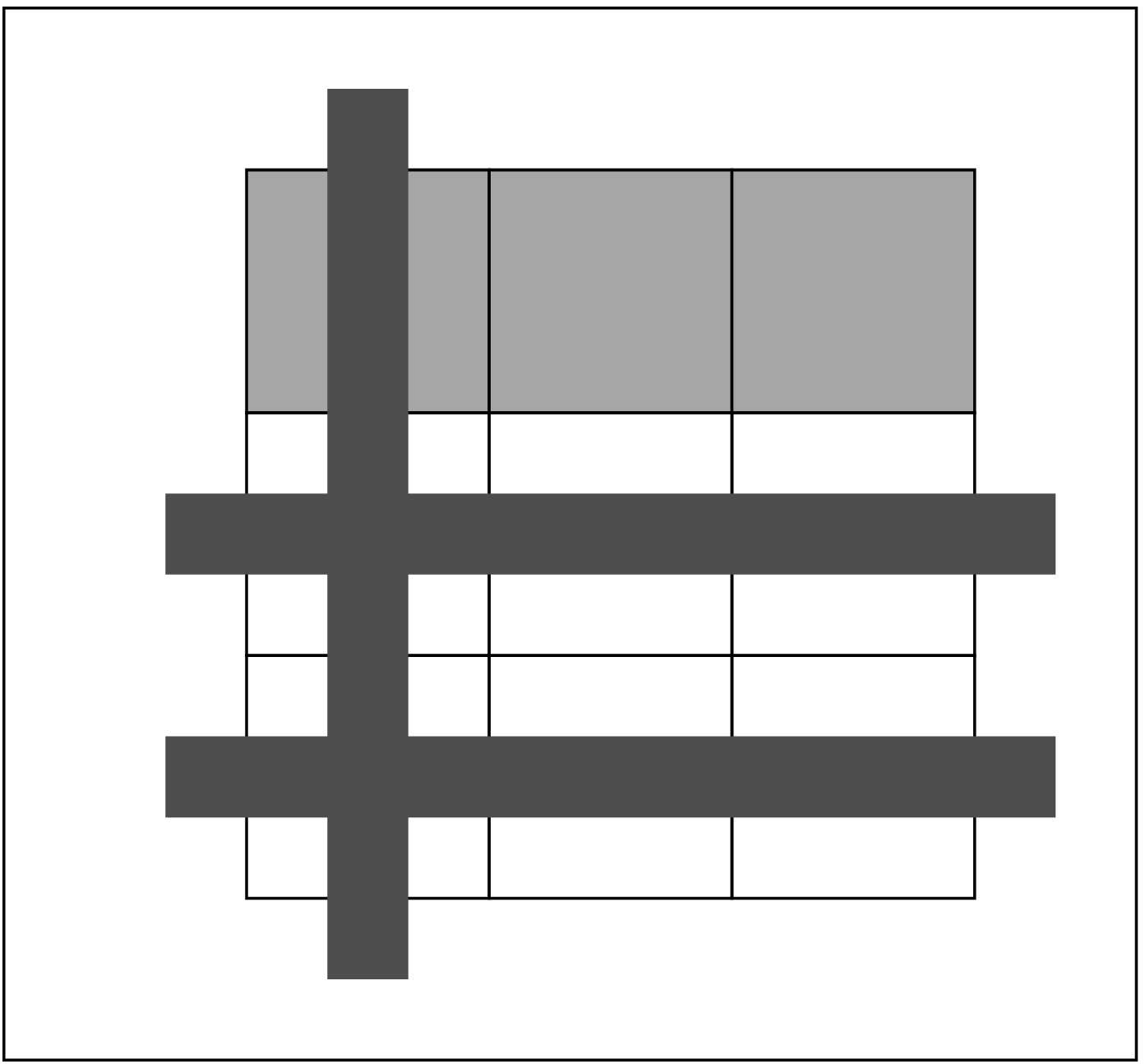}
\includegraphics[width=0.12\linewidth,keepaspectratio]{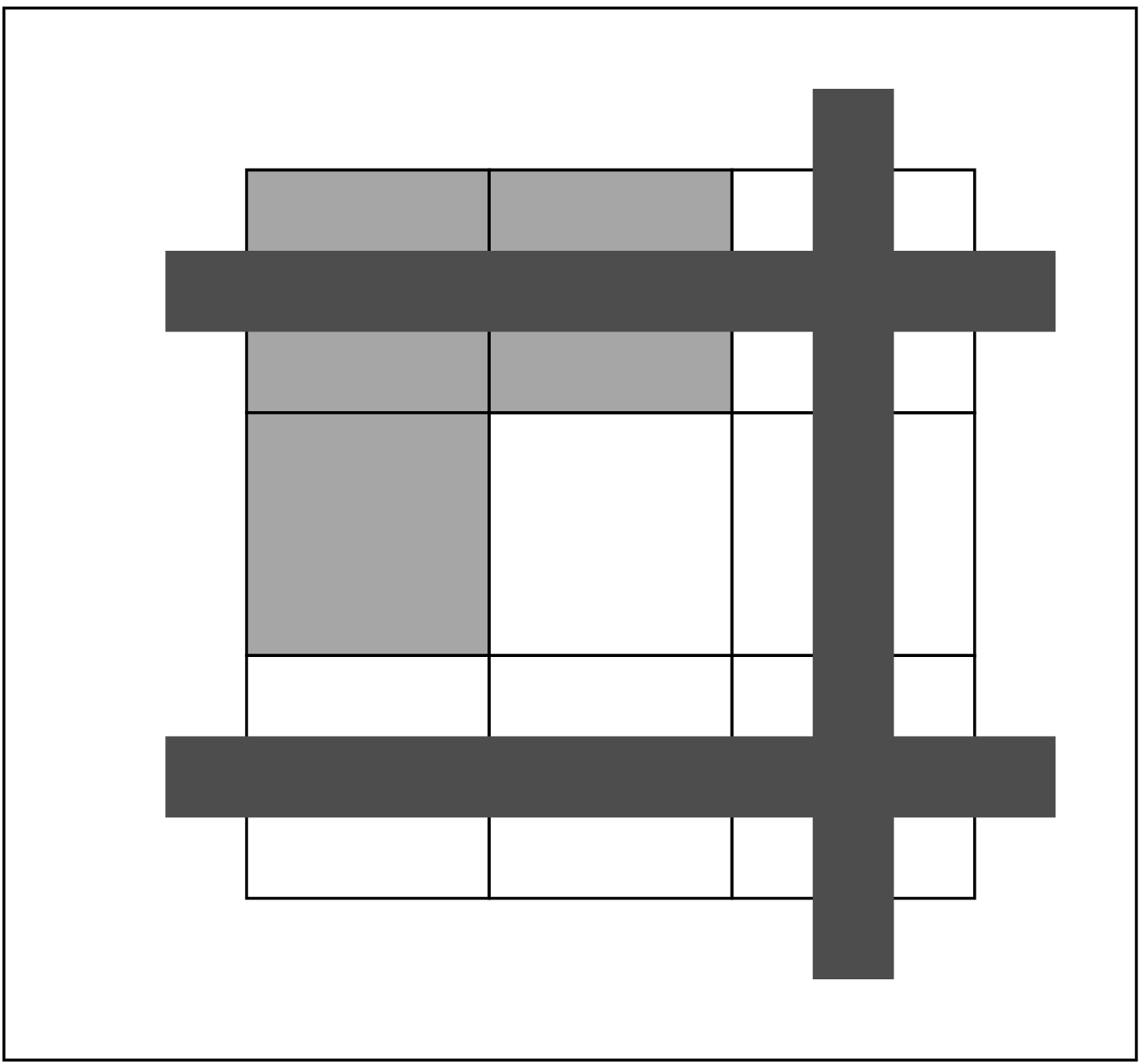}
\includegraphics[width=0.12\linewidth,keepaspectratio]{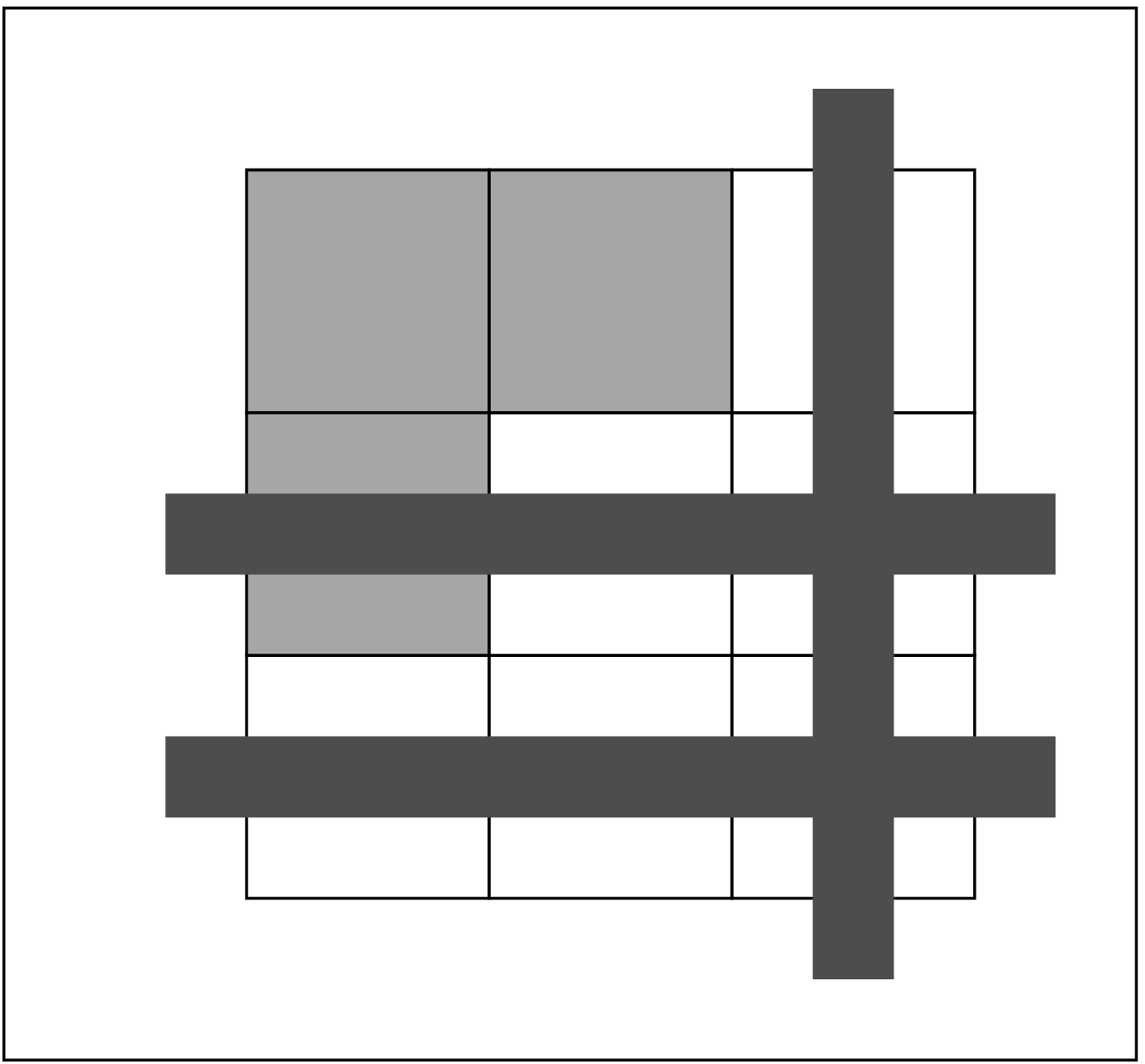}
\includegraphics[width=0.12\linewidth,keepaspectratio]{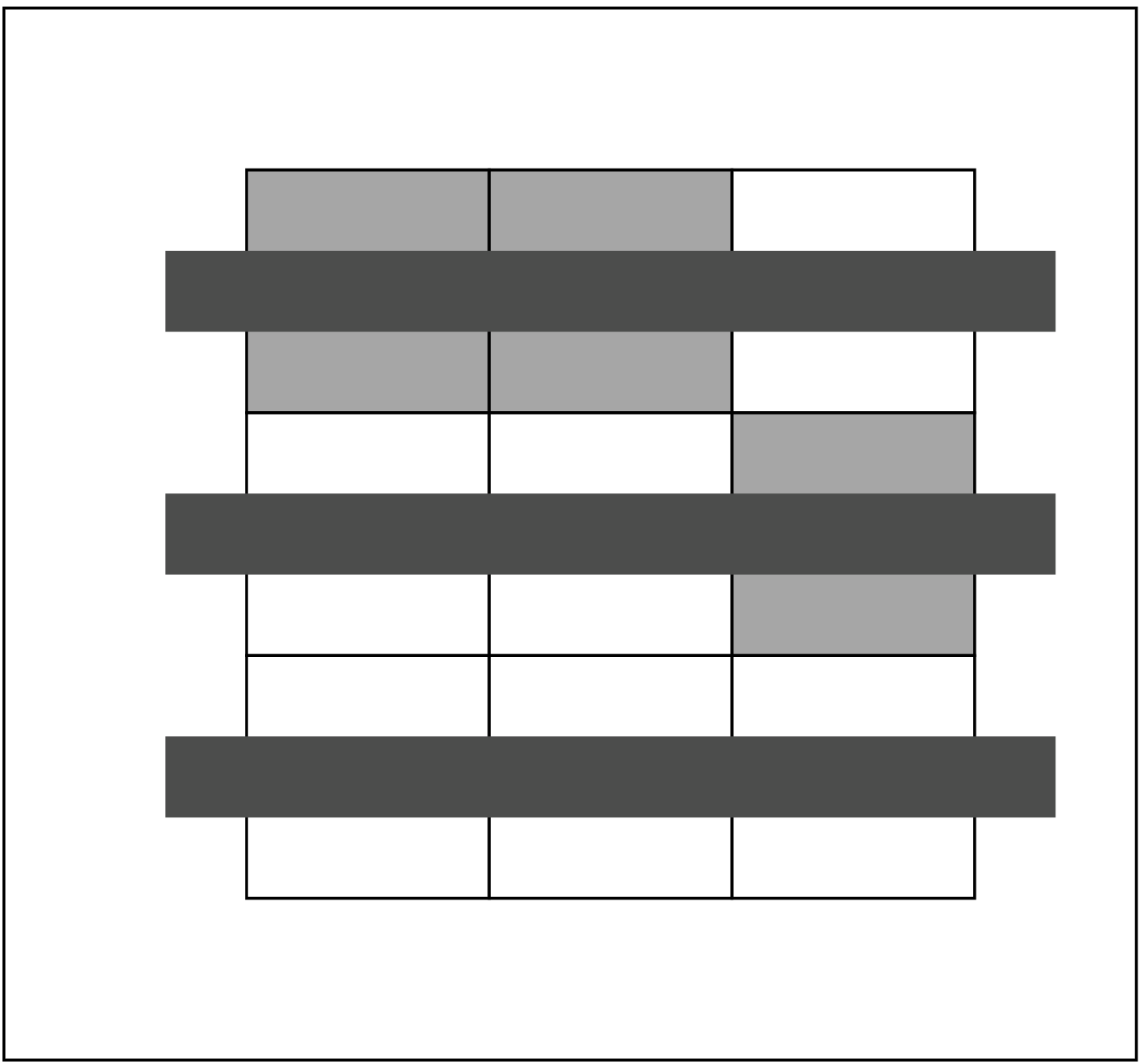}
\includegraphics[width=0.12\linewidth,keepaspectratio]{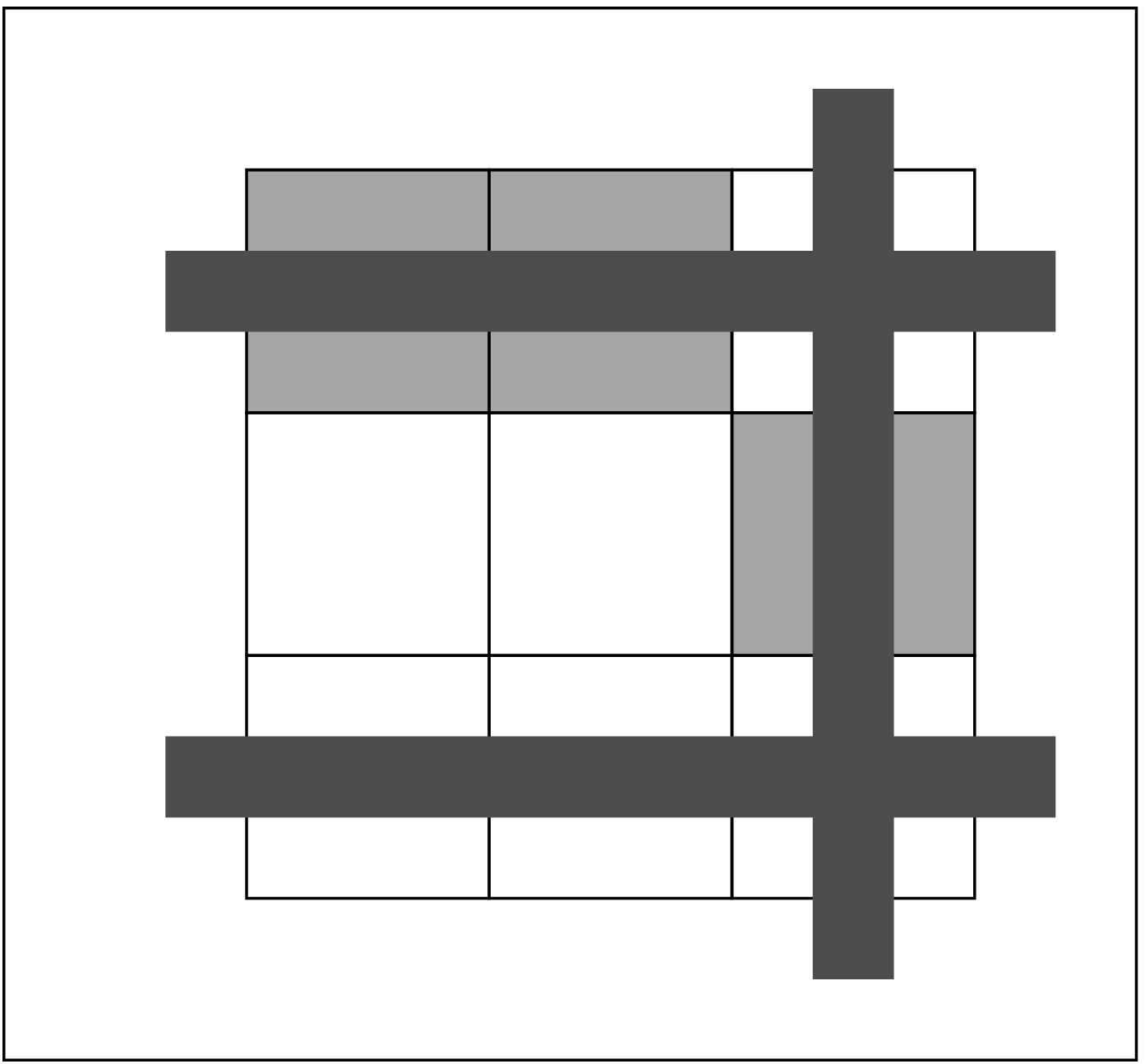}
\includegraphics[width=0.12\linewidth,keepaspectratio]{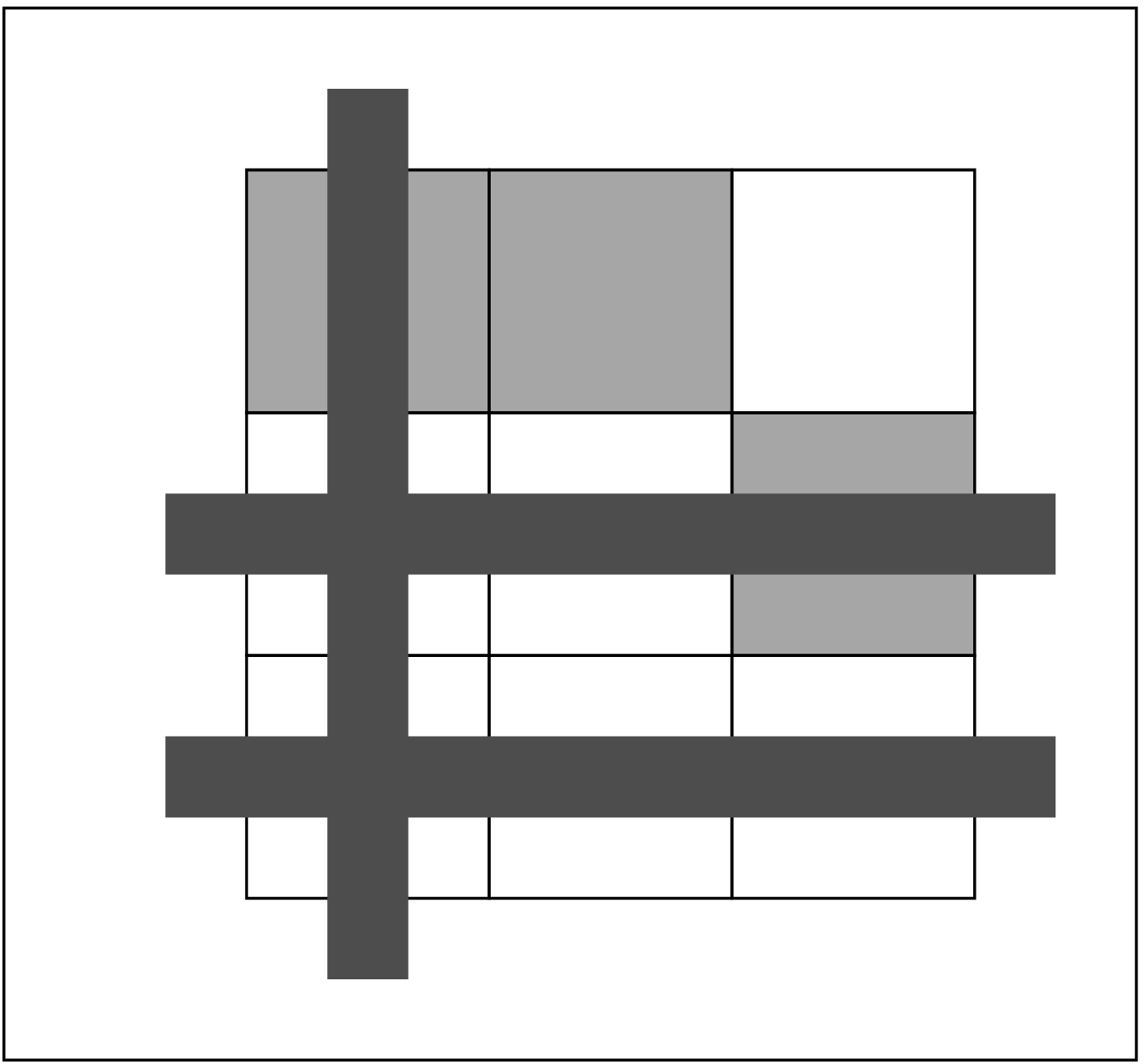}
\includegraphics[width=0.12\linewidth,keepaspectratio]{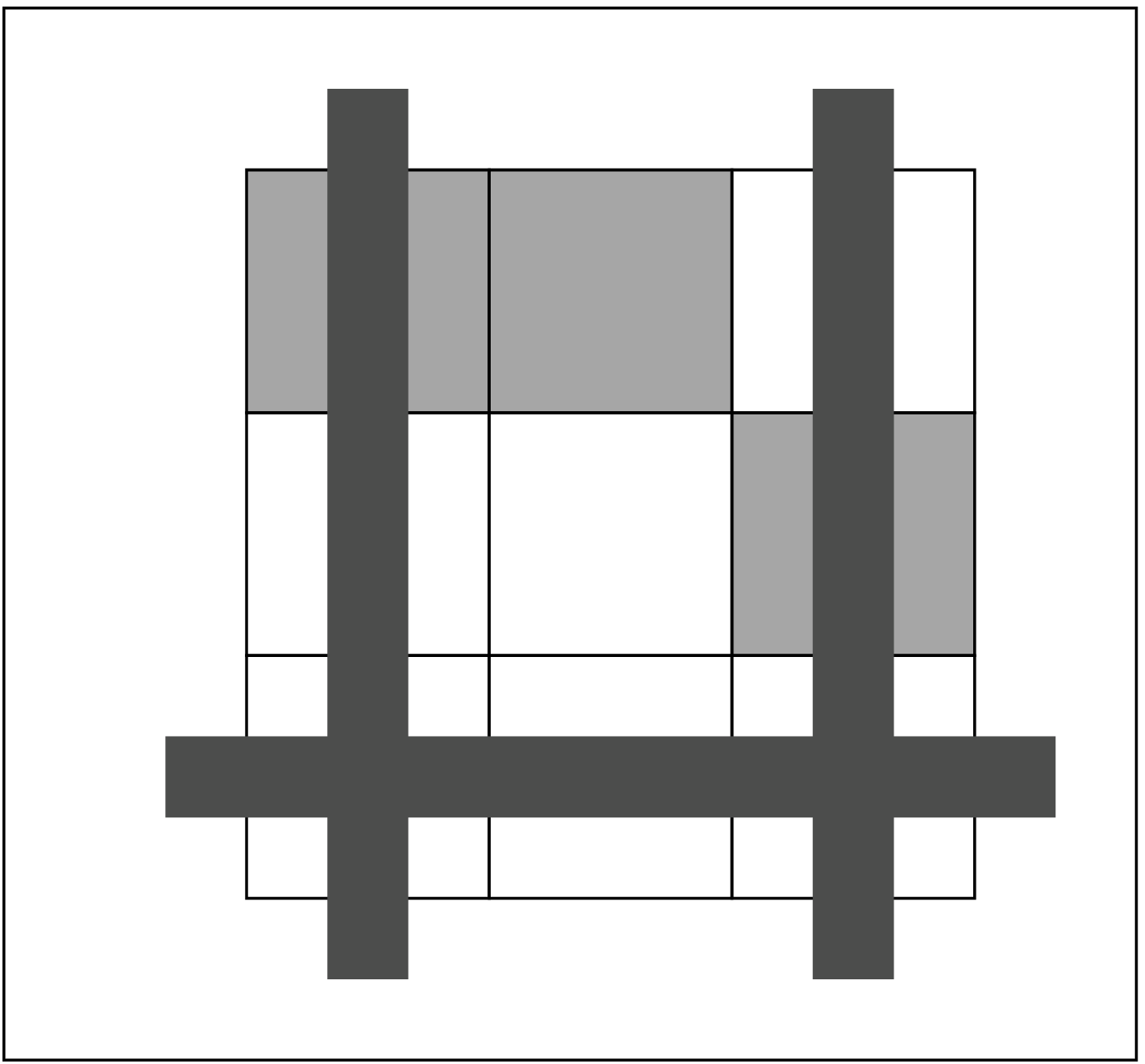}
\includegraphics[width=0.12\linewidth,keepaspectratio]{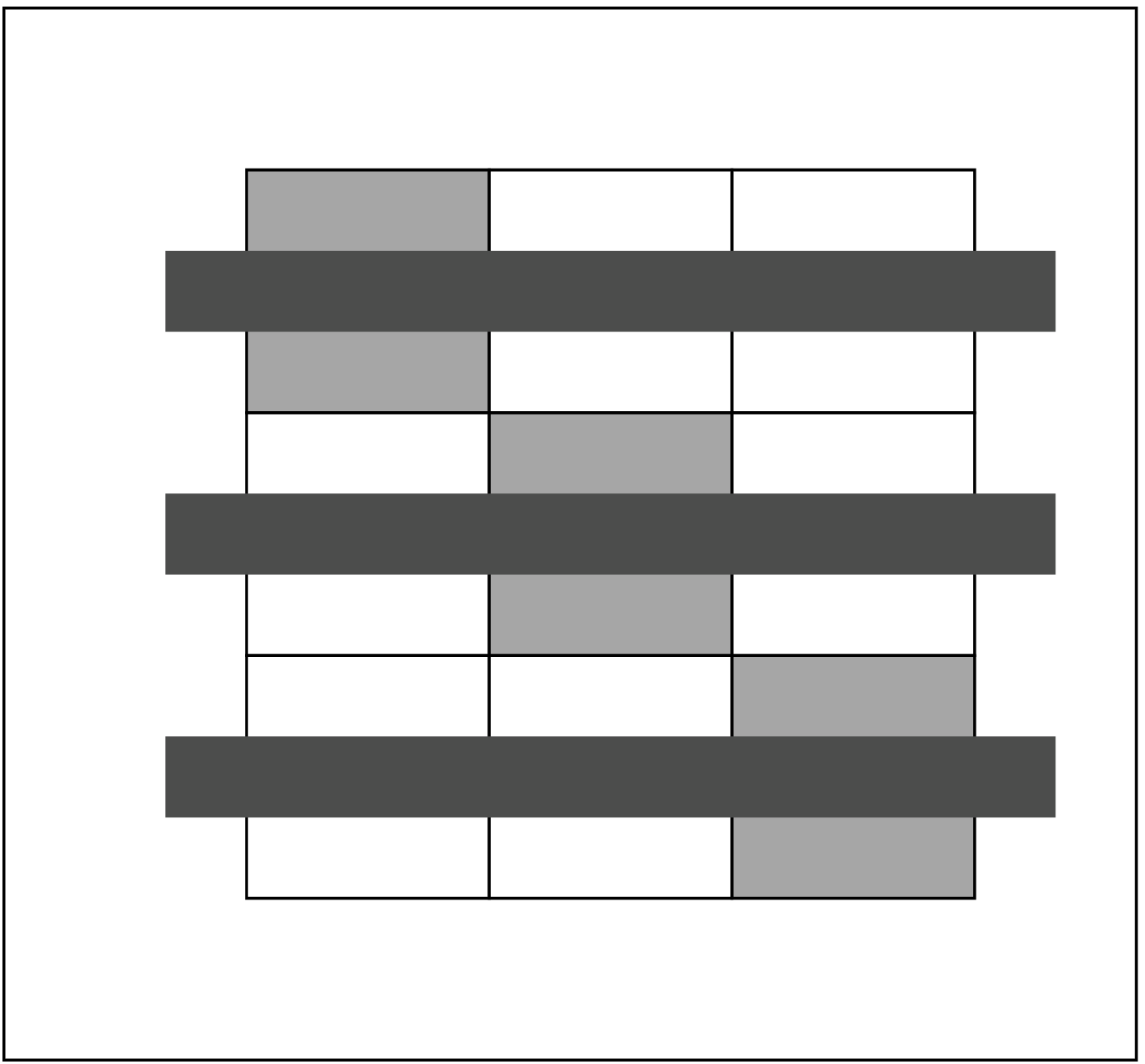}
\includegraphics[width=0.12\linewidth,keepaspectratio]{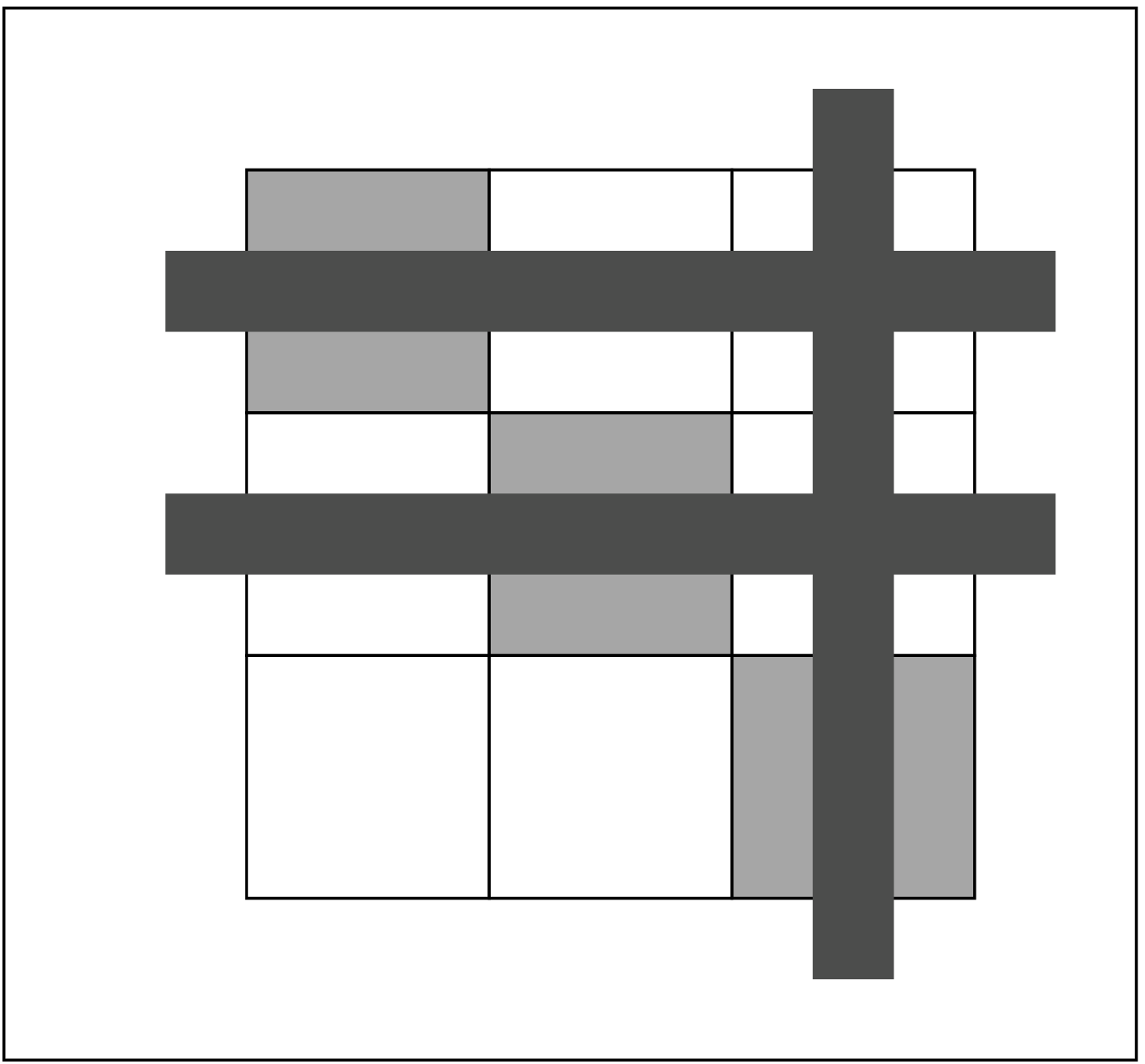}
\includegraphics[width=0.12\linewidth,keepaspectratio]{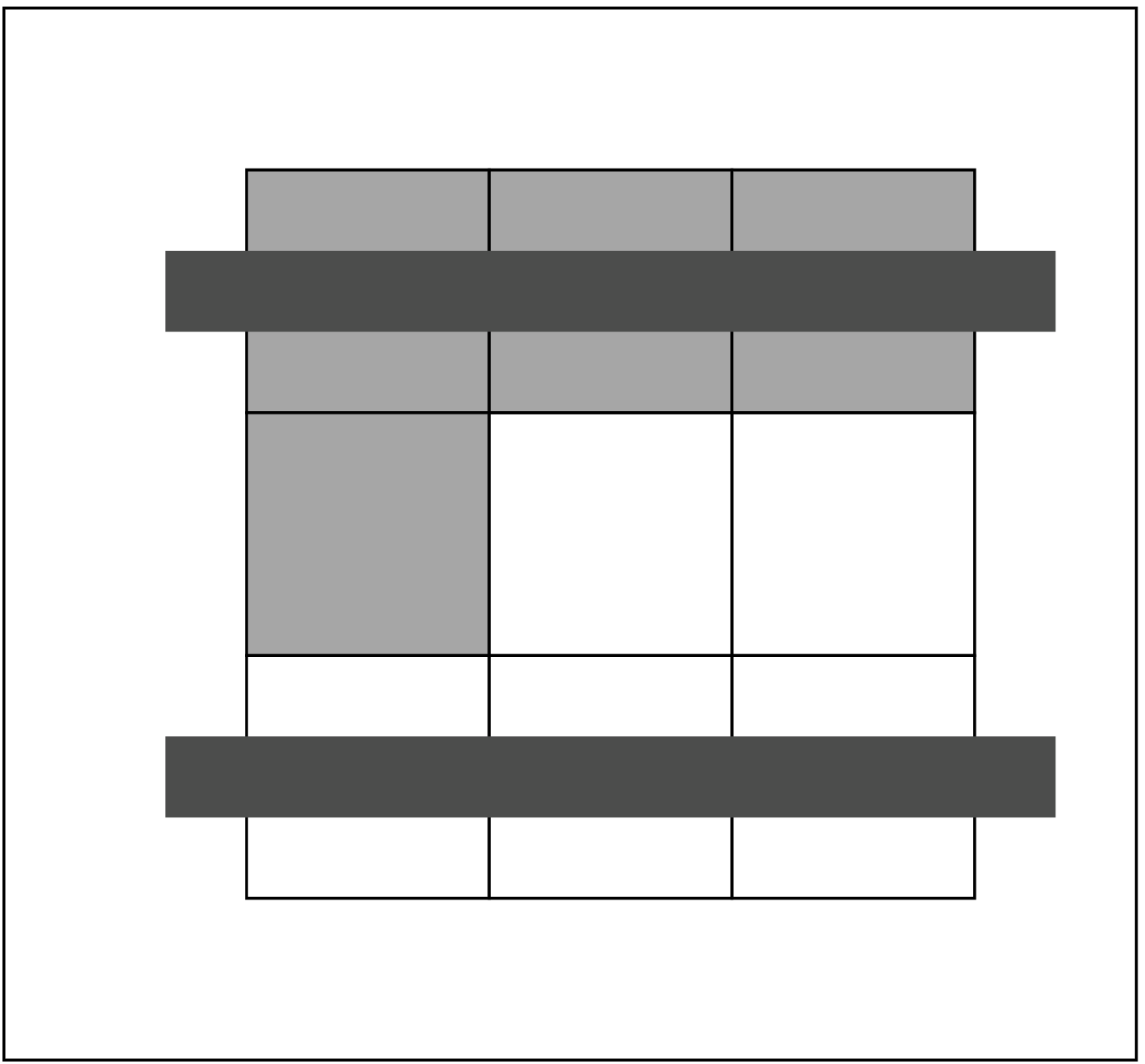}
\includegraphics[width=0.12\linewidth,keepaspectratio]{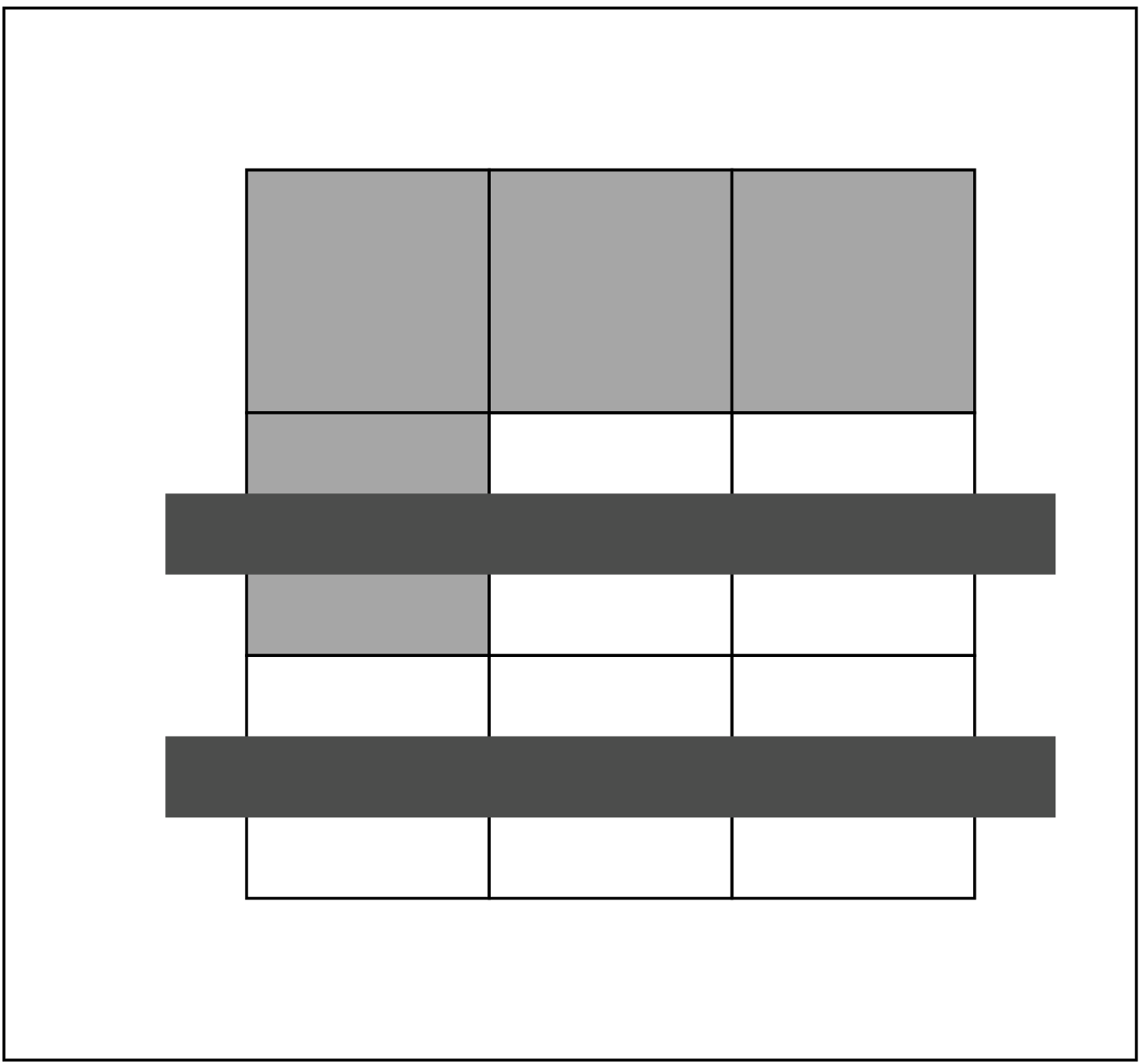}
\includegraphics[width=0.12\linewidth,keepaspectratio]{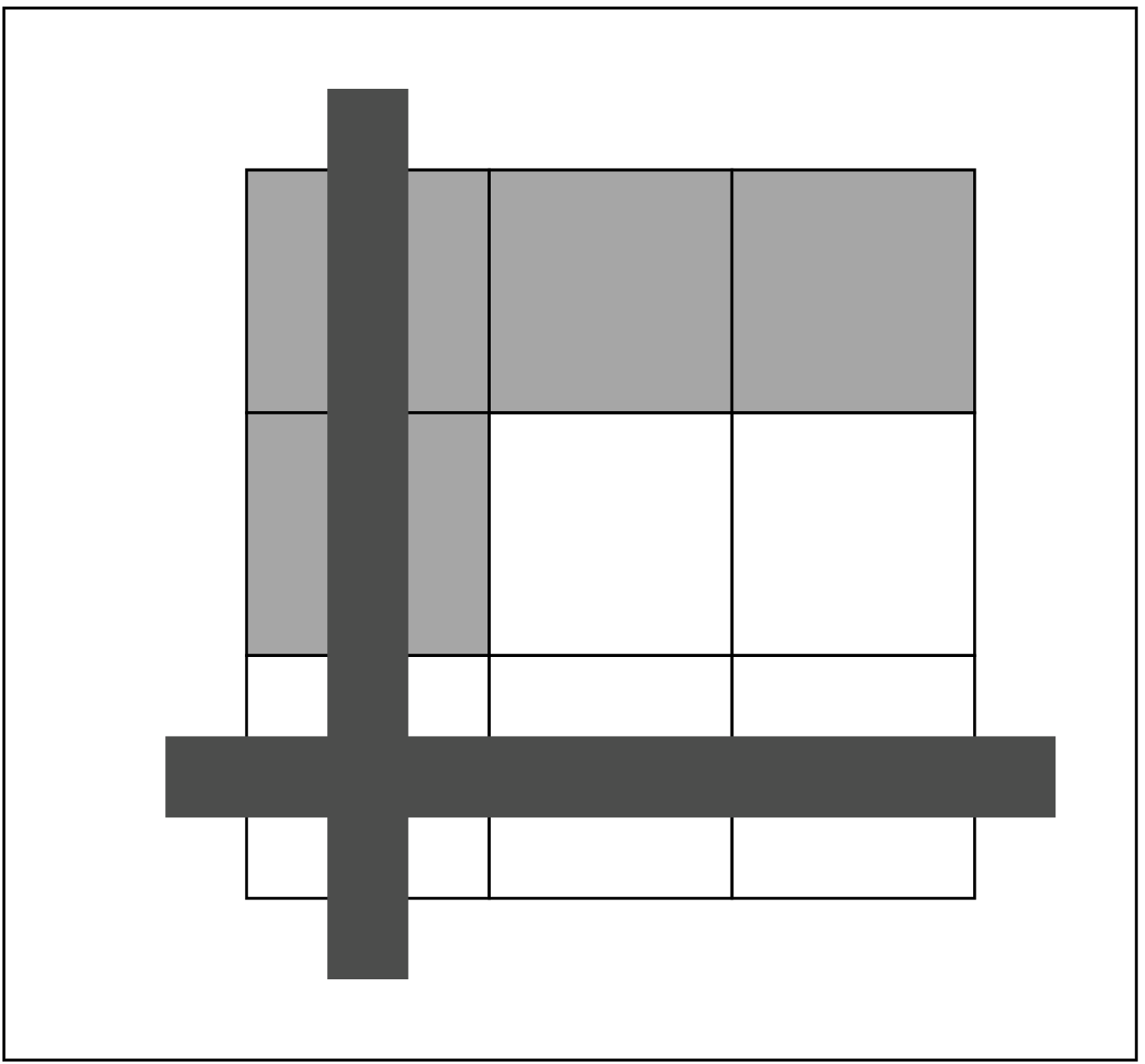}
\includegraphics[width=0.12\linewidth,keepaspectratio]{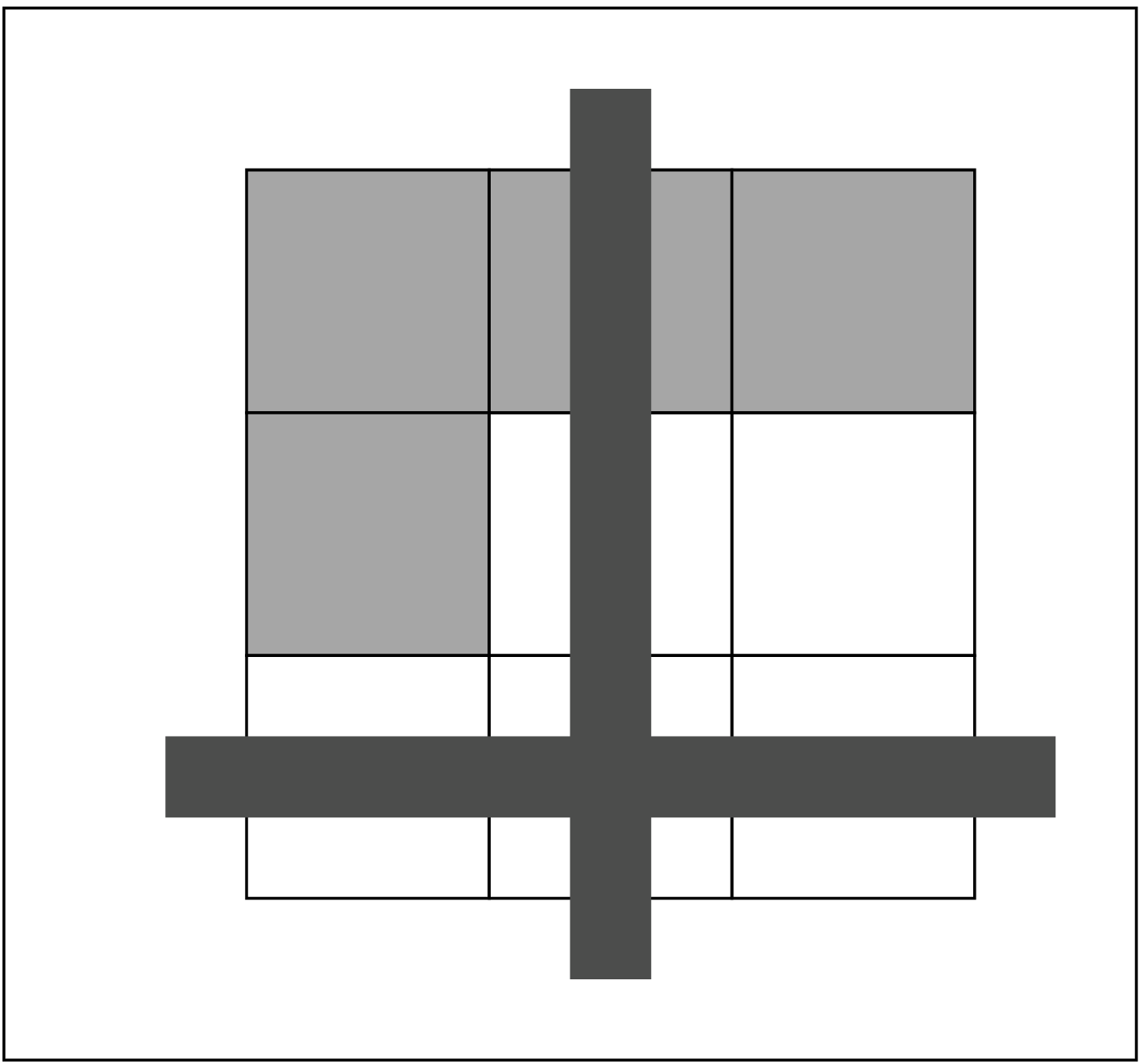}
\includegraphics[width=0.12\linewidth,keepaspectratio]{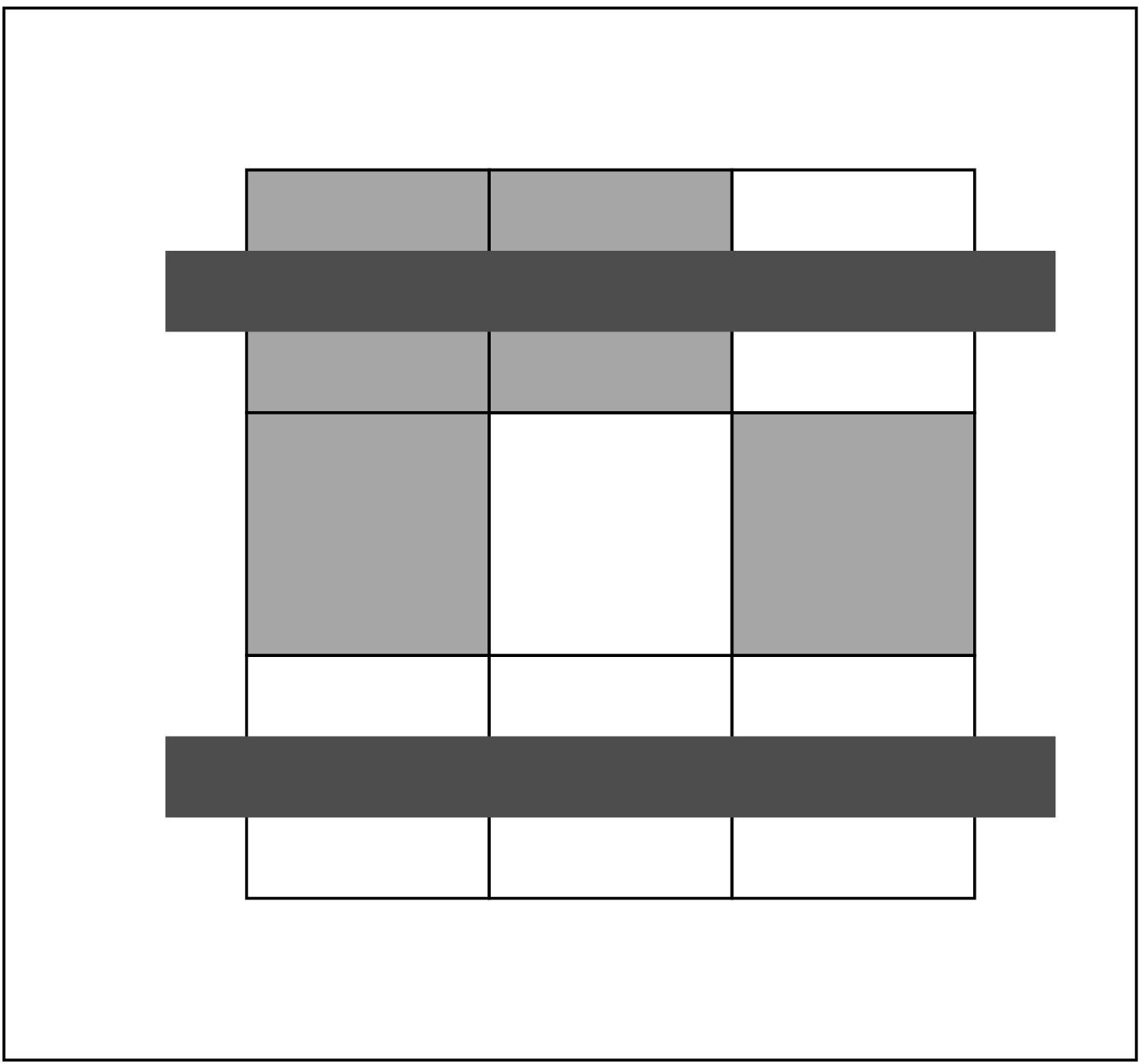}
\includegraphics[width=0.12\linewidth,keepaspectratio]{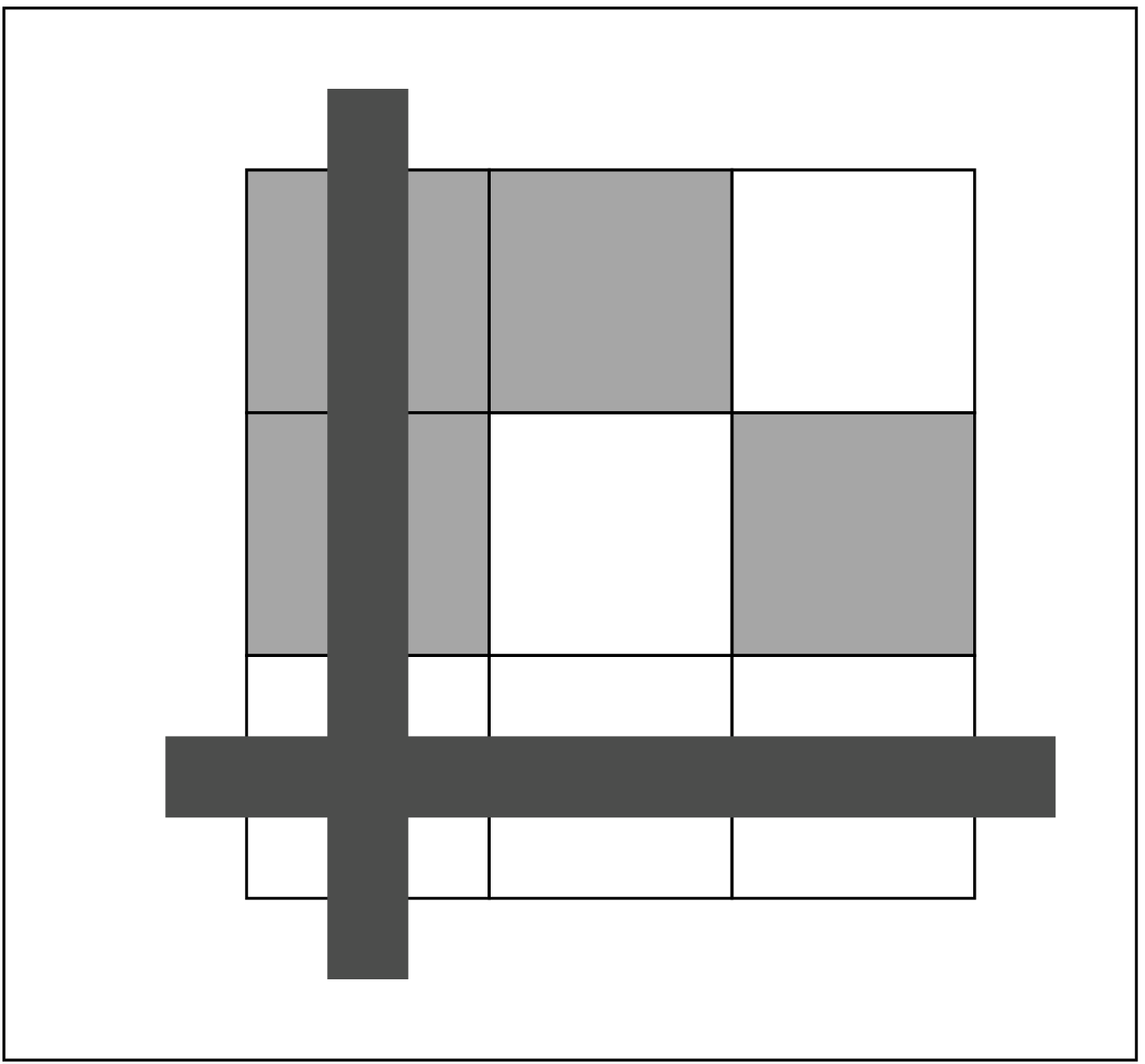}
\includegraphics[width=0.12\linewidth,keepaspectratio]{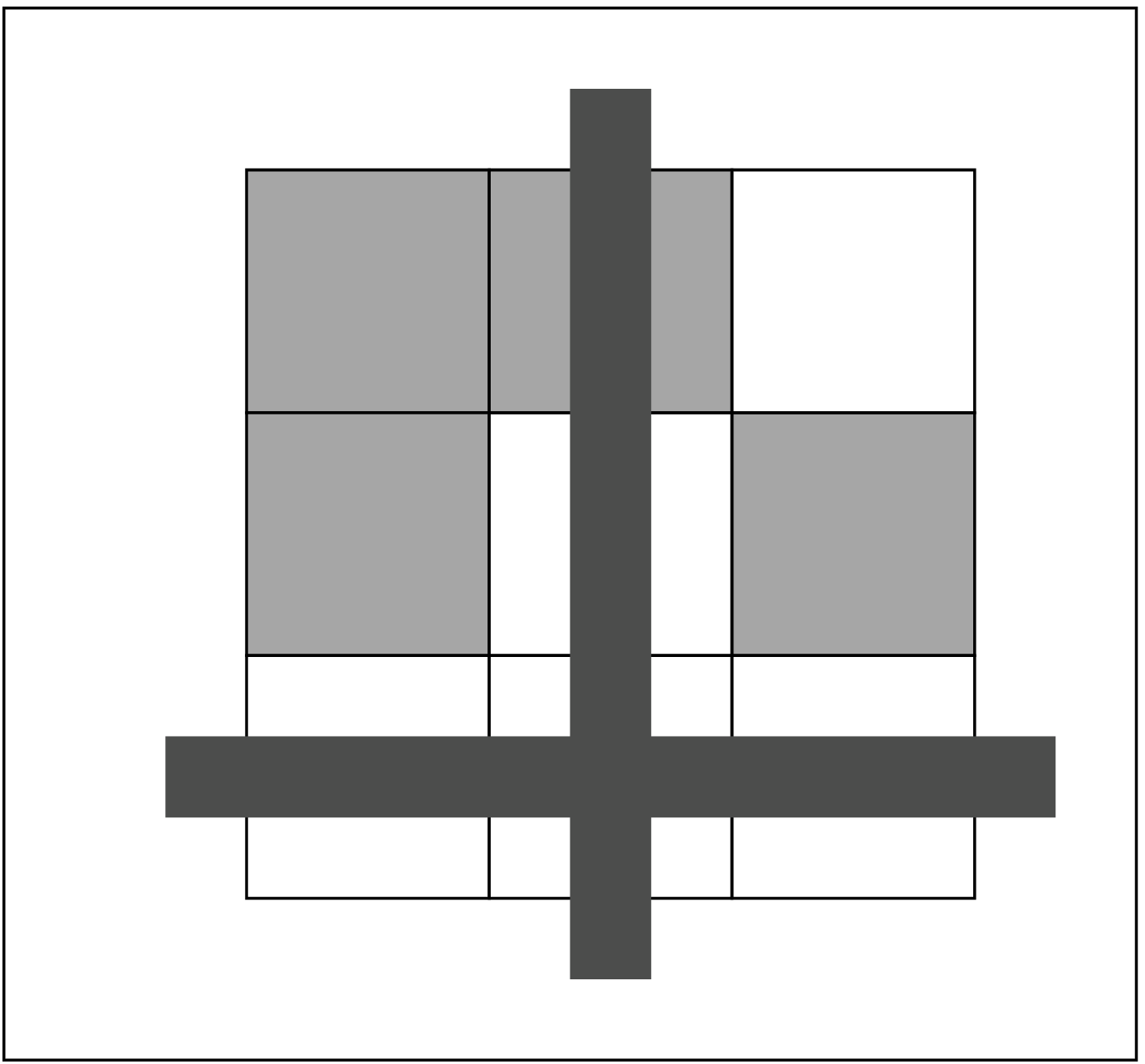}
\includegraphics[width=0.12\linewidth,keepaspectratio]{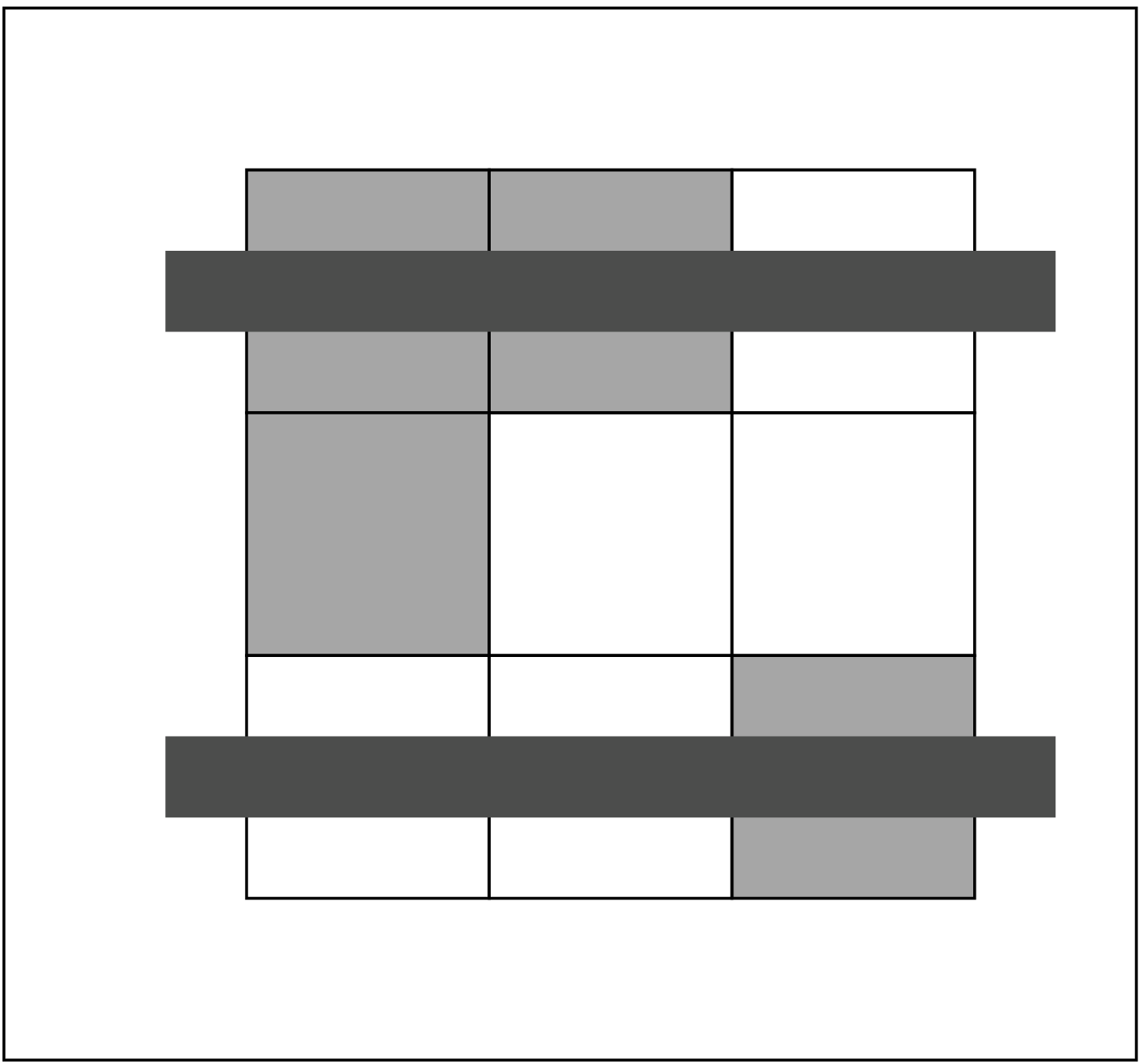}
\includegraphics[width=0.12\linewidth,keepaspectratio]{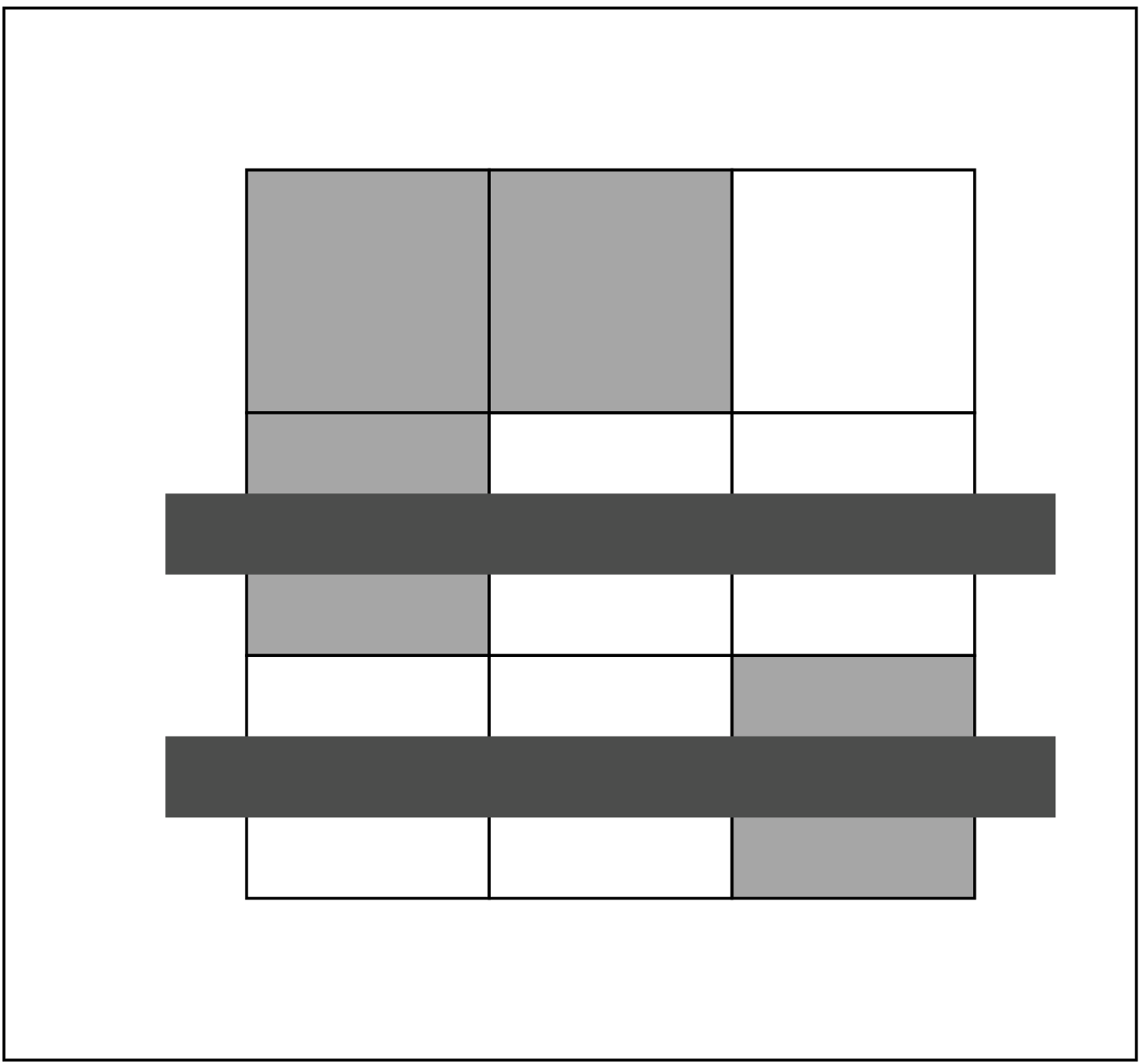}
\includegraphics[width=0.12\linewidth,keepaspectratio]{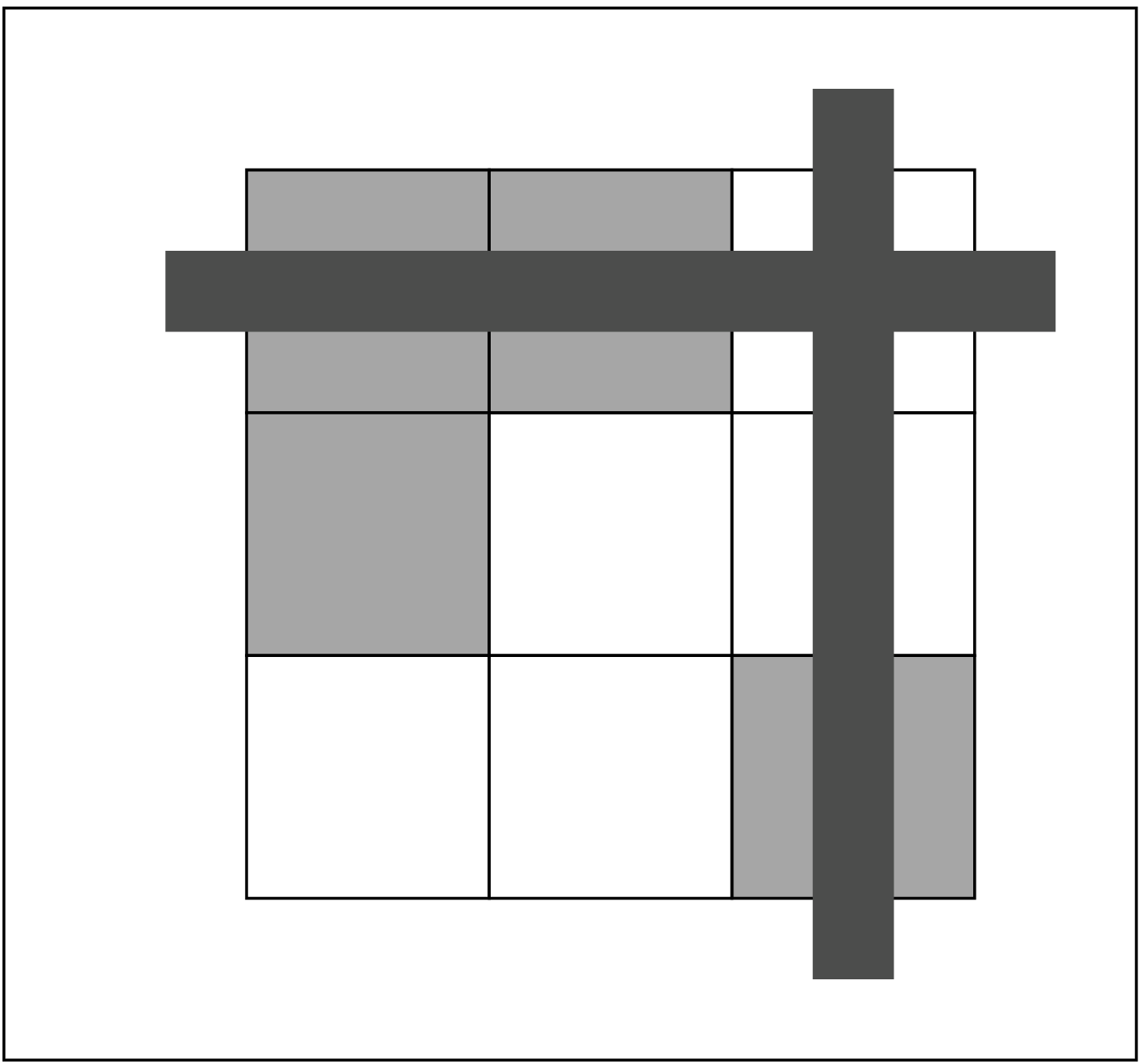}
\includegraphics[width=0.12\linewidth,keepaspectratio]{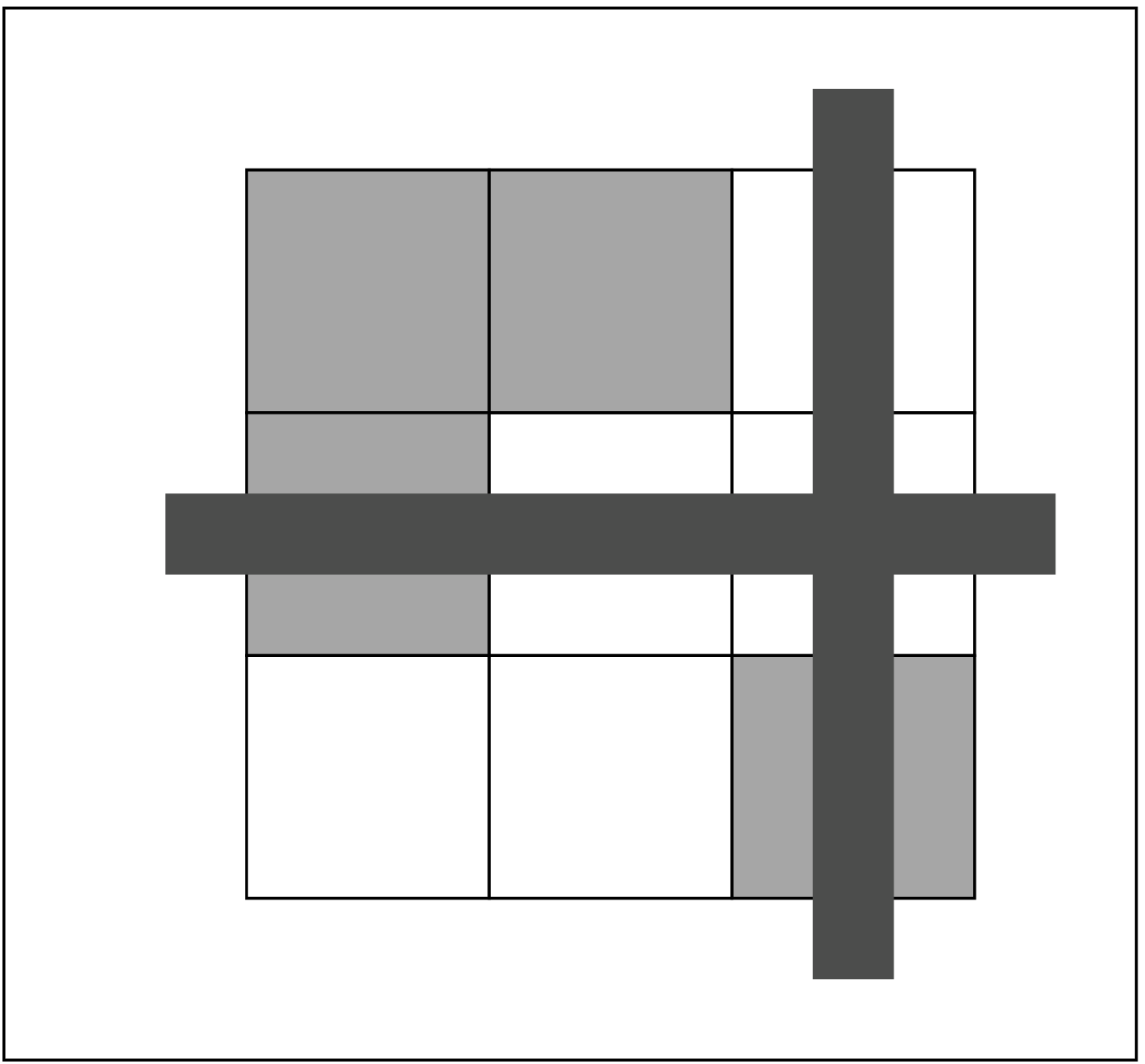}
\includegraphics[width=0.12\linewidth,keepaspectratio]{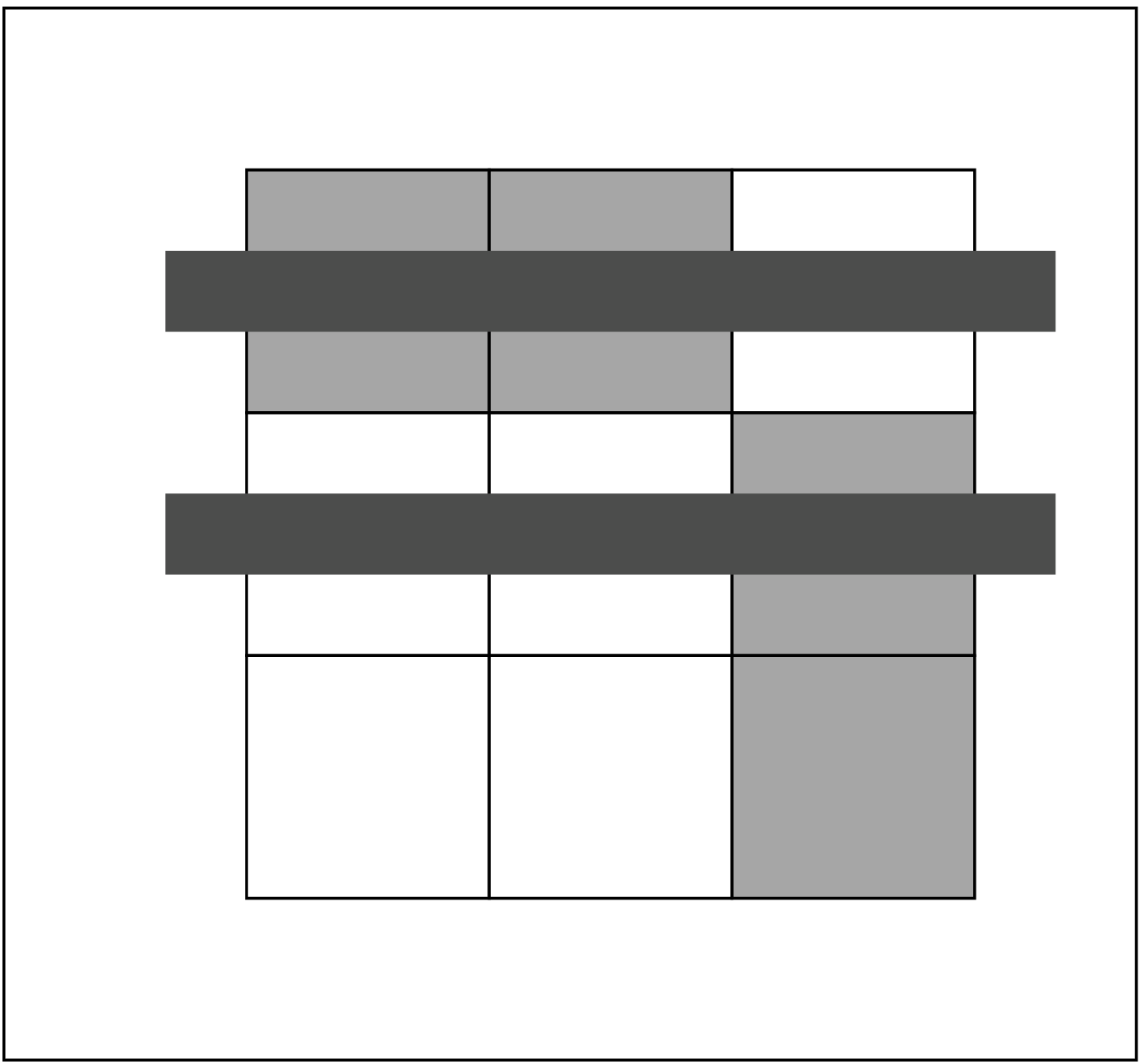}
\includegraphics[width=0.12\linewidth,keepaspectratio]{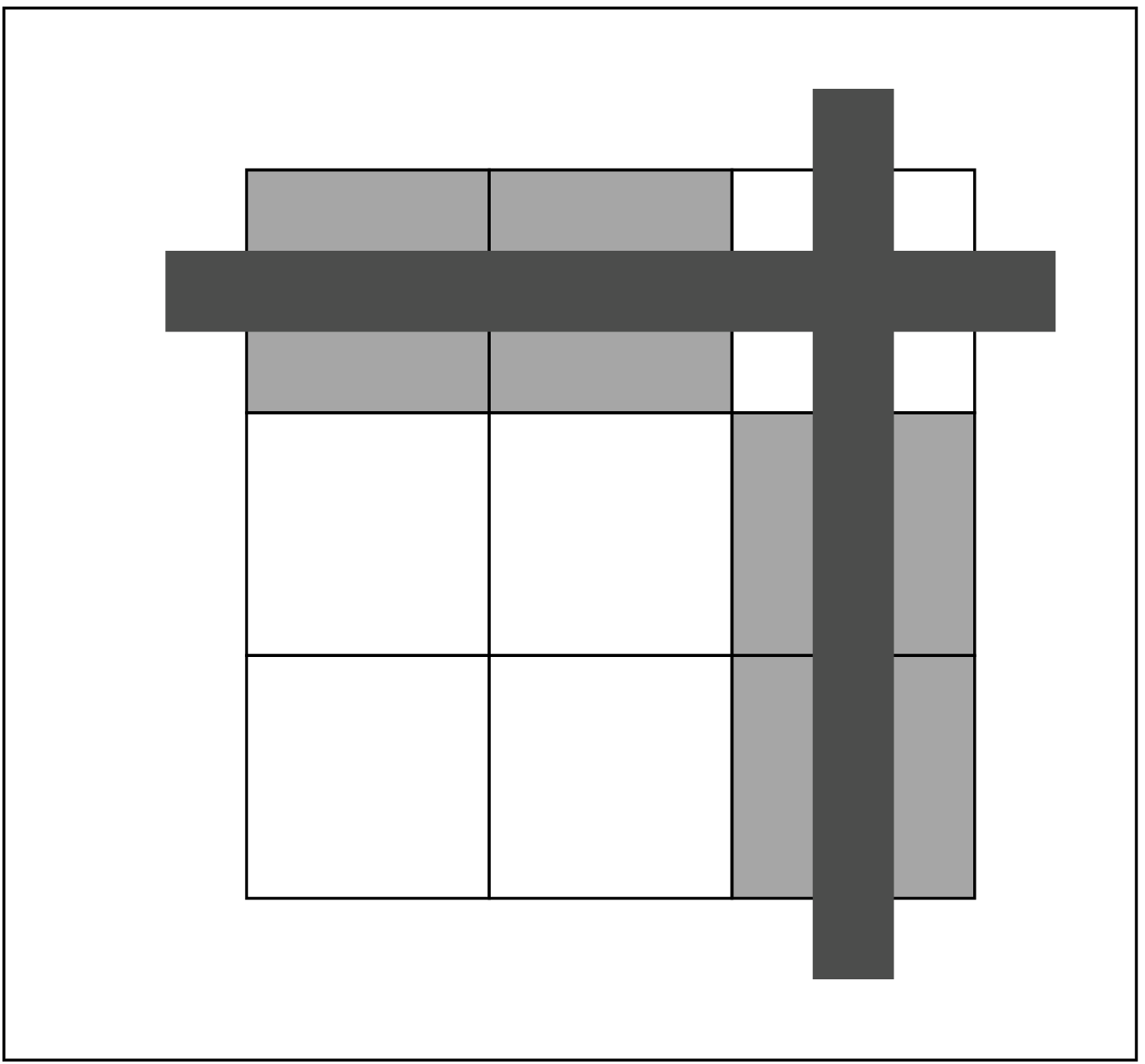}
\includegraphics[width=0.12\linewidth,keepaspectratio]{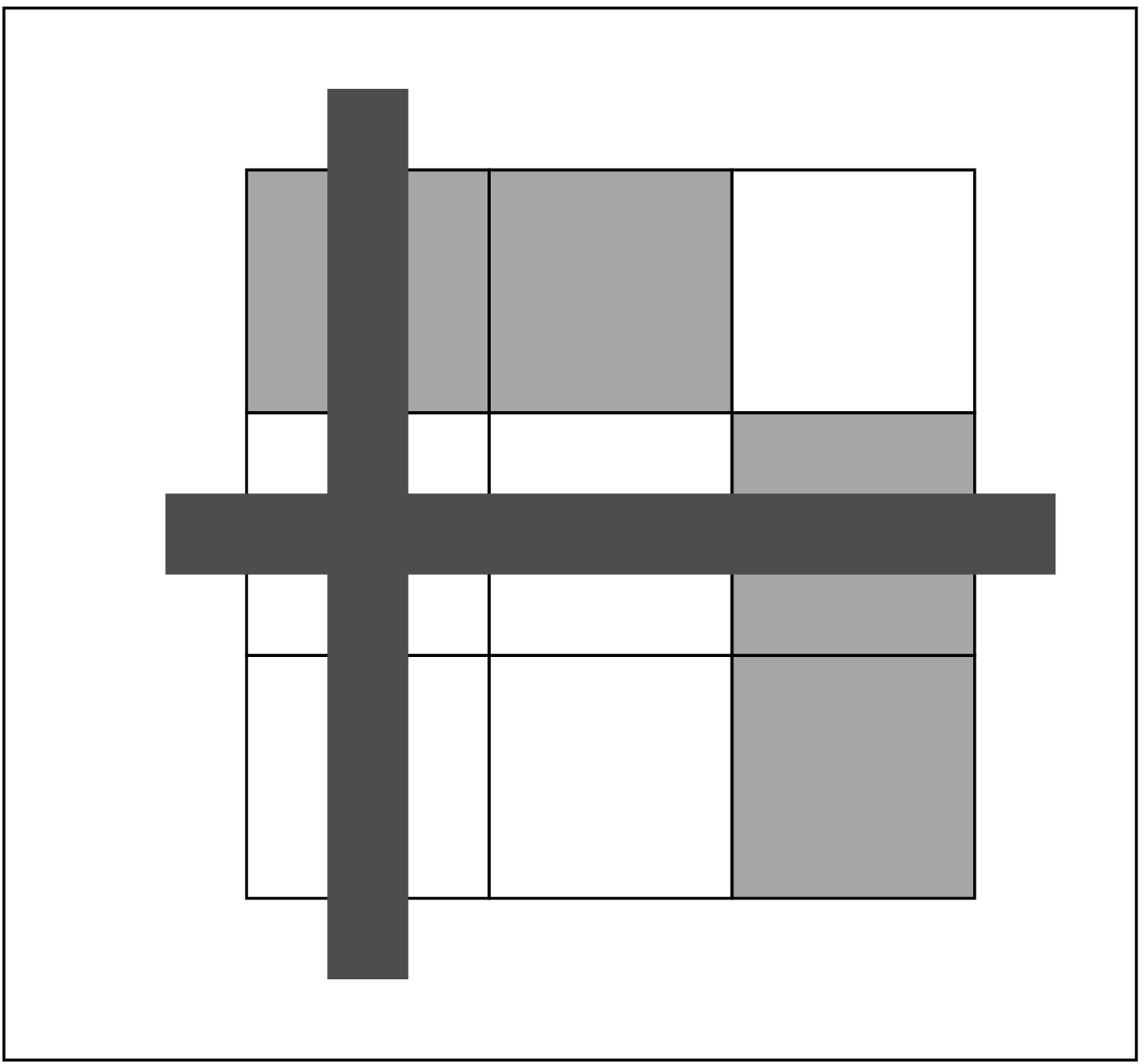}
\includegraphics[width=0.12\linewidth,keepaspectratio]{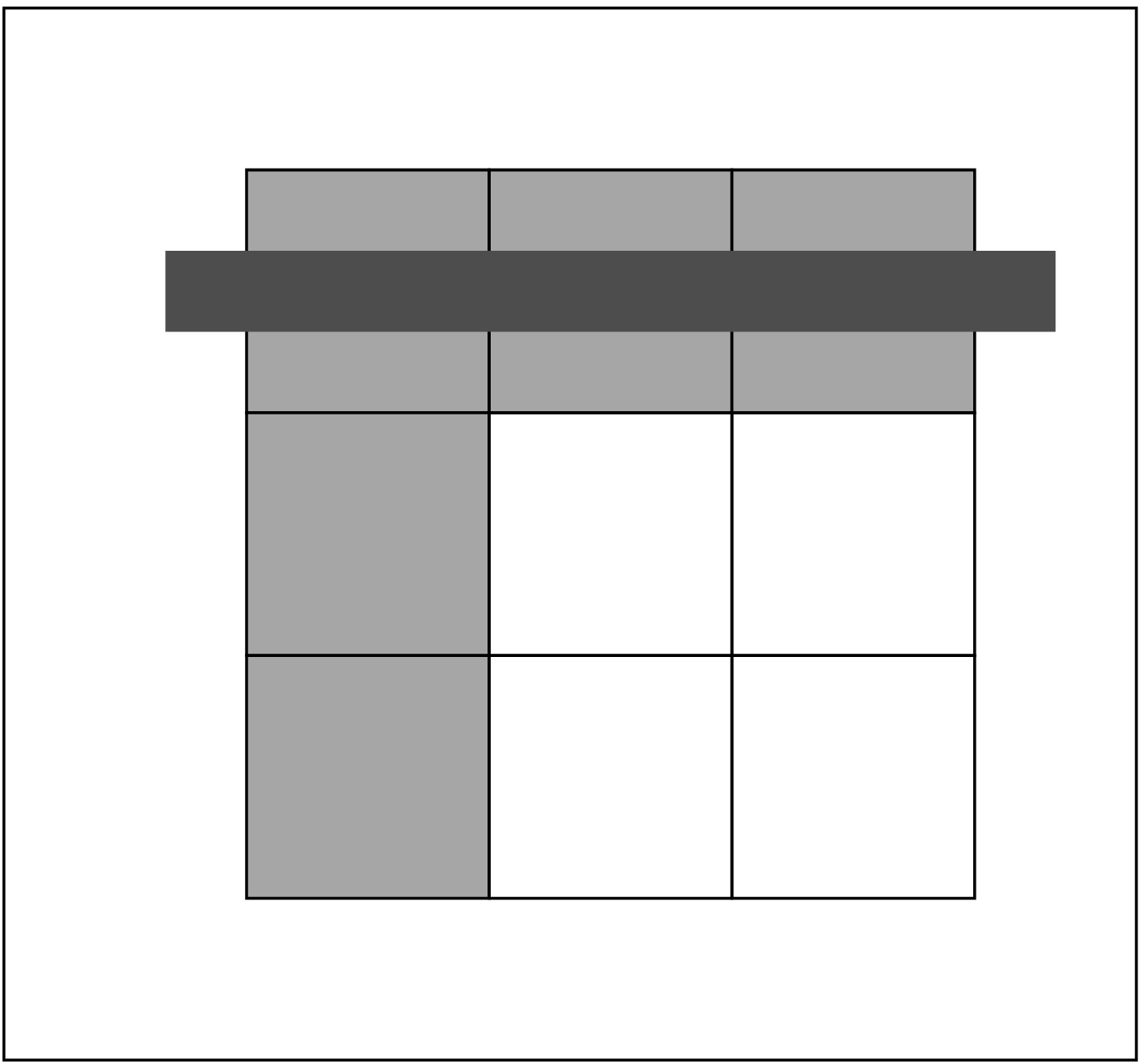}
\includegraphics[width=0.12\linewidth,keepaspectratio]{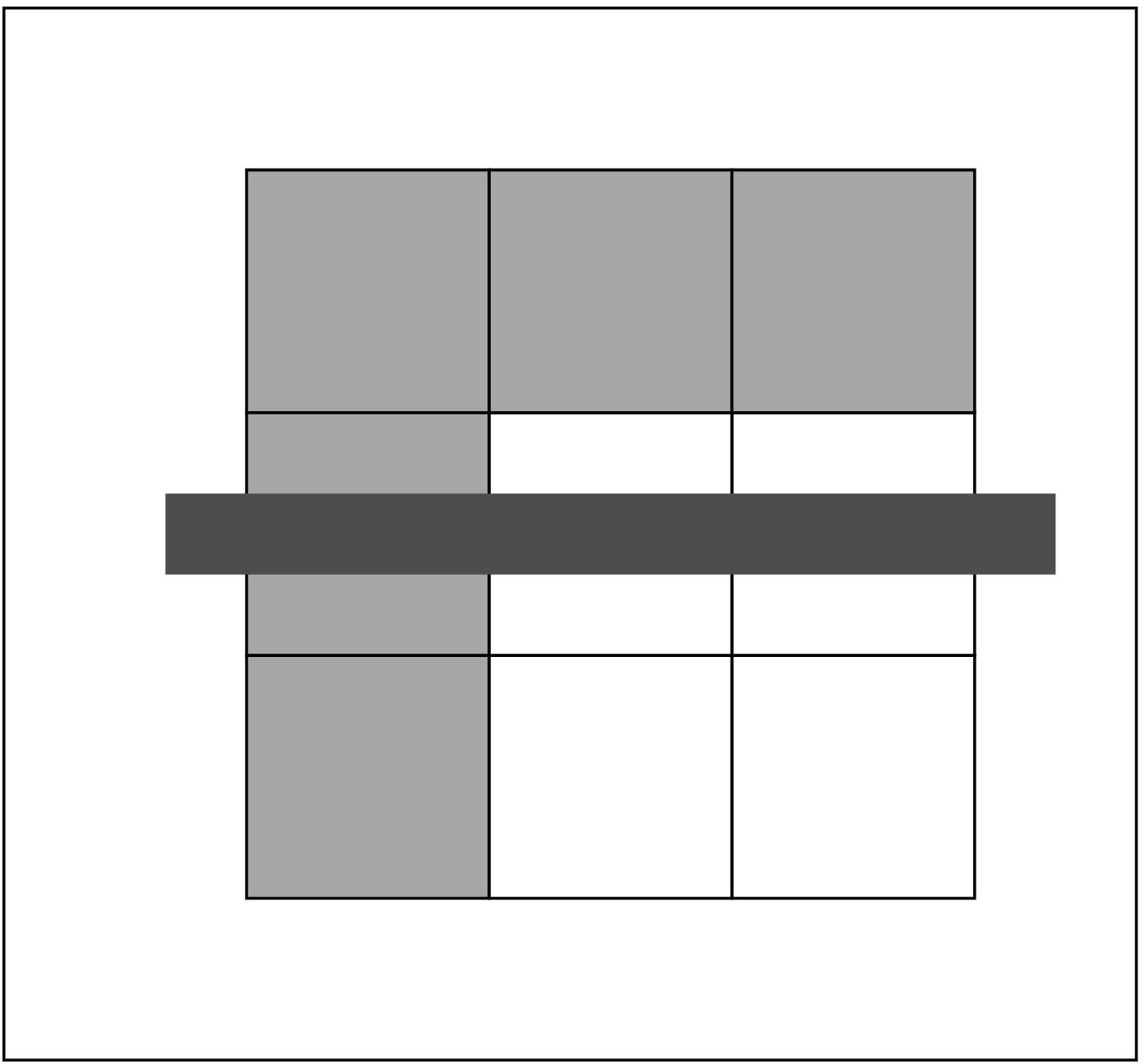}
\includegraphics[width=0.12\linewidth,keepaspectratio]{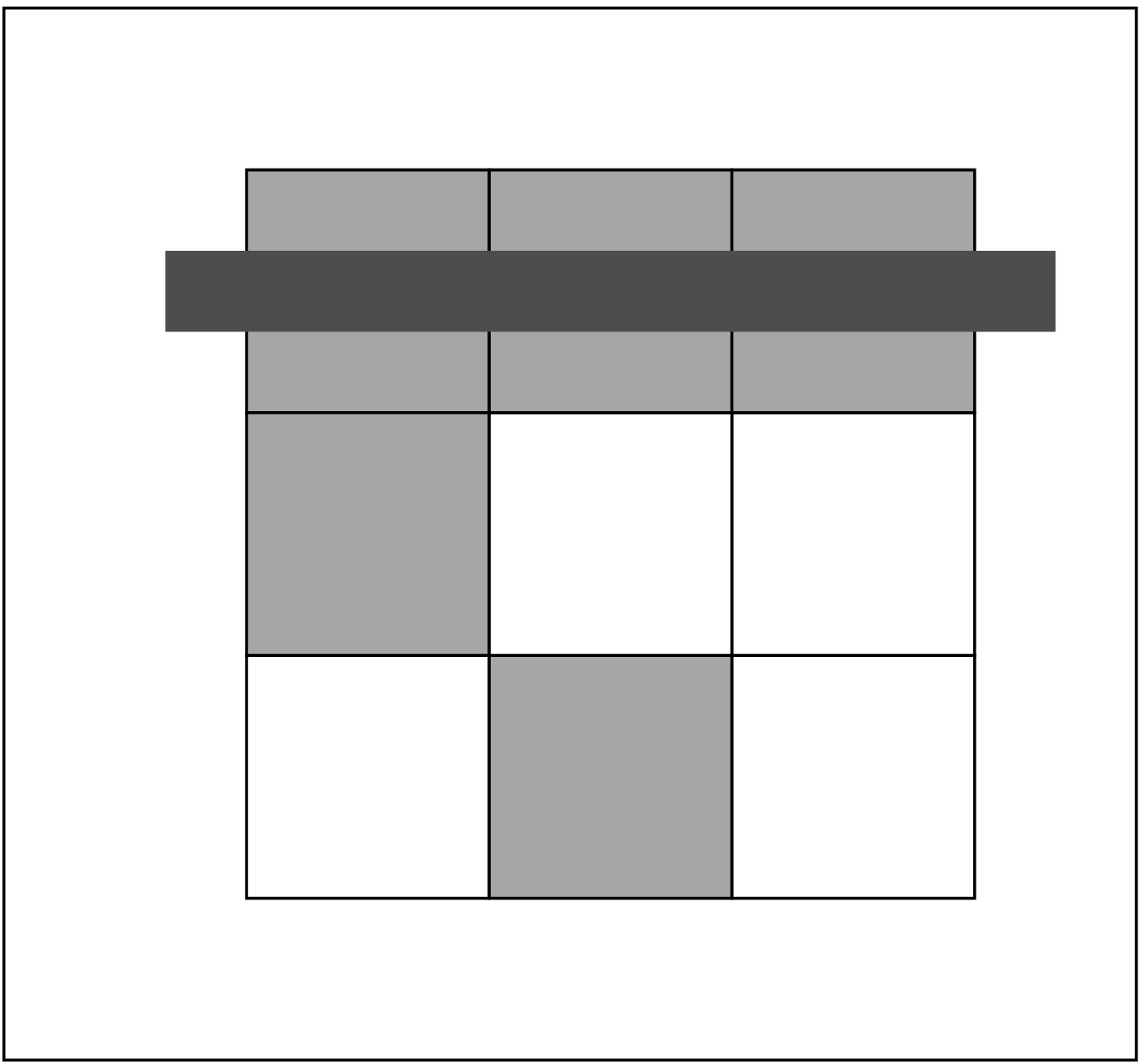}
\includegraphics[width=0.12\linewidth,keepaspectratio]{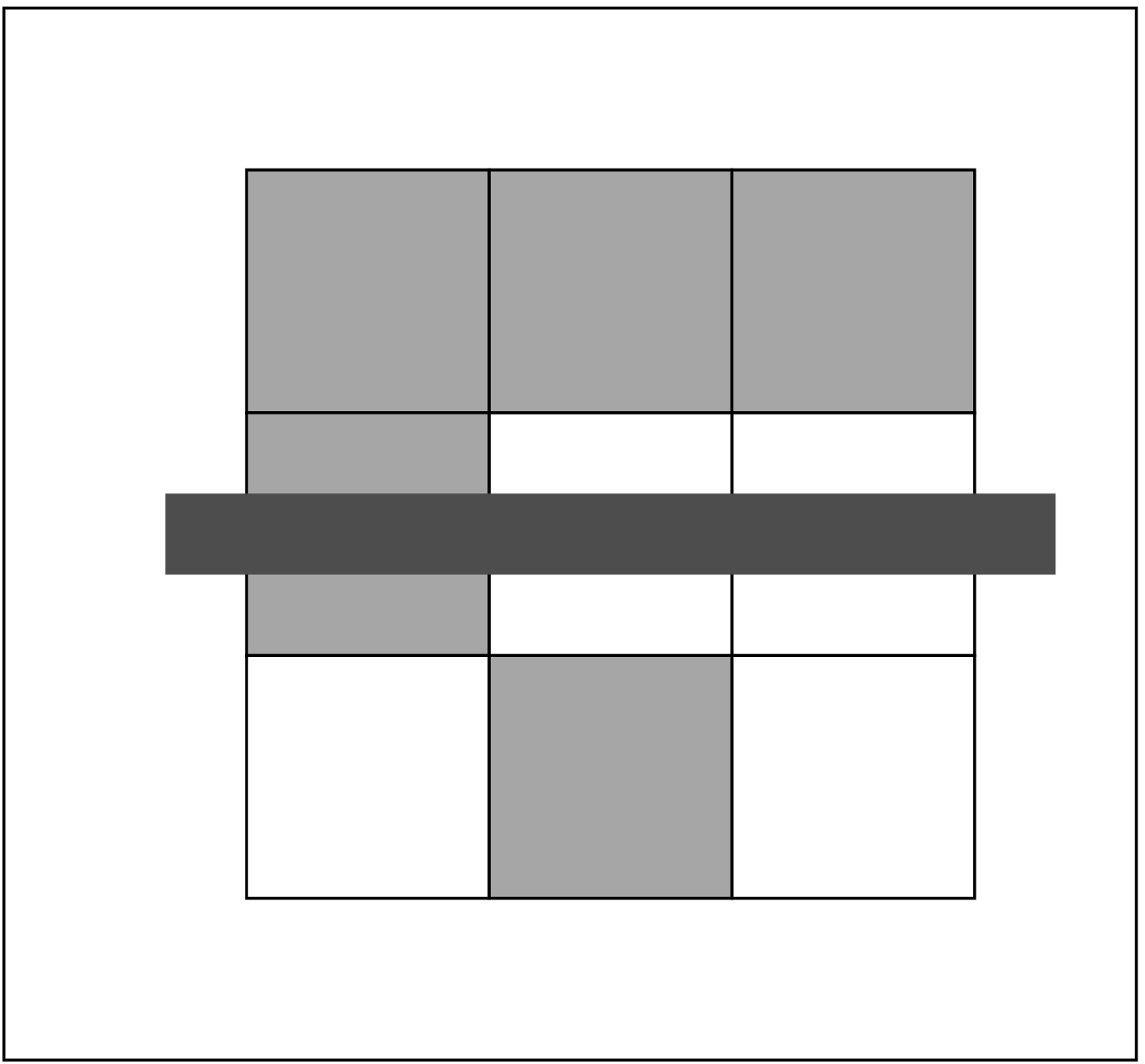}
\includegraphics[width=0.12\linewidth,keepaspectratio]{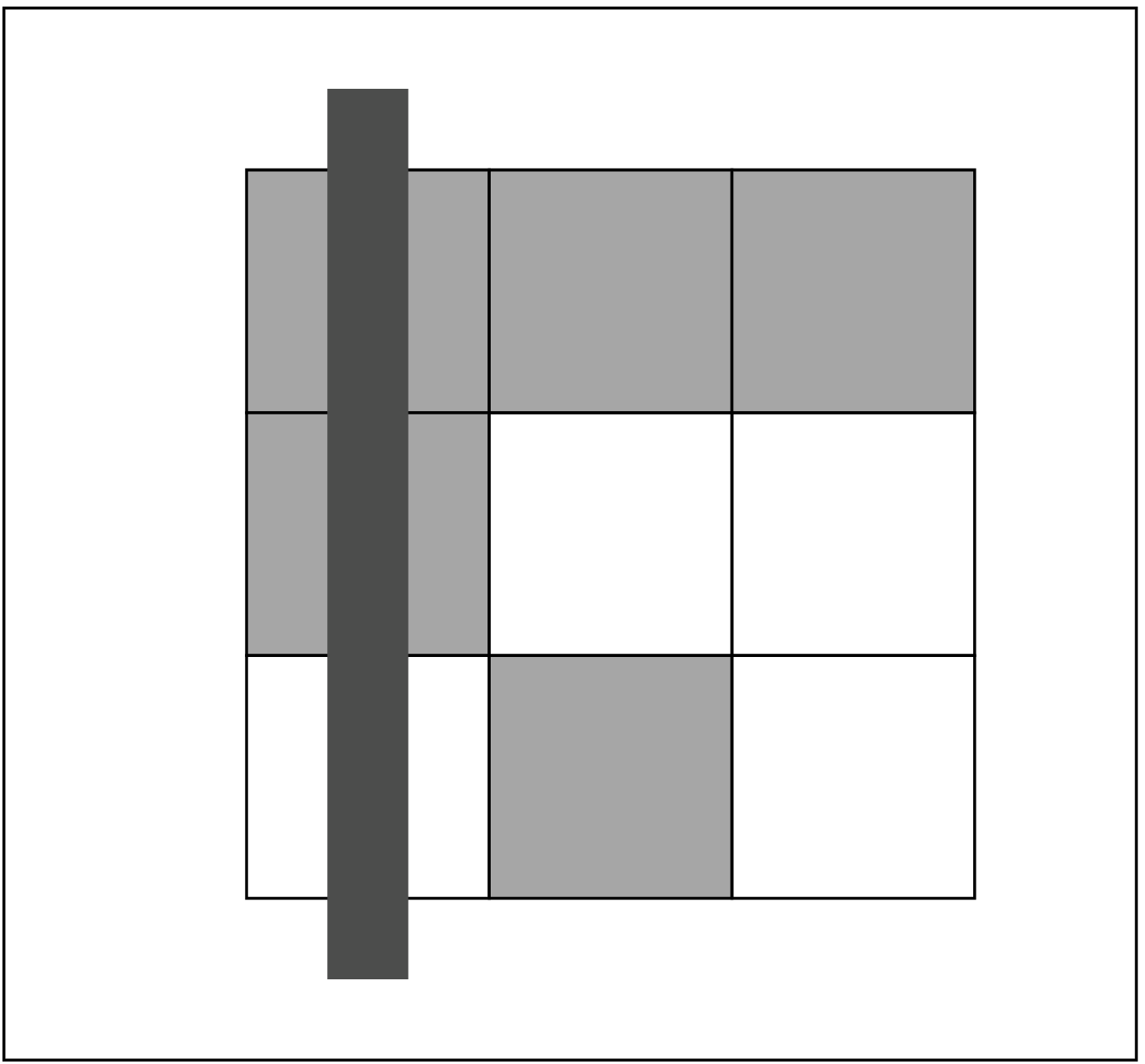}
\includegraphics[width=0.12\linewidth,keepaspectratio]{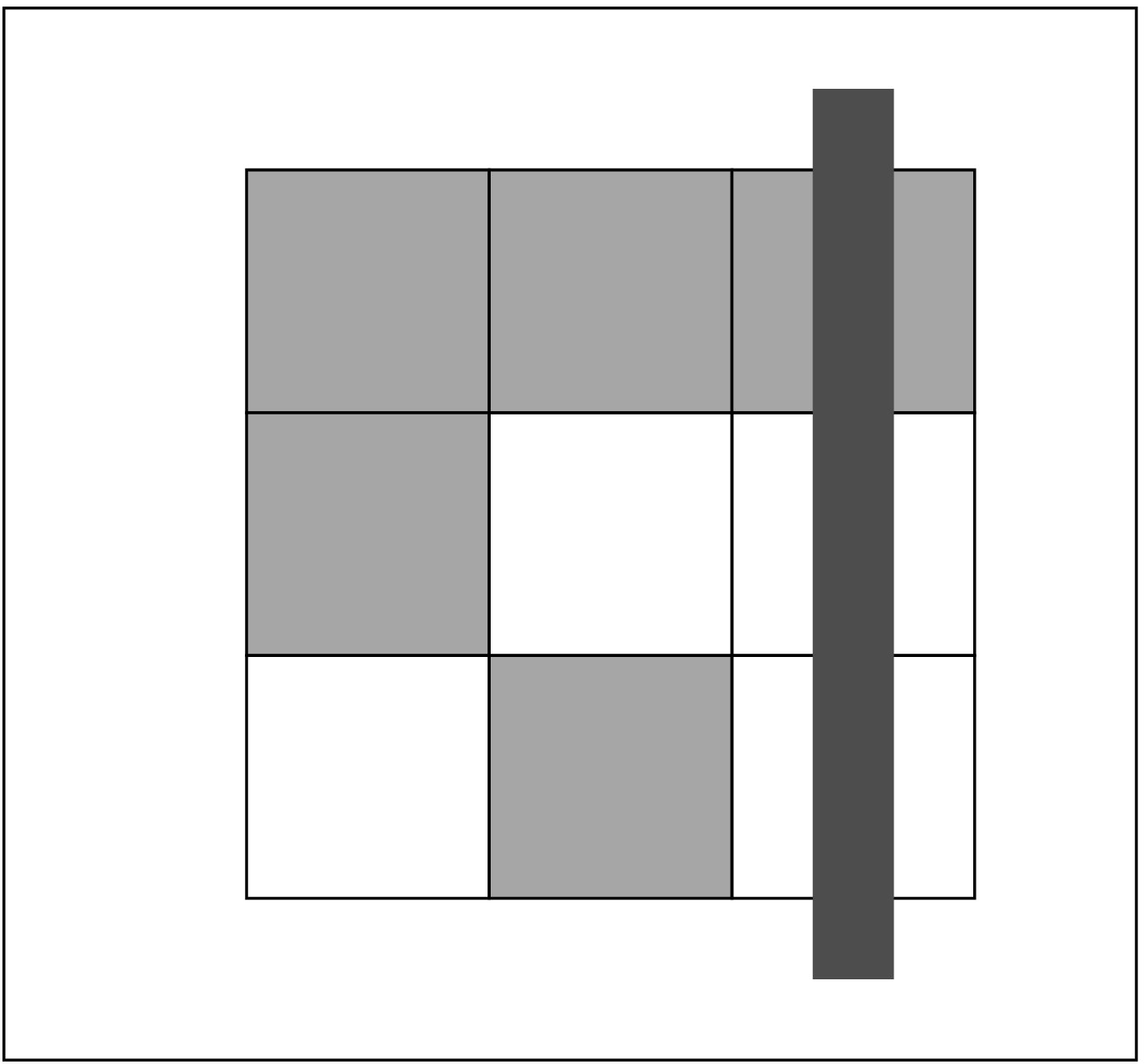}
\includegraphics[width=0.12\linewidth,keepaspectratio]{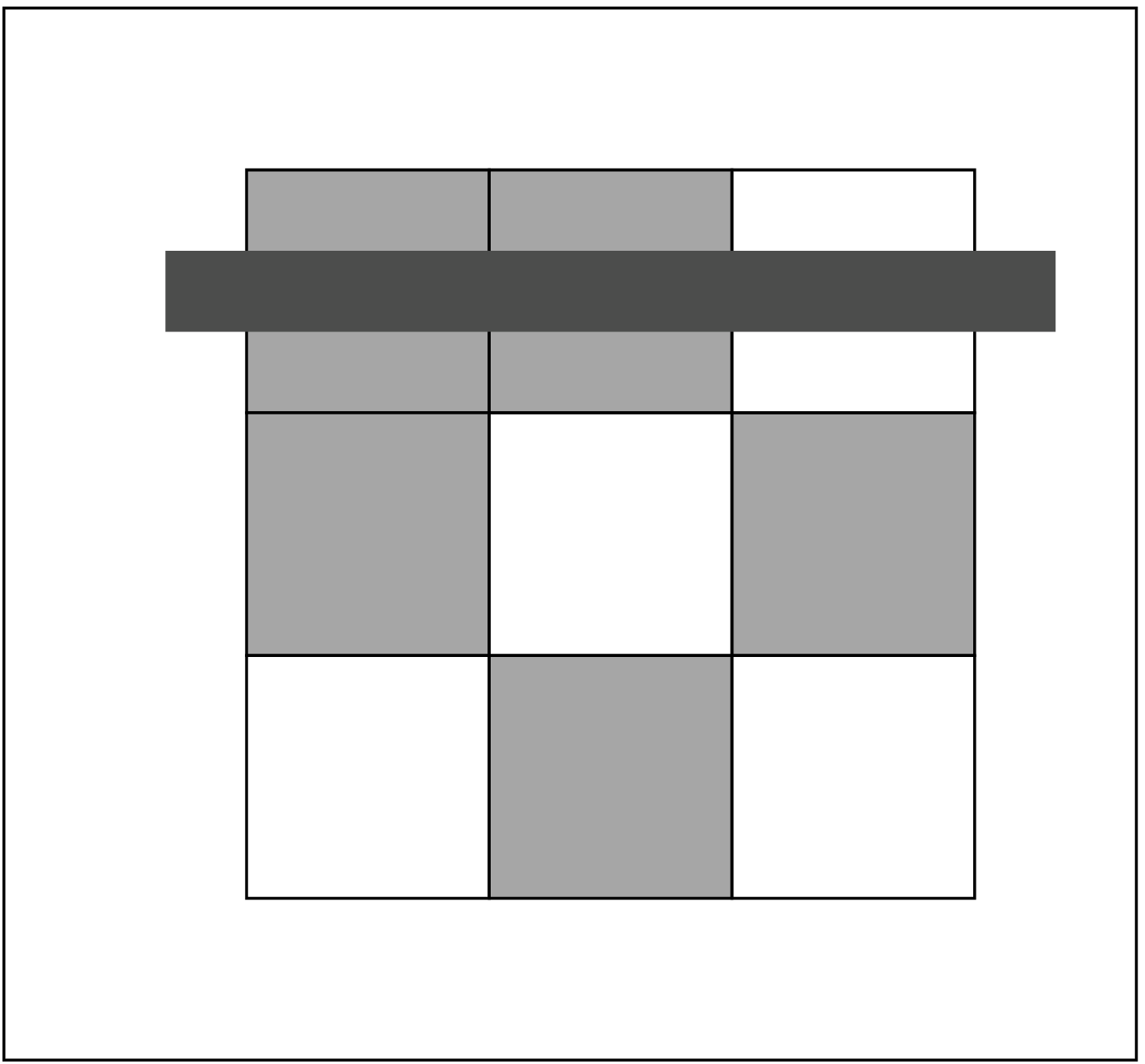}
\includegraphics[width=0.12\linewidth,keepaspectratio]{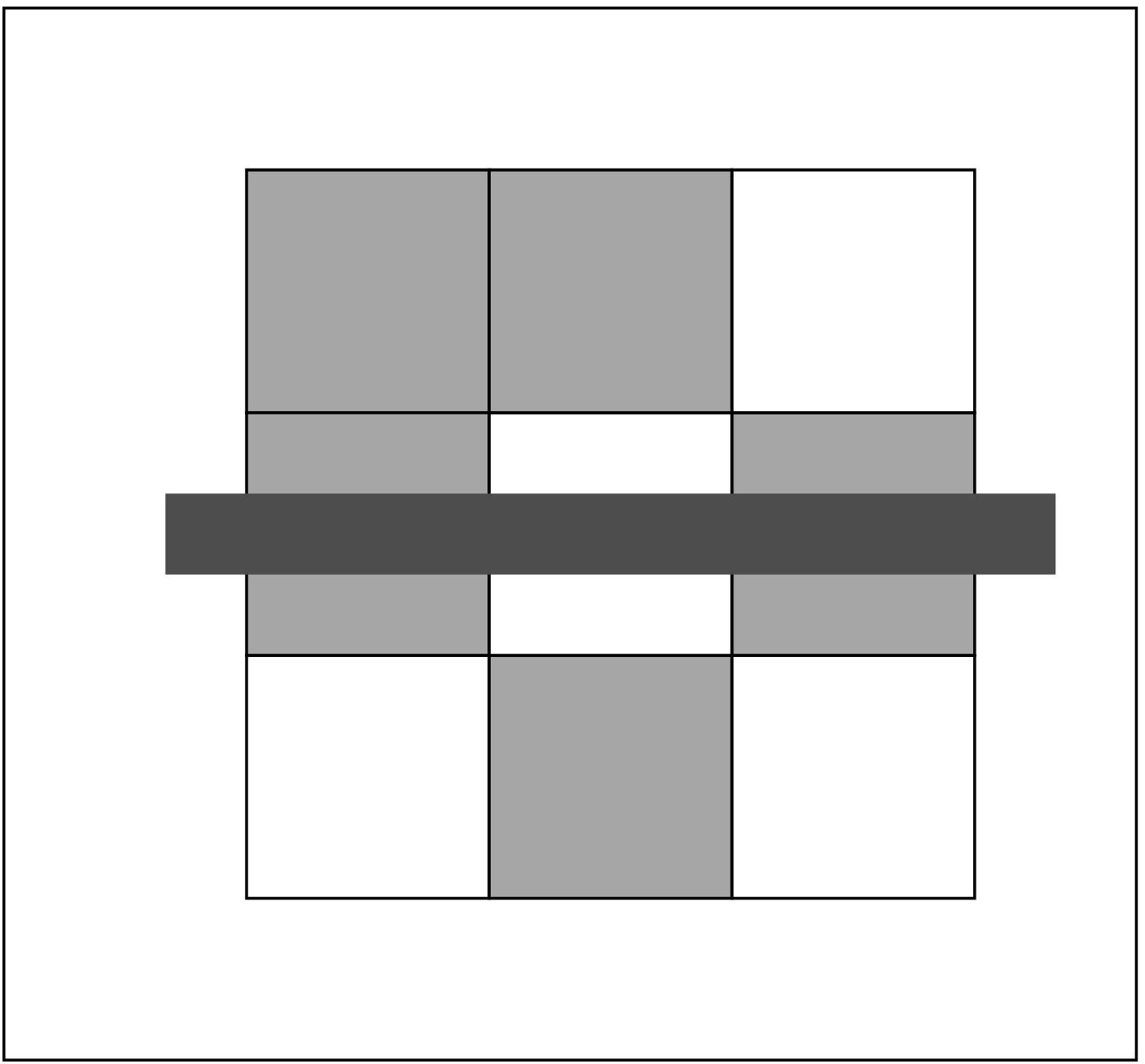}
\includegraphics[width=0.12\linewidth,keepaspectratio]{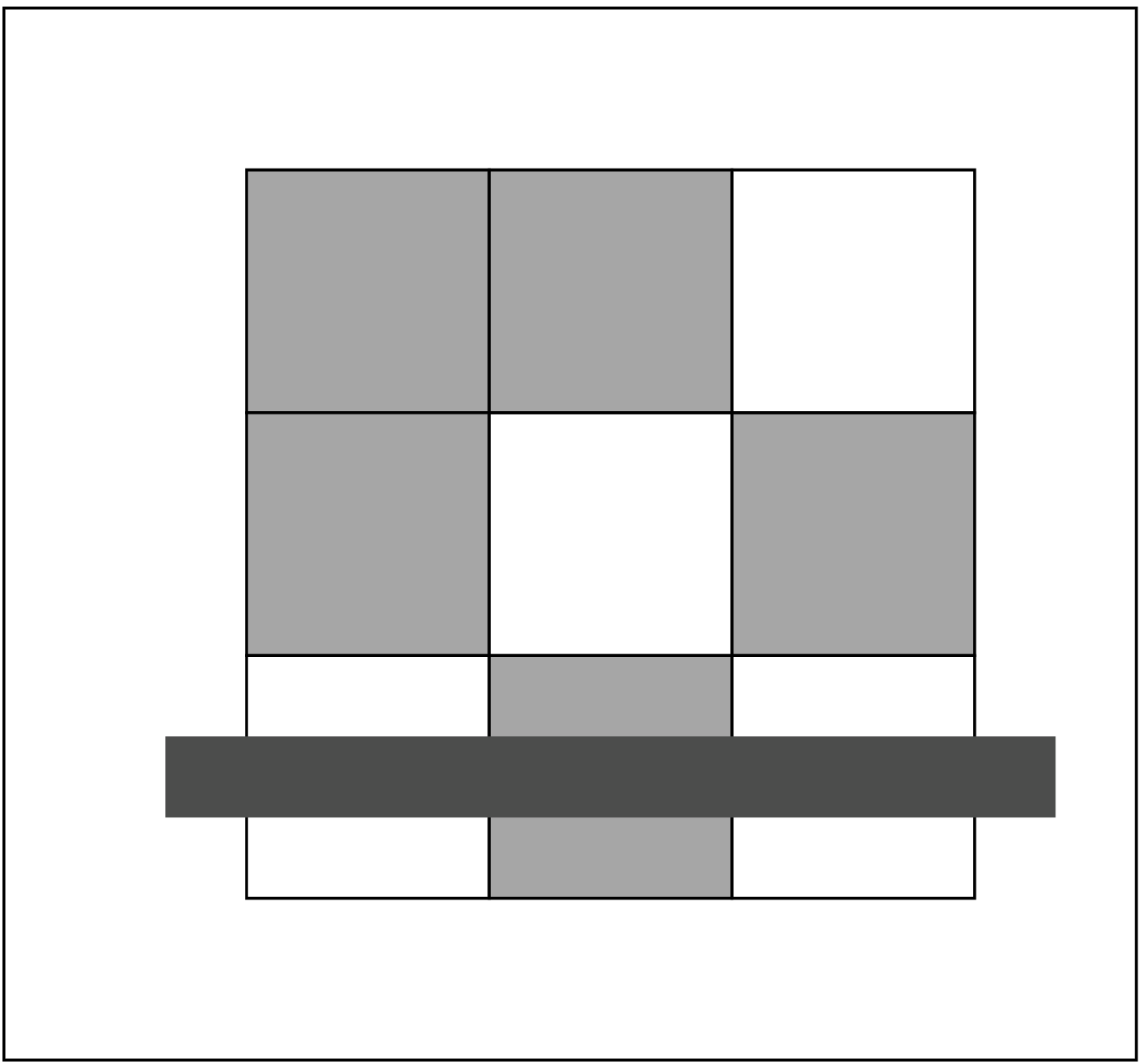}
\includegraphics[width=0.12\linewidth,keepaspectratio]{dummy.eps}

\section*{Appendix II: All {\em non-Sudoku} Models in $\mathit{Missing}(6)$ up to Symmetry} \label{appII}

\noindent
\includegraphics[width=0.12\linewidth,keepaspectratio]{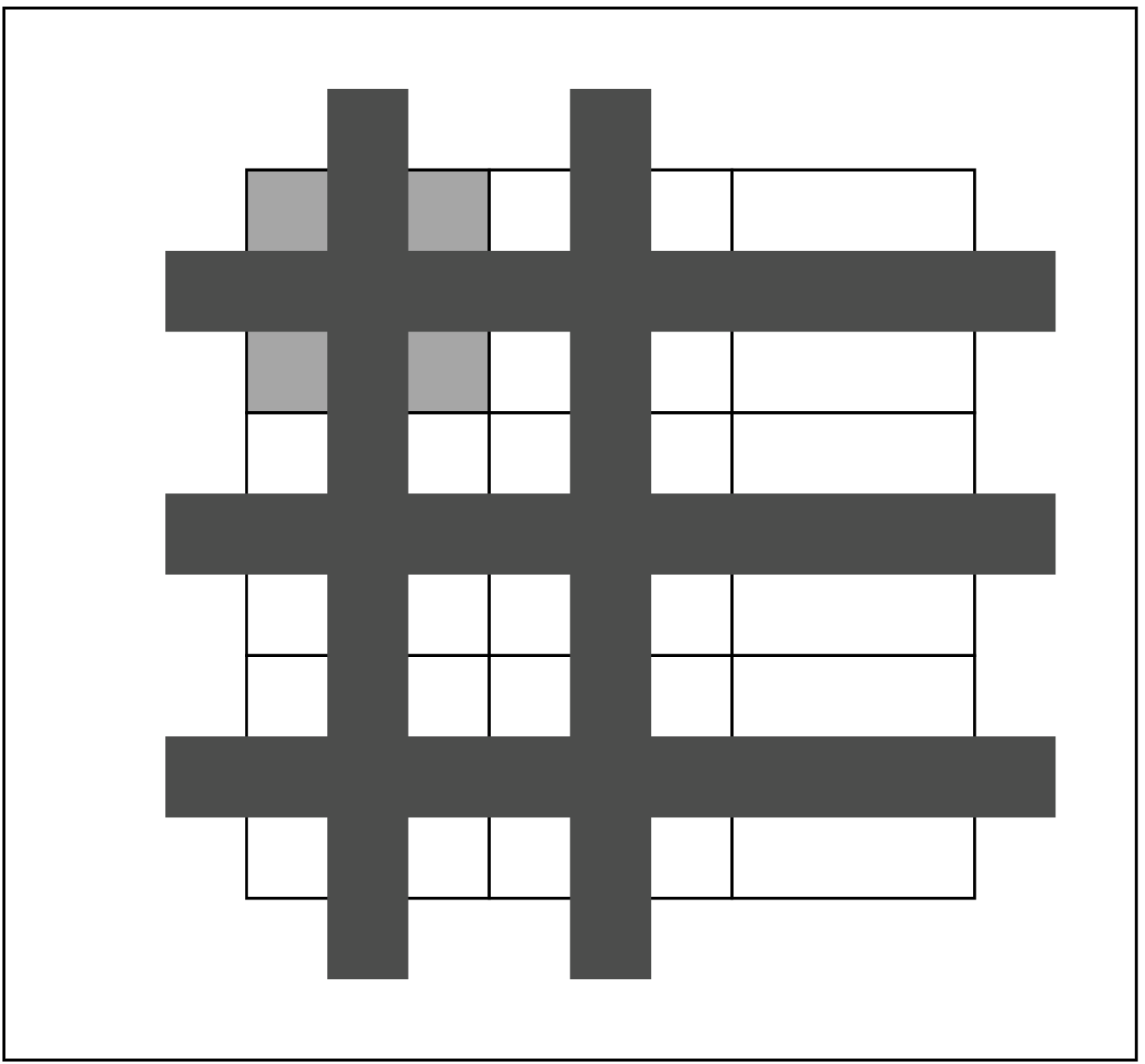}
\includegraphics[width=0.12\linewidth,keepaspectratio]{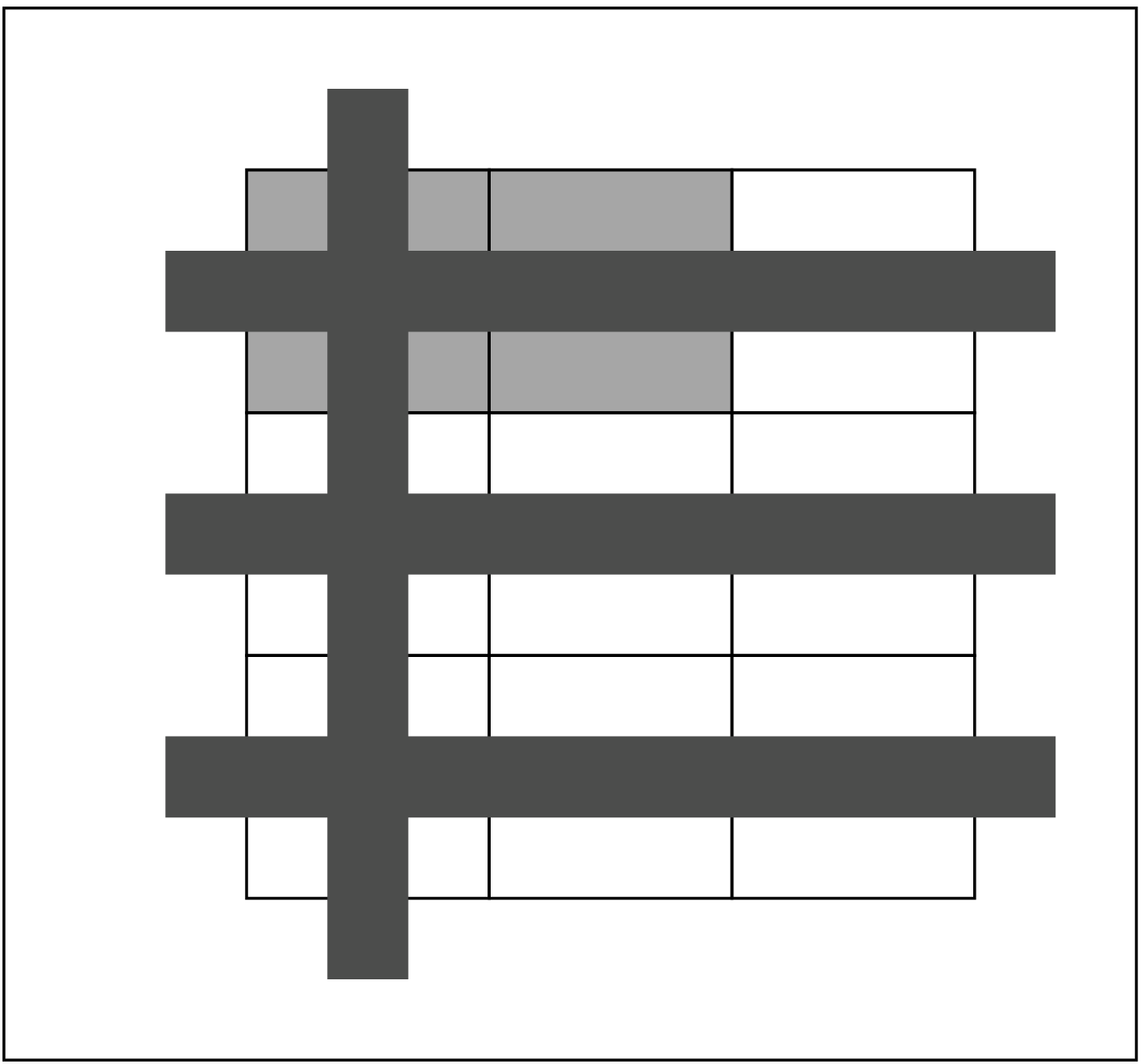}
\includegraphics[width=0.12\linewidth,keepaspectratio]{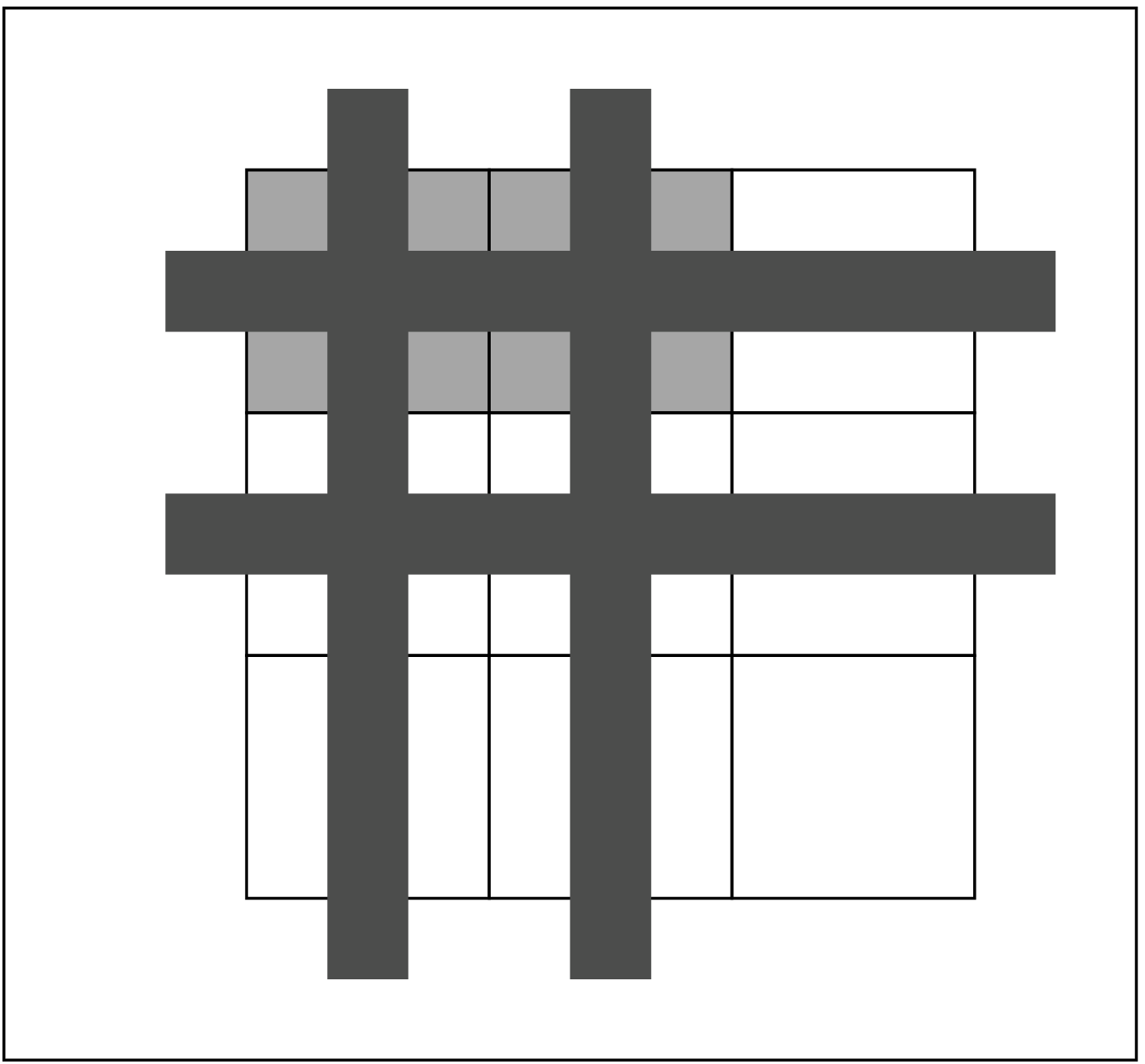}
\includegraphics[width=0.12\linewidth,keepaspectratio]{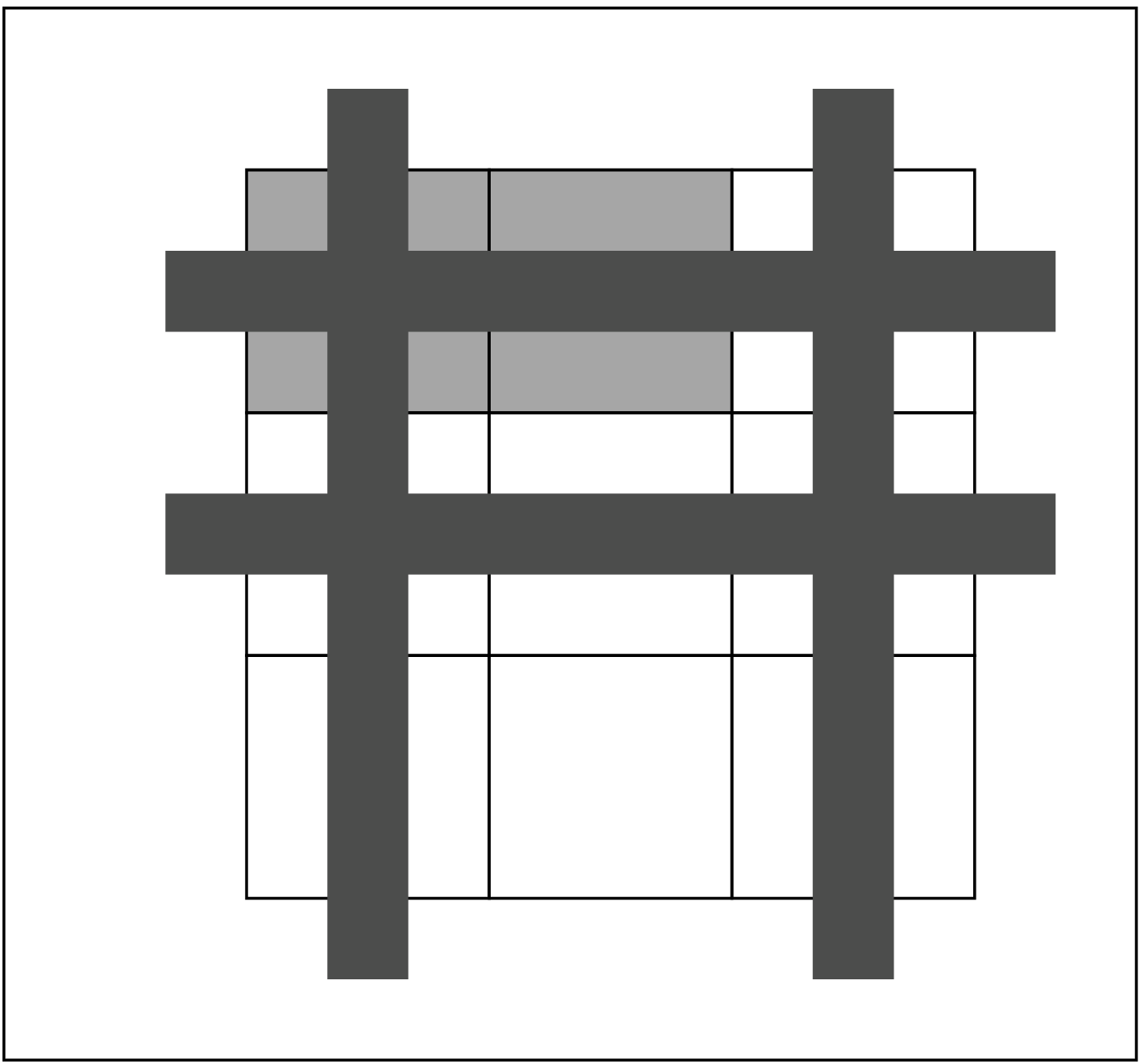}
\includegraphics[width=0.12\linewidth,keepaspectratio]{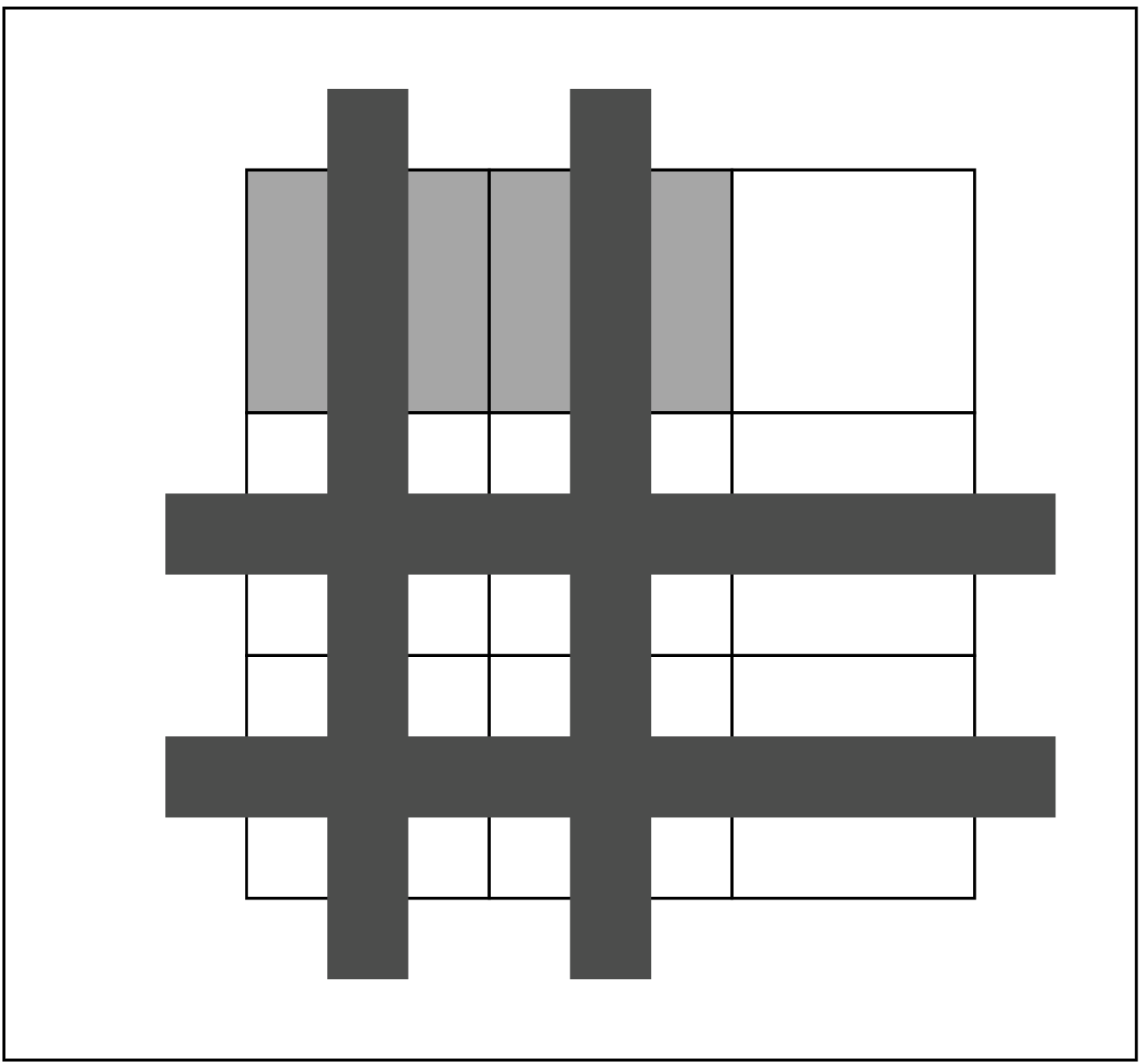}
\includegraphics[width=0.12\linewidth,keepaspectratio]{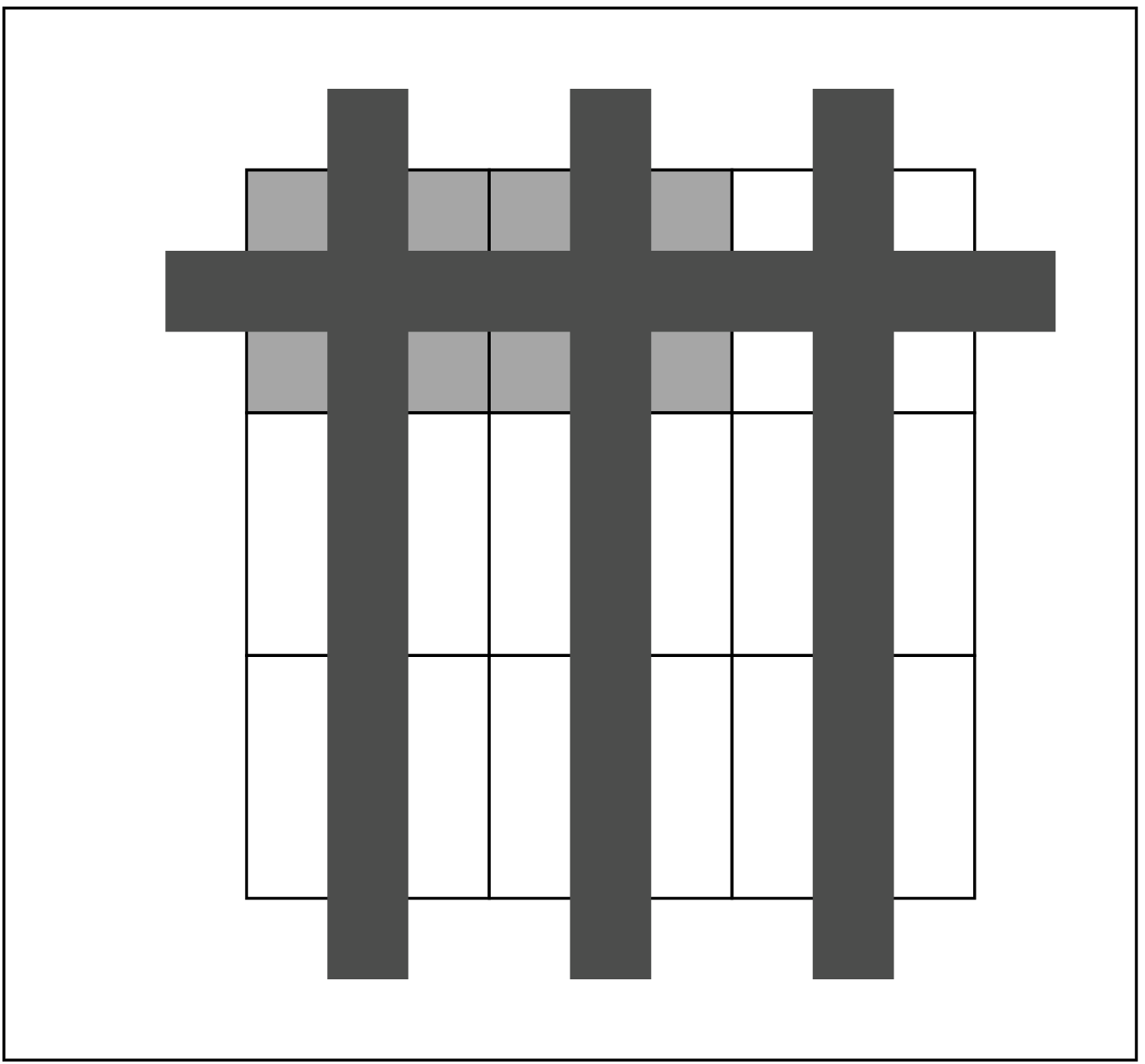}
\includegraphics[width=0.12\linewidth,keepaspectratio]{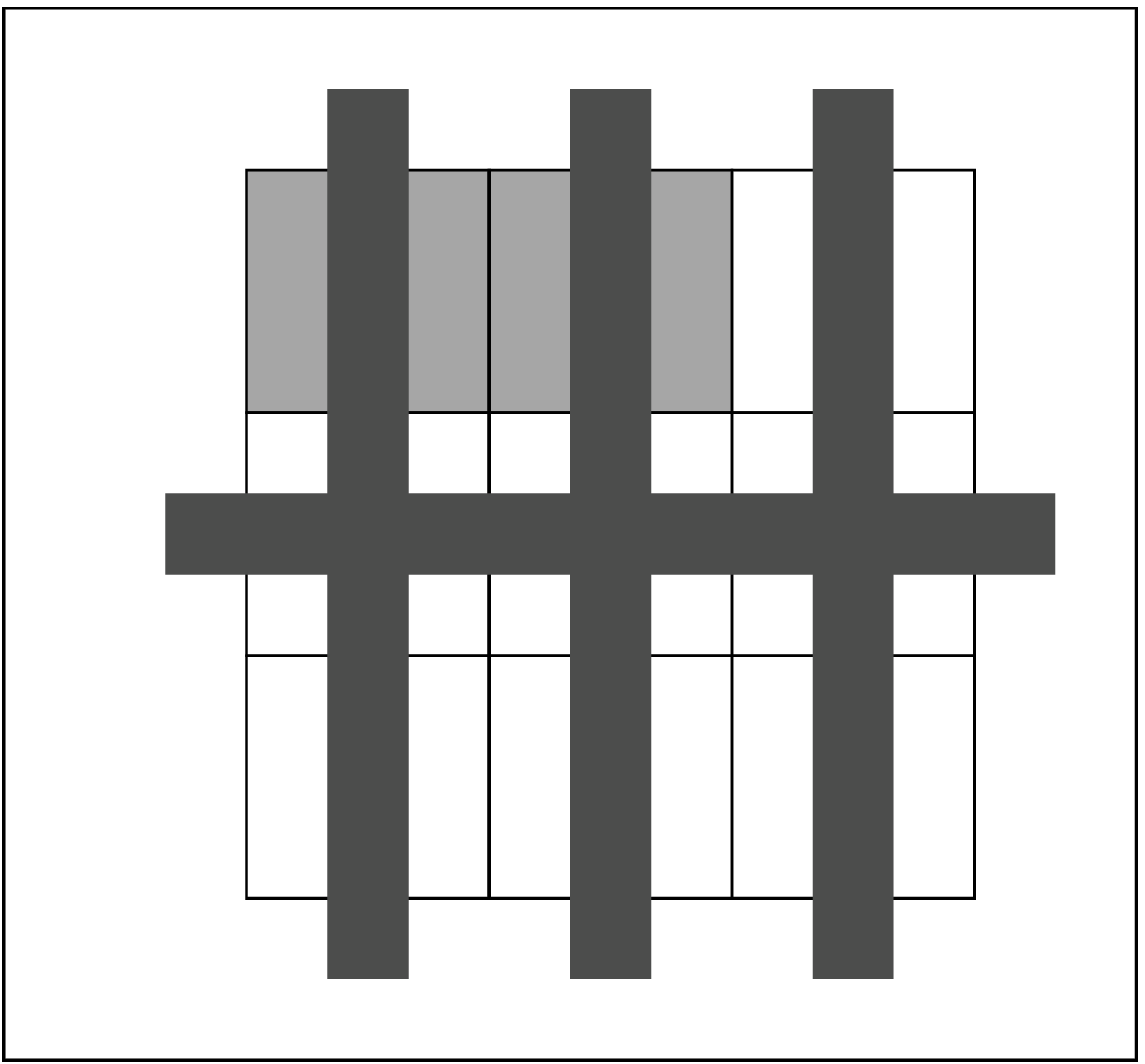}
\includegraphics[width=0.12\linewidth,keepaspectratio]{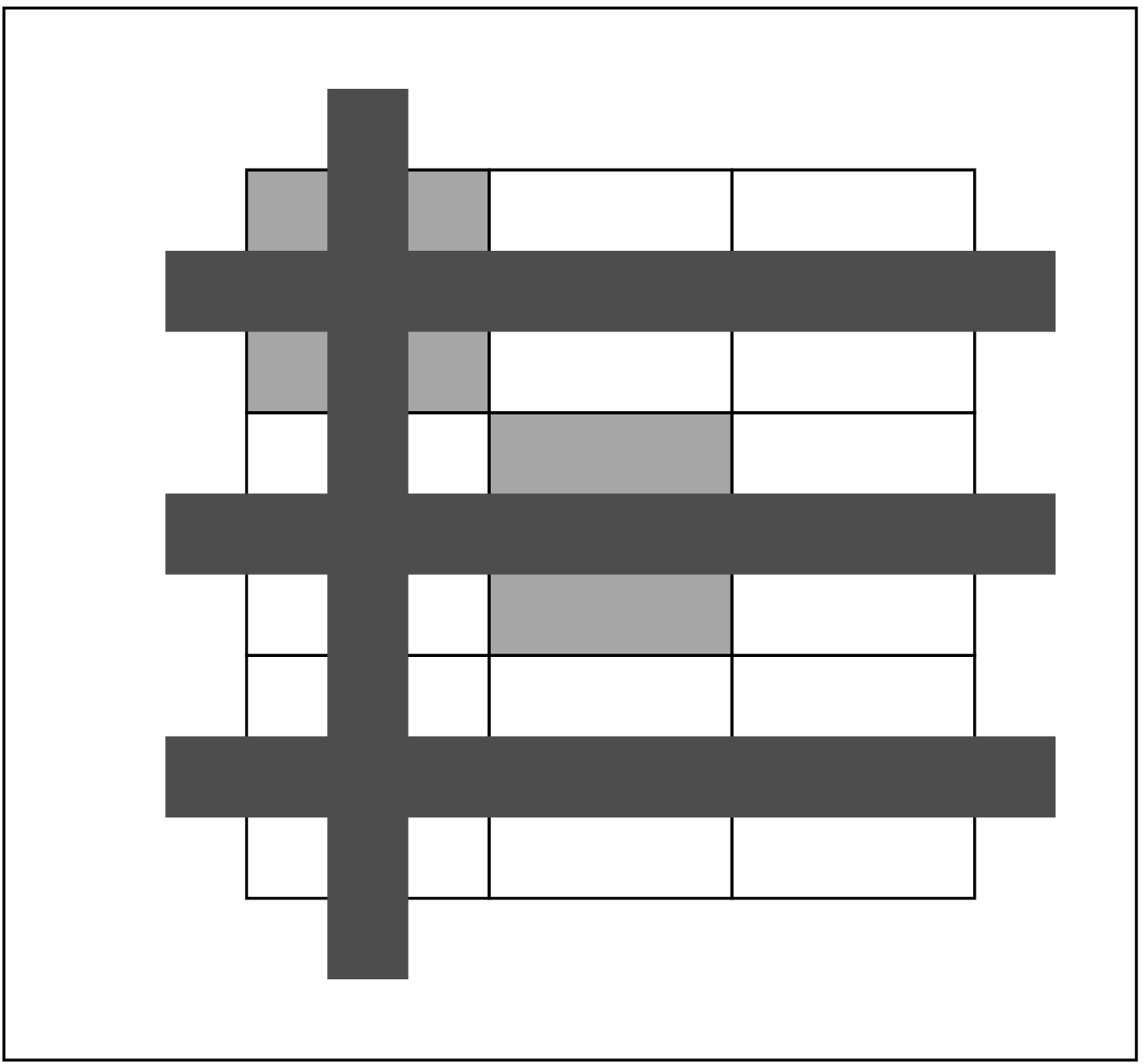}
\includegraphics[width=0.12\linewidth,keepaspectratio]{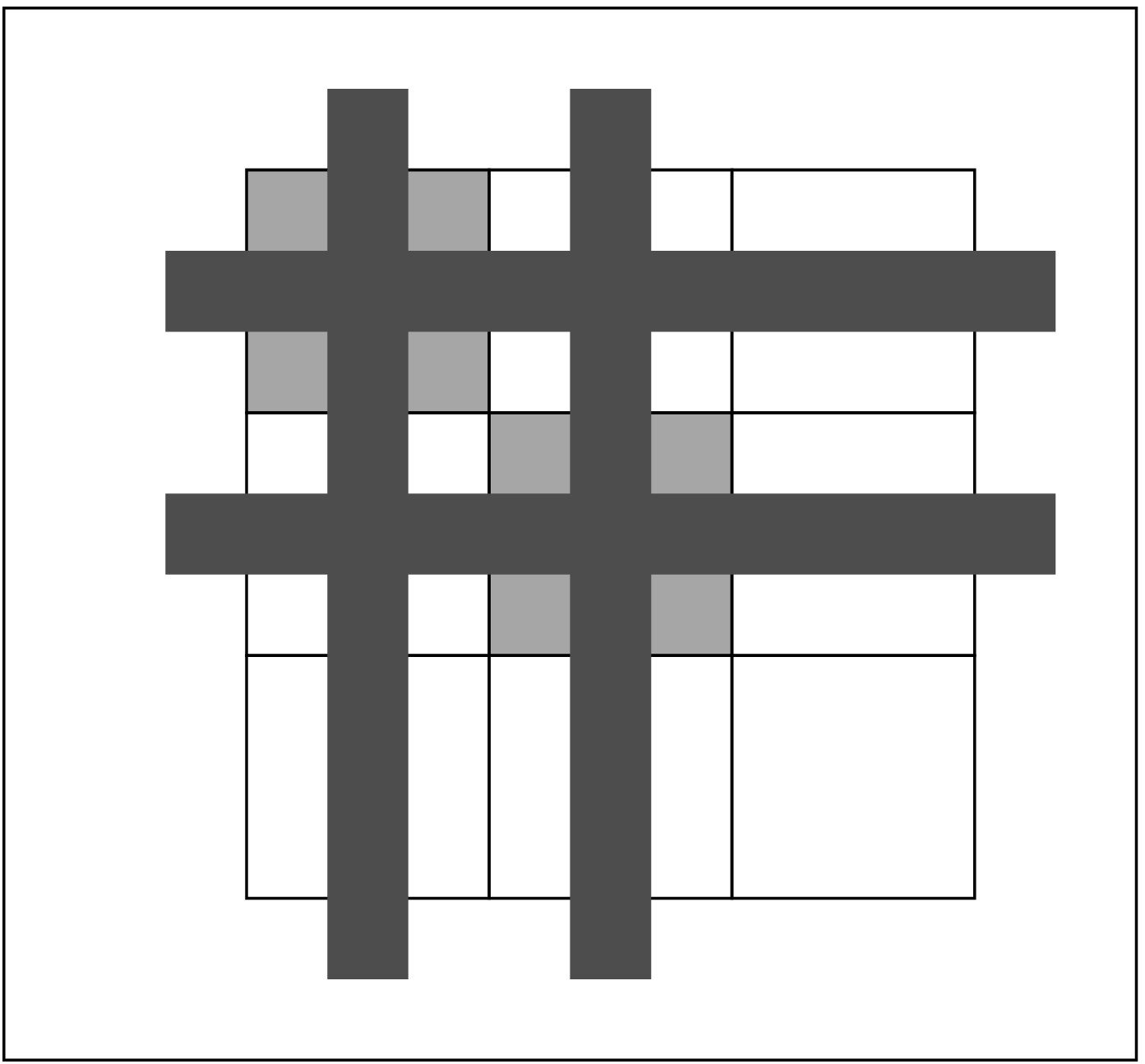}
\includegraphics[width=0.12\linewidth,keepaspectratio]{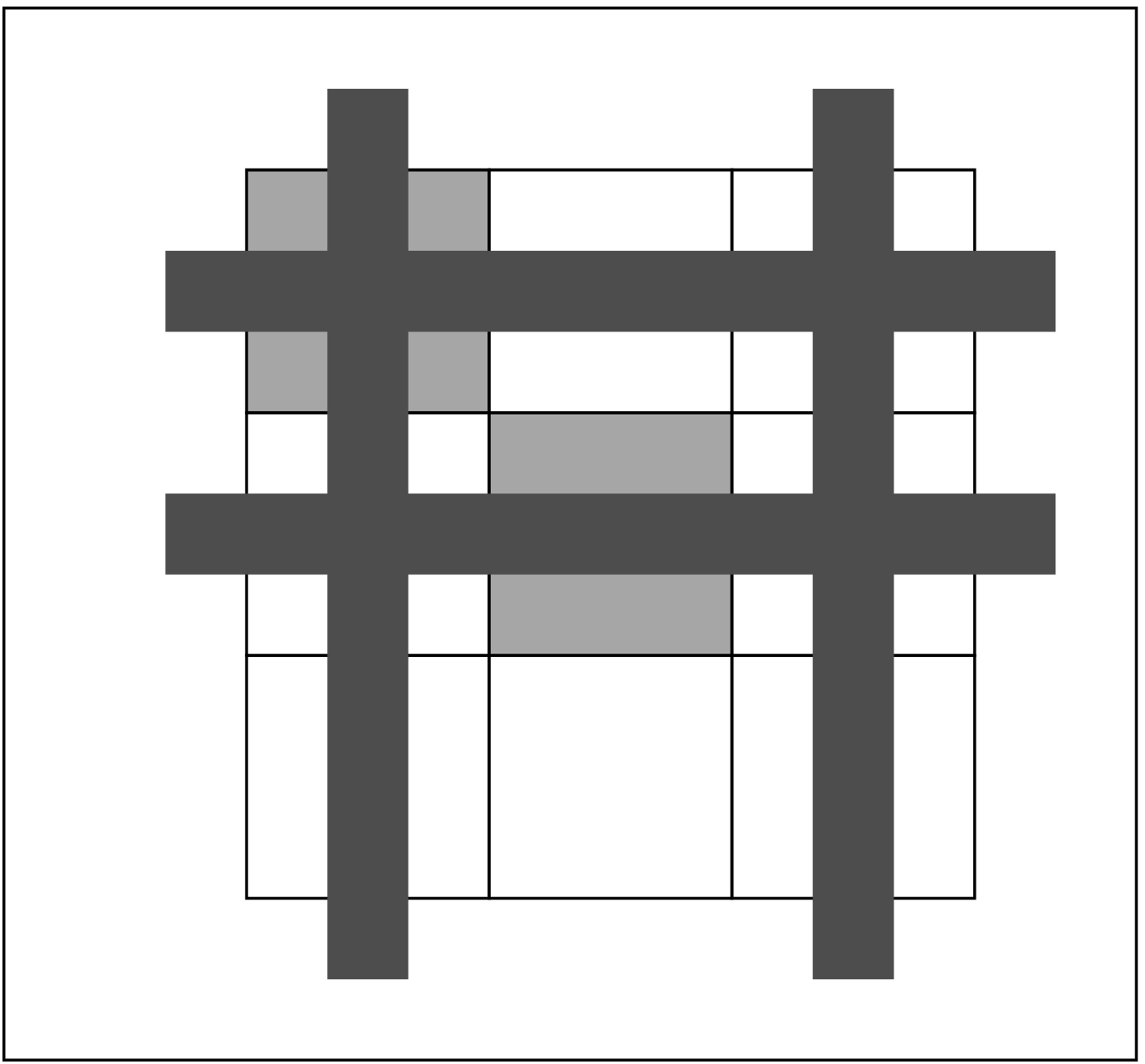}
\includegraphics[width=0.12\linewidth,keepaspectratio]{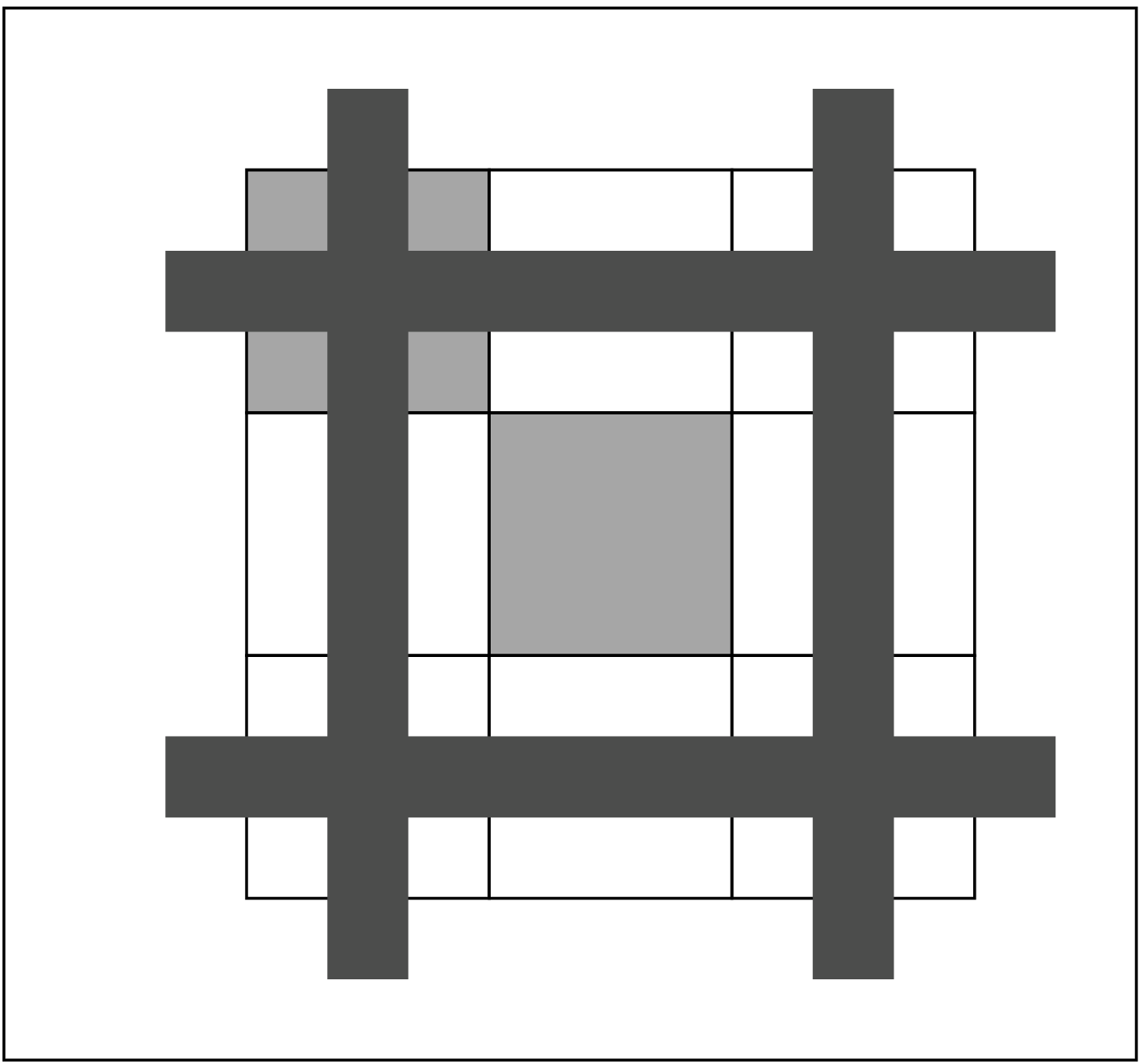}
\includegraphics[width=0.12\linewidth,keepaspectratio]{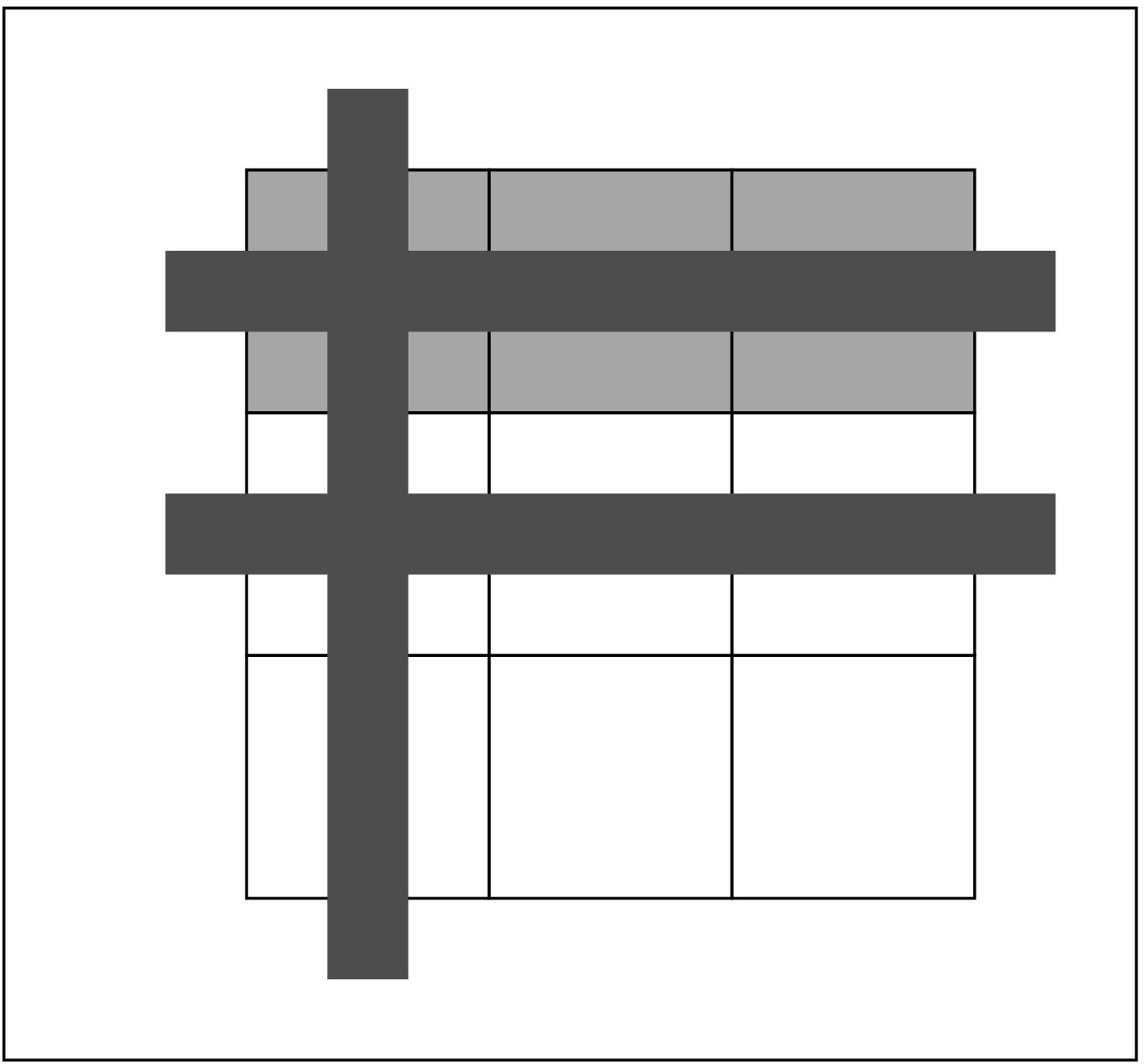}
\includegraphics[width=0.12\linewidth,keepaspectratio]{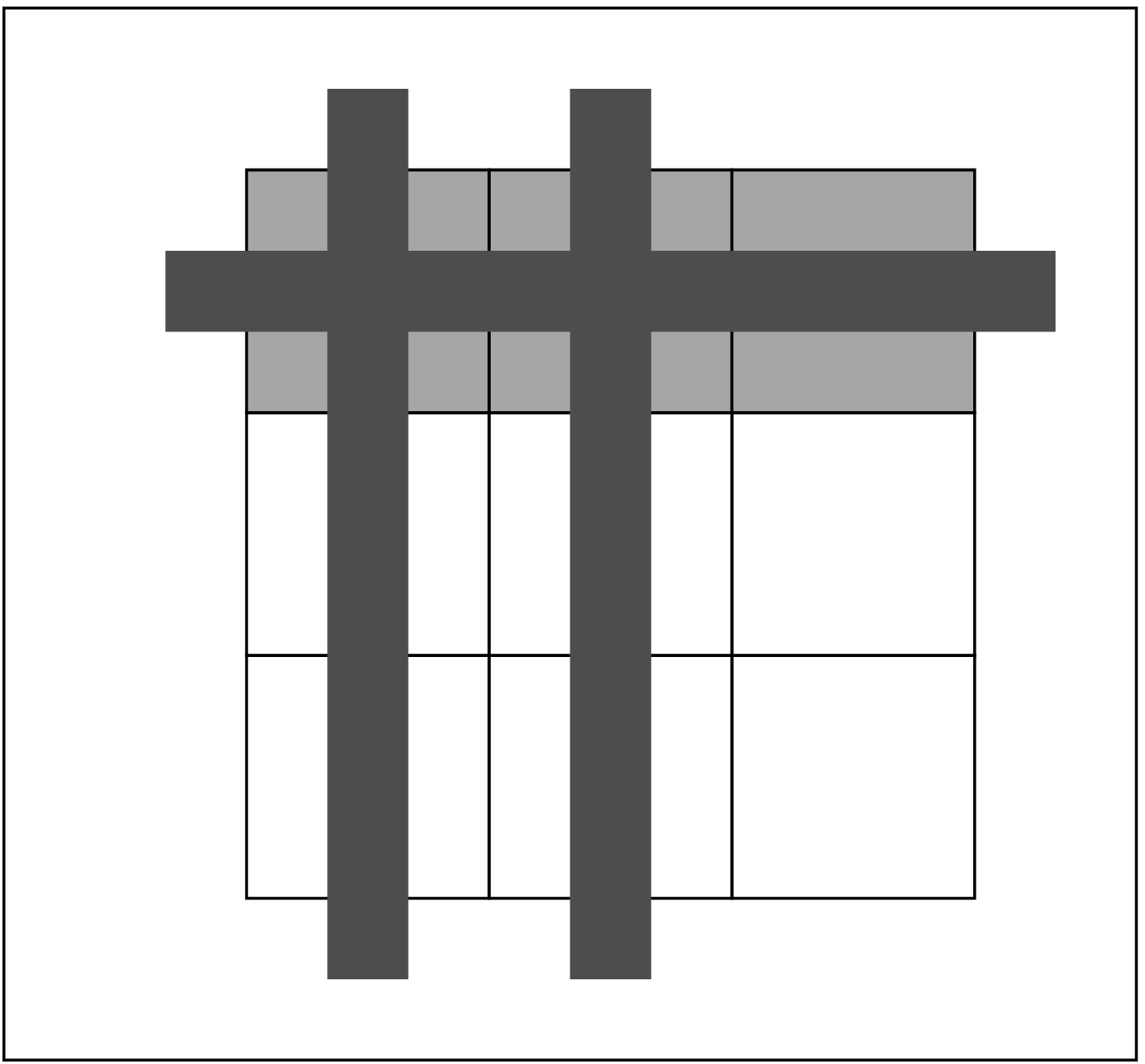}
\includegraphics[width=0.12\linewidth,keepaspectratio]{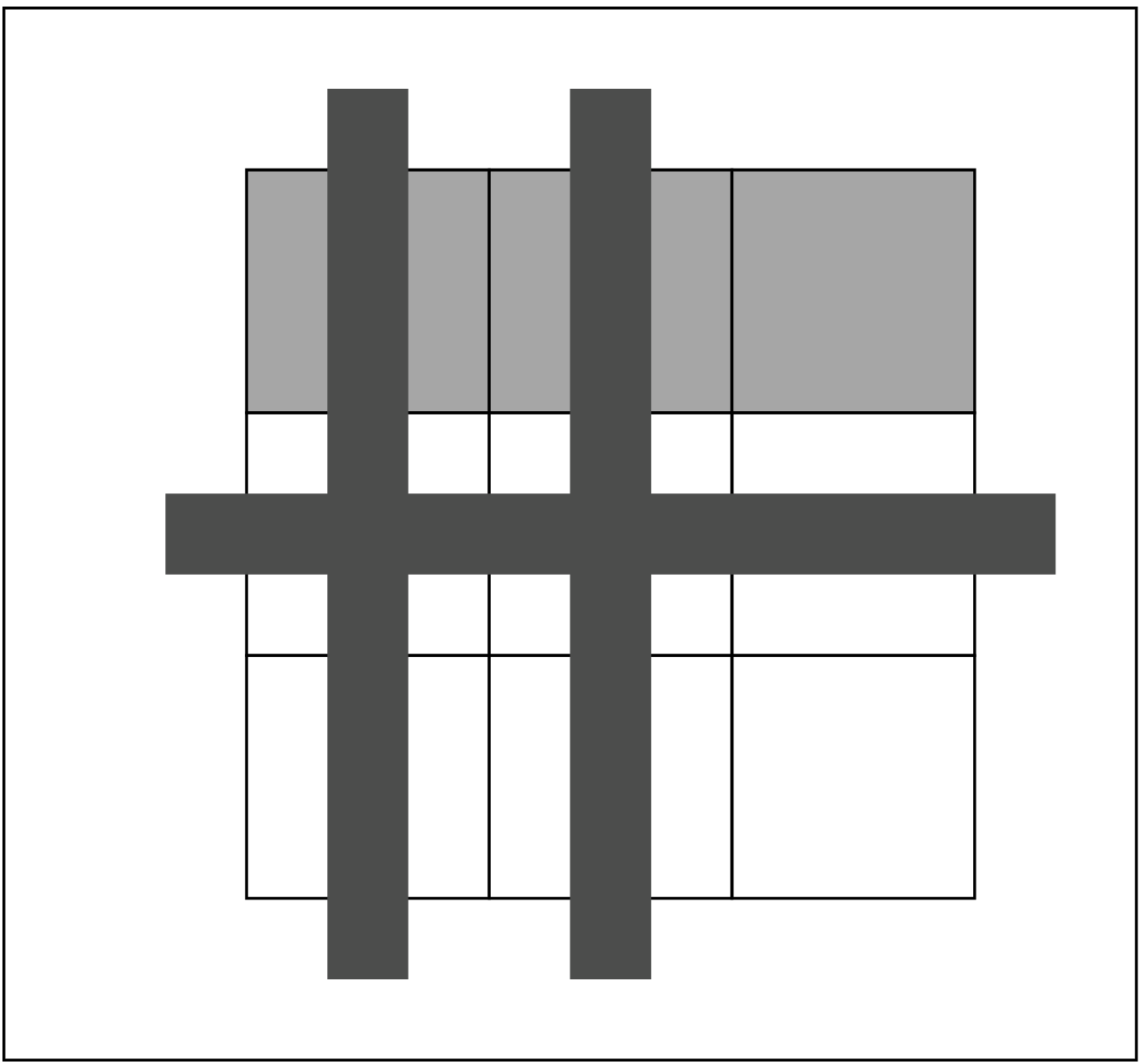}
\includegraphics[width=0.12\linewidth,keepaspectratio]{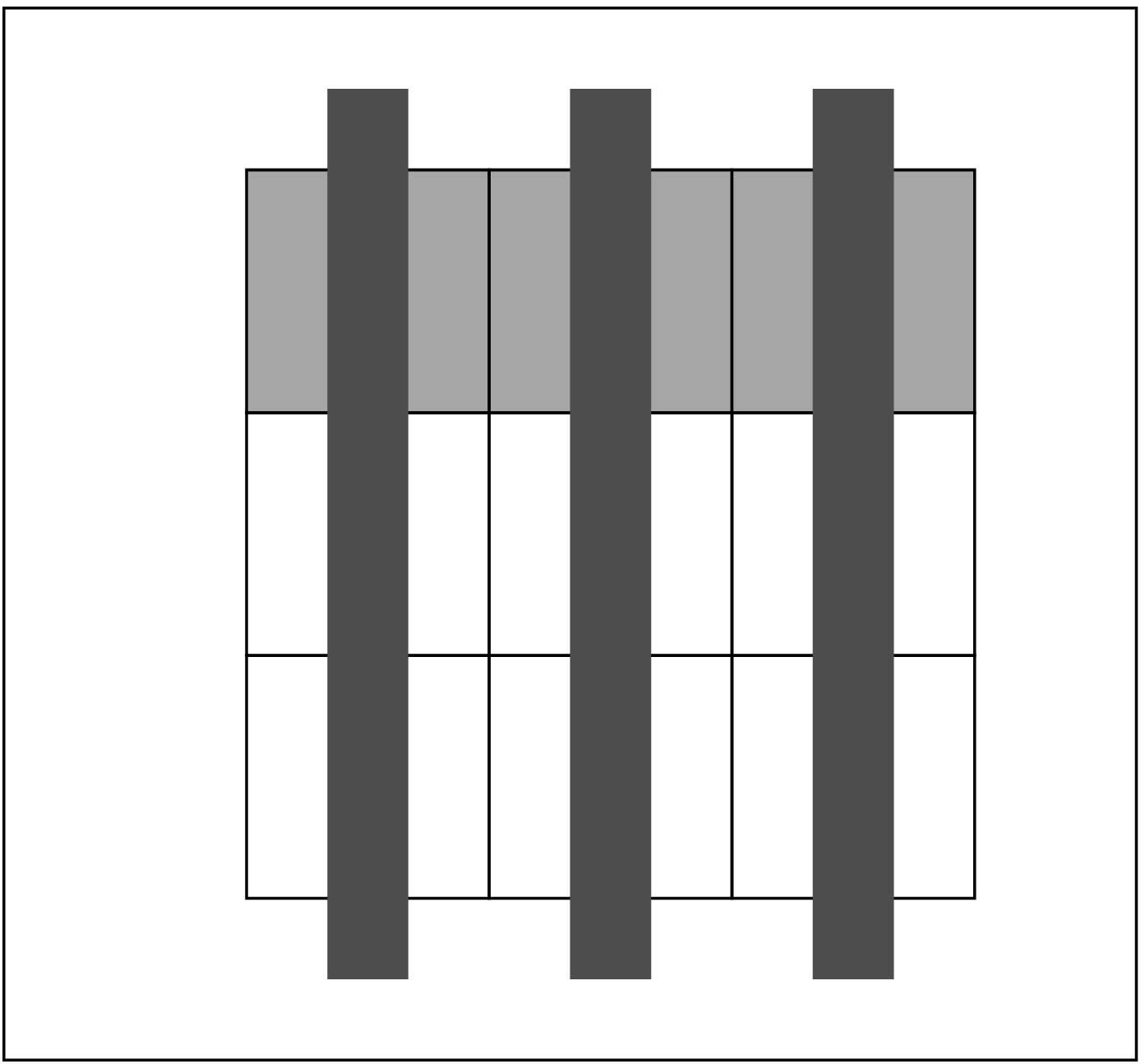}
\includegraphics[width=0.12\linewidth,keepaspectratio]{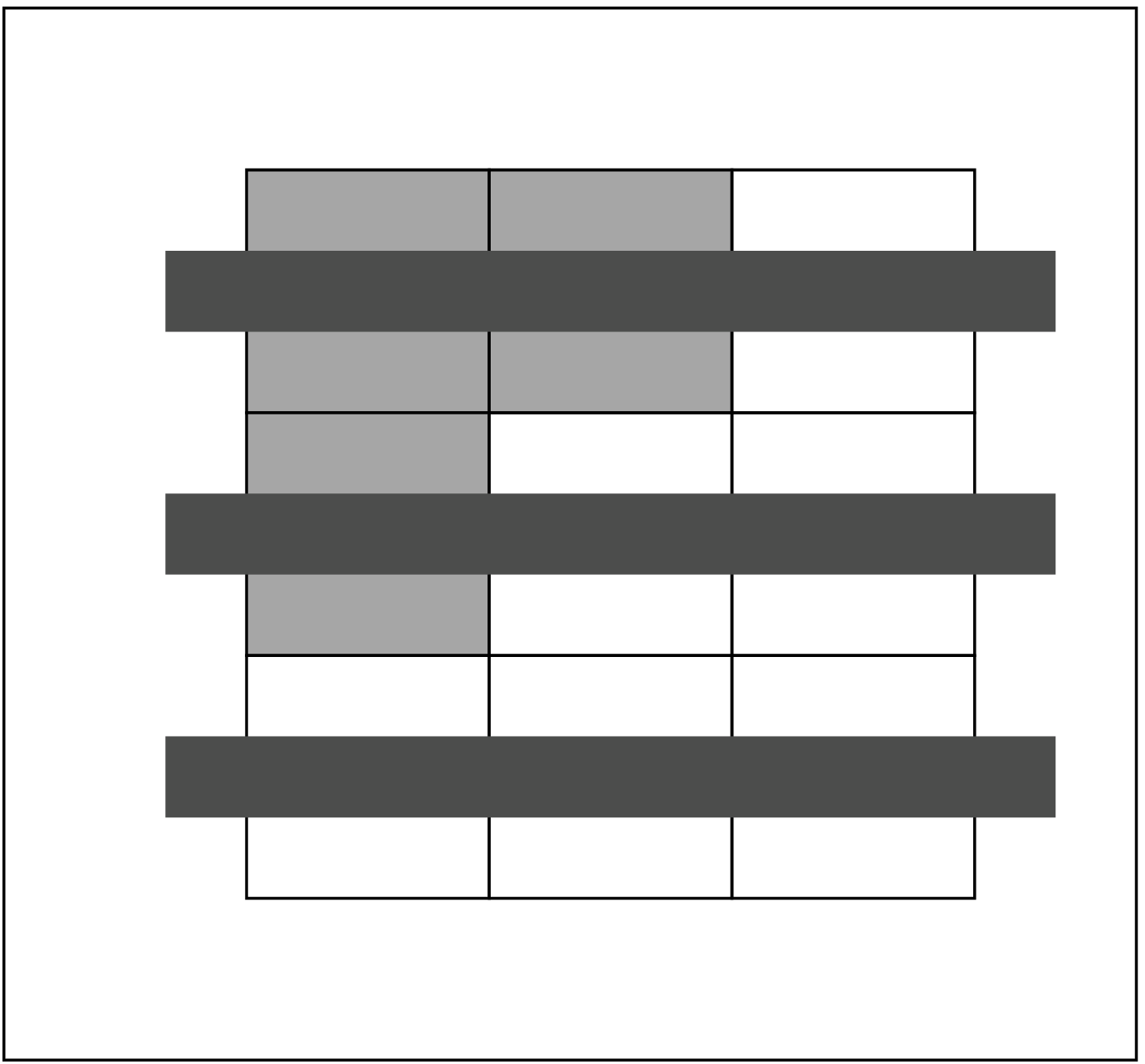}
\includegraphics[width=0.12\linewidth,keepaspectratio]{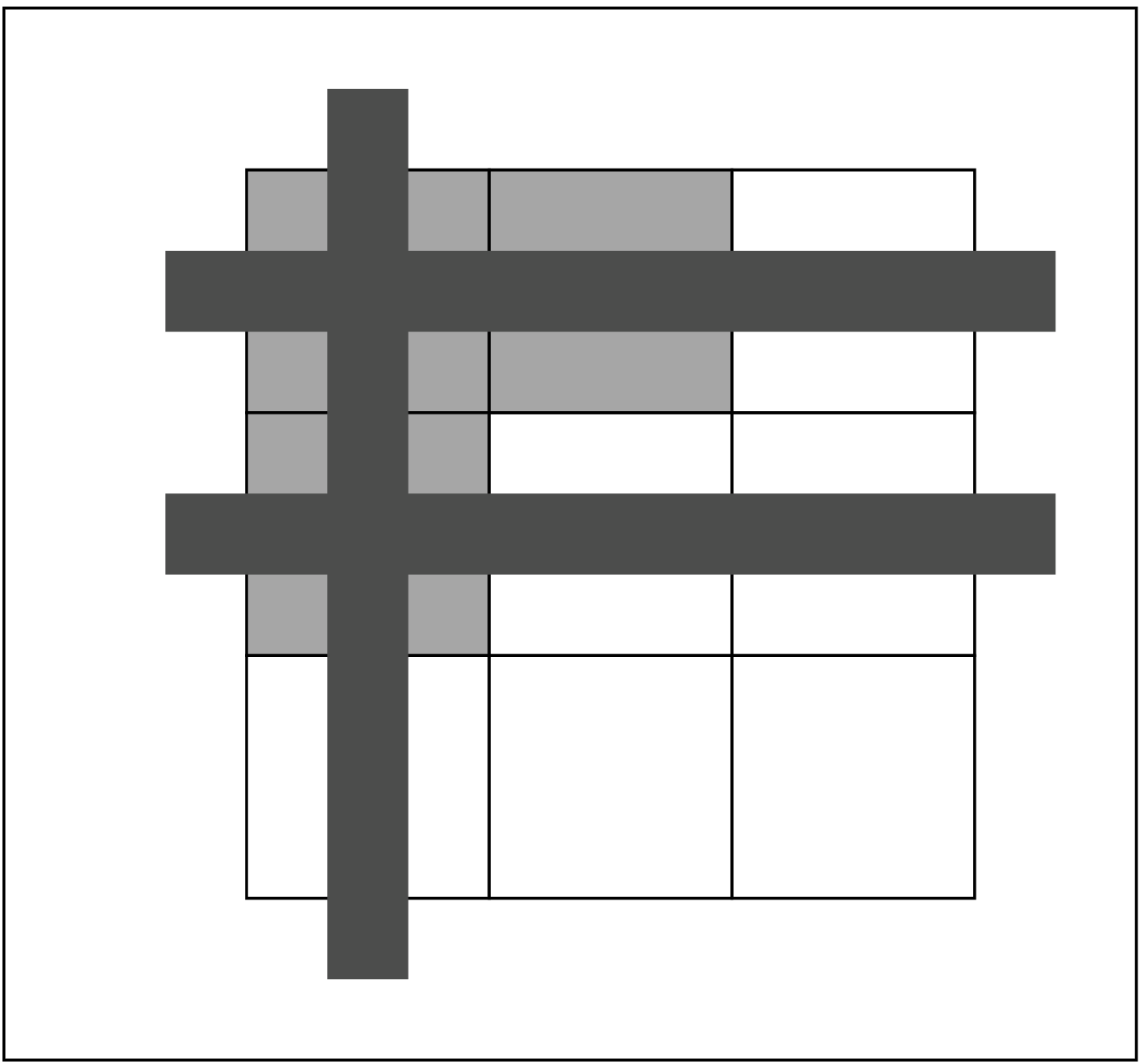}
\includegraphics[width=0.12\linewidth,keepaspectratio]{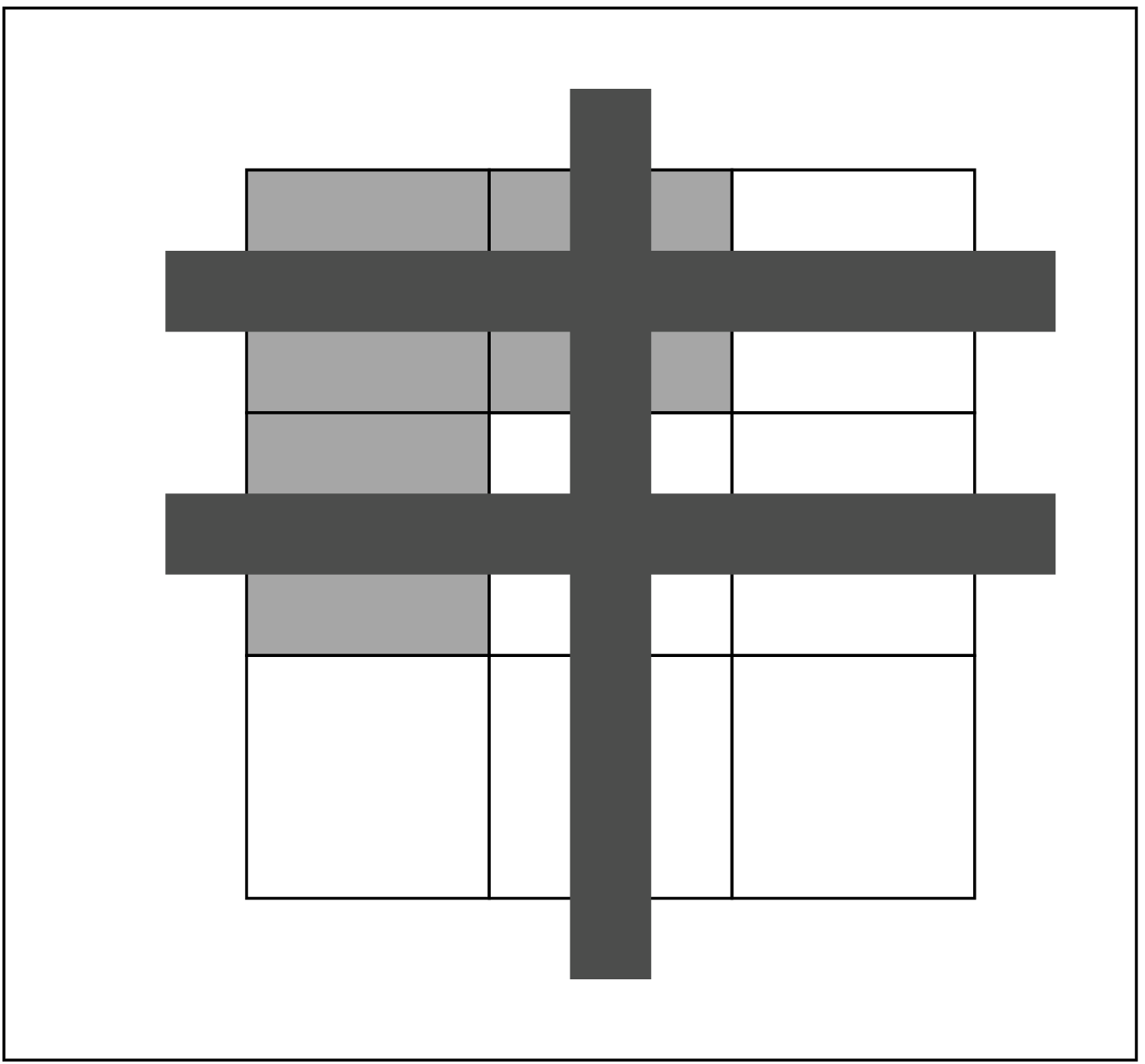}
\includegraphics[width=0.12\linewidth,keepaspectratio]{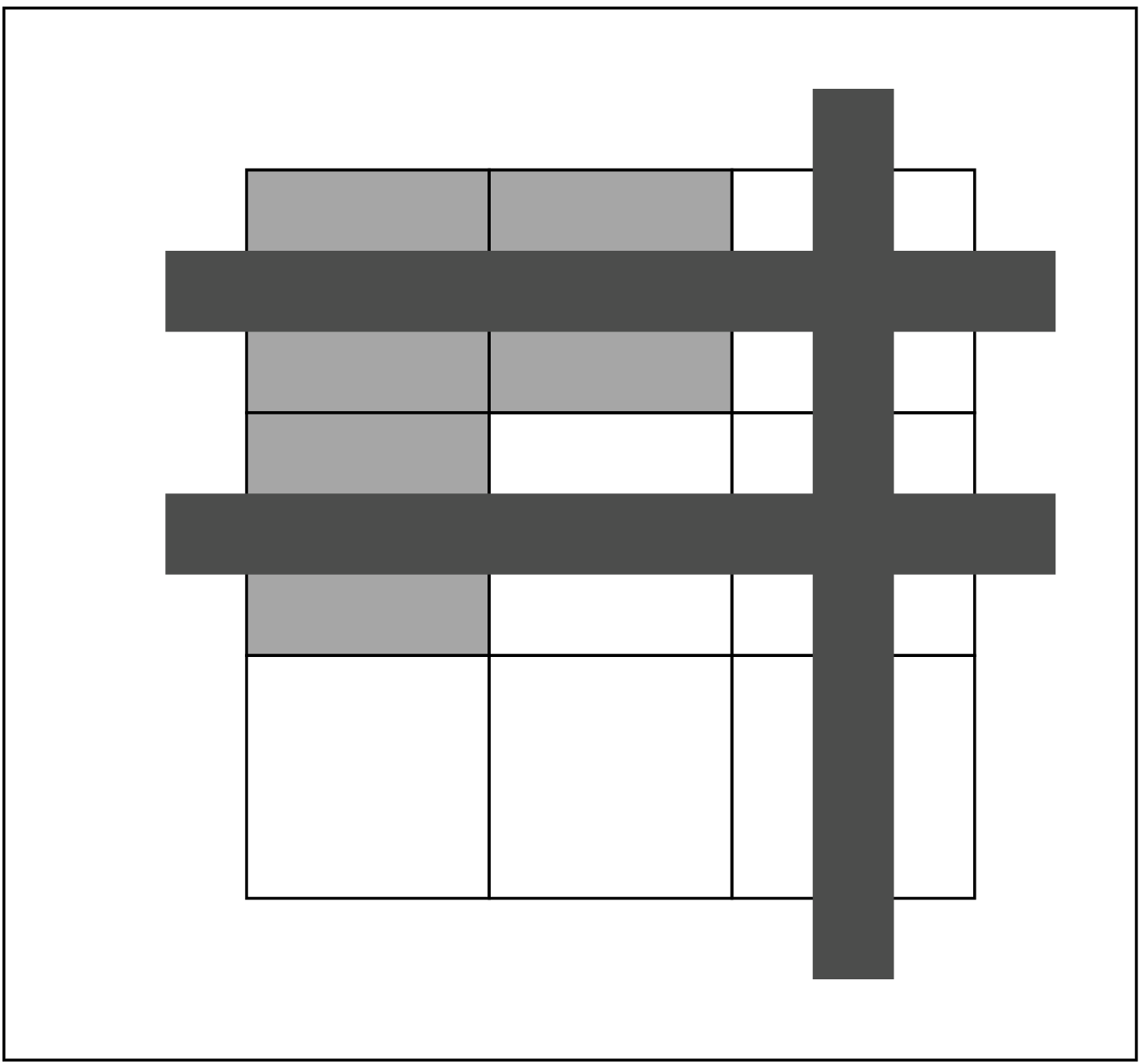}
\includegraphics[width=0.12\linewidth,keepaspectratio]{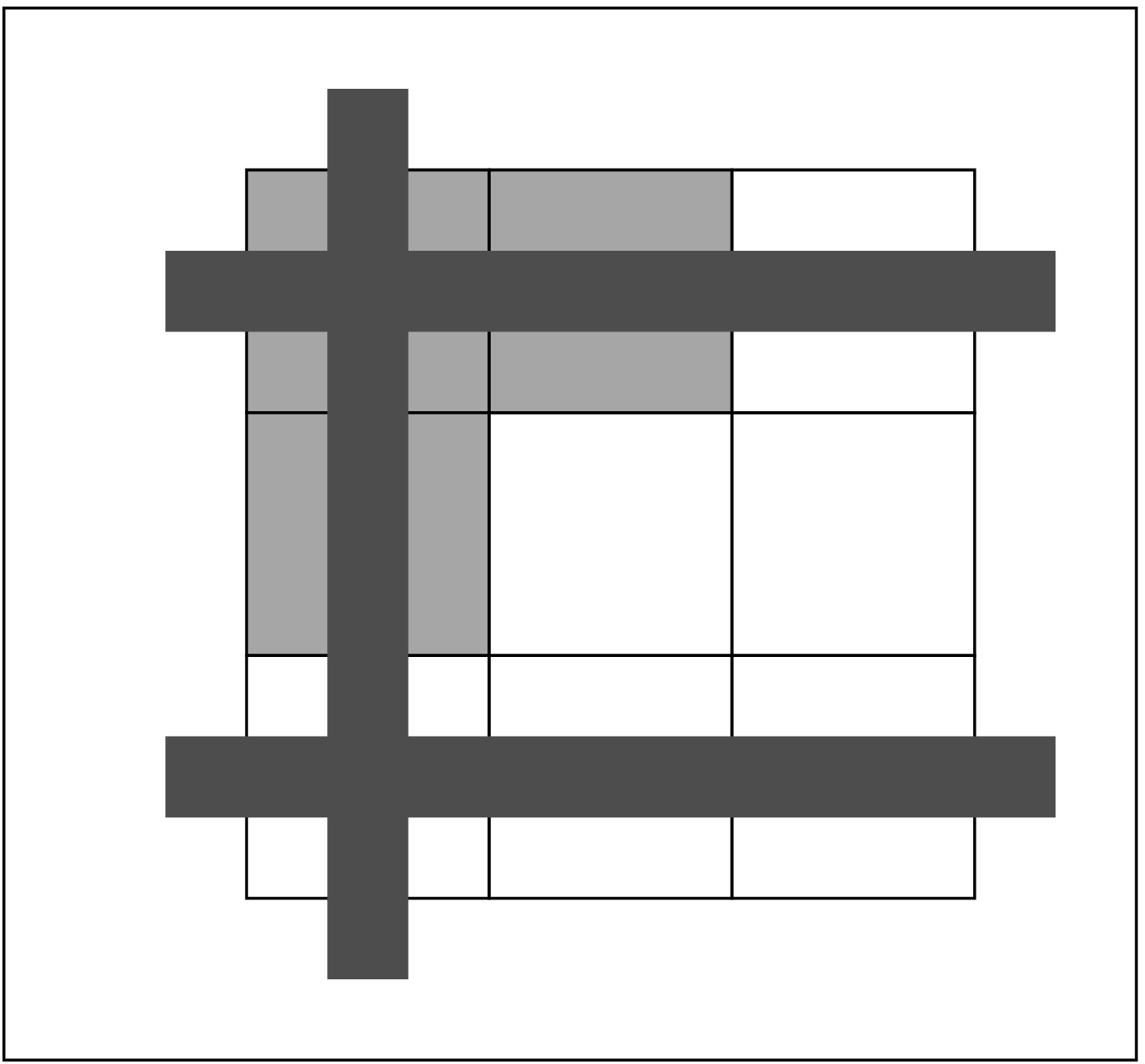}
\includegraphics[width=0.12\linewidth,keepaspectratio]{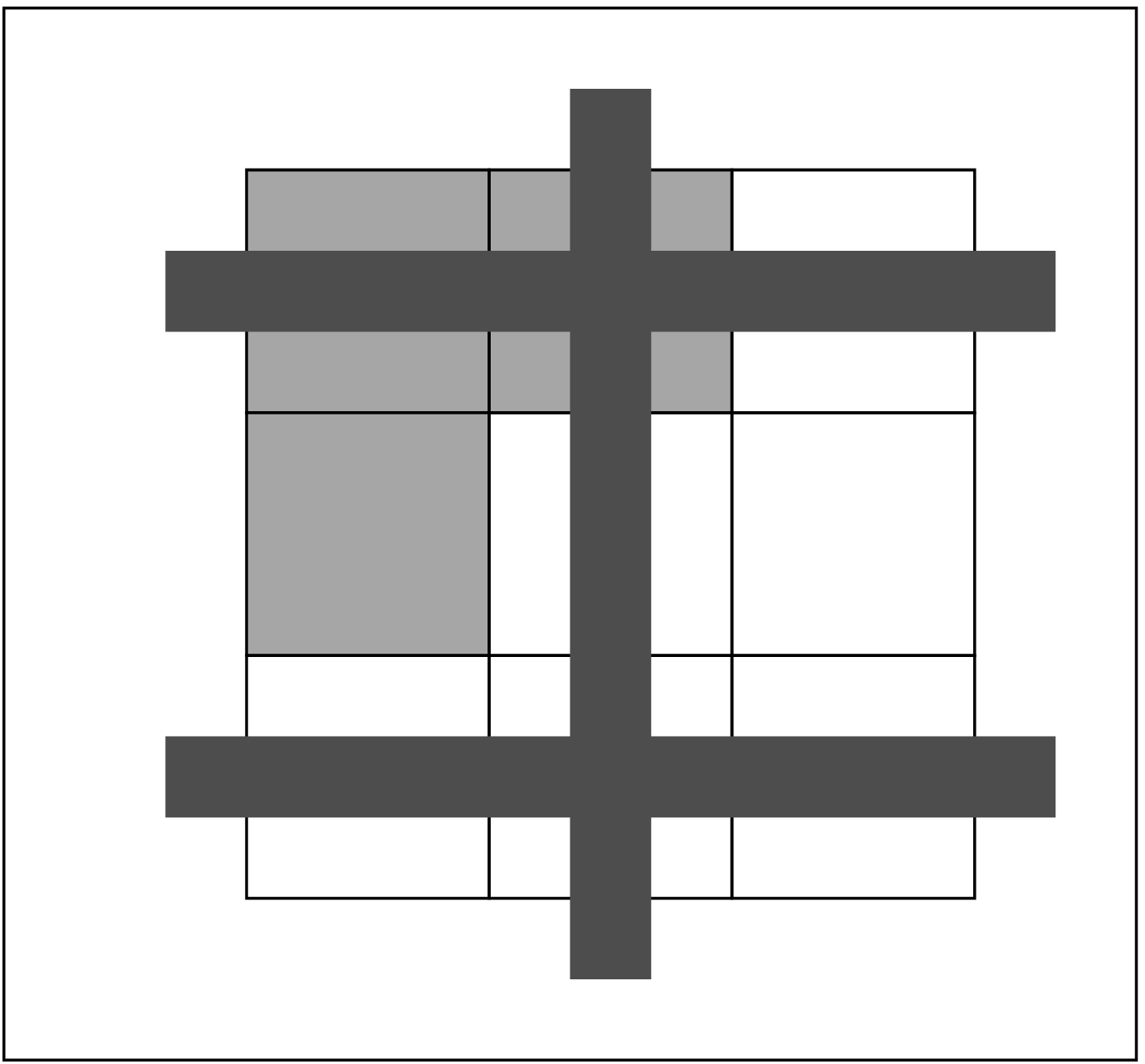}
\includegraphics[width=0.12\linewidth,keepaspectratio]{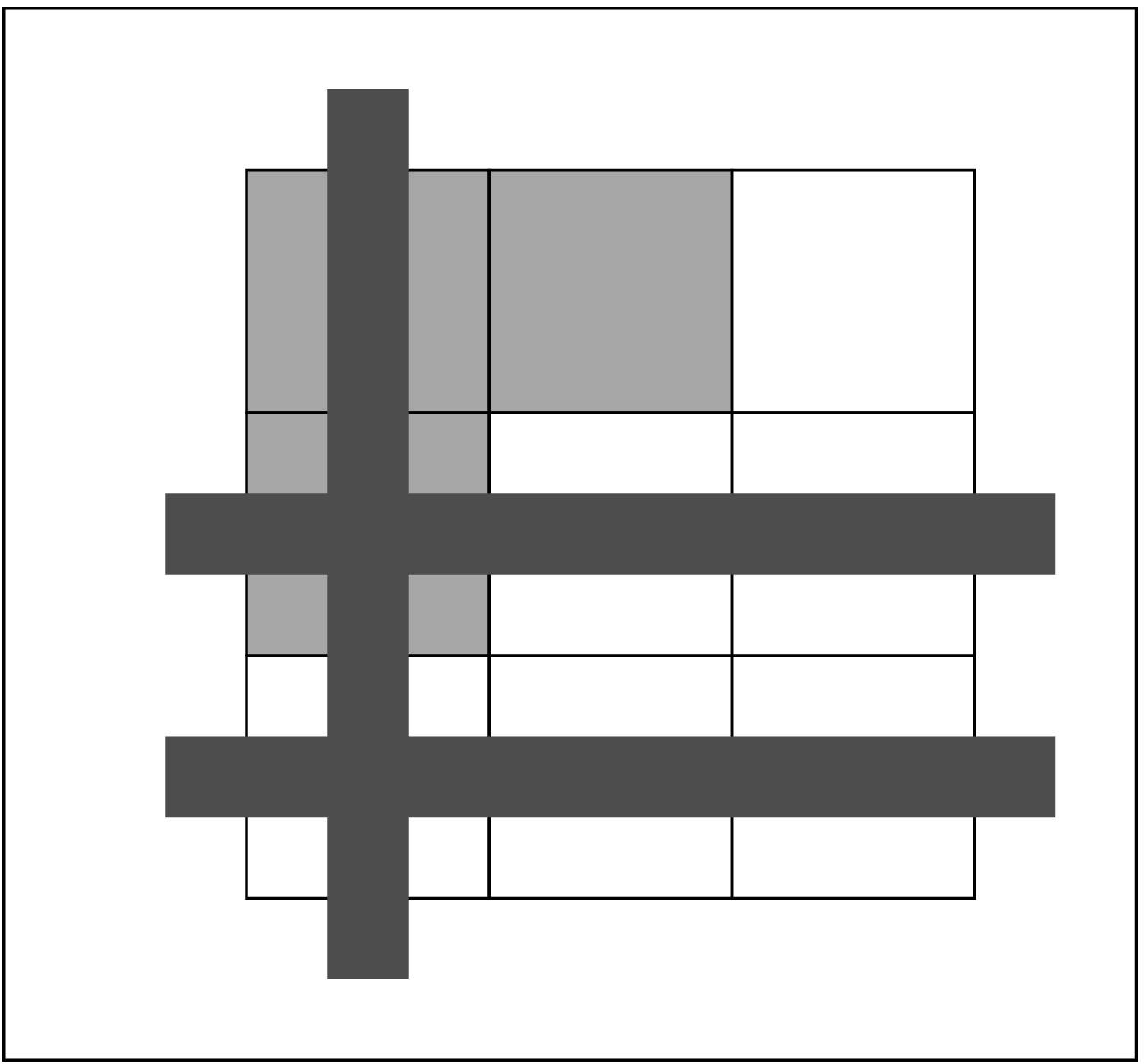}
\includegraphics[width=0.12\linewidth,keepaspectratio]{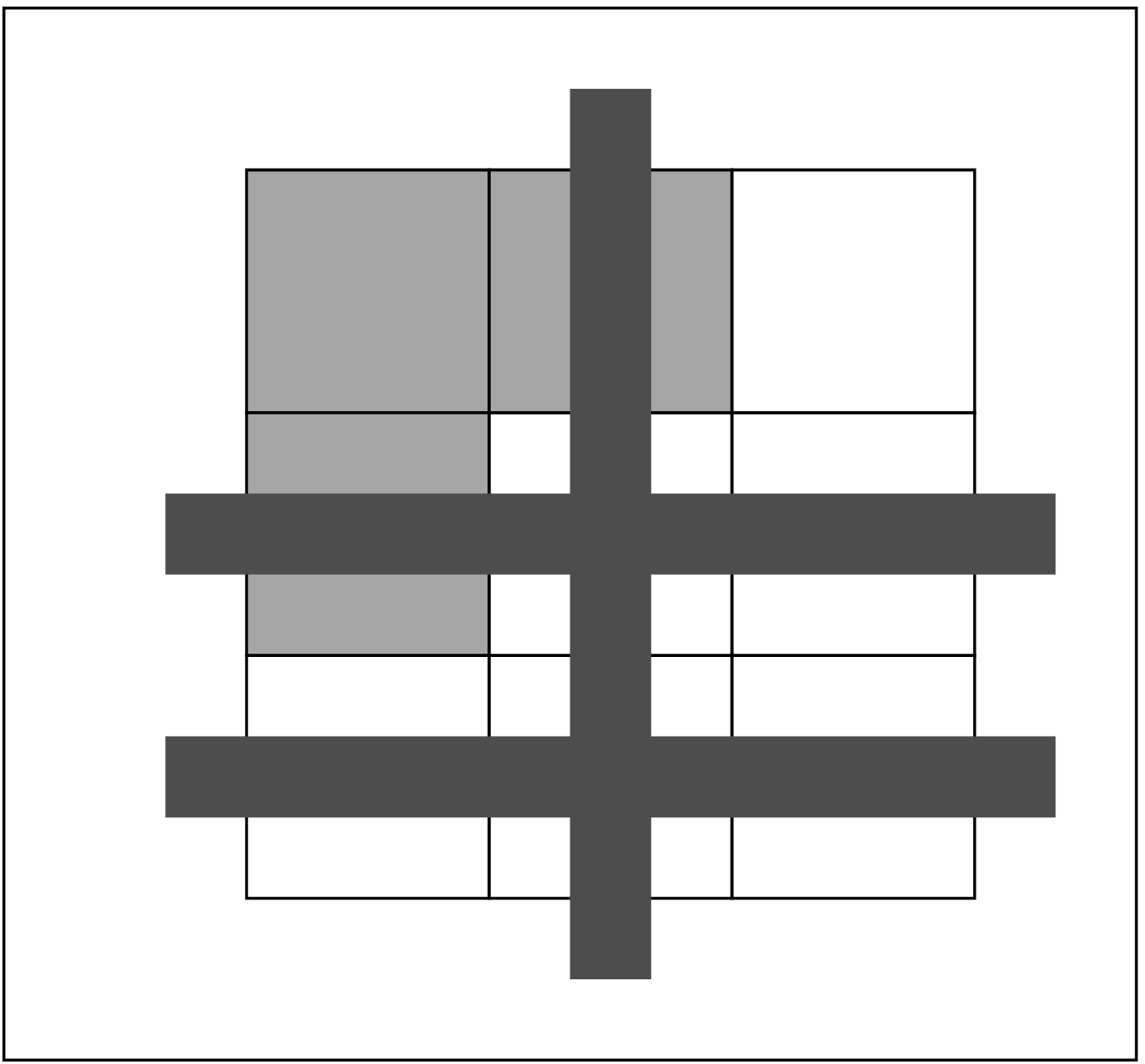}
\includegraphics[width=0.12\linewidth,keepaspectratio]{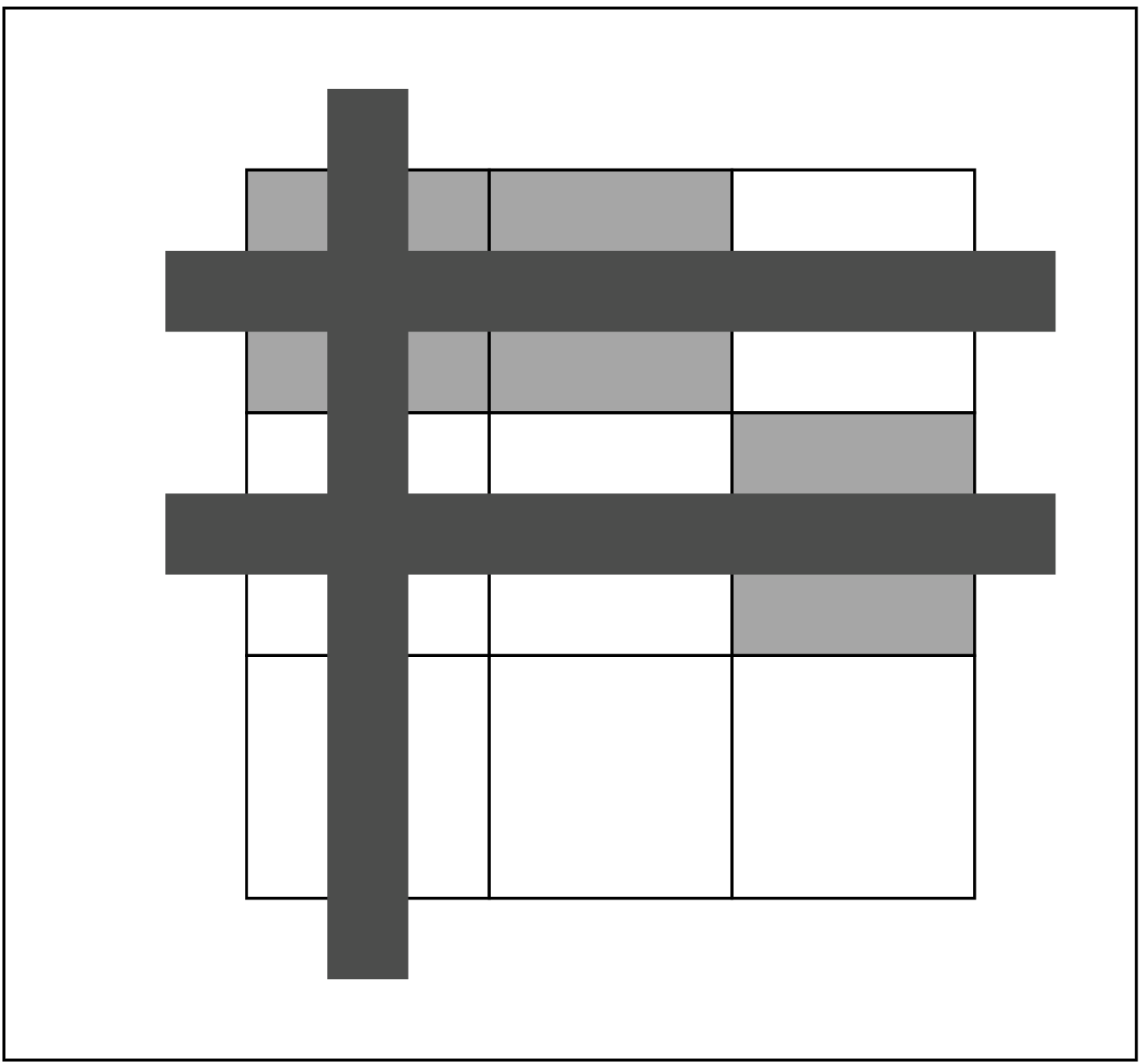}
\includegraphics[width=0.12\linewidth,keepaspectratio]{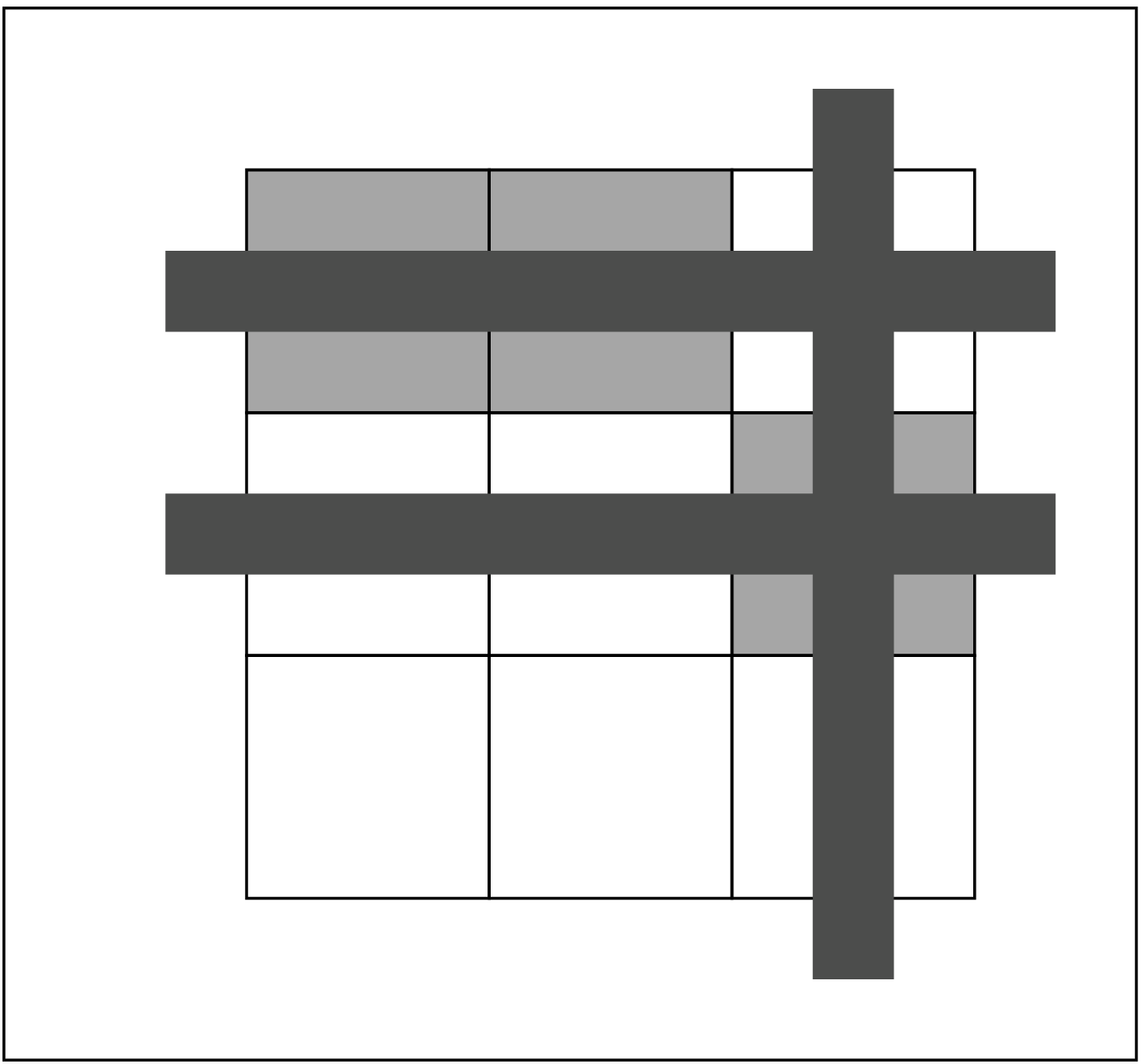}
\includegraphics[width=0.12\linewidth,keepaspectratio]{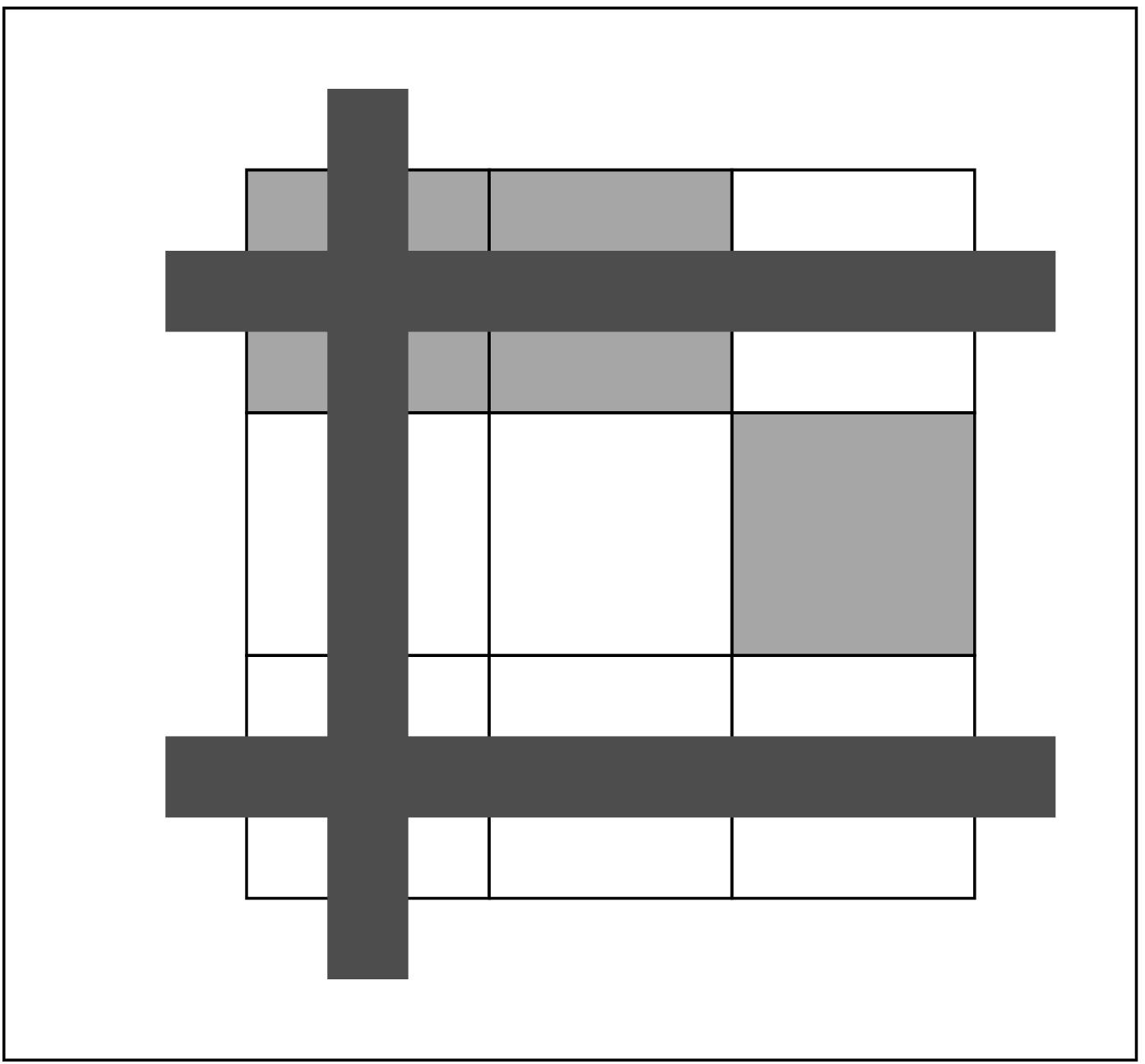}
\includegraphics[width=0.12\linewidth,keepaspectratio]{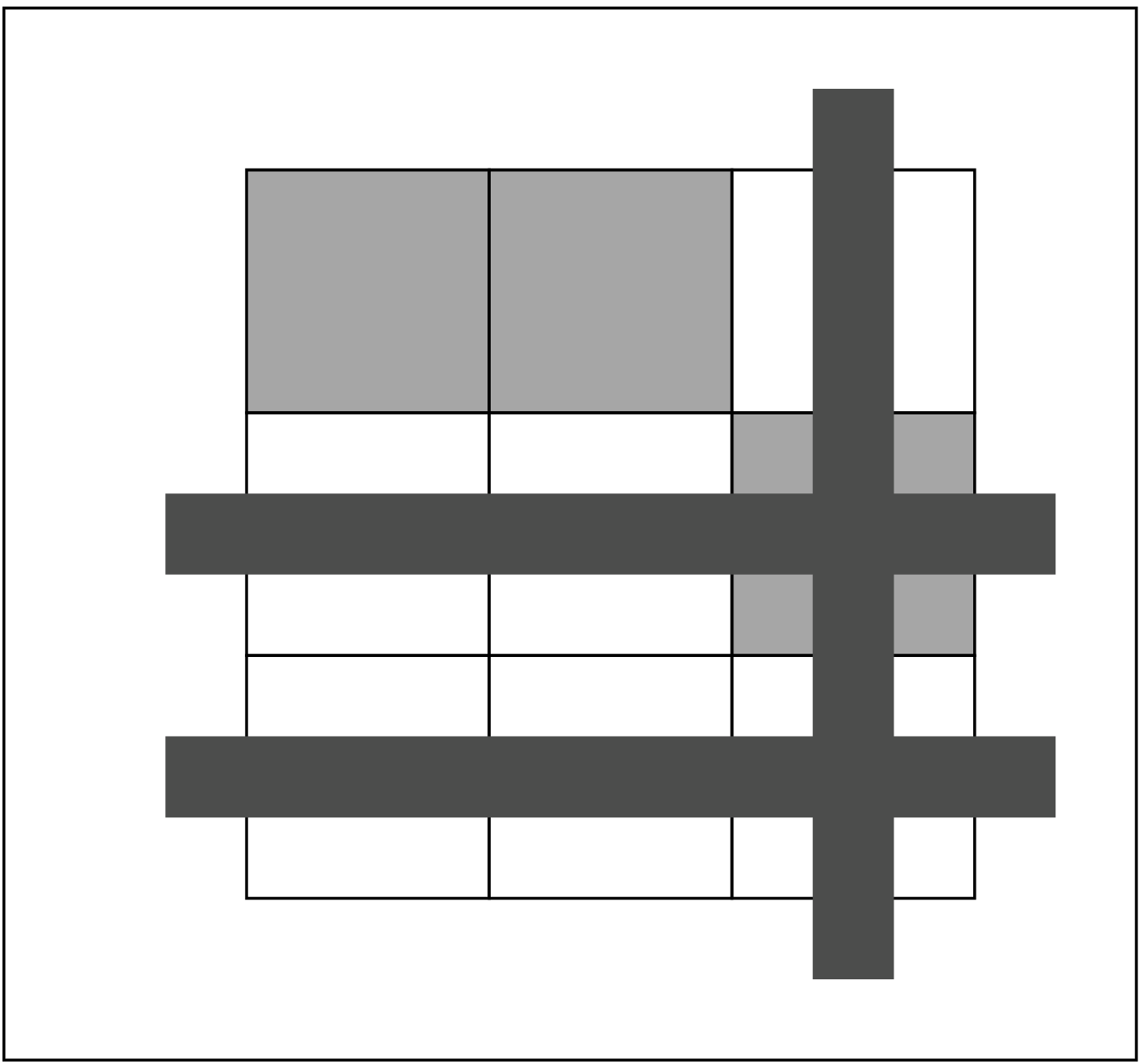}
\includegraphics[width=0.12\linewidth,keepaspectratio]{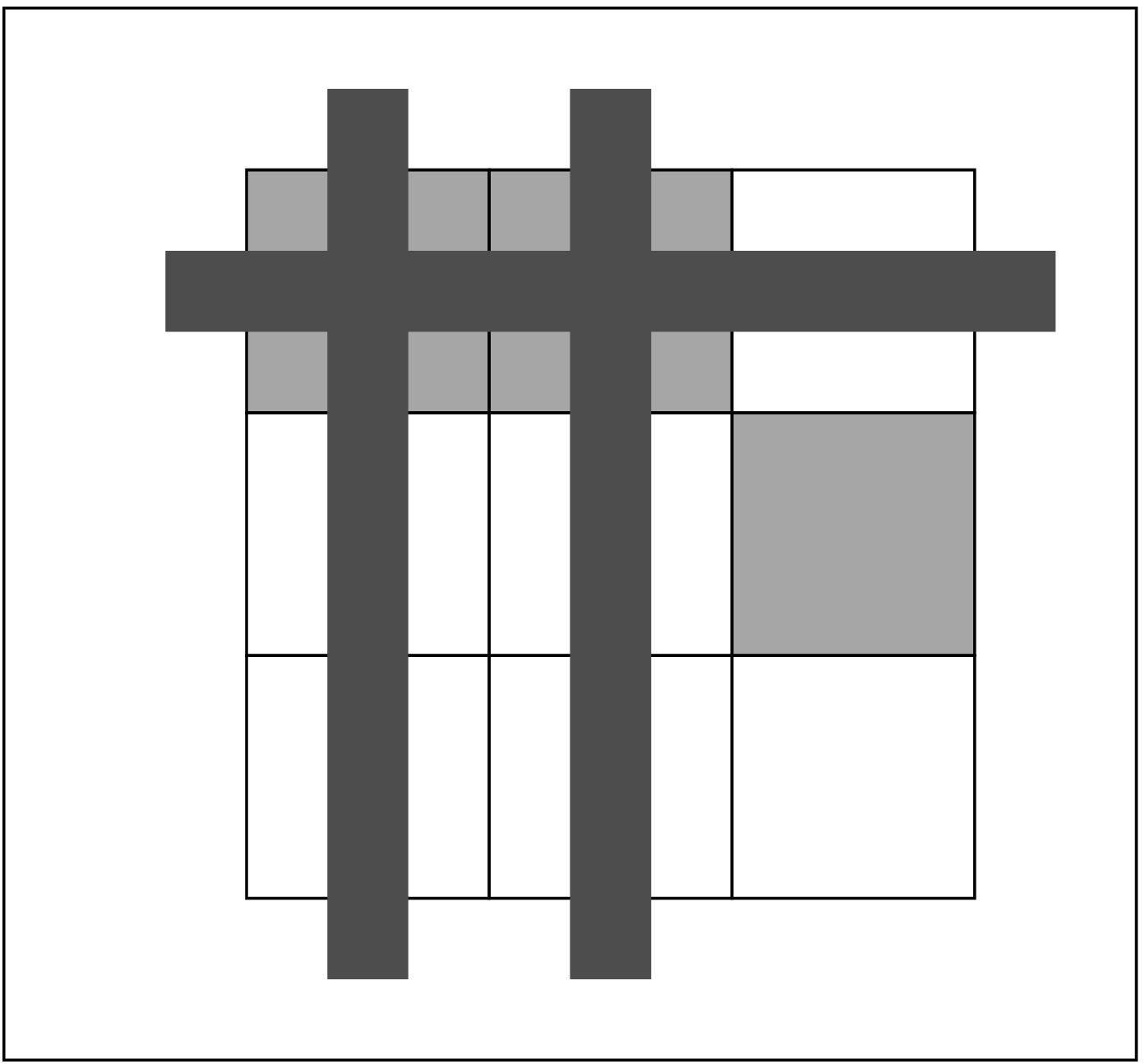}
\includegraphics[width=0.12\linewidth,keepaspectratio]{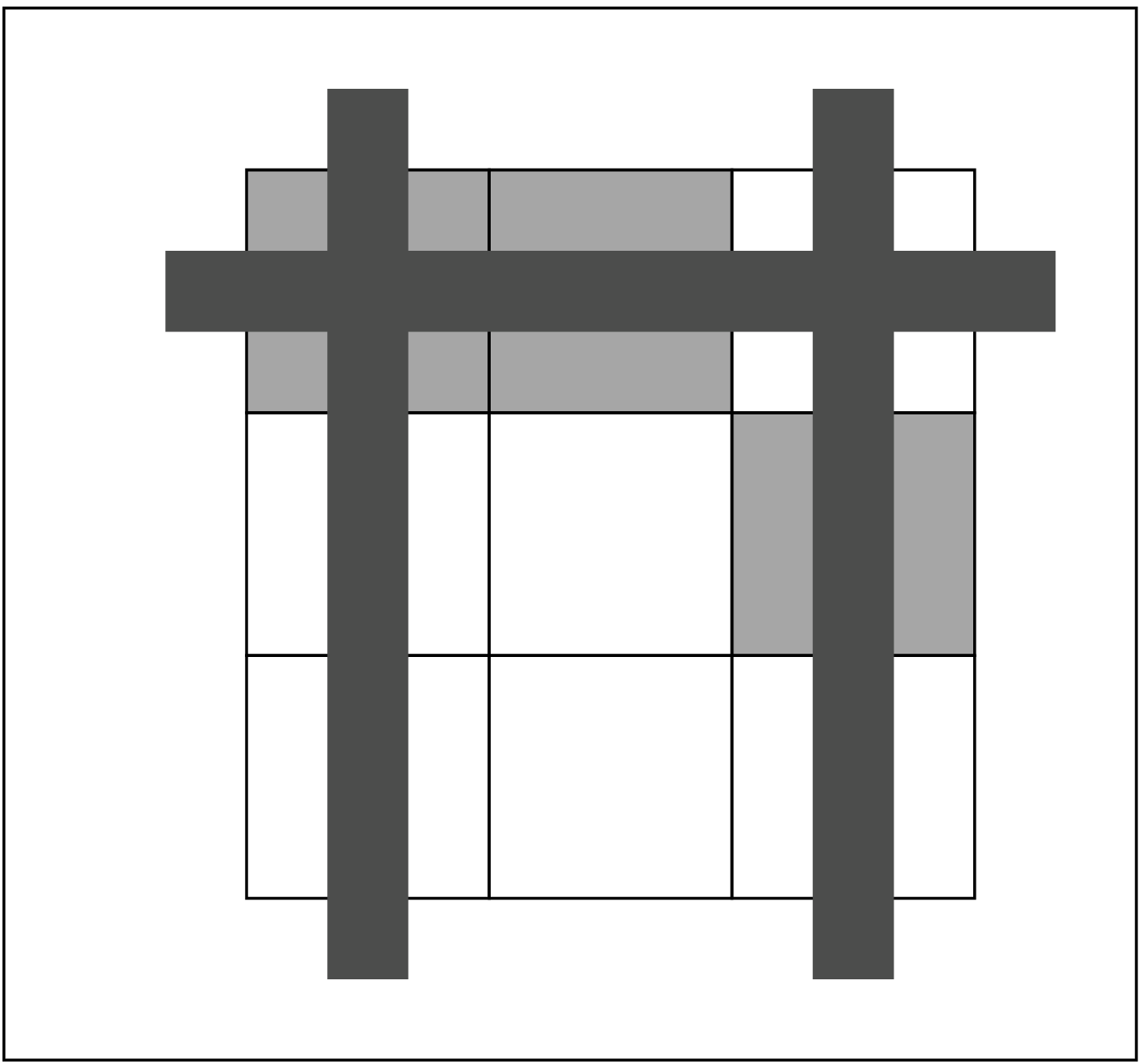}
\includegraphics[width=0.12\linewidth,keepaspectratio]{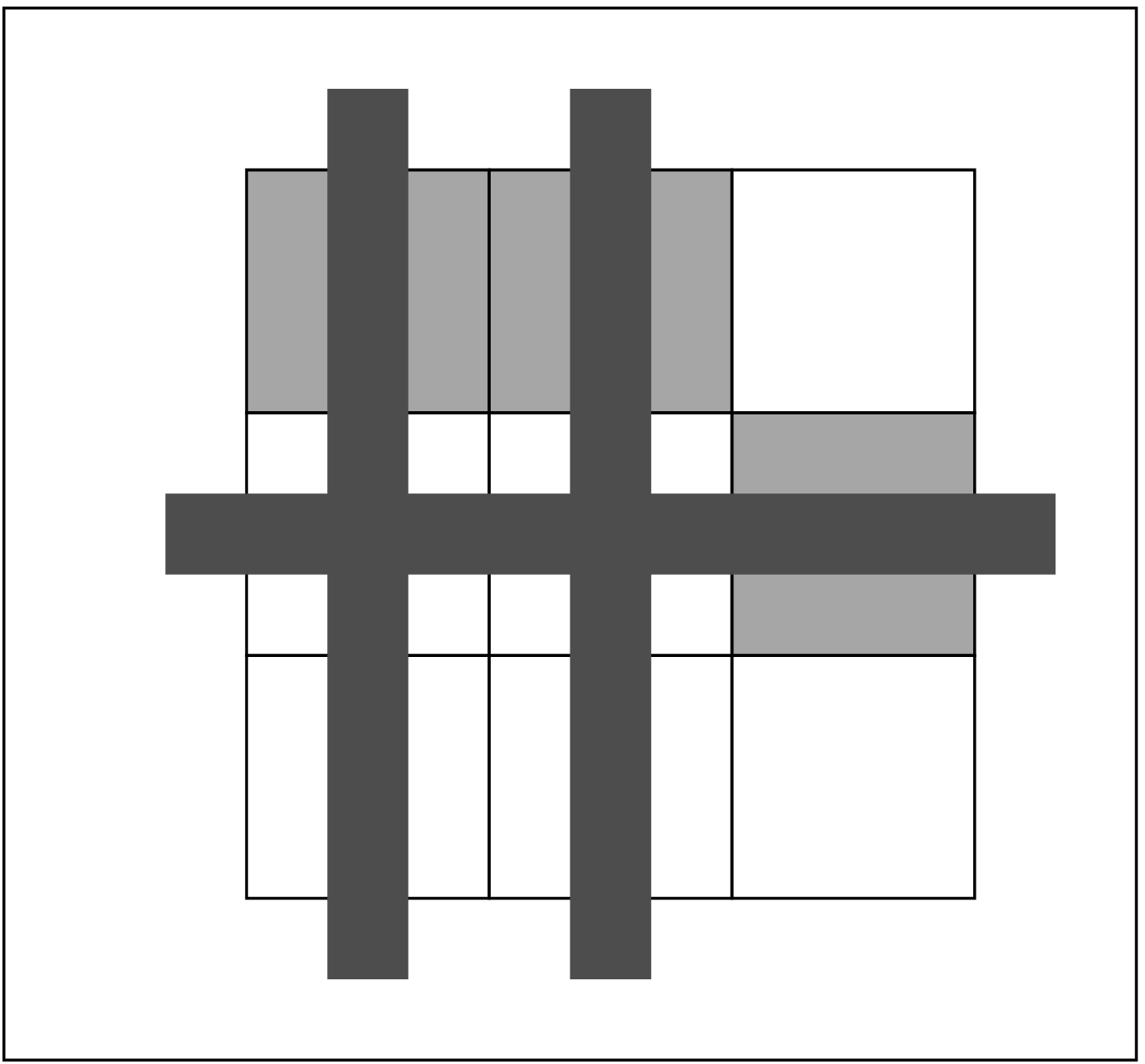}
\includegraphics[width=0.12\linewidth,keepaspectratio]{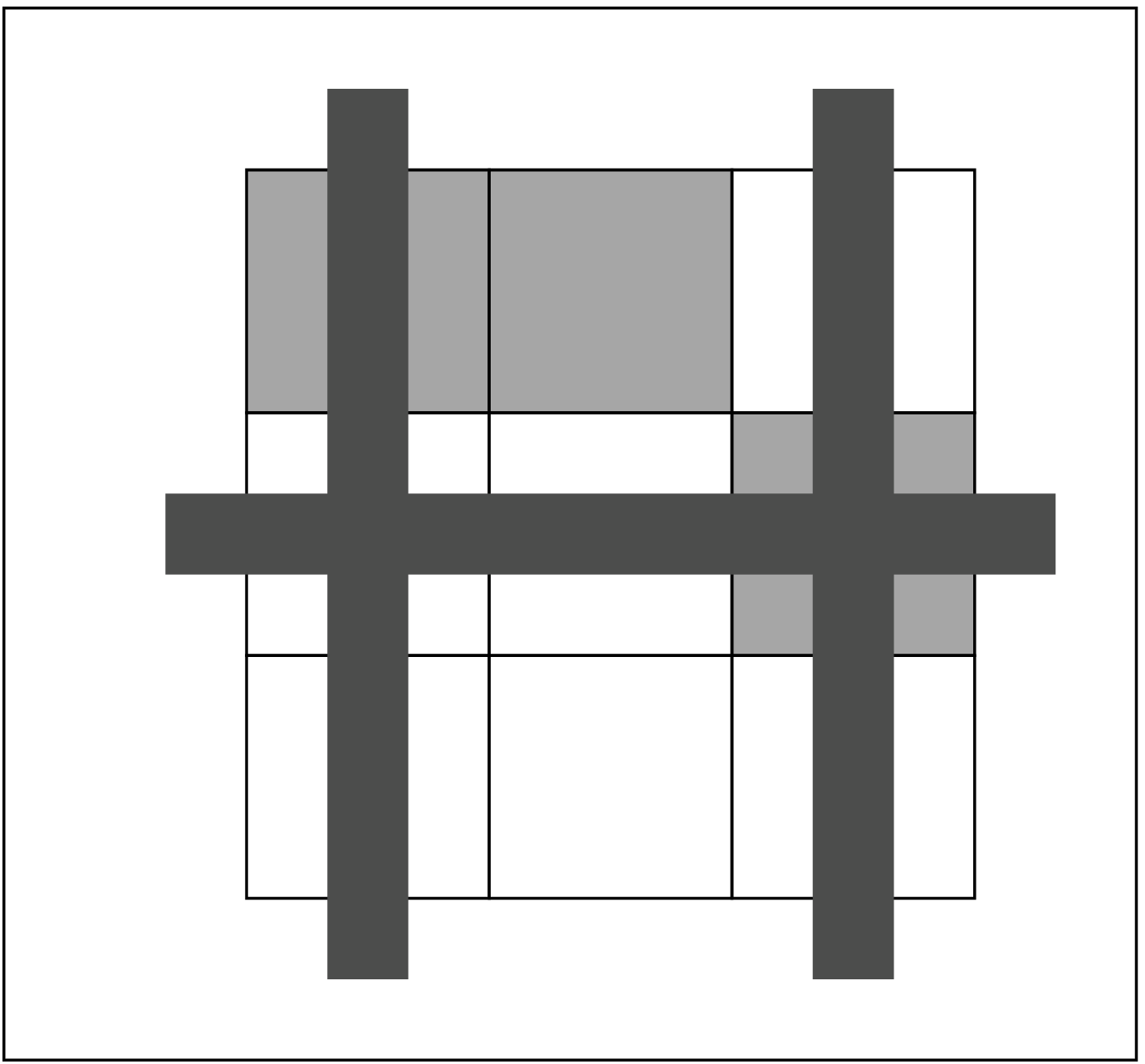}
\includegraphics[width=0.12\linewidth,keepaspectratio]{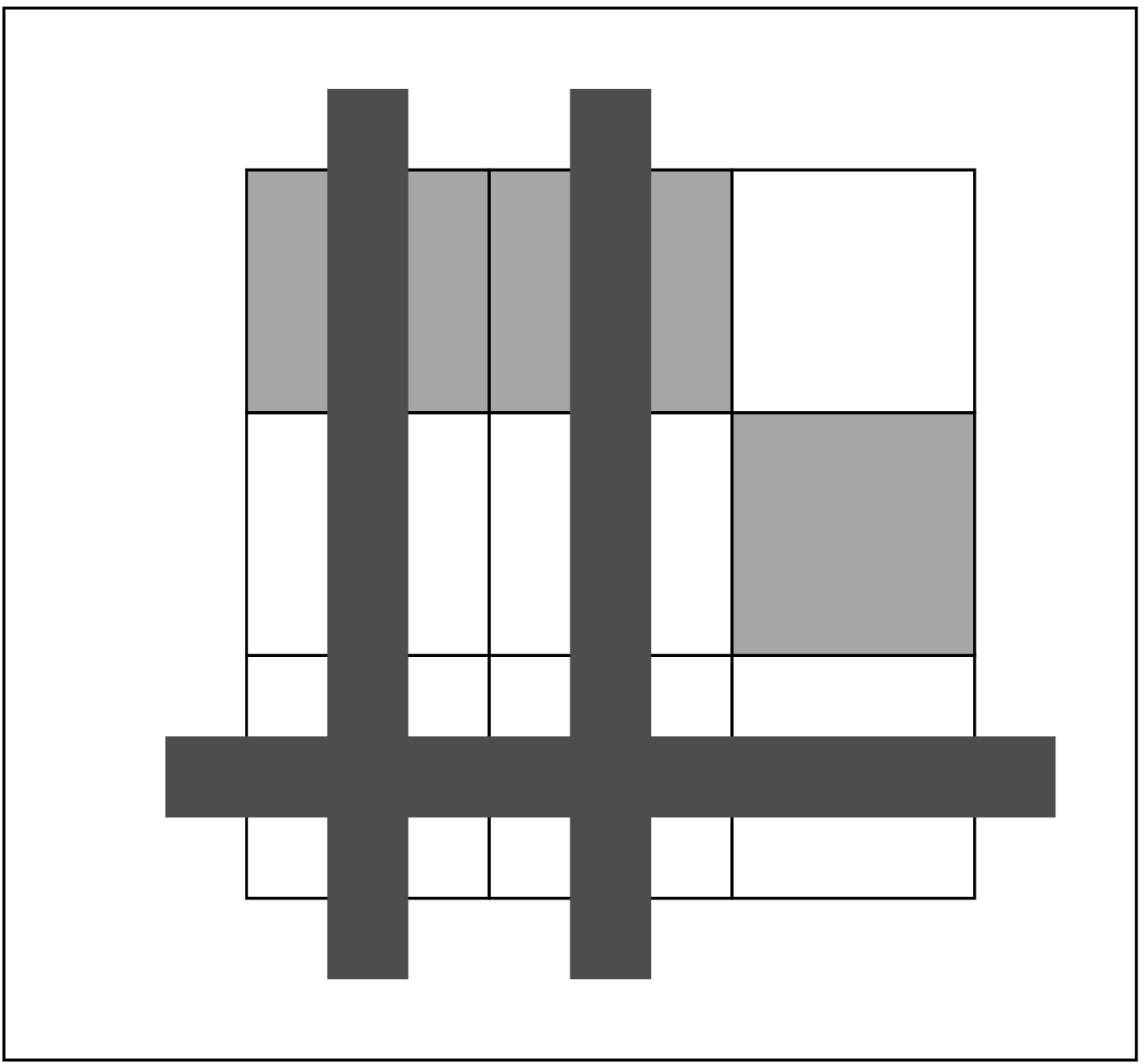}
\includegraphics[width=0.12\linewidth,keepaspectratio]{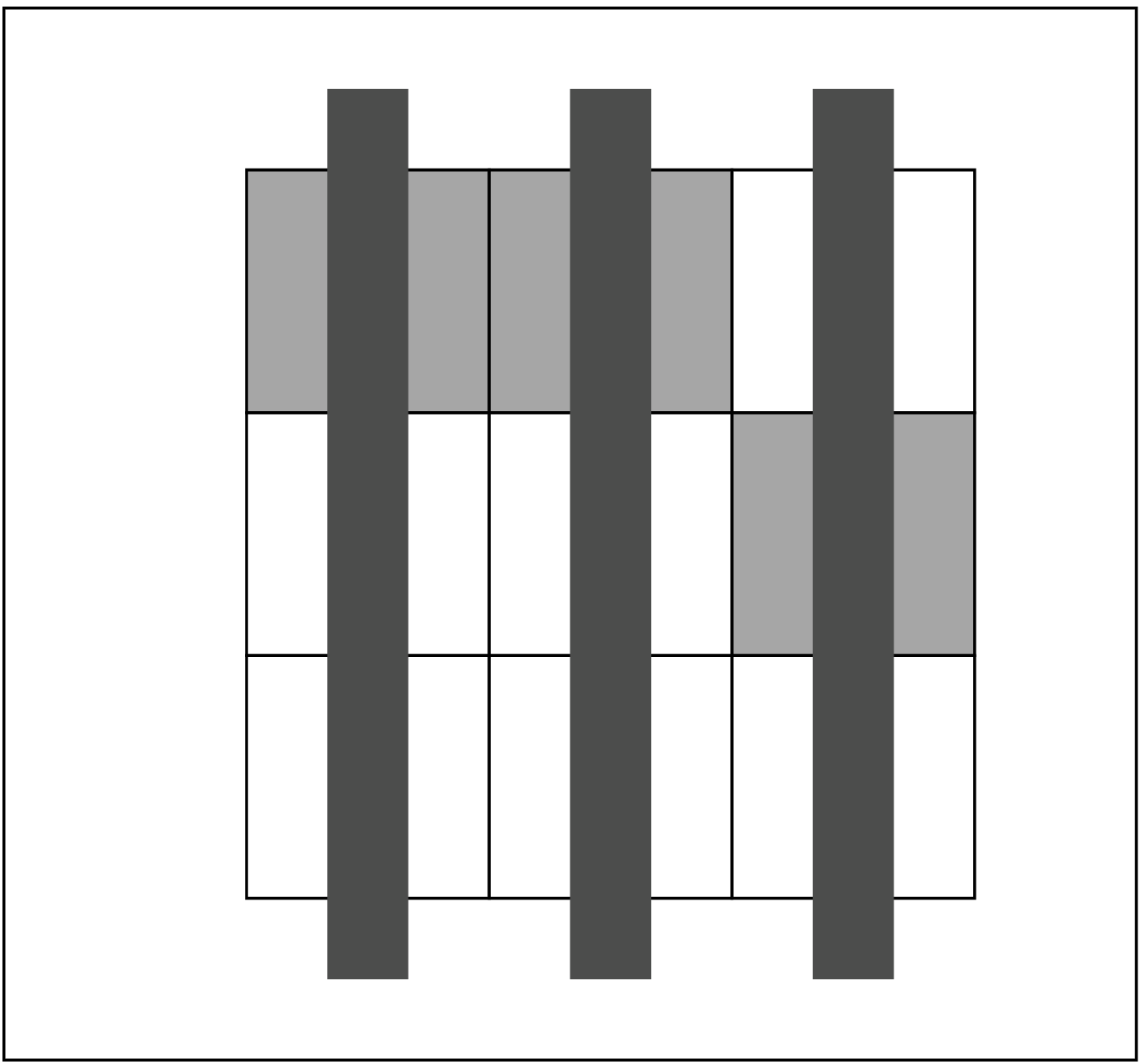}
\includegraphics[width=0.12\linewidth,keepaspectratio]{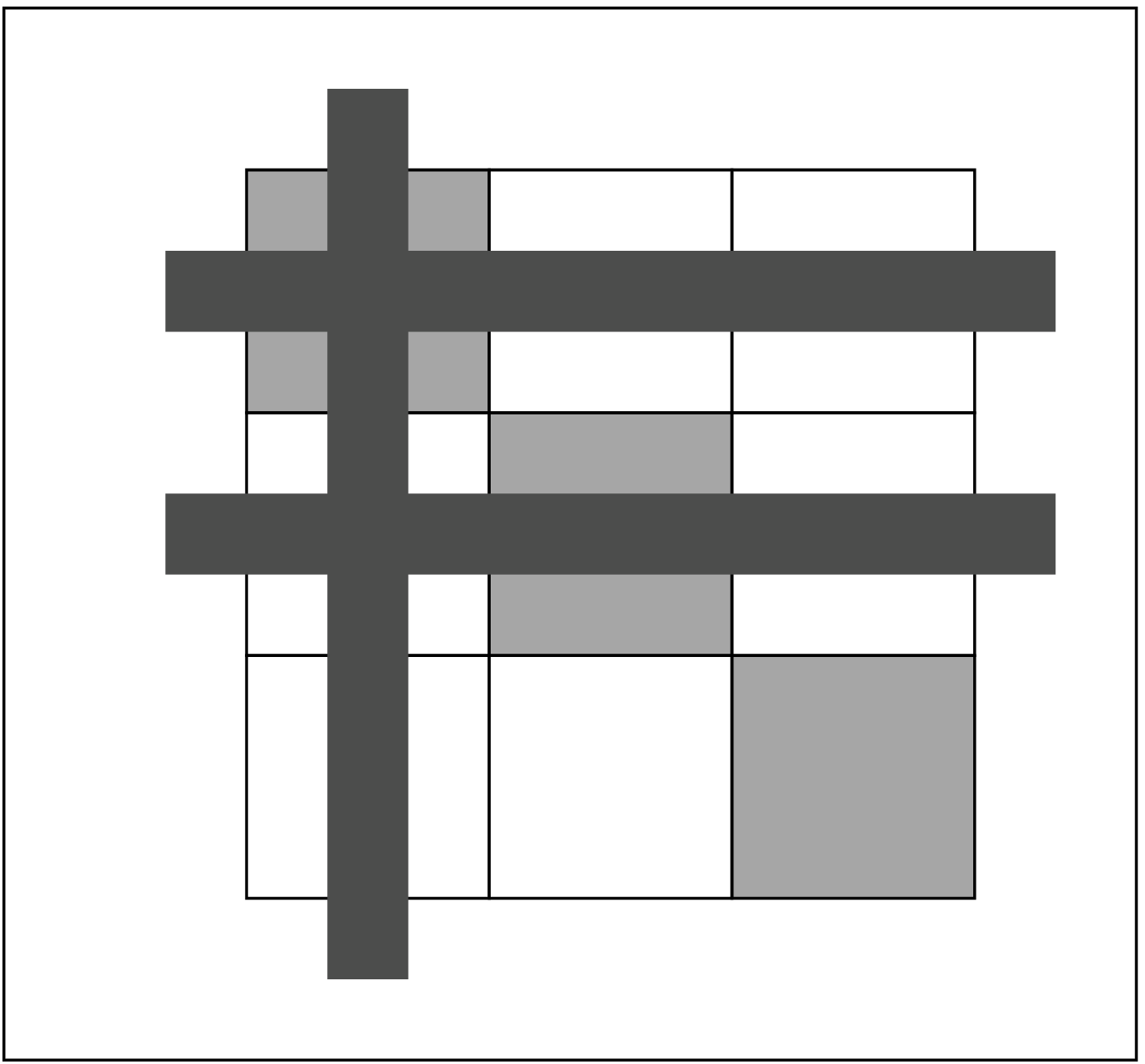}
\includegraphics[width=0.12\linewidth,keepaspectratio]{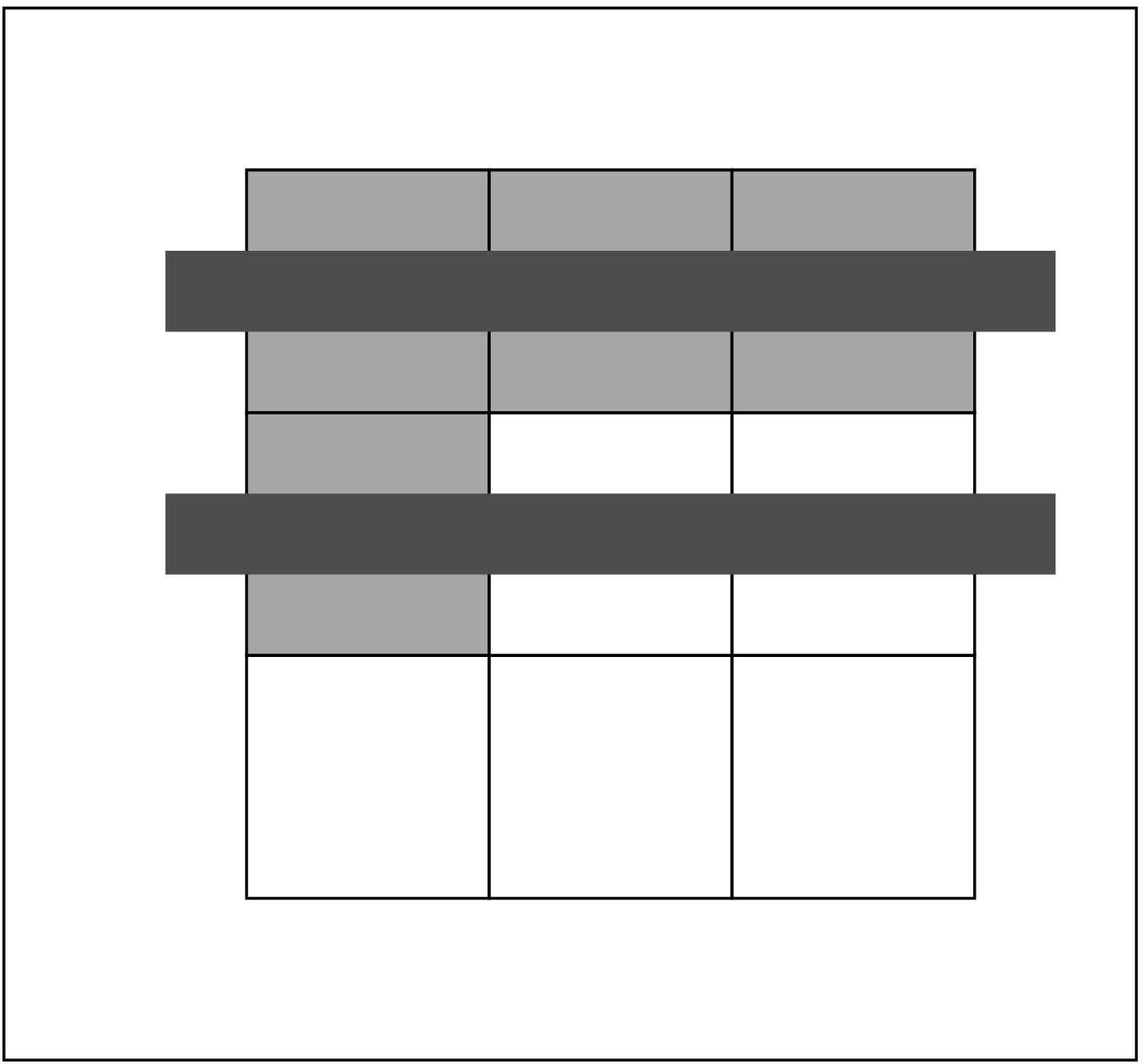}
\includegraphics[width=0.12\linewidth,keepaspectratio]{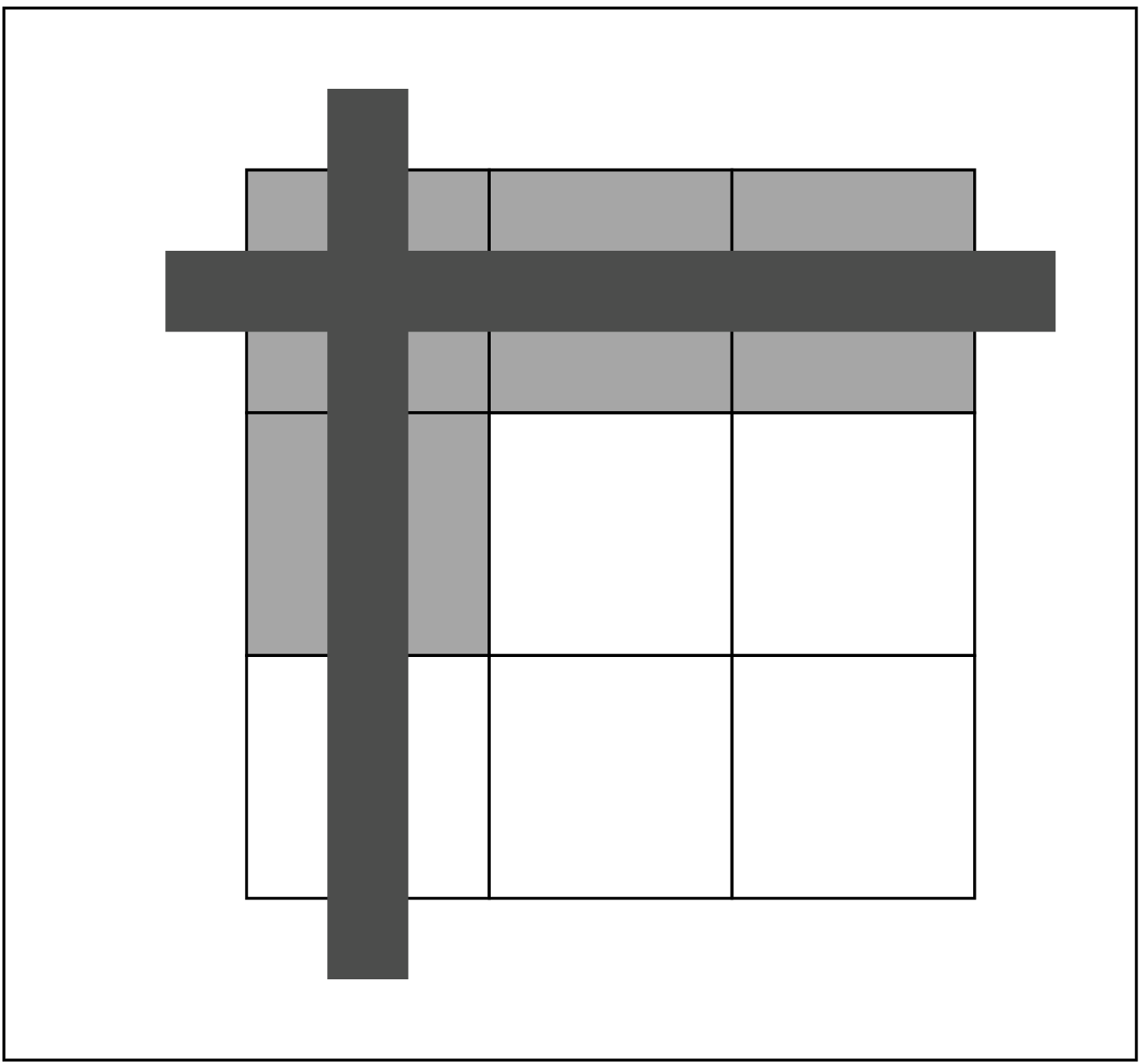}
\includegraphics[width=0.12\linewidth,keepaspectratio]{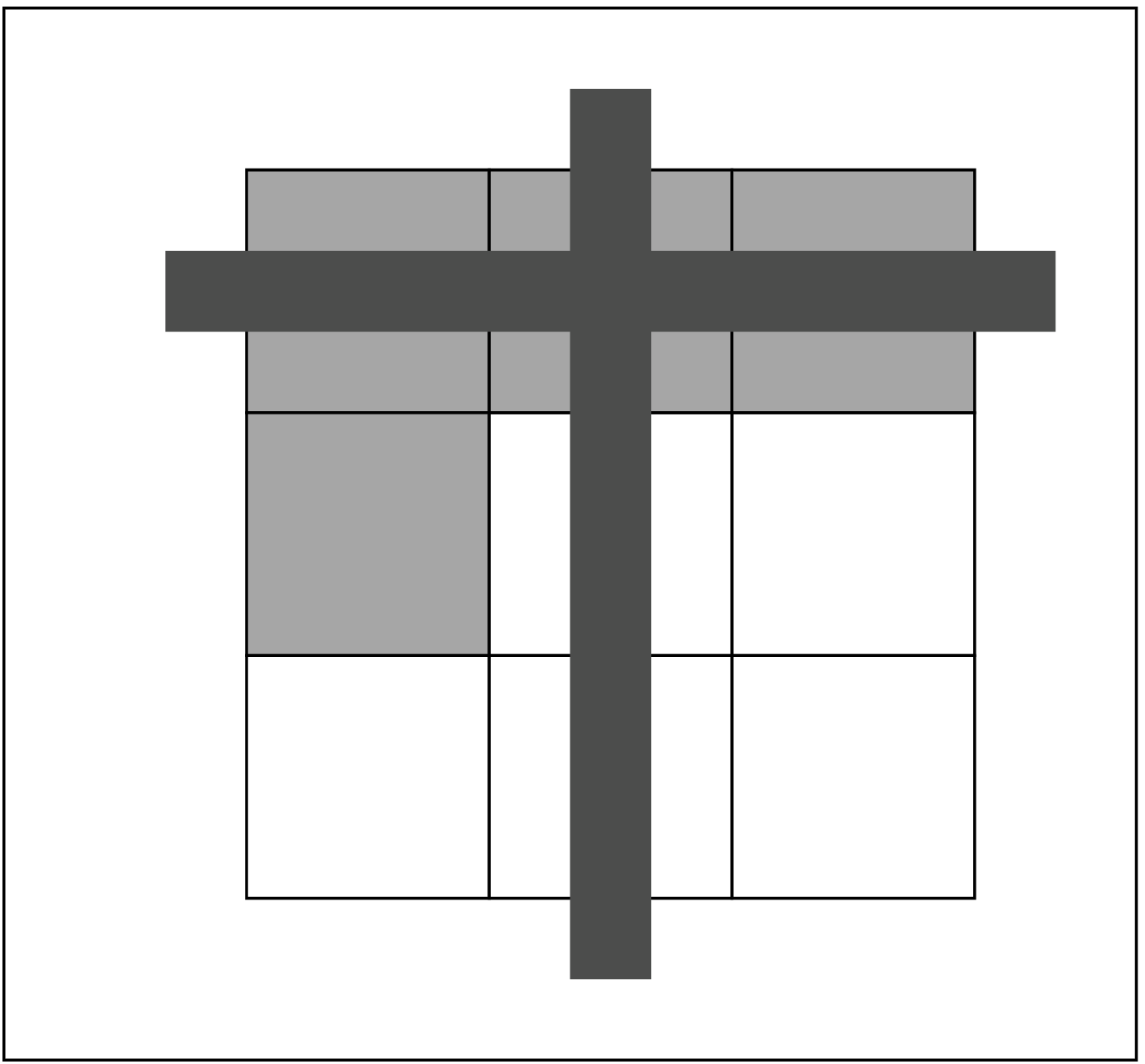}
\includegraphics[width=0.12\linewidth,keepaspectratio]{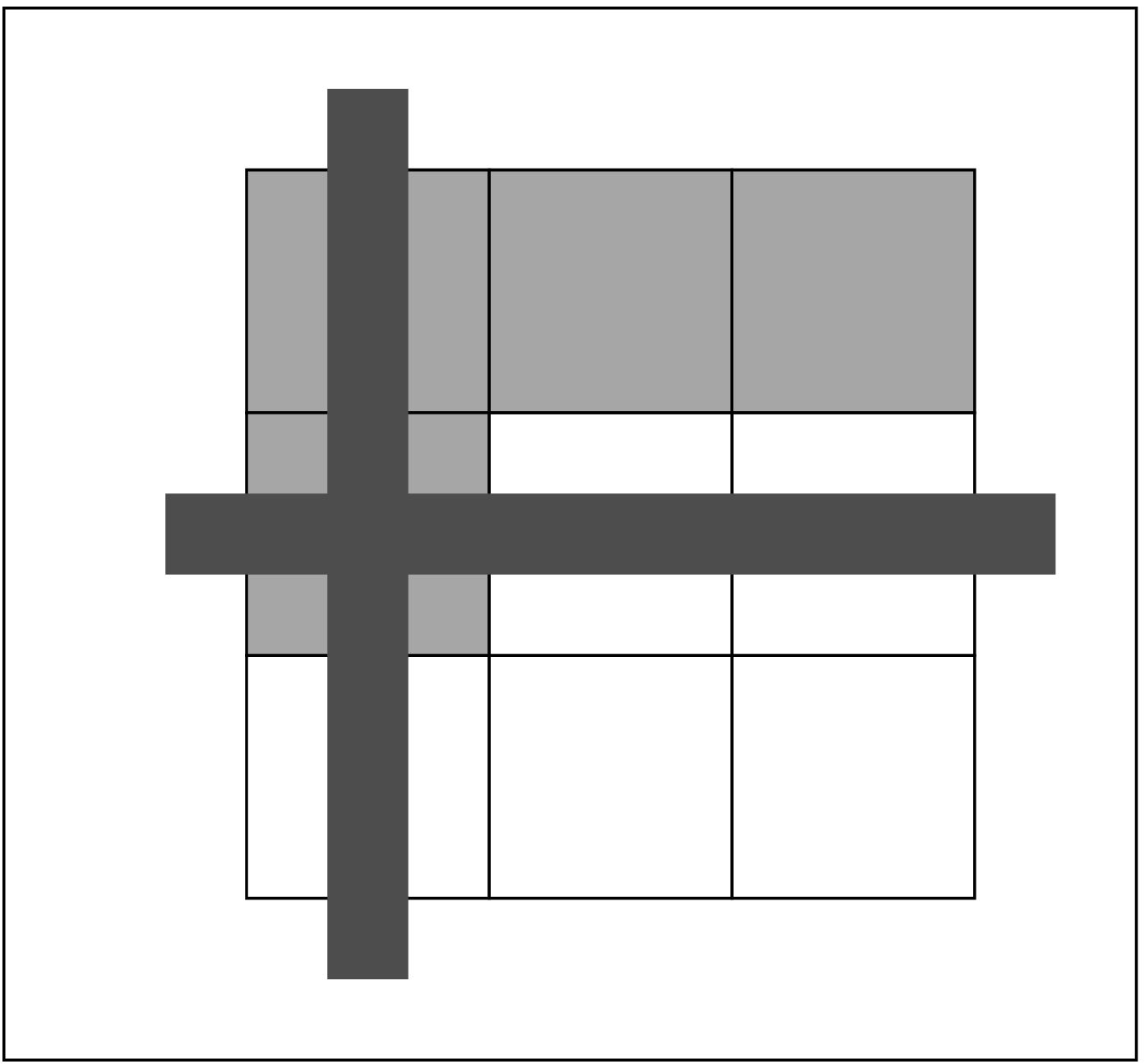}
\includegraphics[width=0.12\linewidth,keepaspectratio]{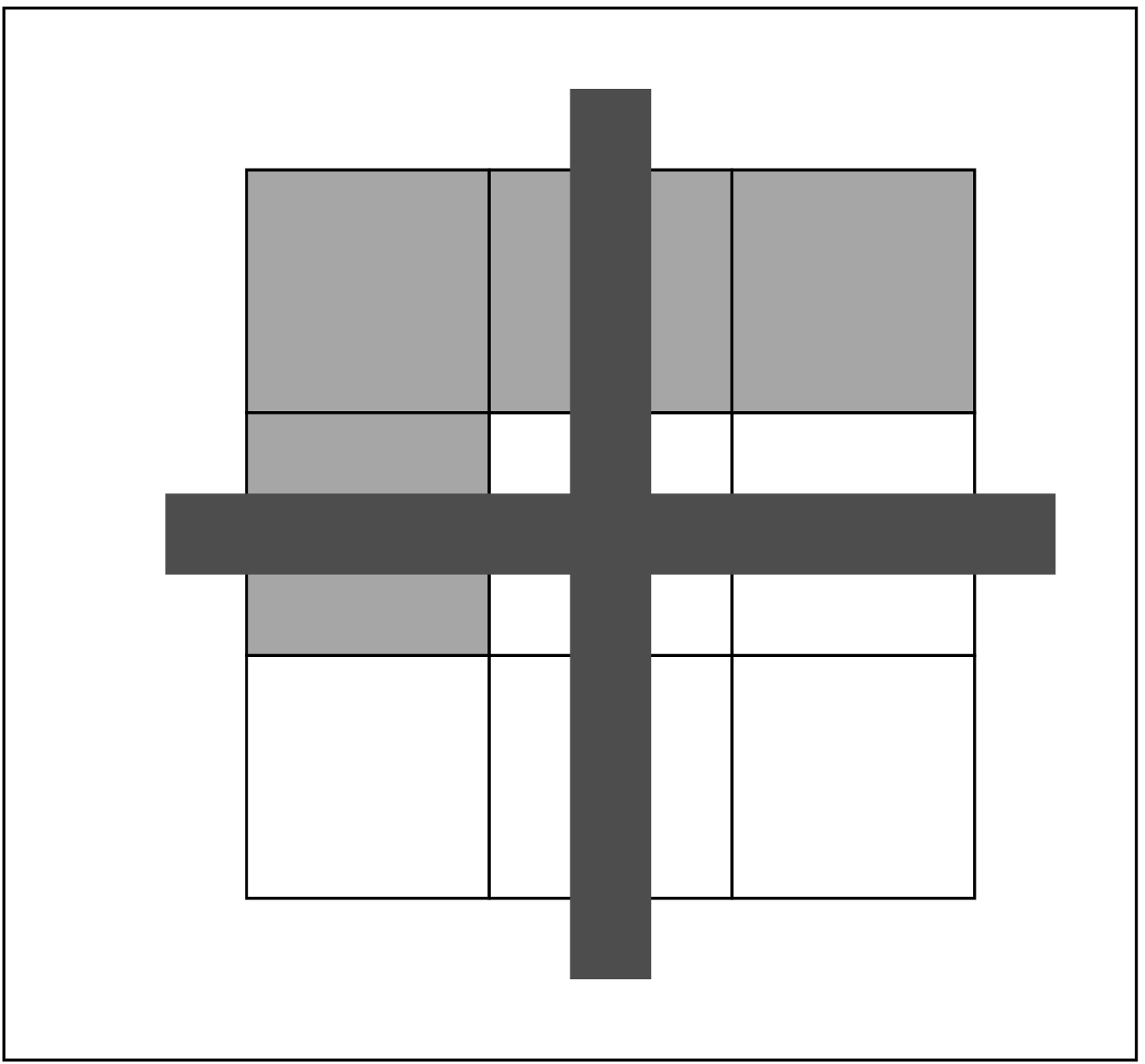}
\includegraphics[width=0.12\linewidth,keepaspectratio]{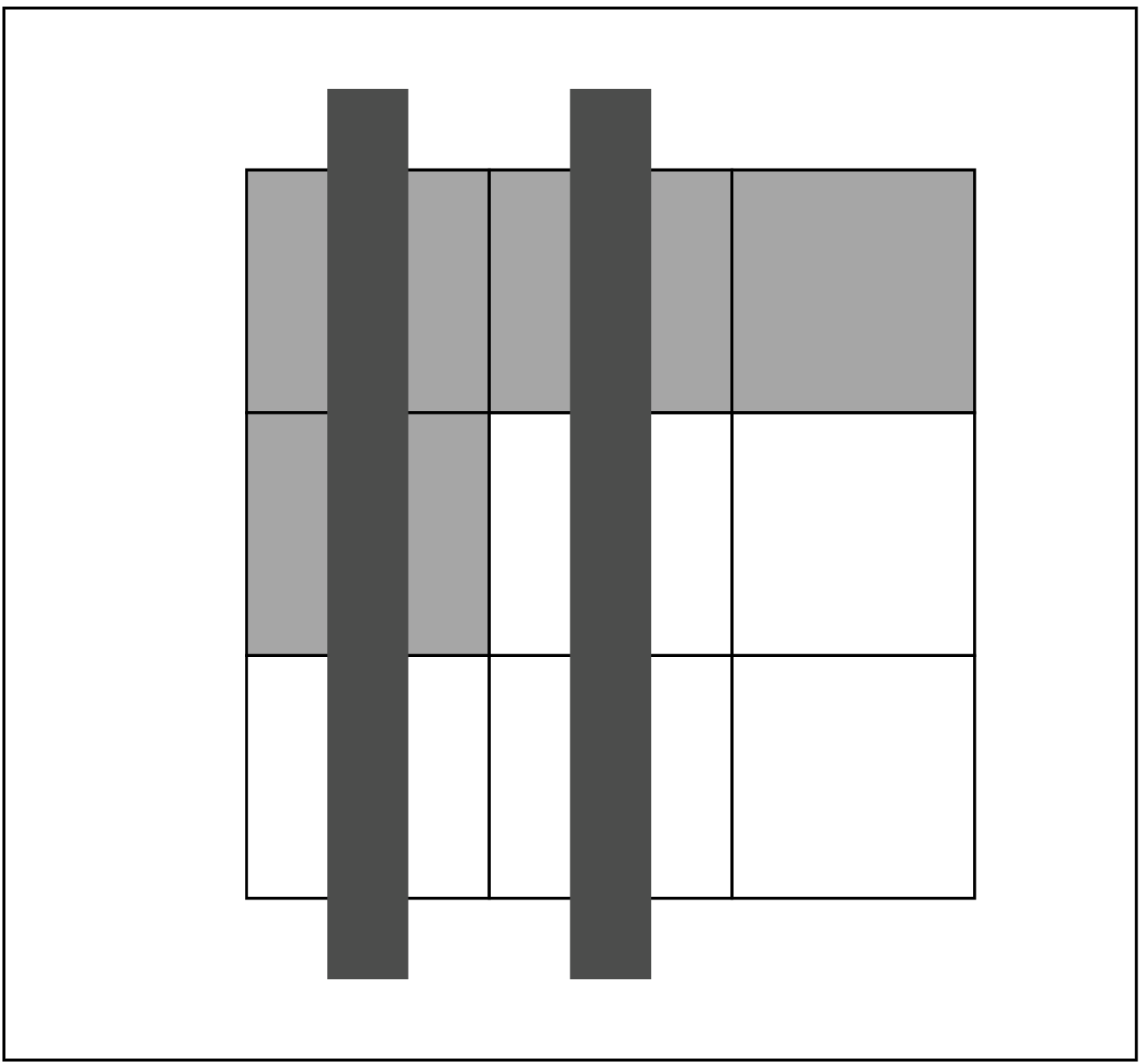}
\includegraphics[width=0.12\linewidth,keepaspectratio]{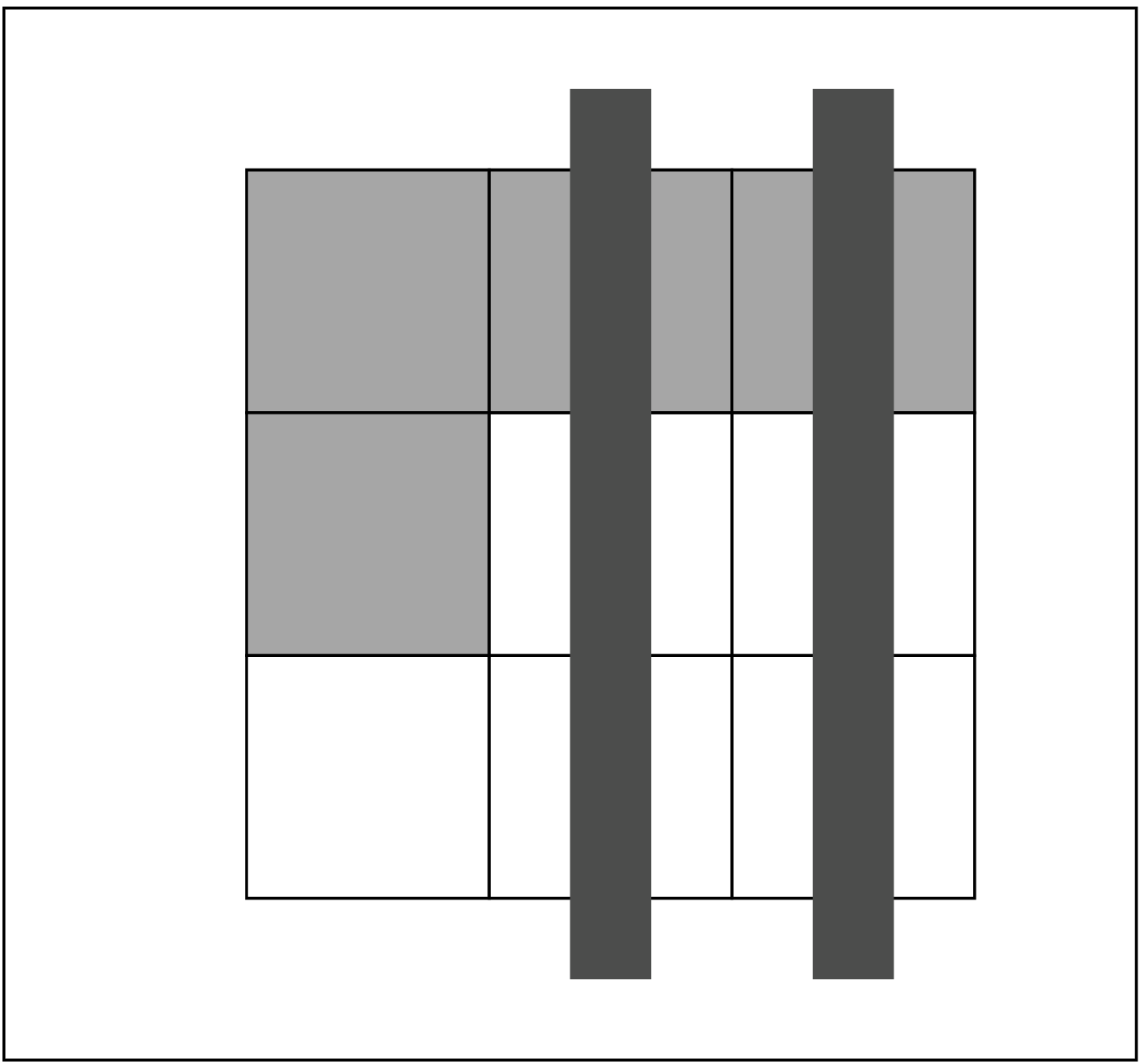}
\includegraphics[width=0.12\linewidth,keepaspectratio]{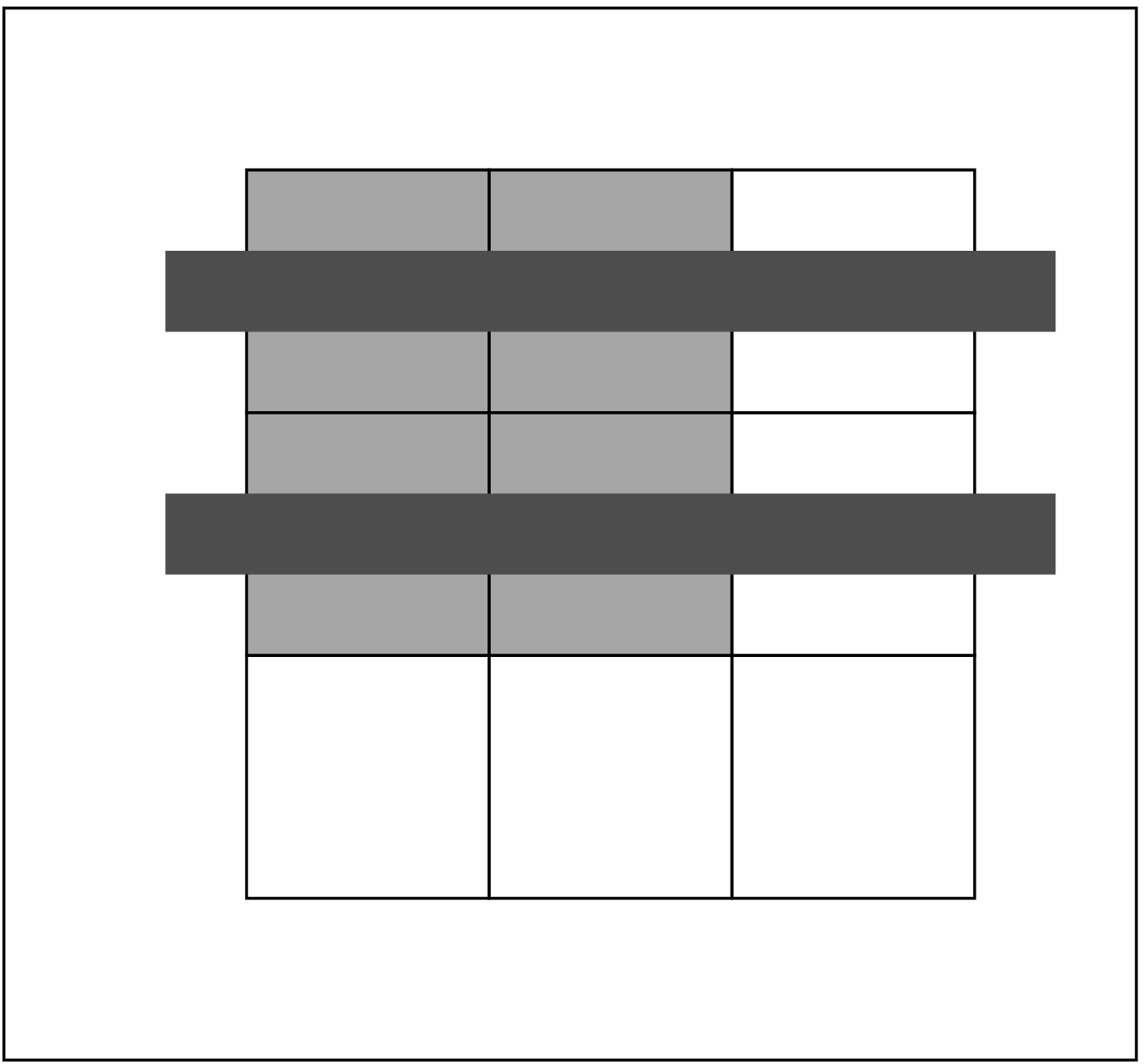}
\includegraphics[width=0.12\linewidth,keepaspectratio]{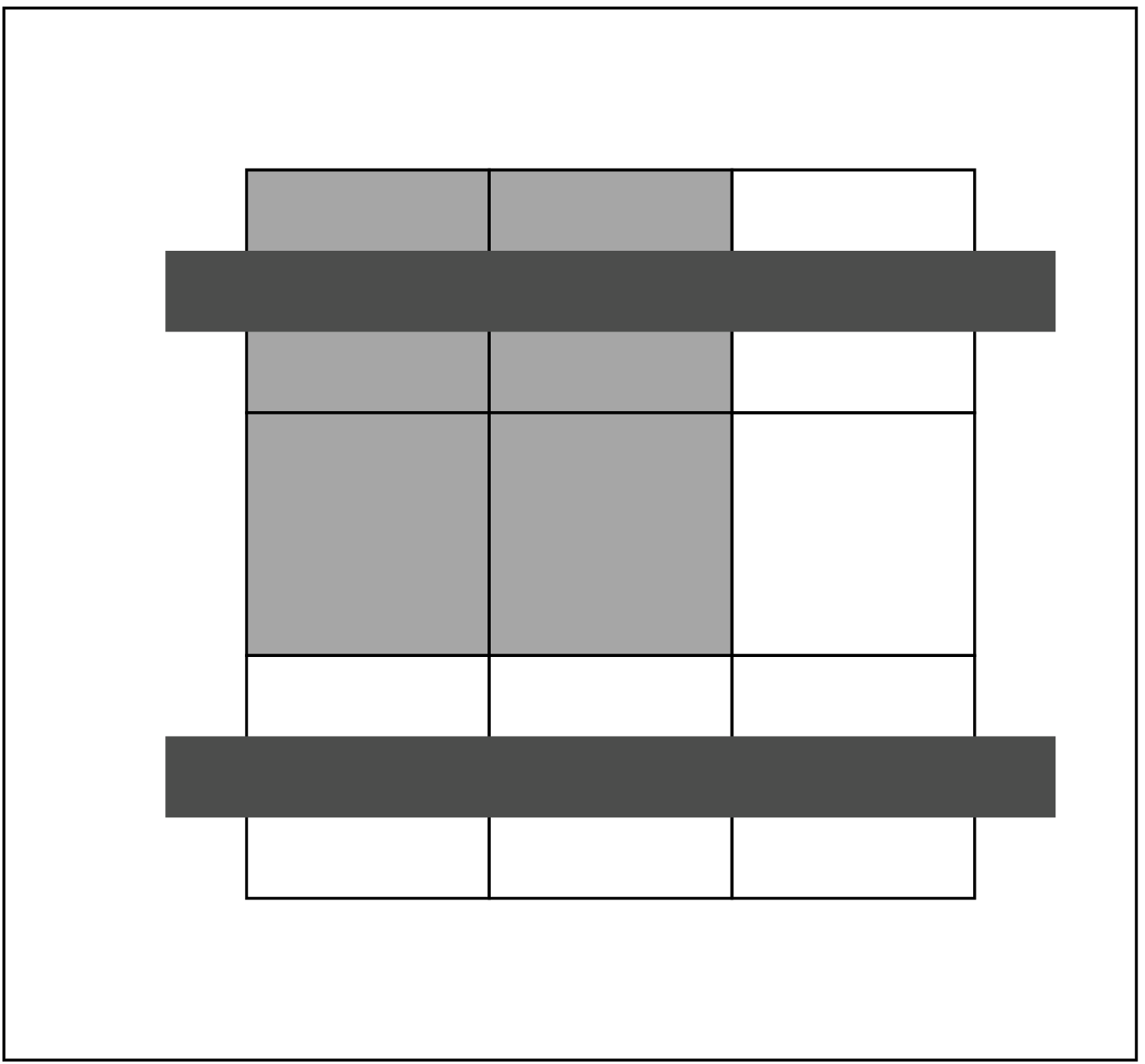}
\includegraphics[width=0.12\linewidth,keepaspectratio]{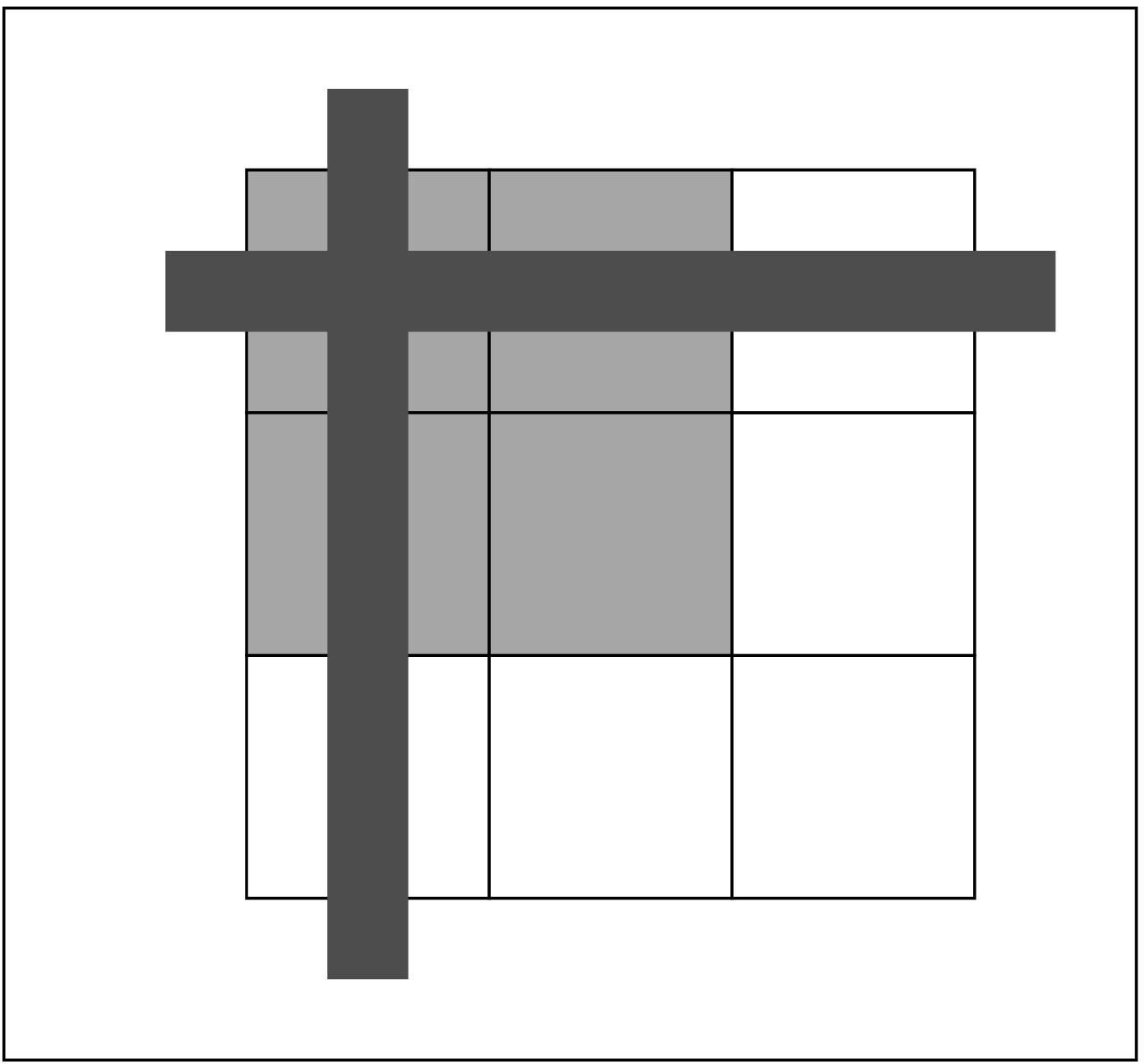}
\includegraphics[width=0.12\linewidth,keepaspectratio]{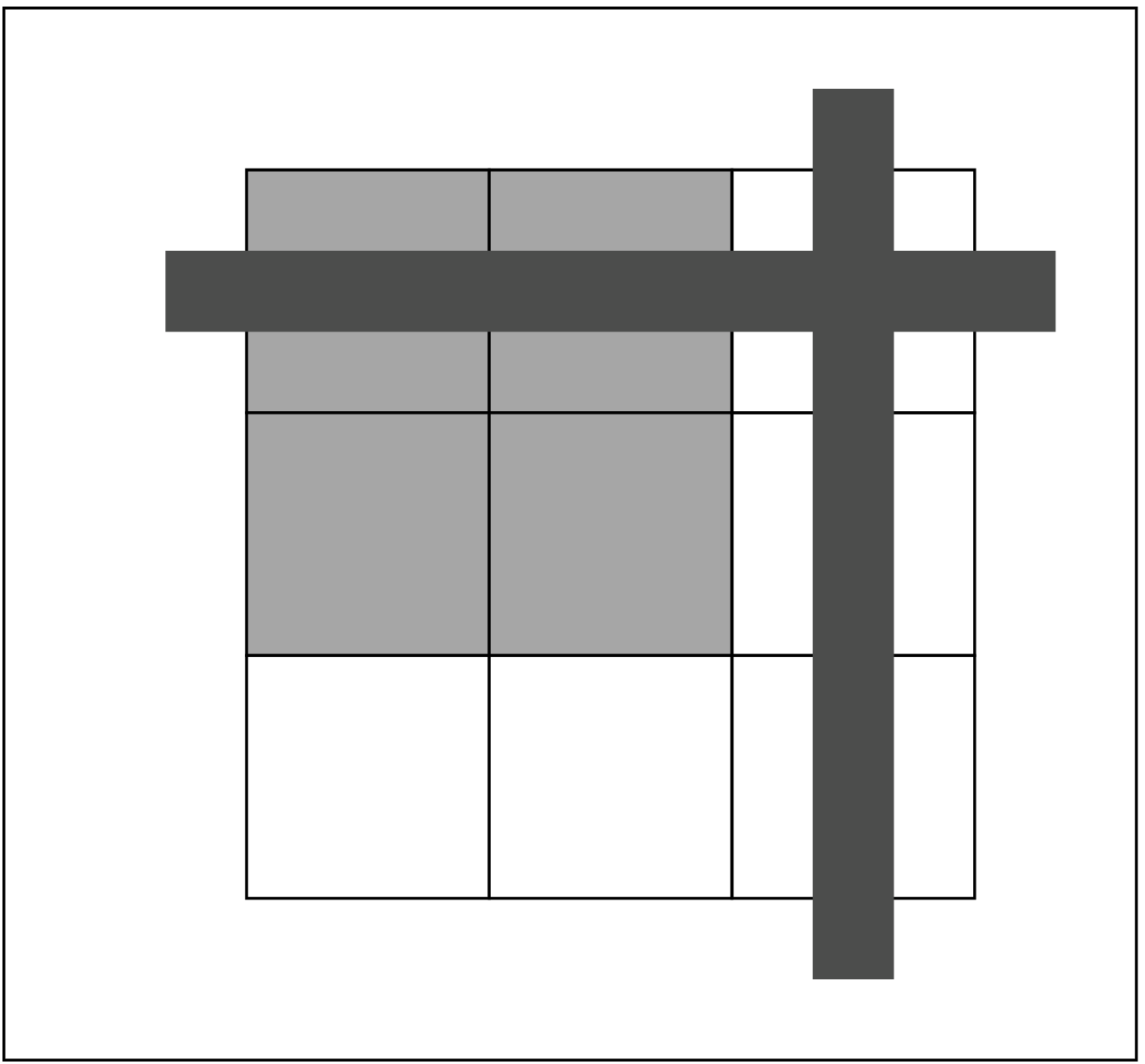}
\includegraphics[width=0.12\linewidth,keepaspectratio]{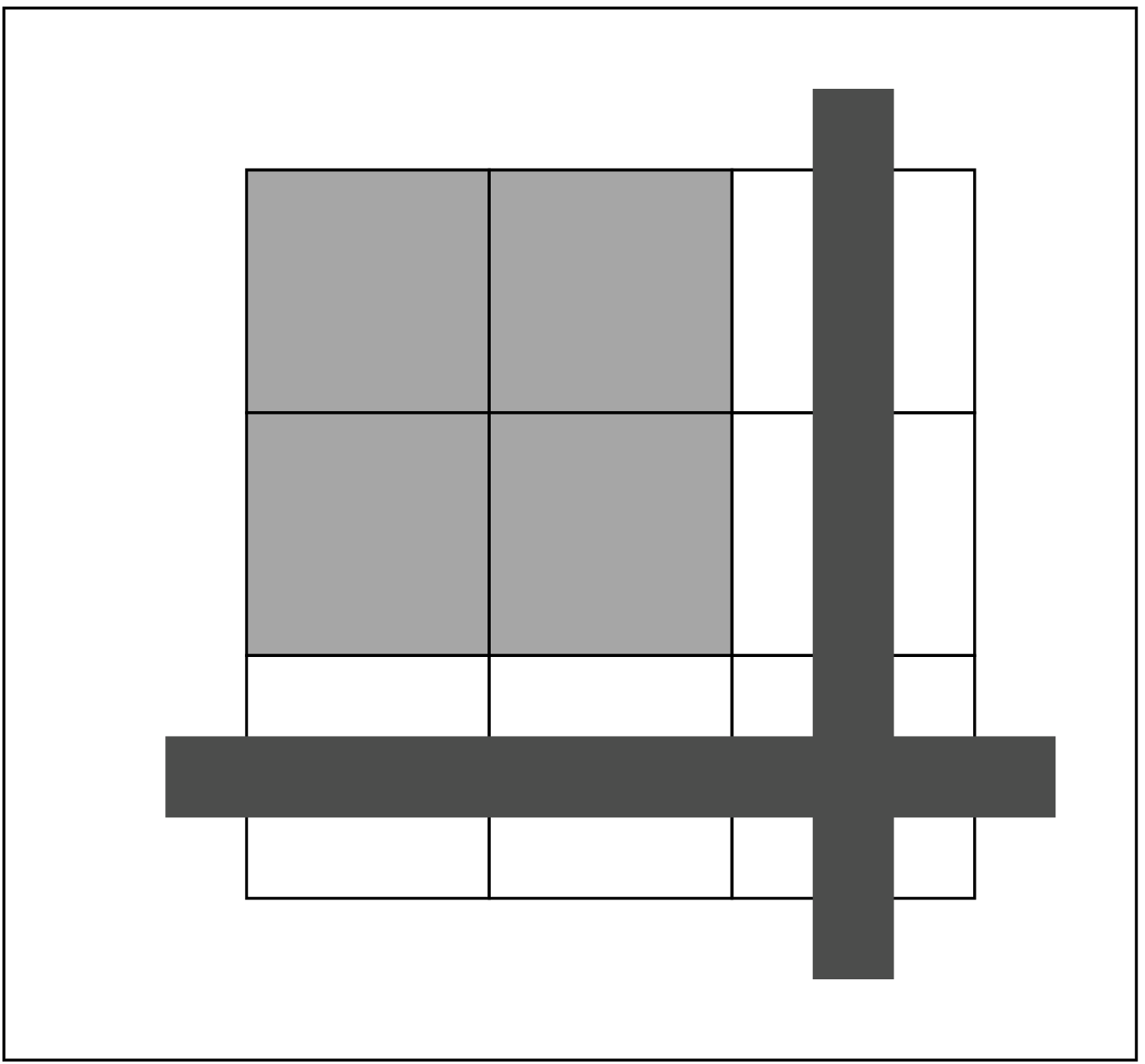}
\includegraphics[width=0.12\linewidth,keepaspectratio]{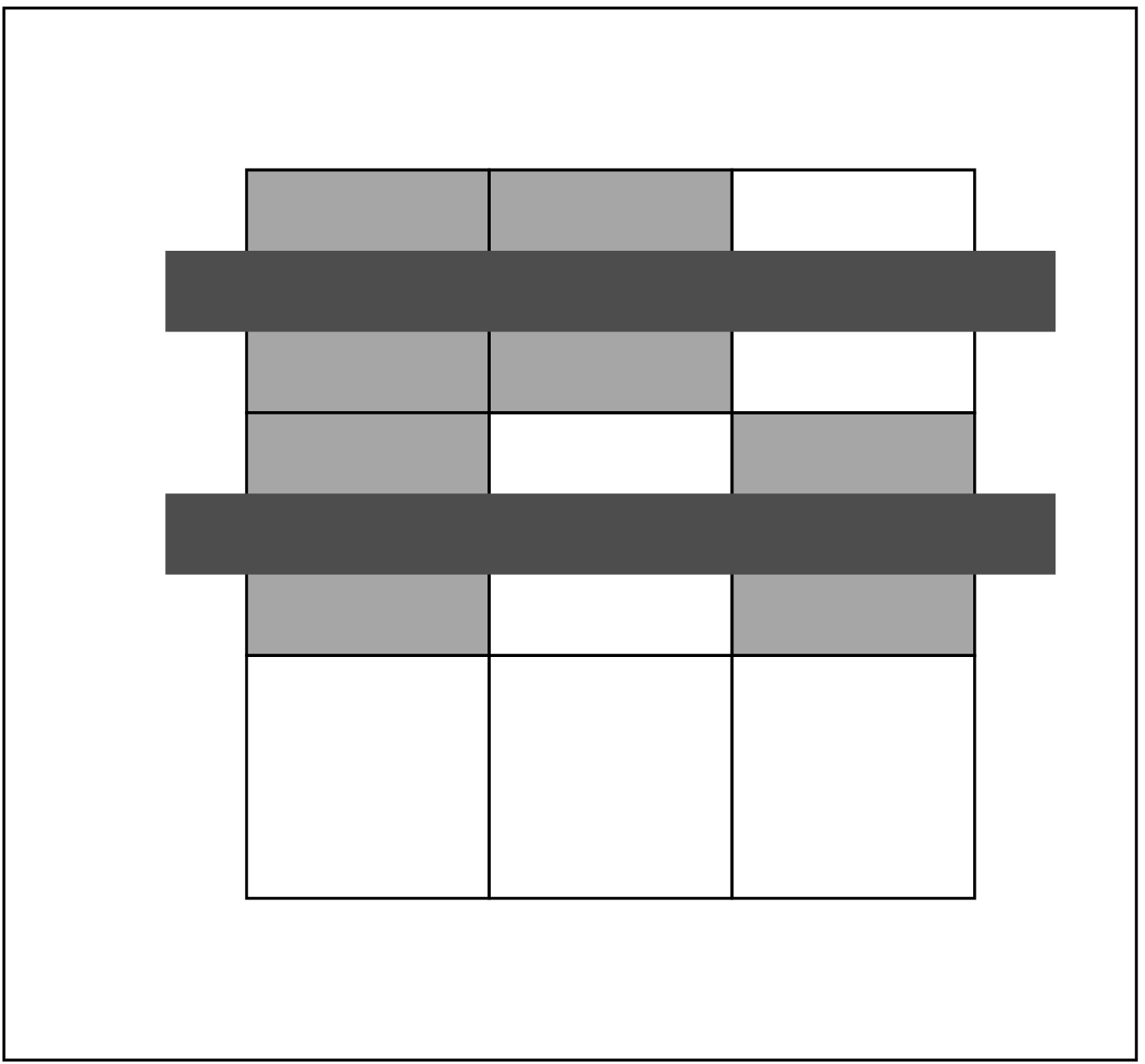}
\includegraphics[width=0.12\linewidth,keepaspectratio]{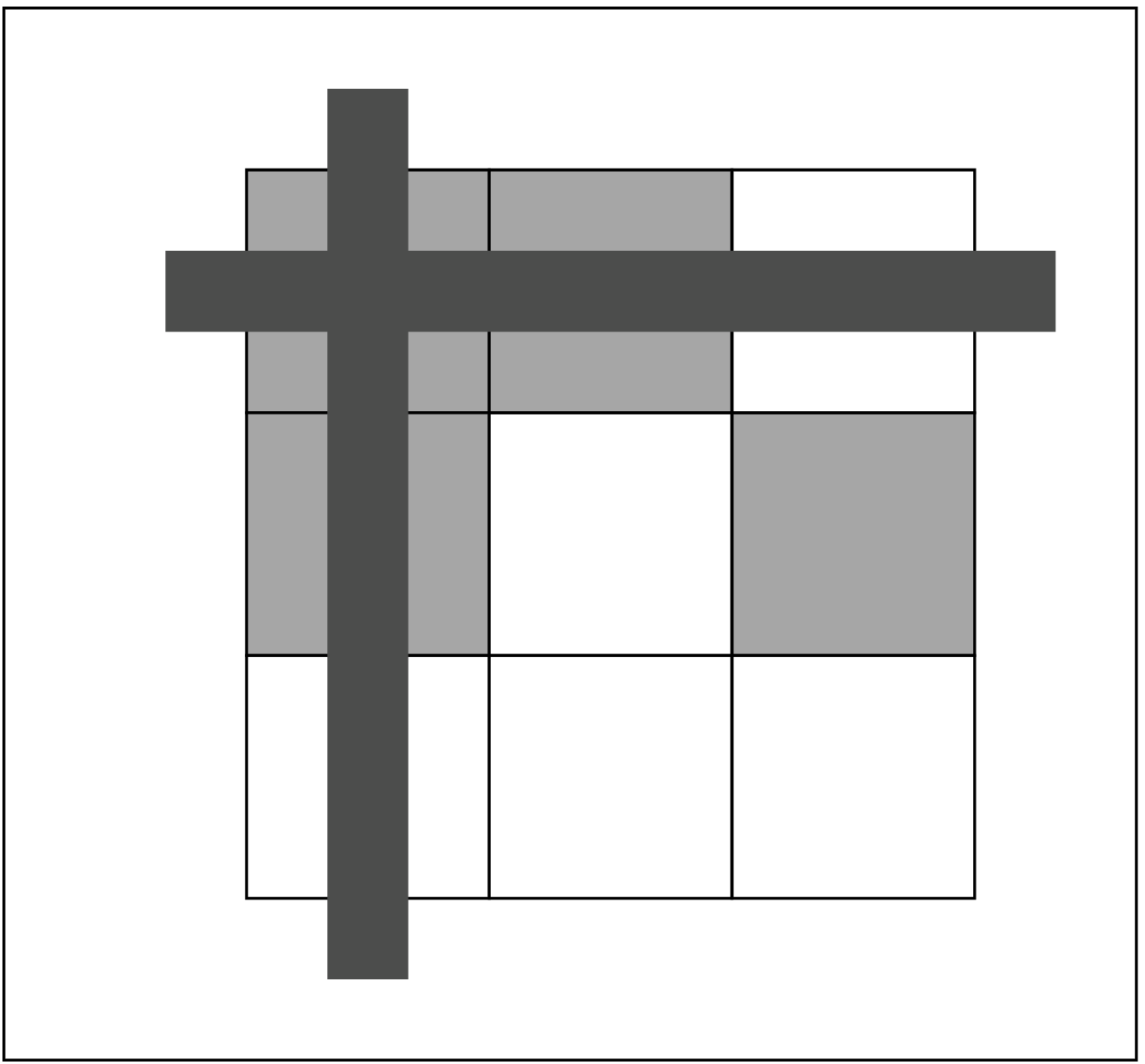}
\includegraphics[width=0.12\linewidth,keepaspectratio]{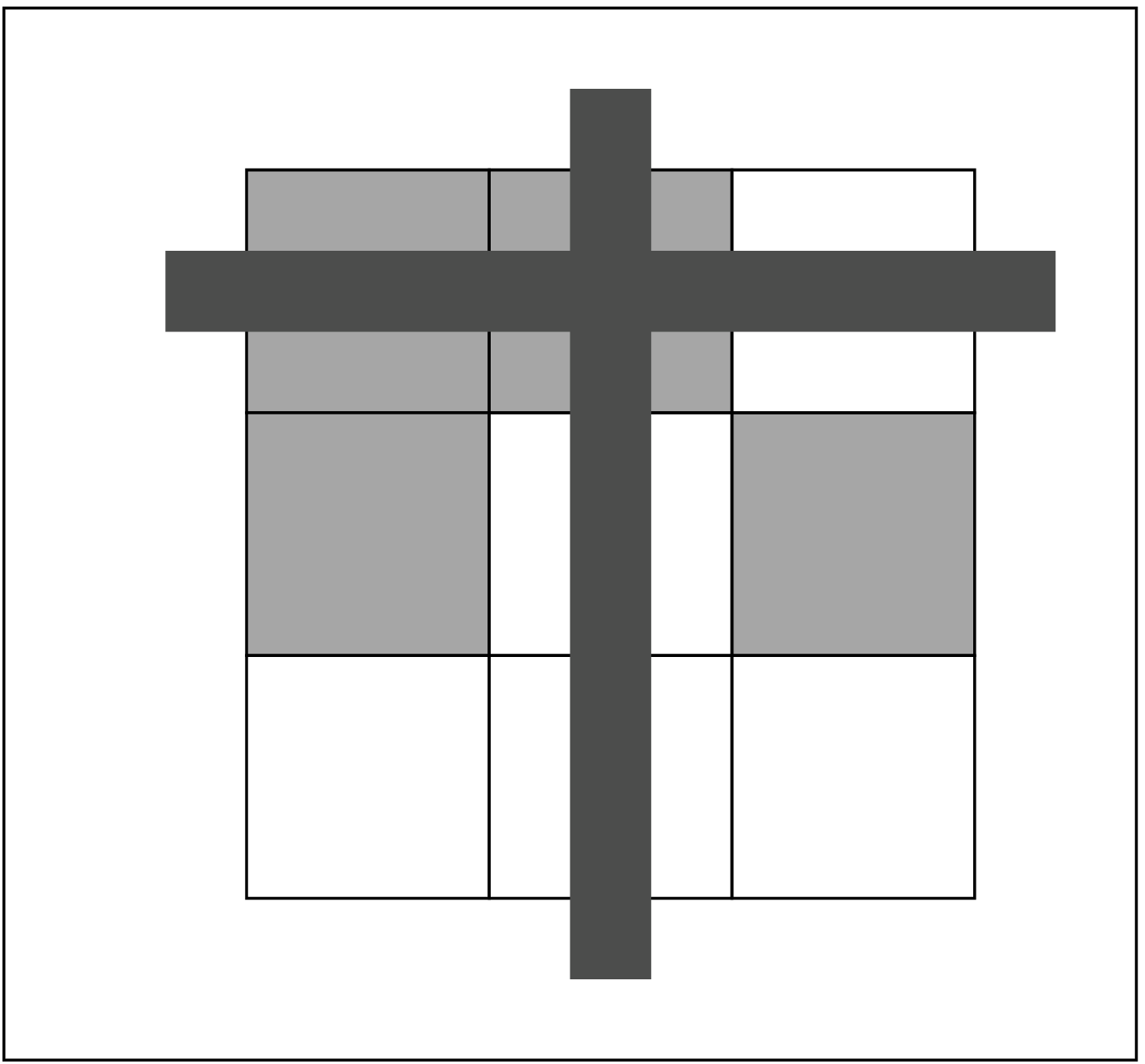}
\includegraphics[width=0.12\linewidth,keepaspectratio]{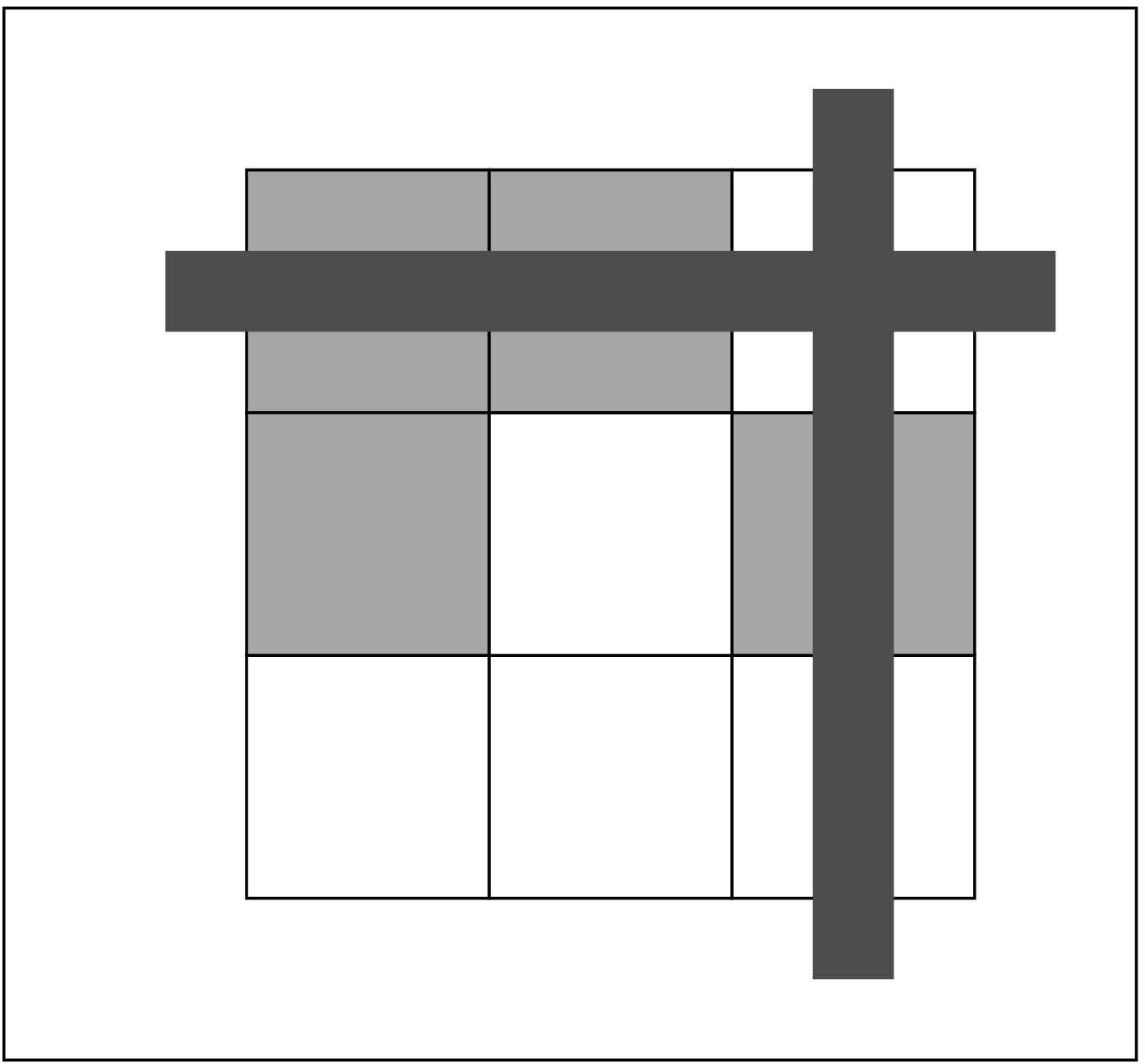}
\includegraphics[width=0.12\linewidth,keepaspectratio]{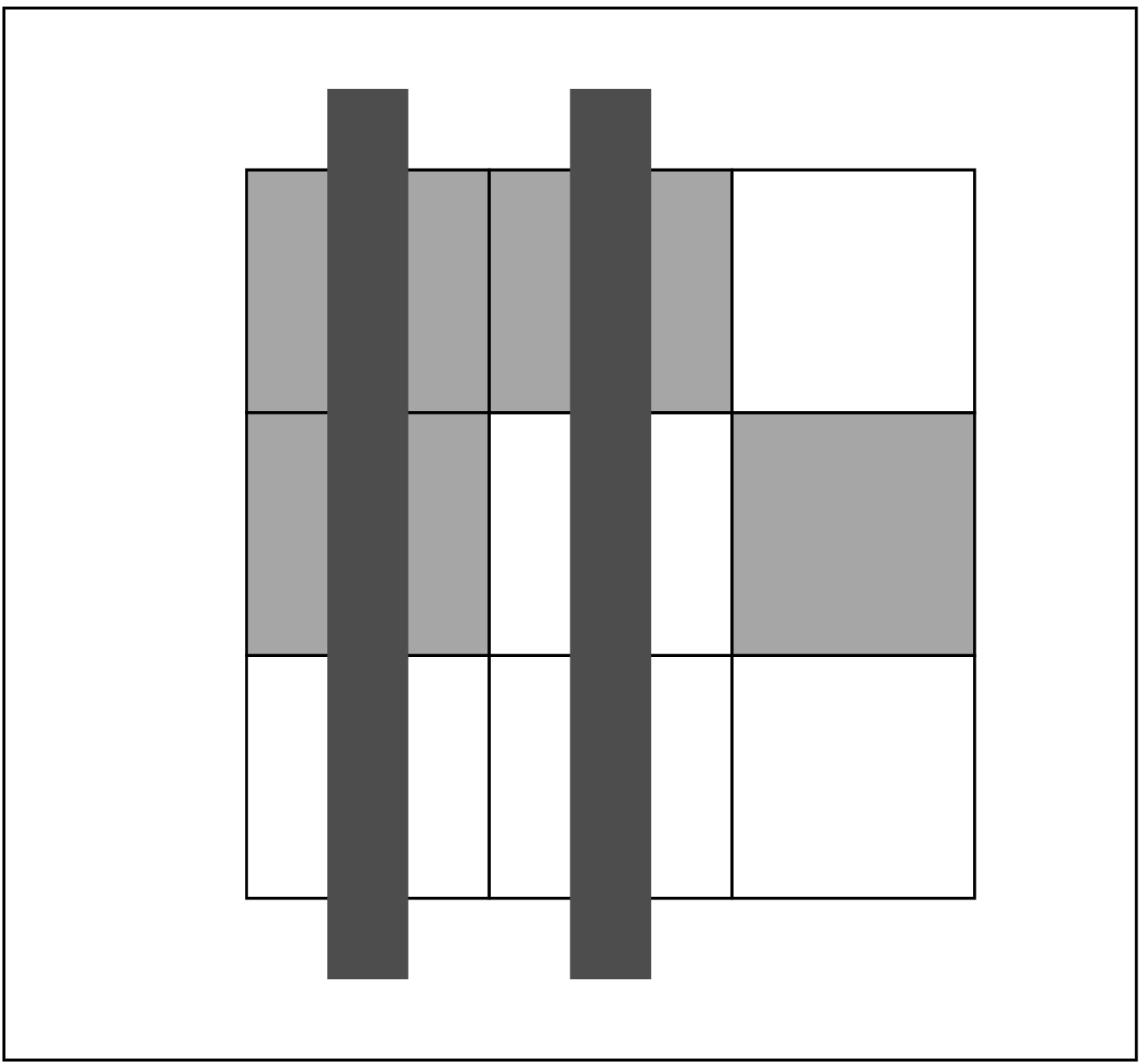}
\includegraphics[width=0.12\linewidth,keepaspectratio]{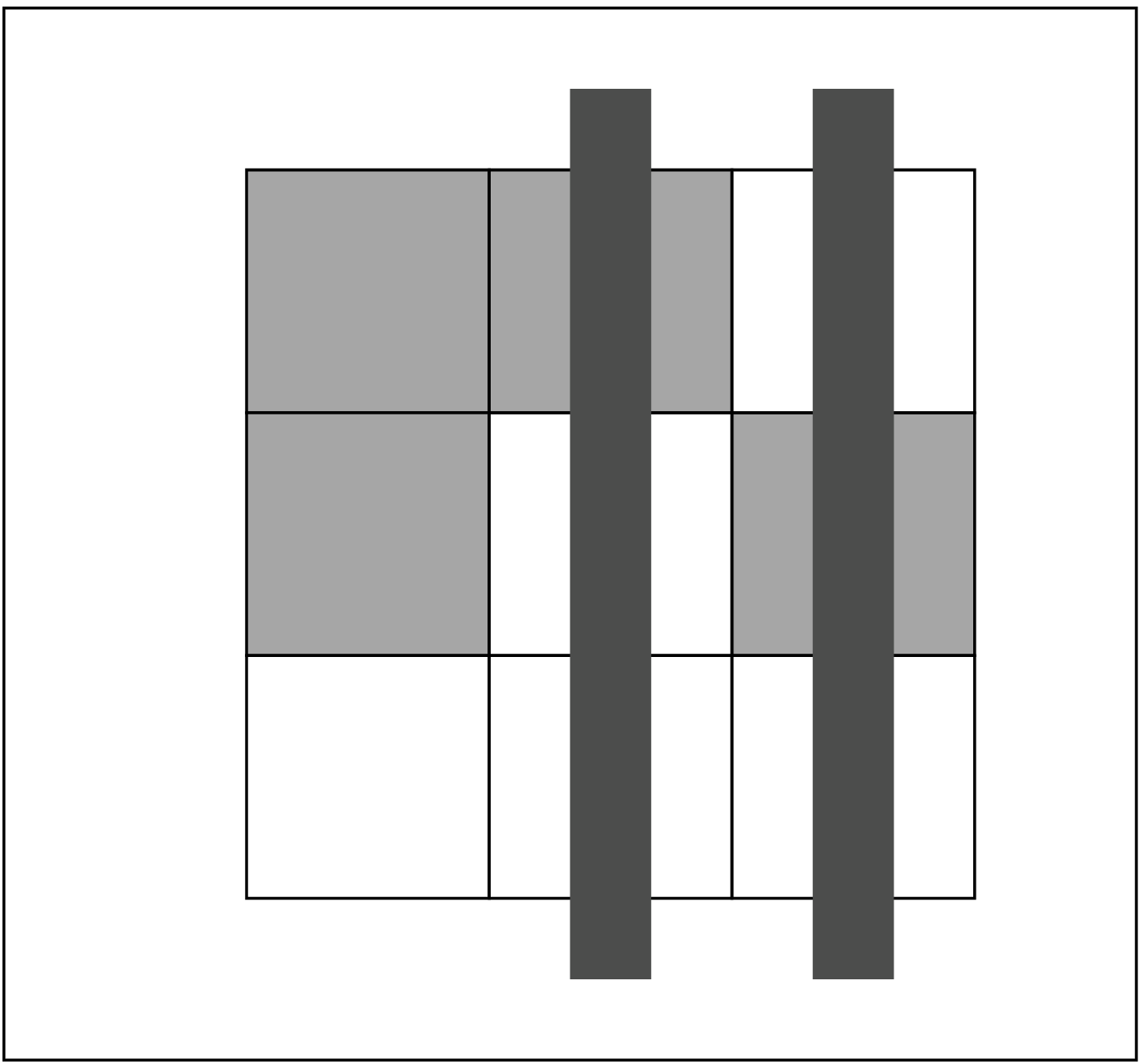}
\includegraphics[width=0.12\linewidth,keepaspectratio]{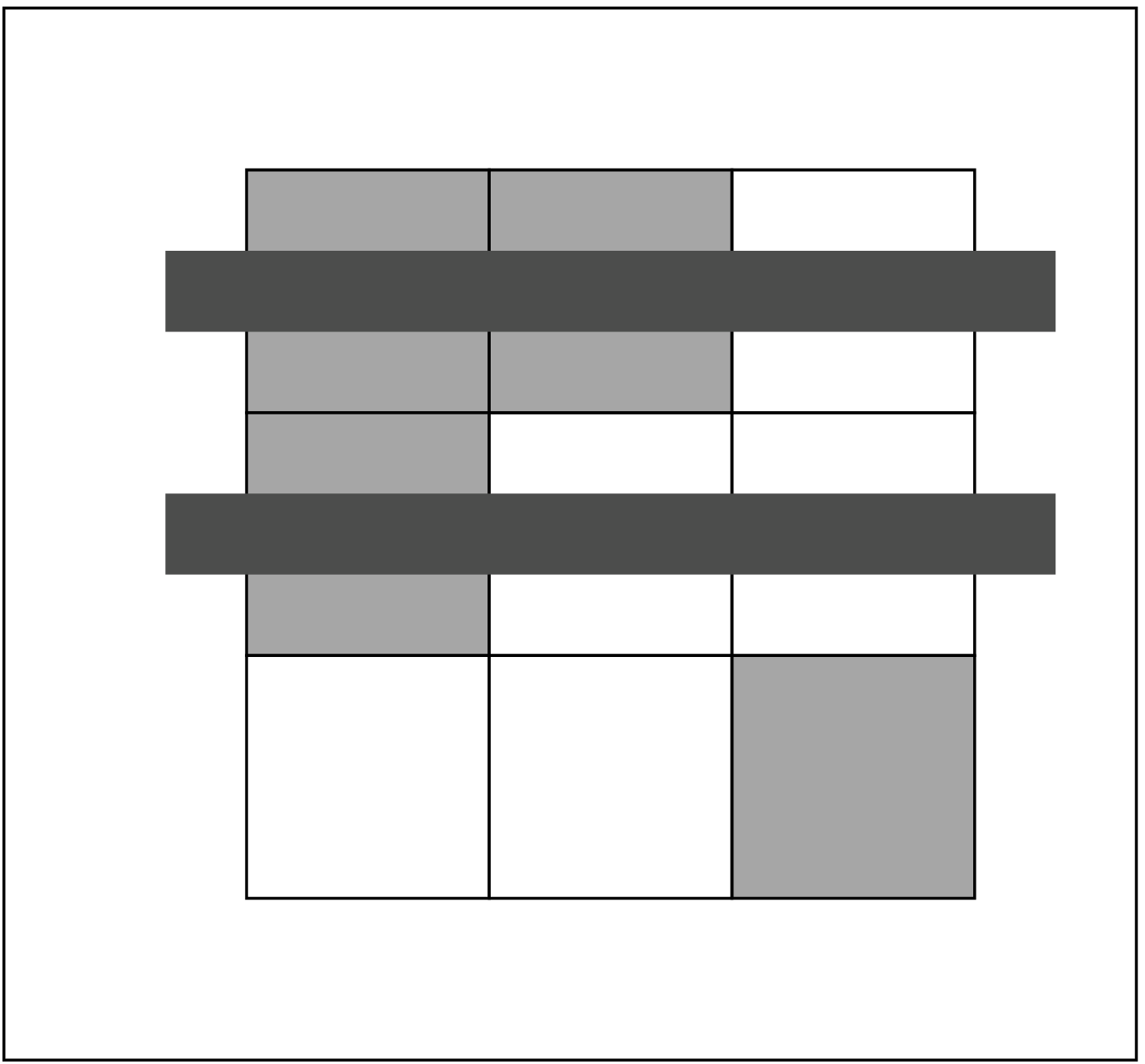}
\includegraphics[width=0.12\linewidth,keepaspectratio]{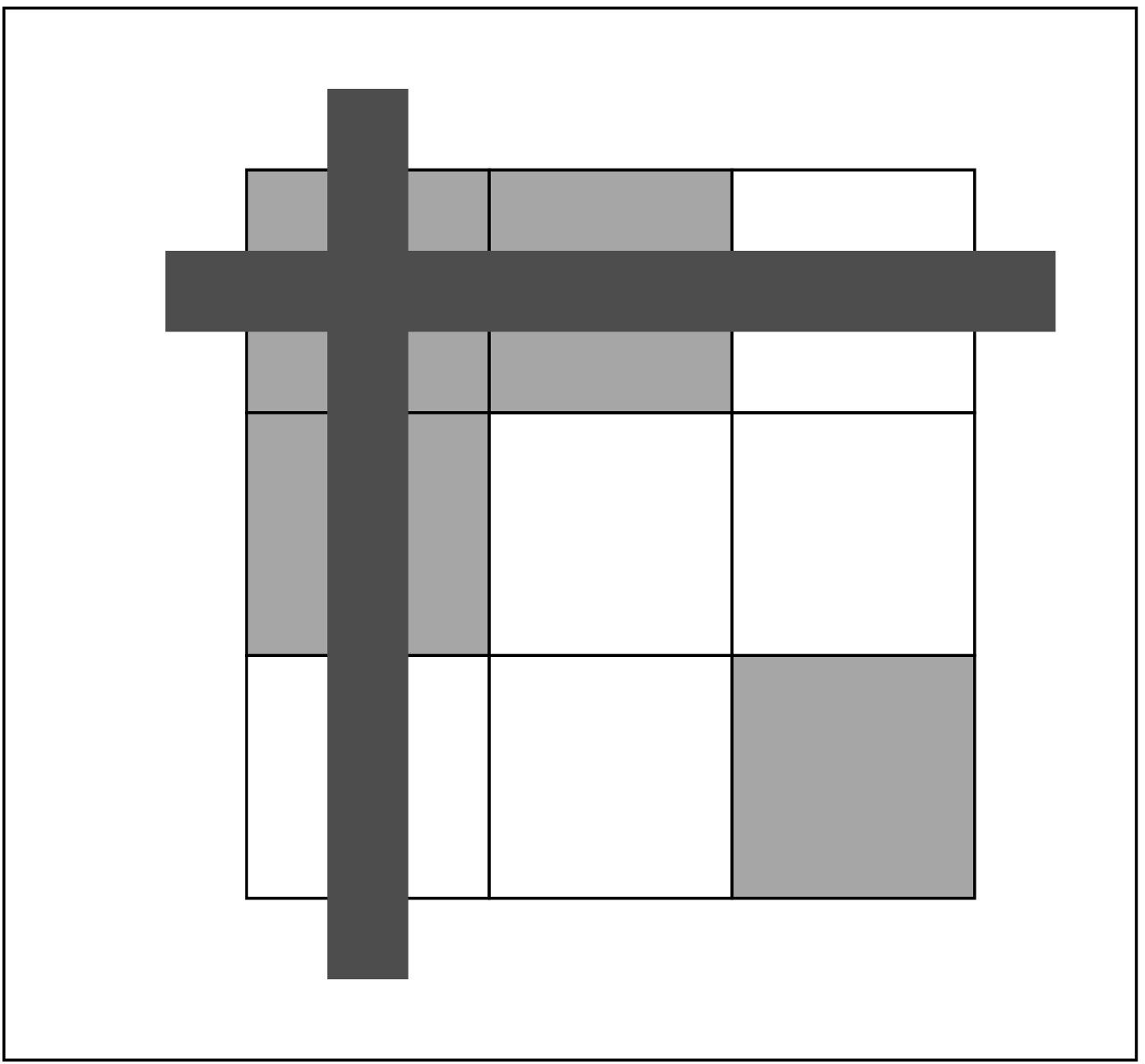}
\includegraphics[width=0.12\linewidth,keepaspectratio]{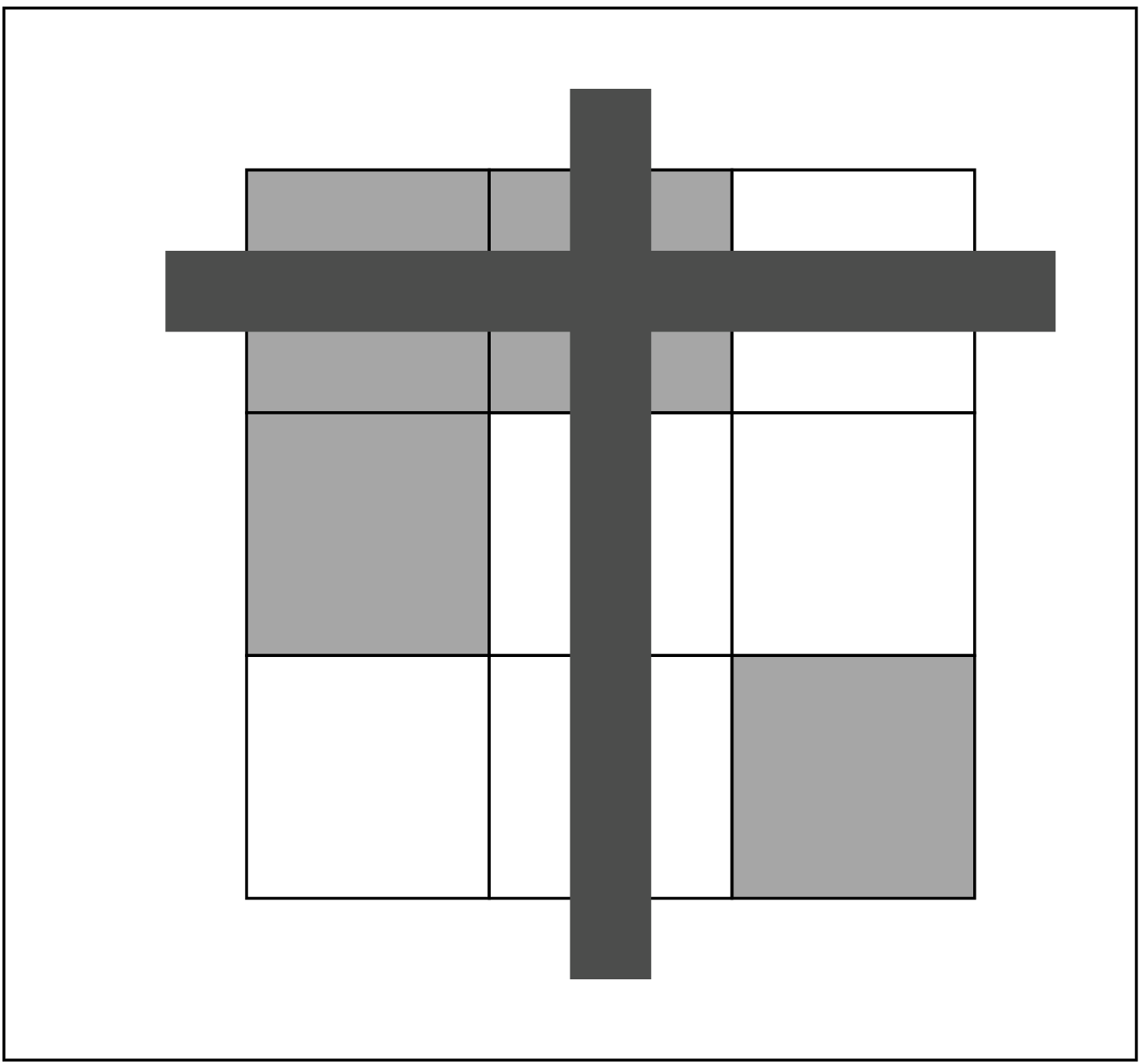}
\includegraphics[width=0.12\linewidth,keepaspectratio]{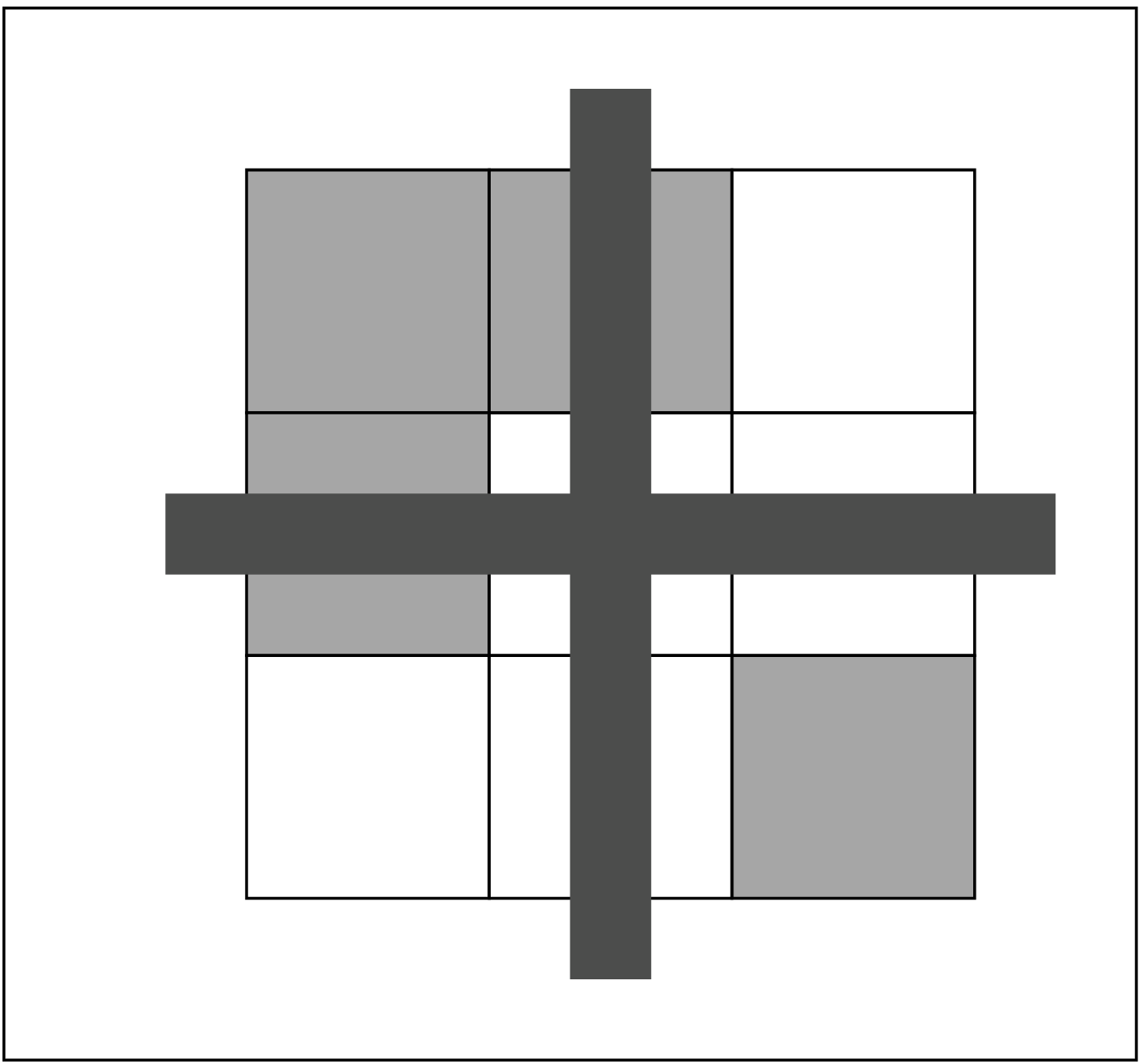}
\includegraphics[width=0.12\linewidth,keepaspectratio]{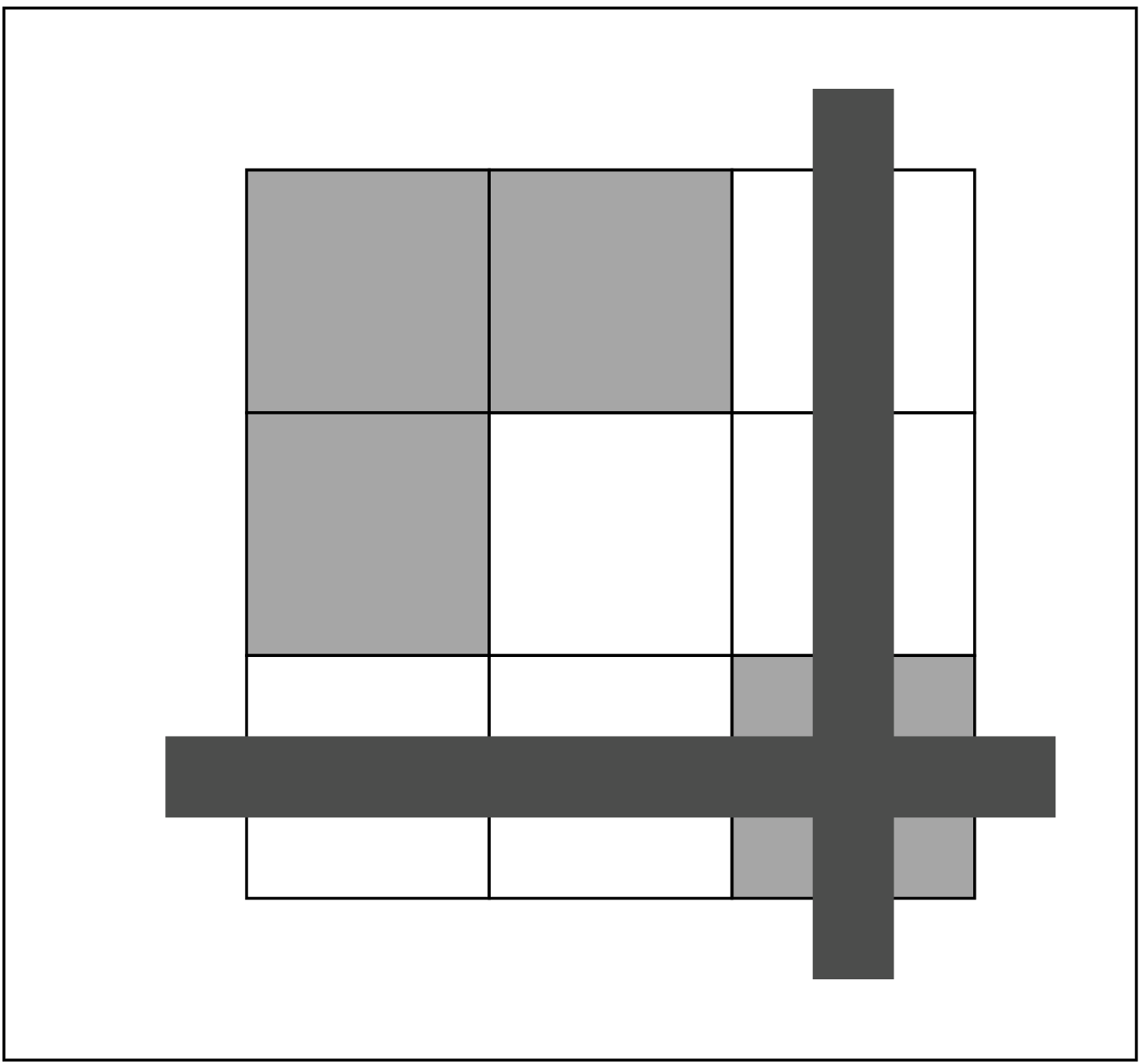}
\includegraphics[width=0.12\linewidth,keepaspectratio]{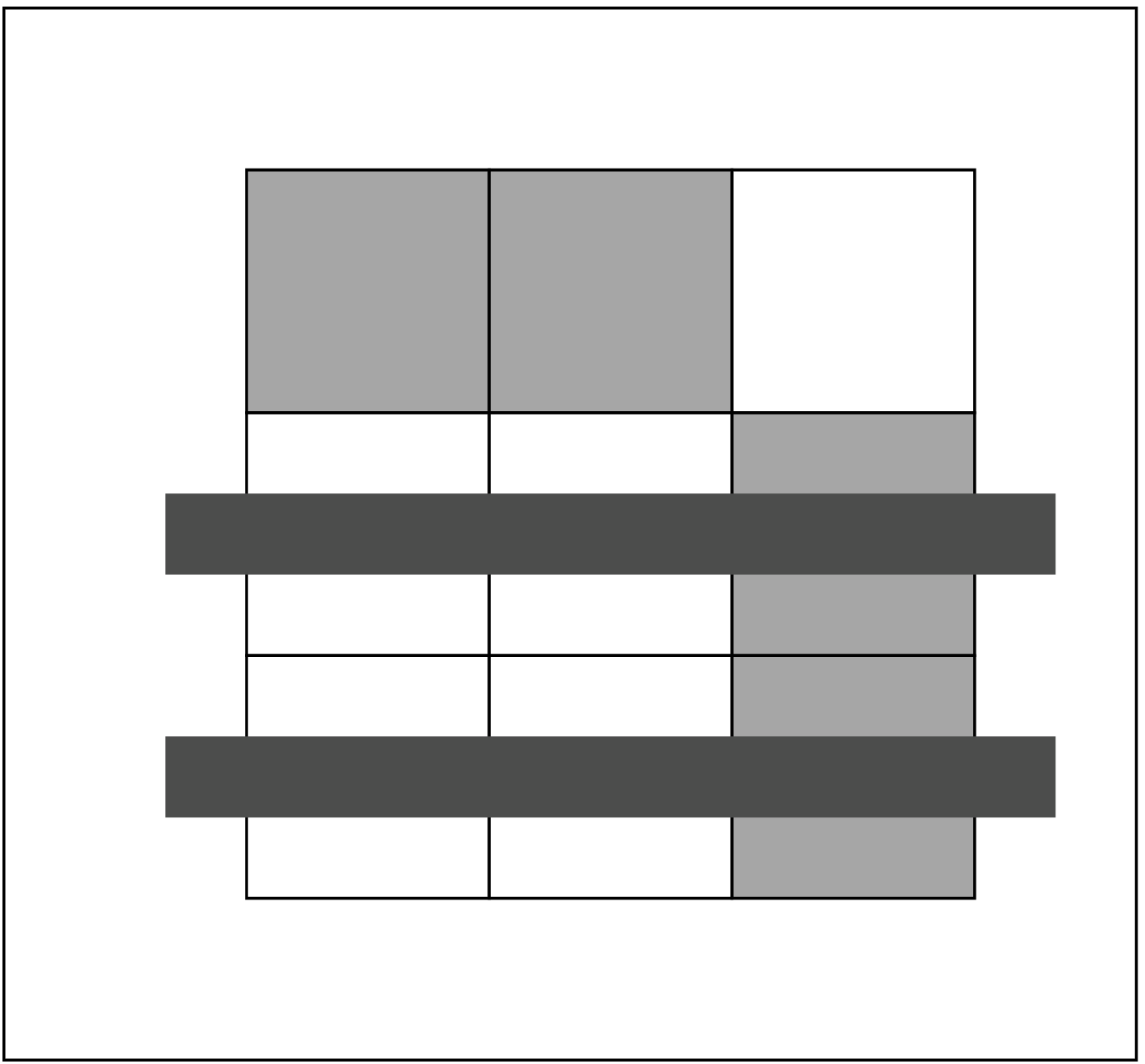}
\includegraphics[width=0.12\linewidth,keepaspectratio]{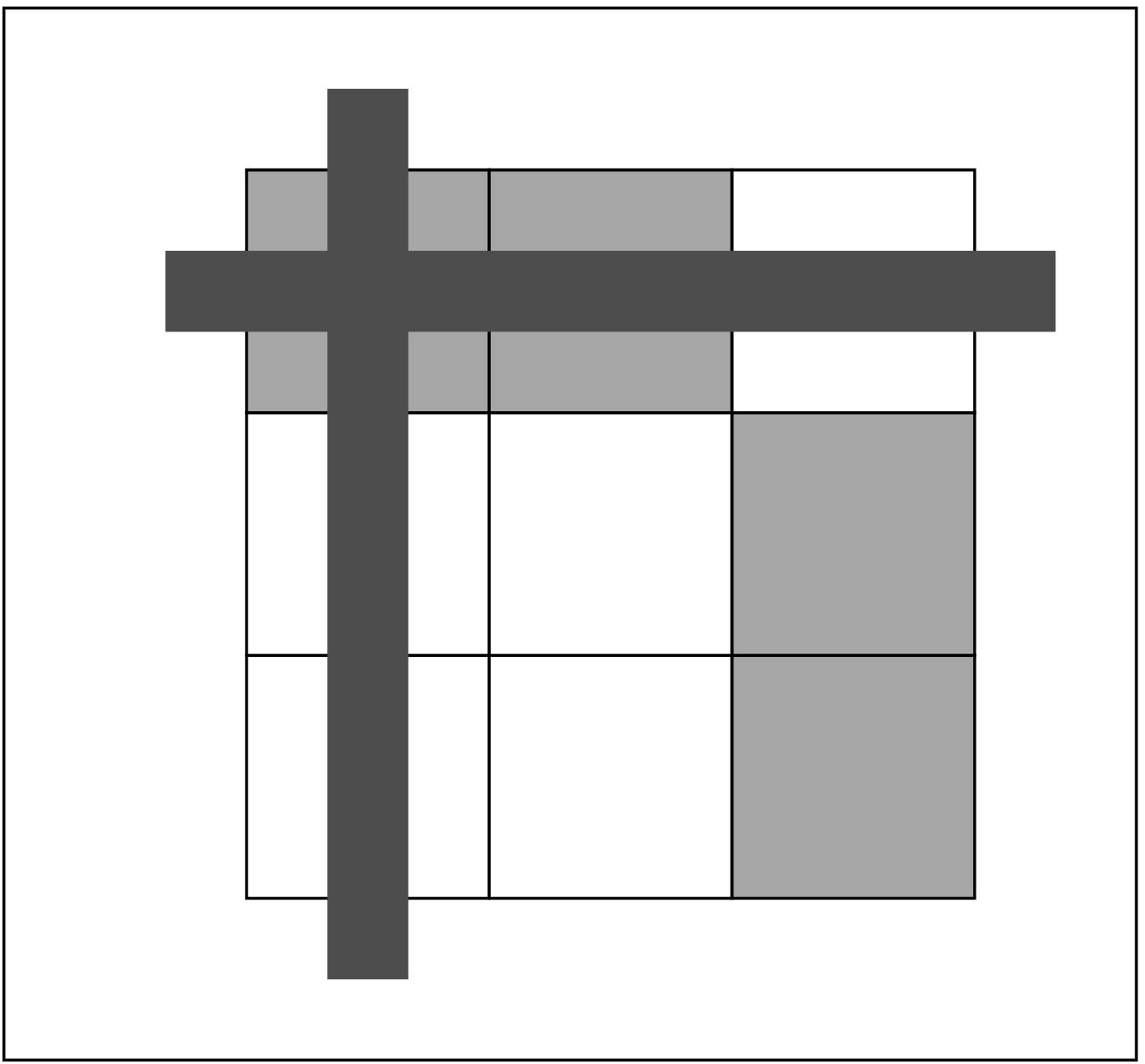}
\includegraphics[width=0.12\linewidth,keepaspectratio]{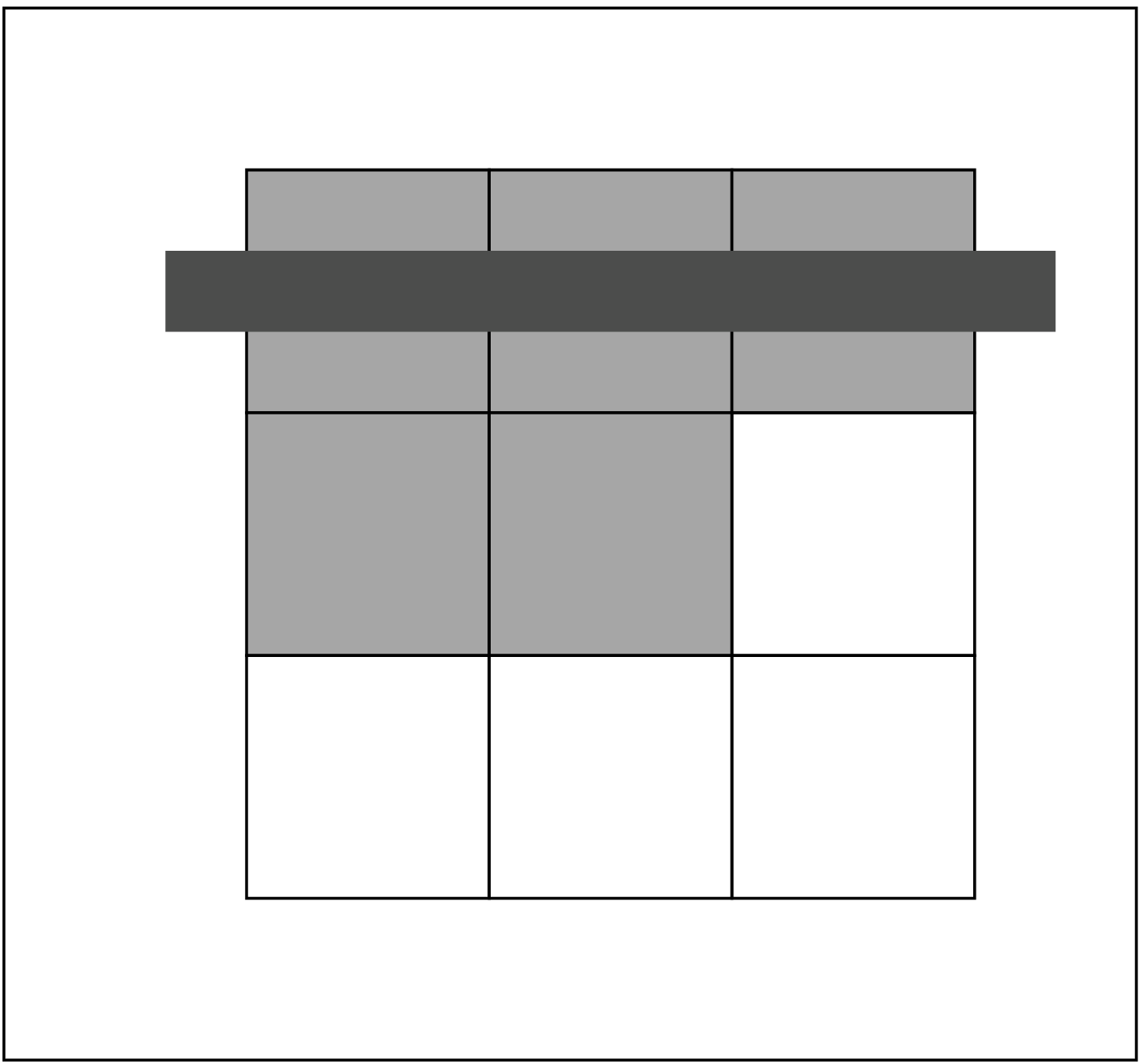}
\includegraphics[width=0.12\linewidth,keepaspectratio]{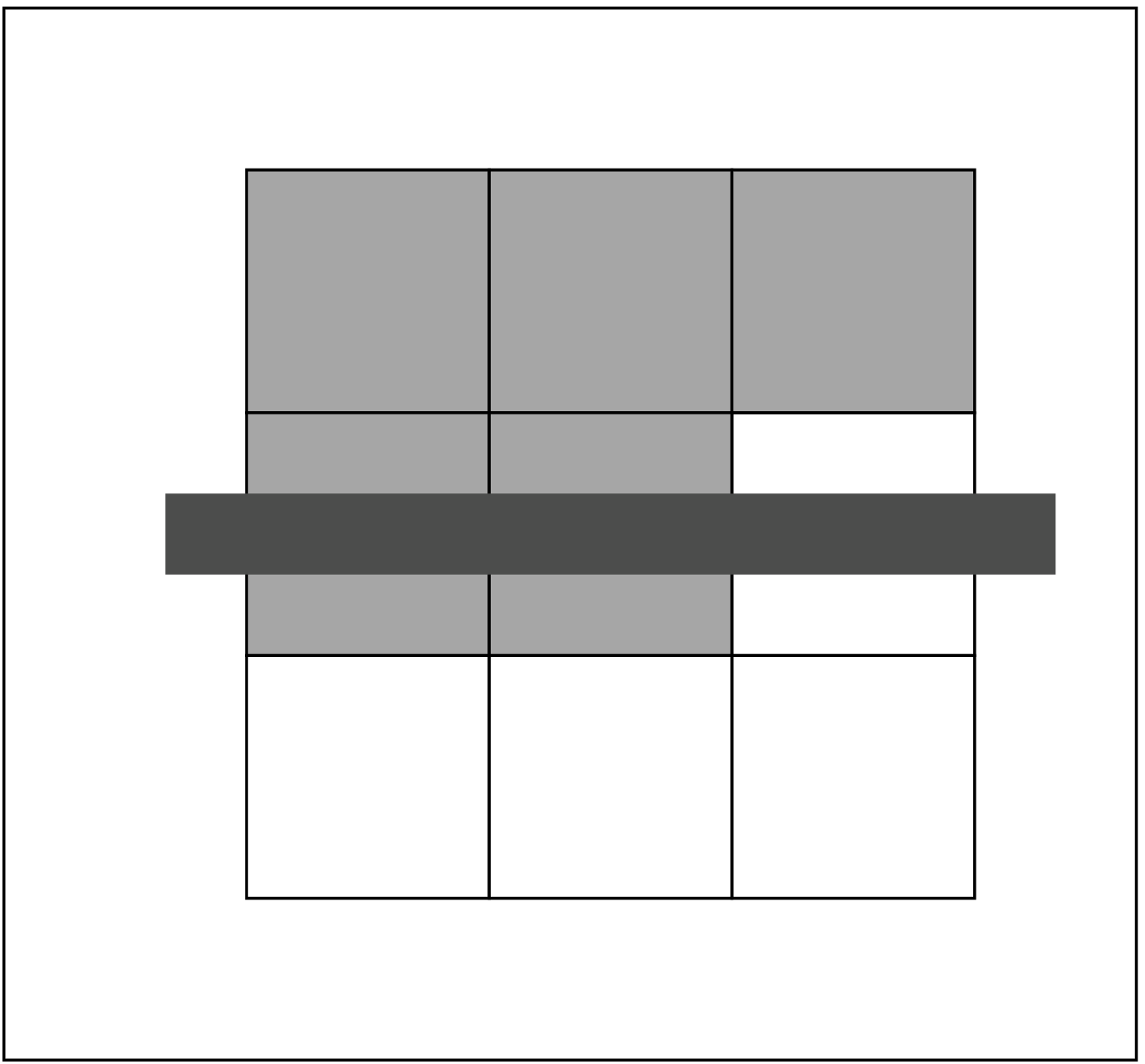}
\includegraphics[width=0.12\linewidth,keepaspectratio]{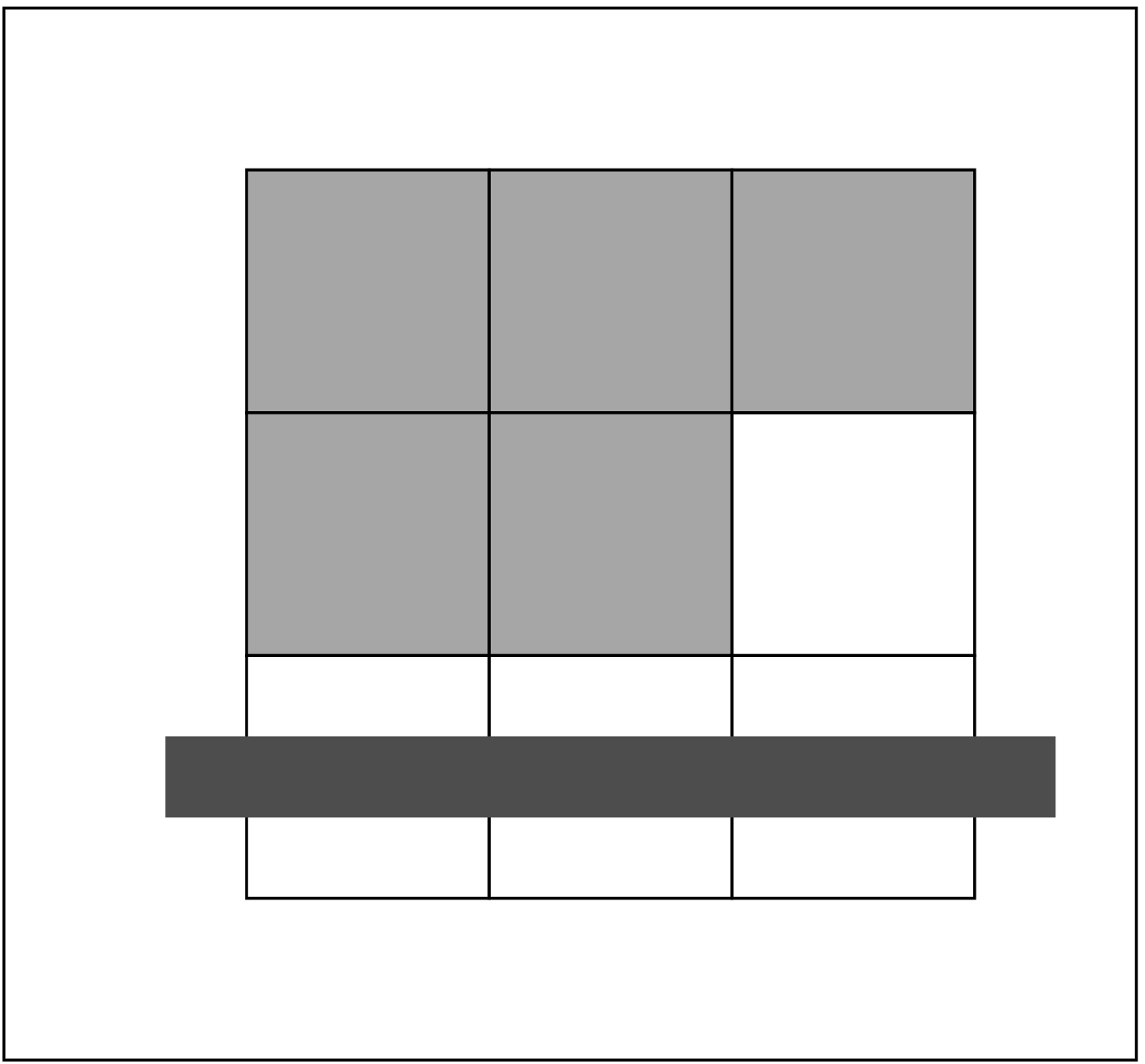}
\includegraphics[width=0.12\linewidth,keepaspectratio]{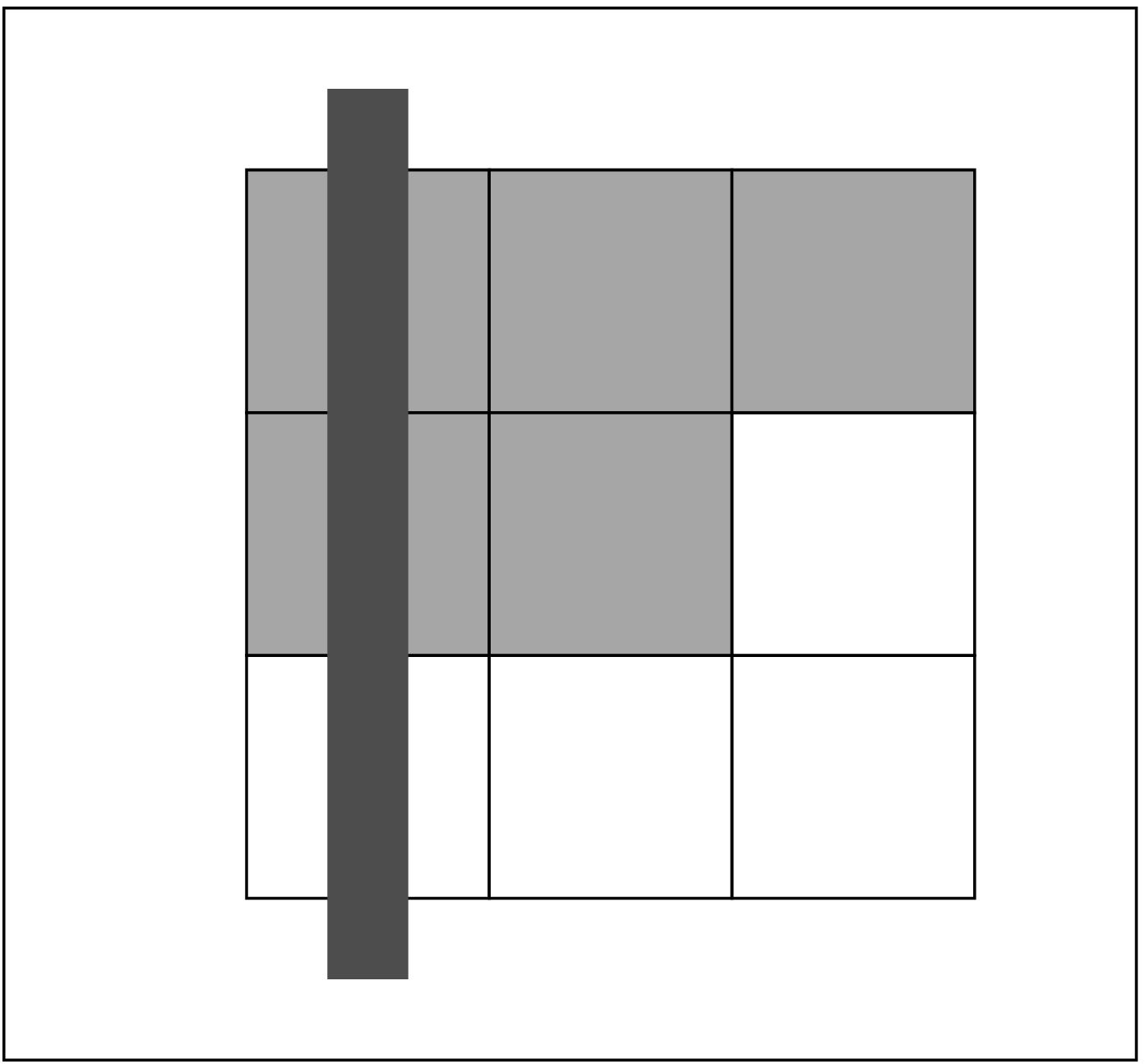}
\includegraphics[width=0.12\linewidth,keepaspectratio]{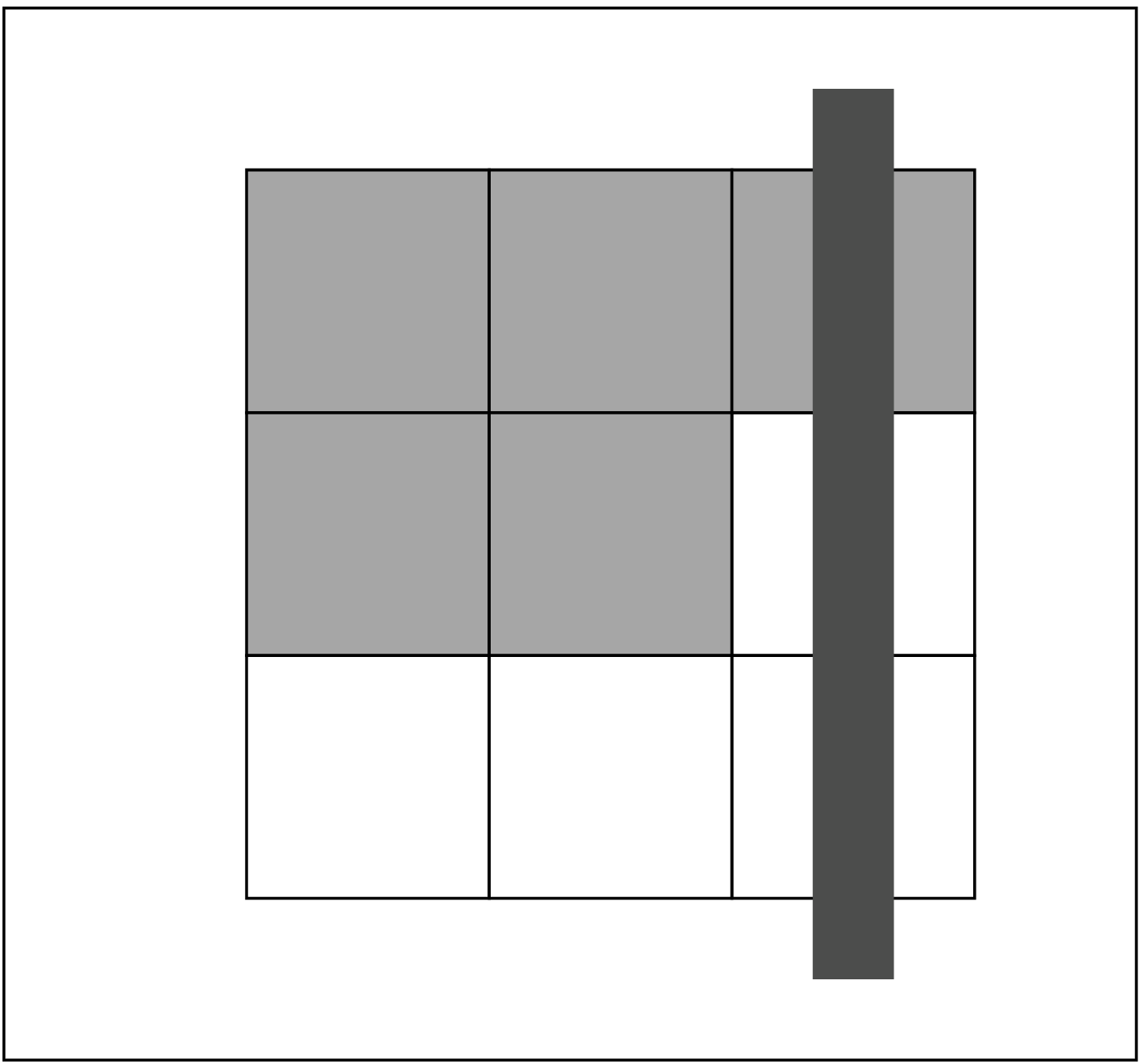}
\includegraphics[width=0.12\linewidth,keepaspectratio]{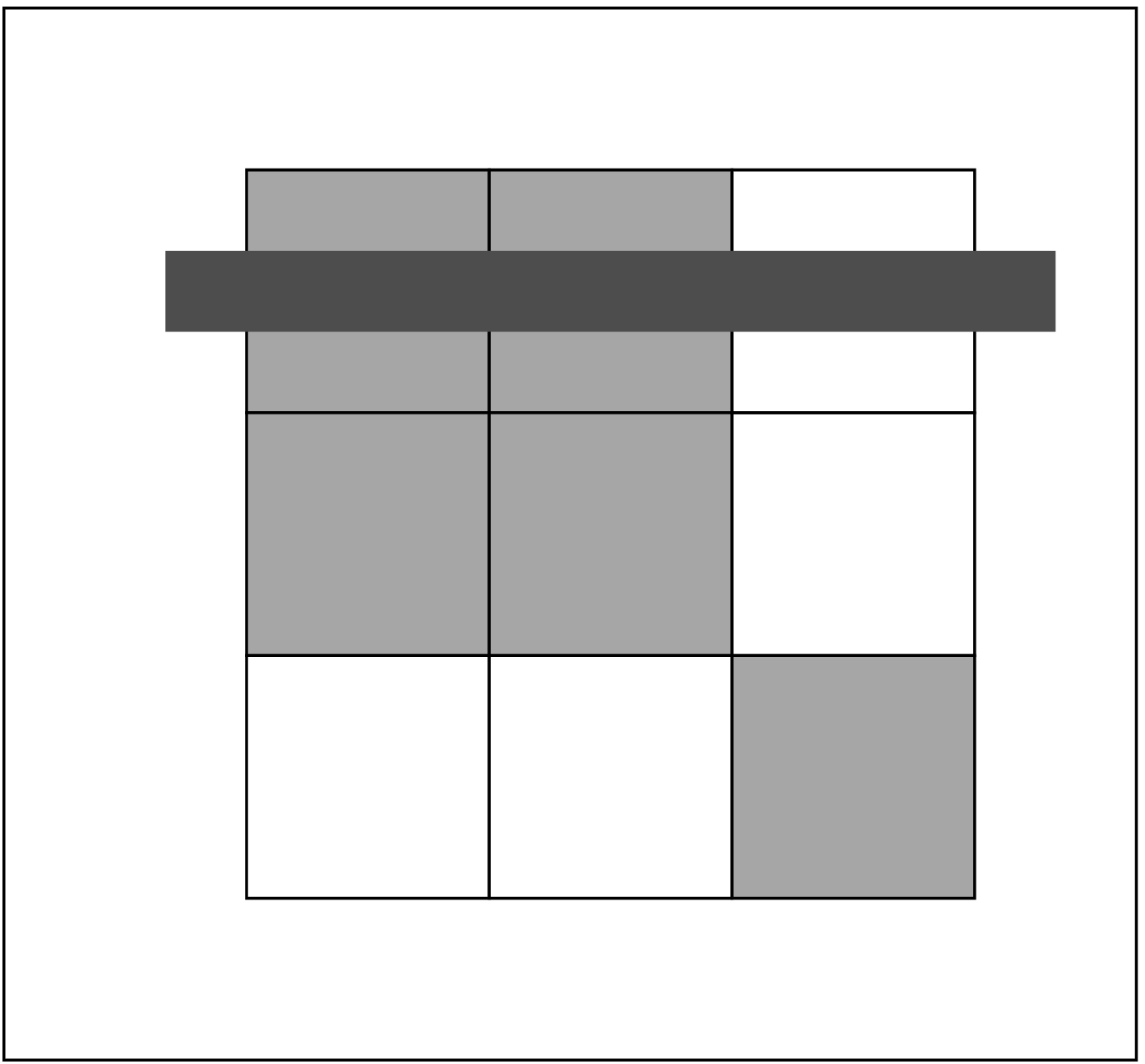}
\includegraphics[width=0.12\linewidth,keepaspectratio]{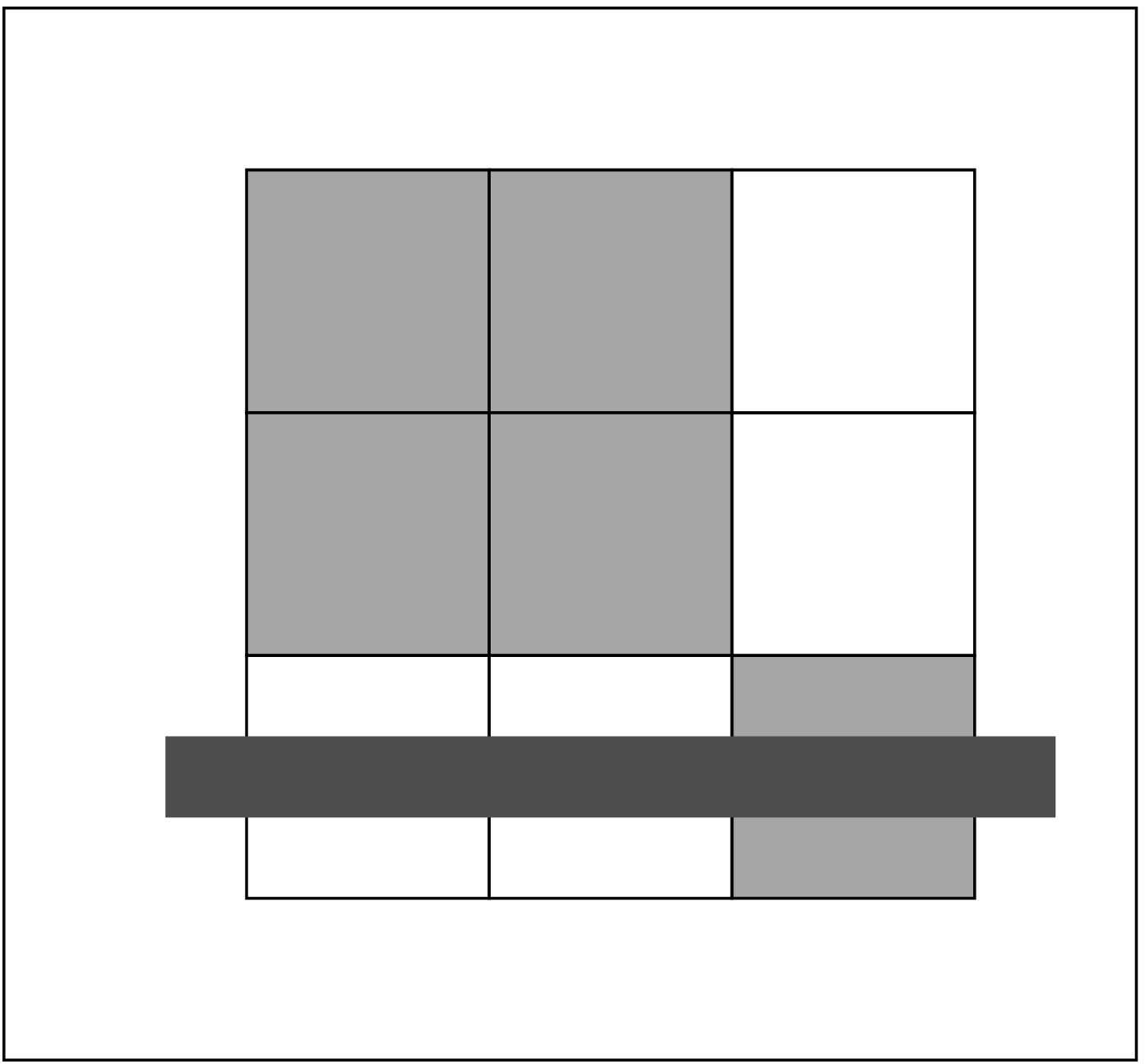}
\includegraphics[width=0.12\linewidth,keepaspectratio]{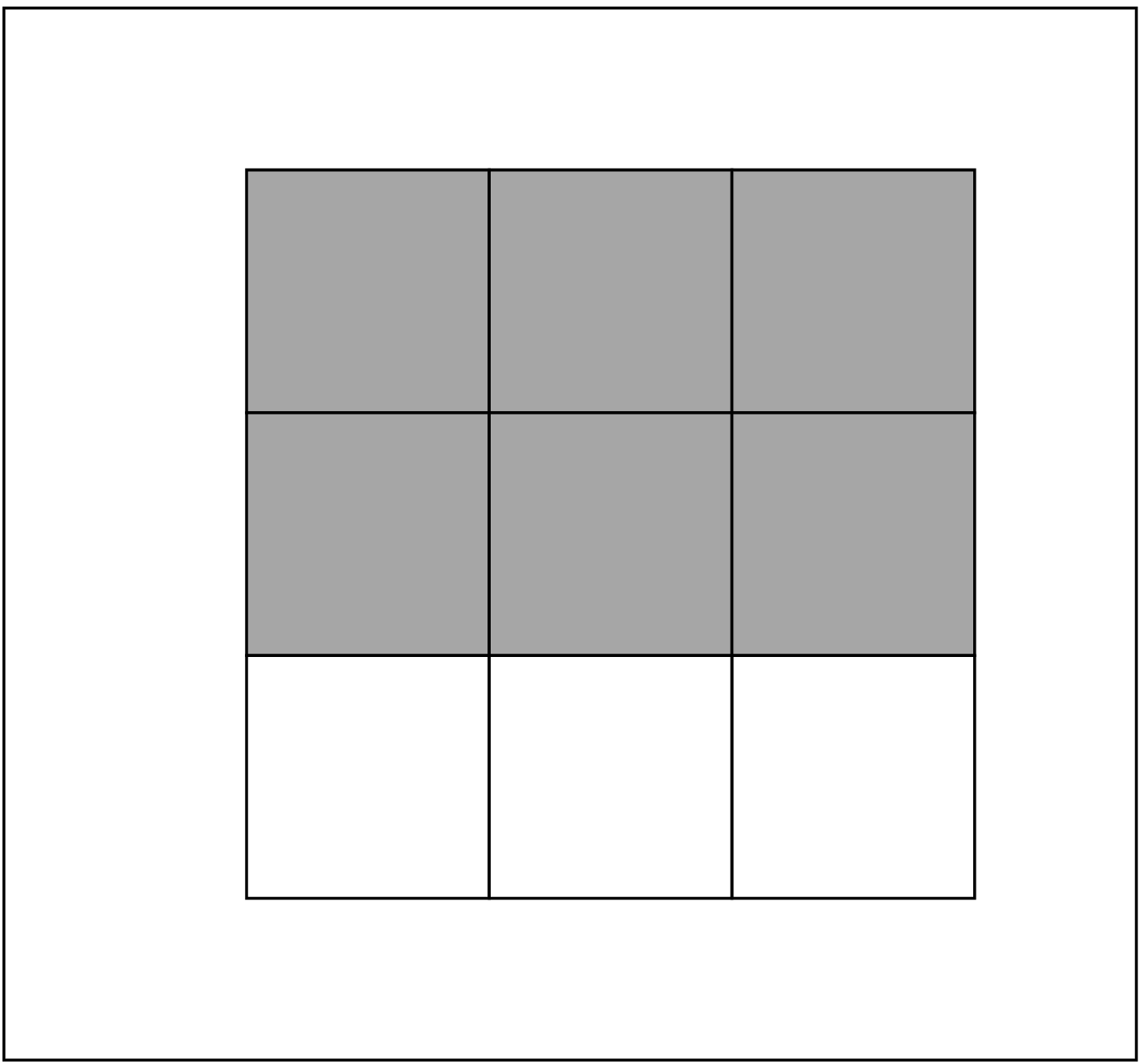}
\includegraphics[width=0.12\linewidth,keepaspectratio]{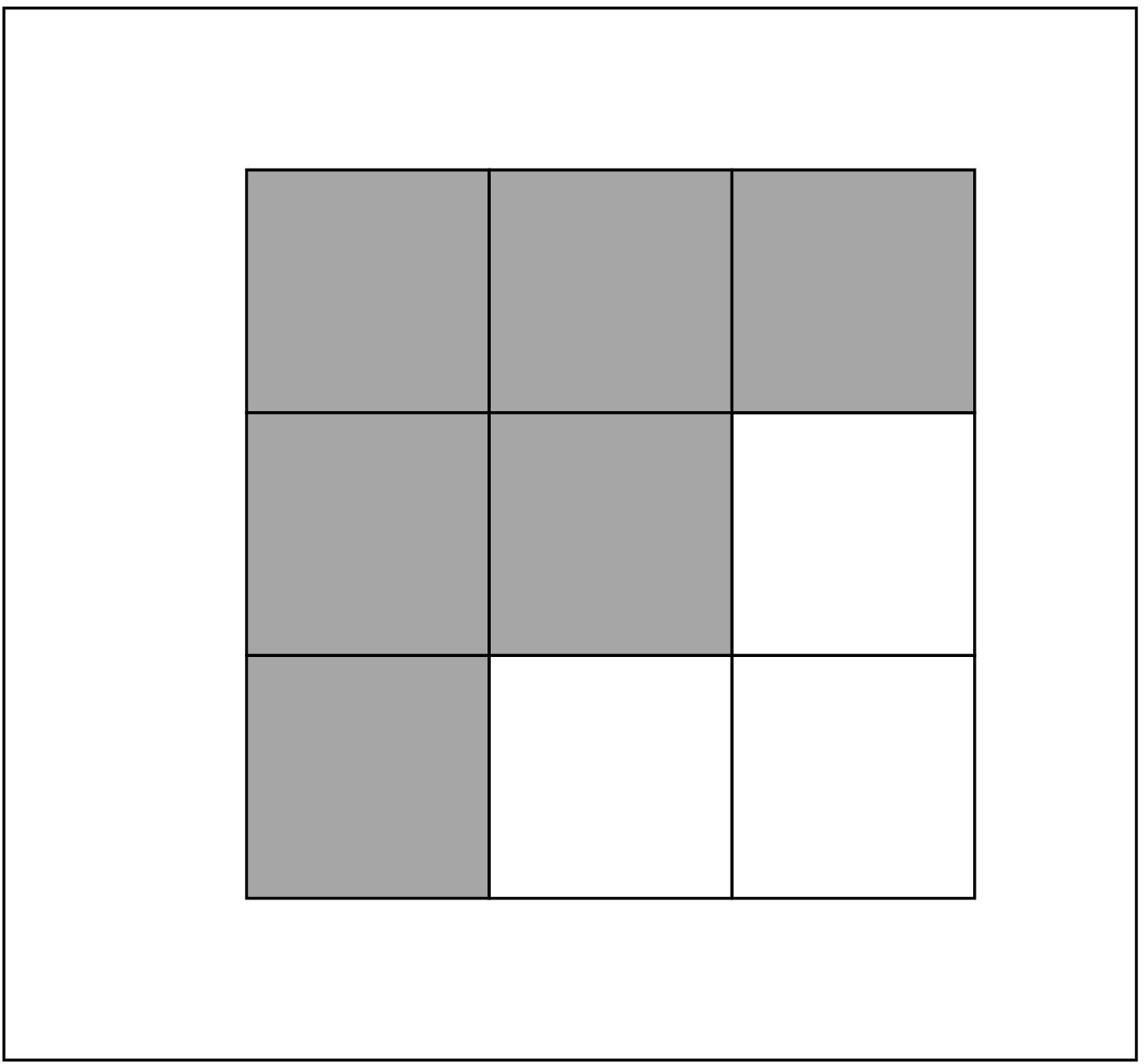}
\includegraphics[width=0.12\linewidth,keepaspectratio]{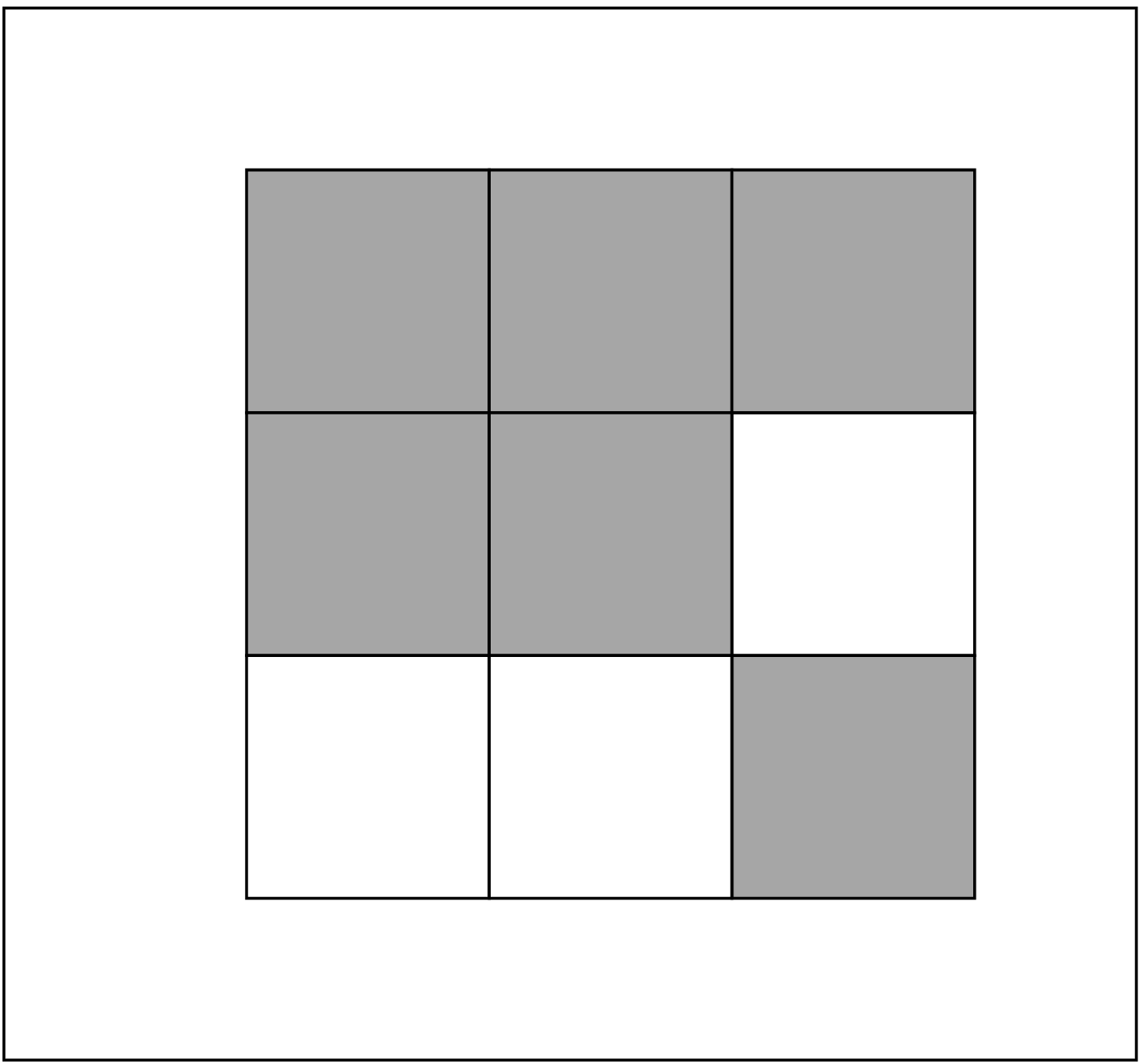}
\includegraphics[width=0.12\linewidth,keepaspectratio]{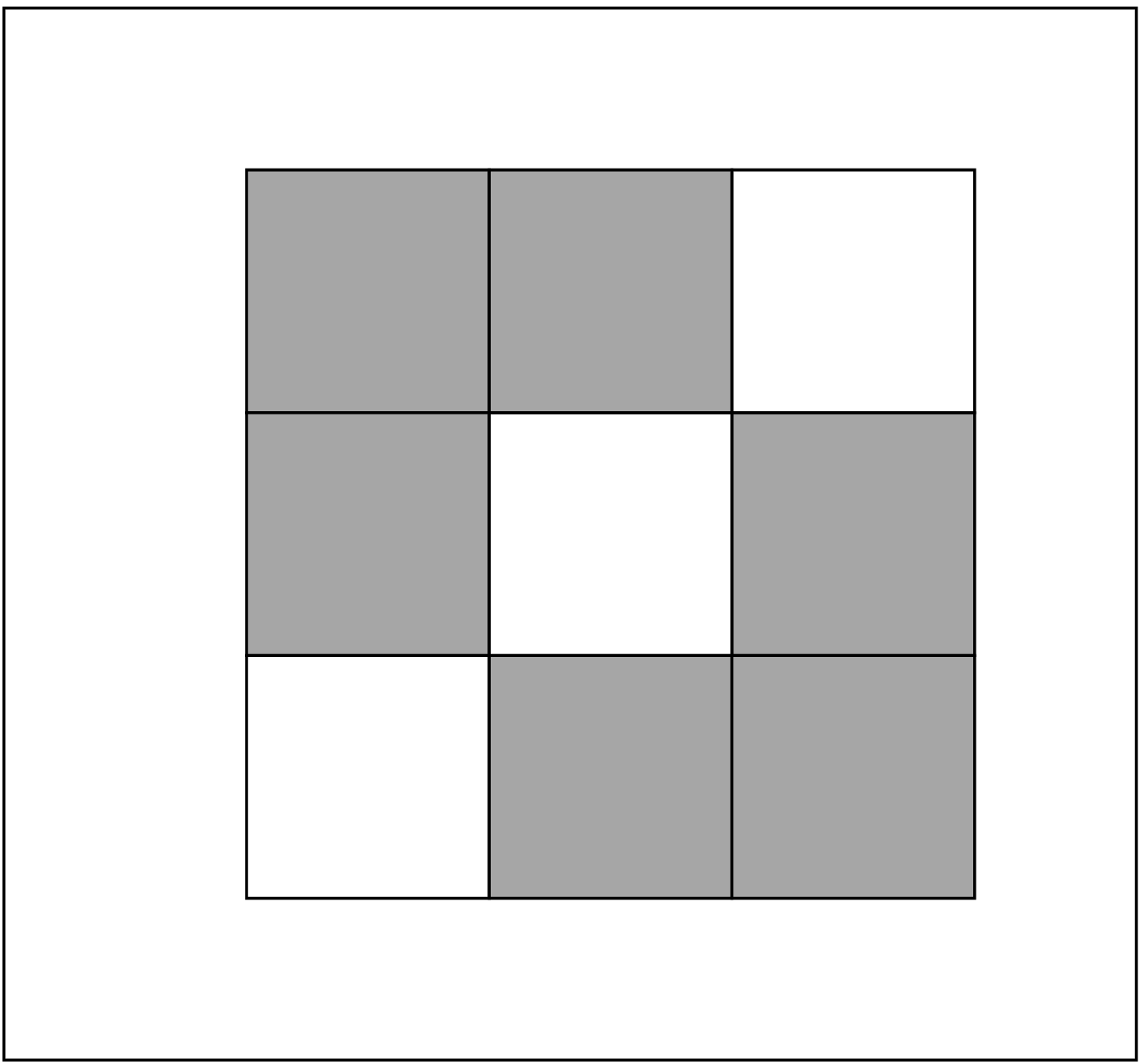}
\includegraphics[width=0.12\linewidth,keepaspectratio]{dummy.eps}
\includegraphics[width=0.12\linewidth,keepaspectratio]{dummy.eps}

\section*{Appendix III: Initial solved puzzles for Lemma~\ref{lemmatrivial} up to Lemma~\ref{lemmaon7}} \label{appIII}

\begin{center}
\includegraphics[height=0.13\textheight,keepaspectratio]{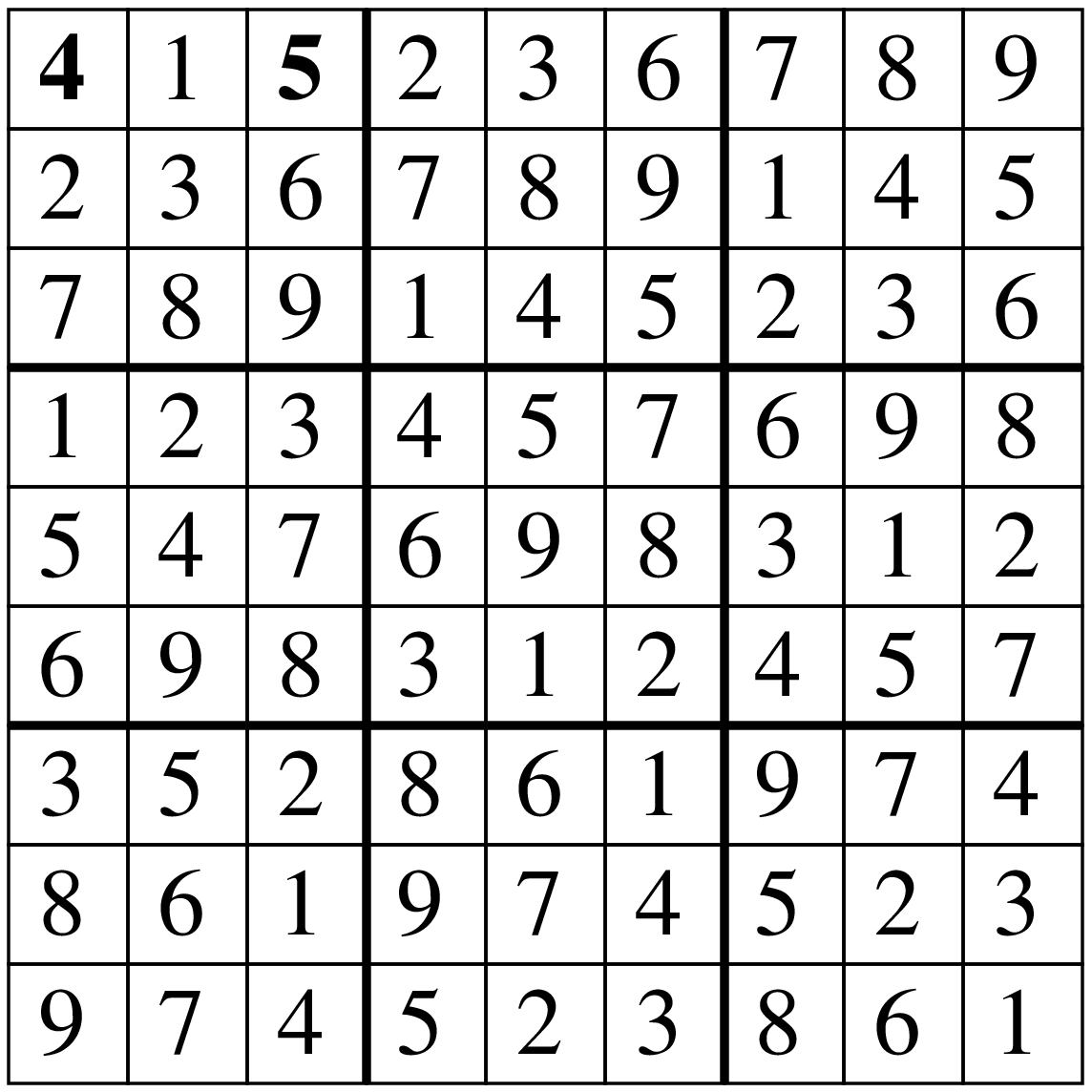}
~~~
\includegraphics[height=0.13\textheight,keepaspectratio]{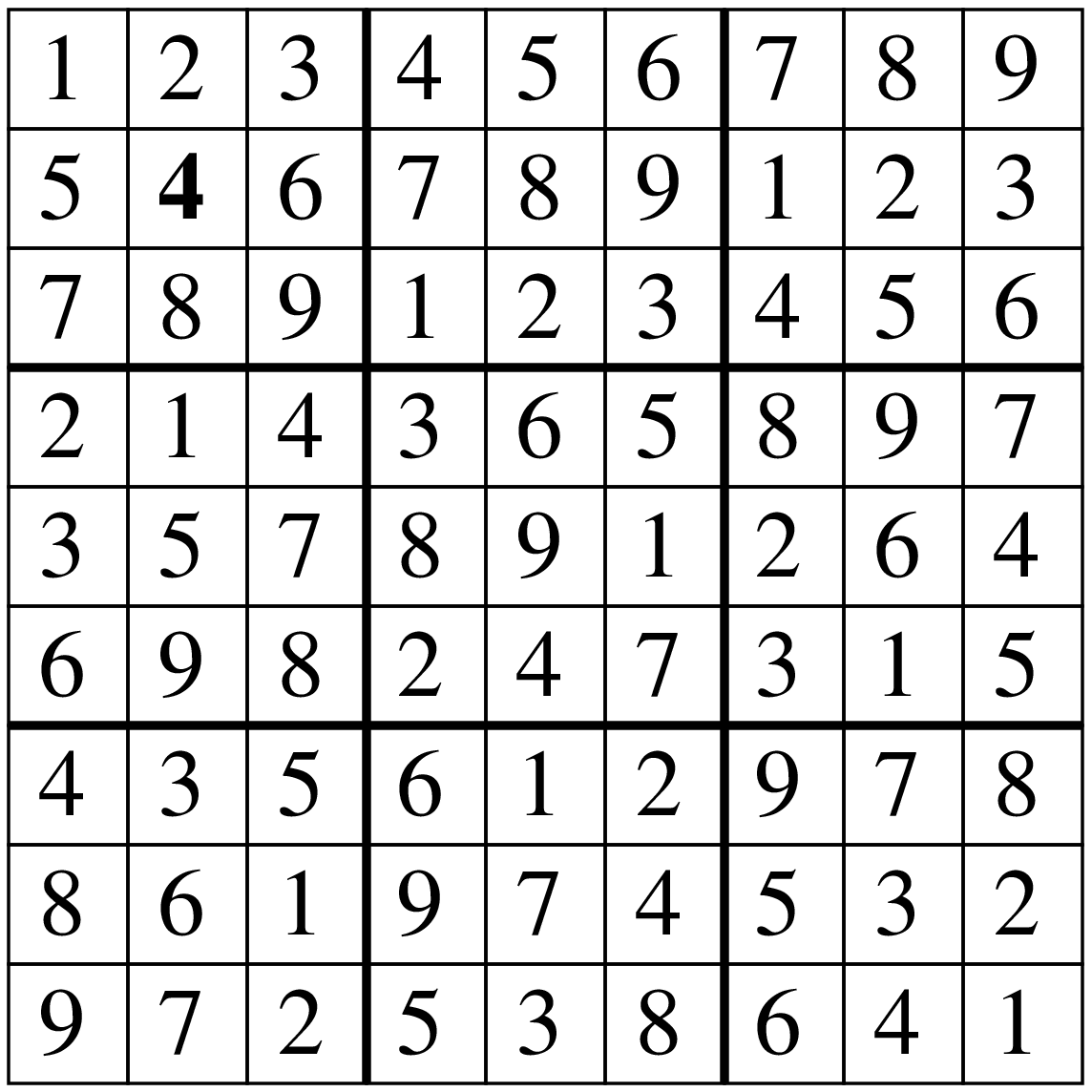}
~~~
\includegraphics[height=0.13\textheight,keepaspectratio]{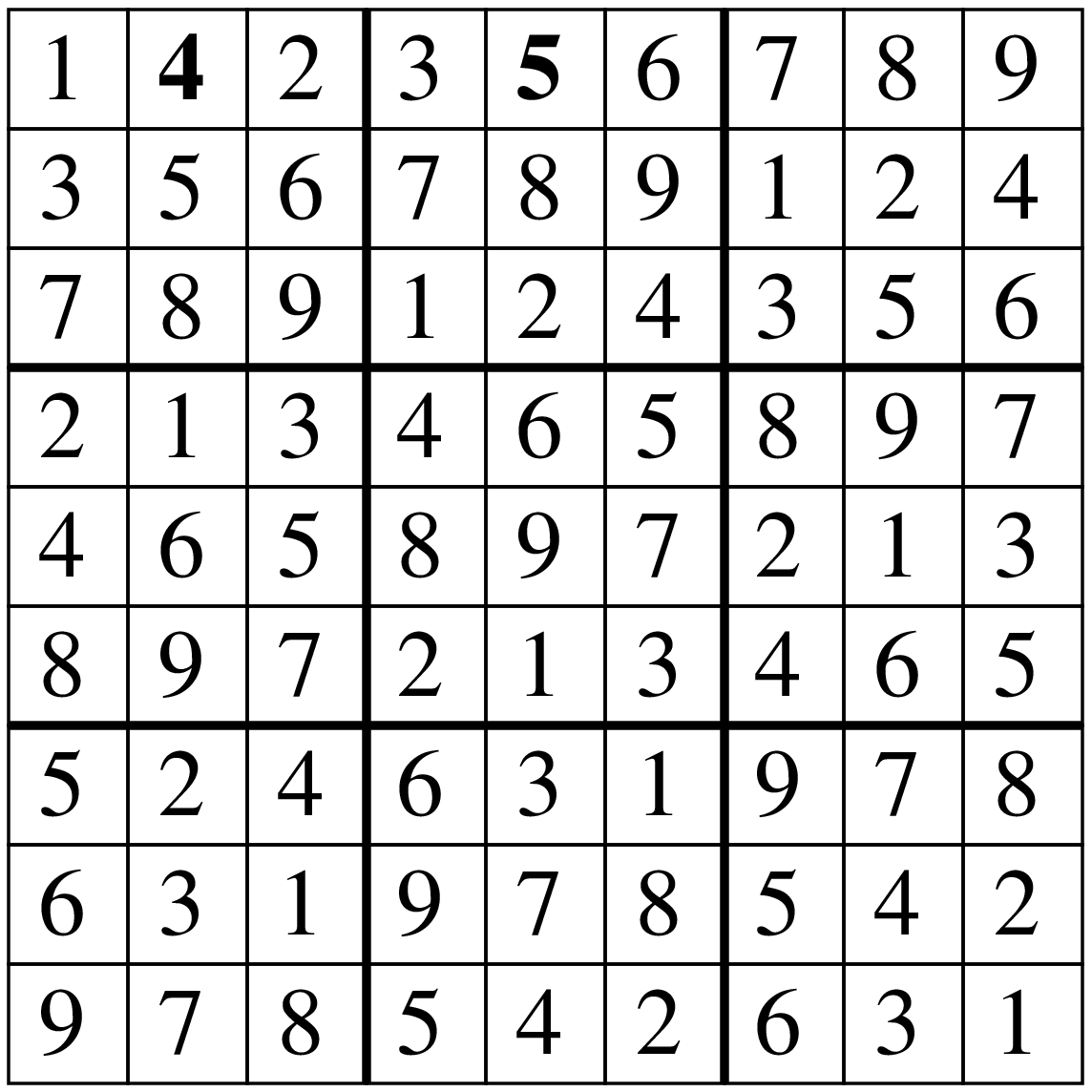}
~~~
\includegraphics[height=0.13\textheight,keepaspectratio]{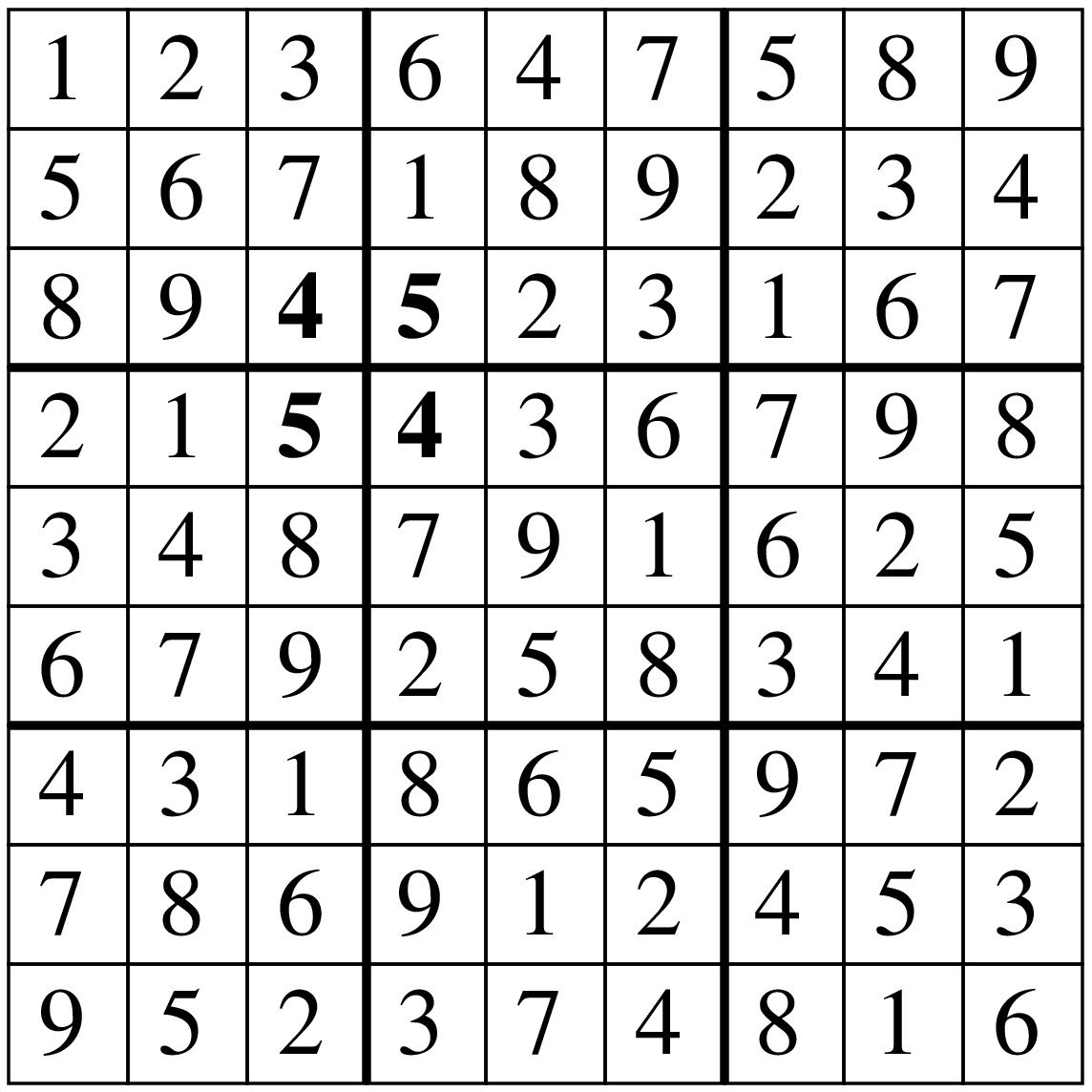}
\end{center}

\medskip
\begin{center}
\includegraphics[height=0.13\textheight,keepaspectratio]{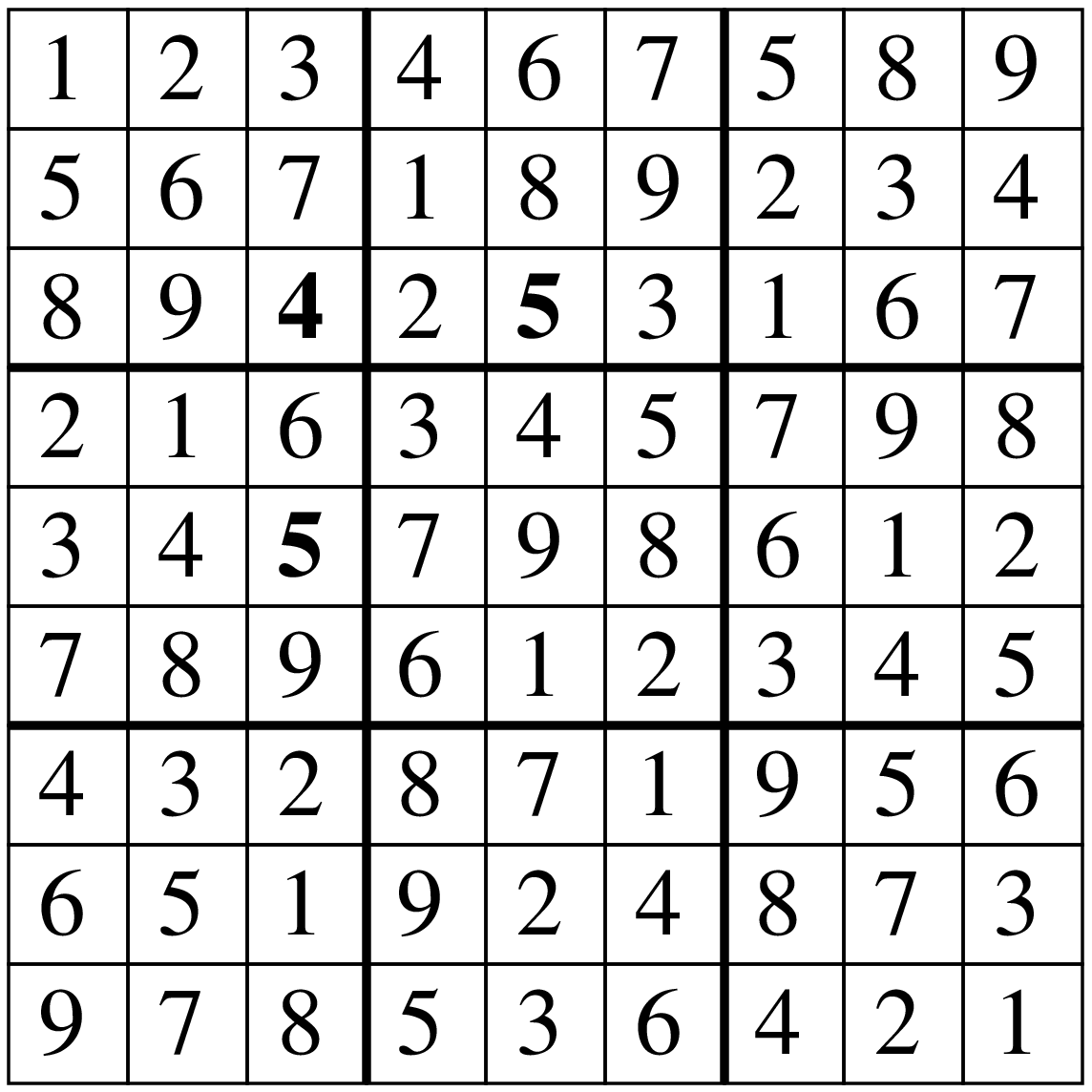}
~~~
\includegraphics[height=0.13\textheight,keepaspectratio]{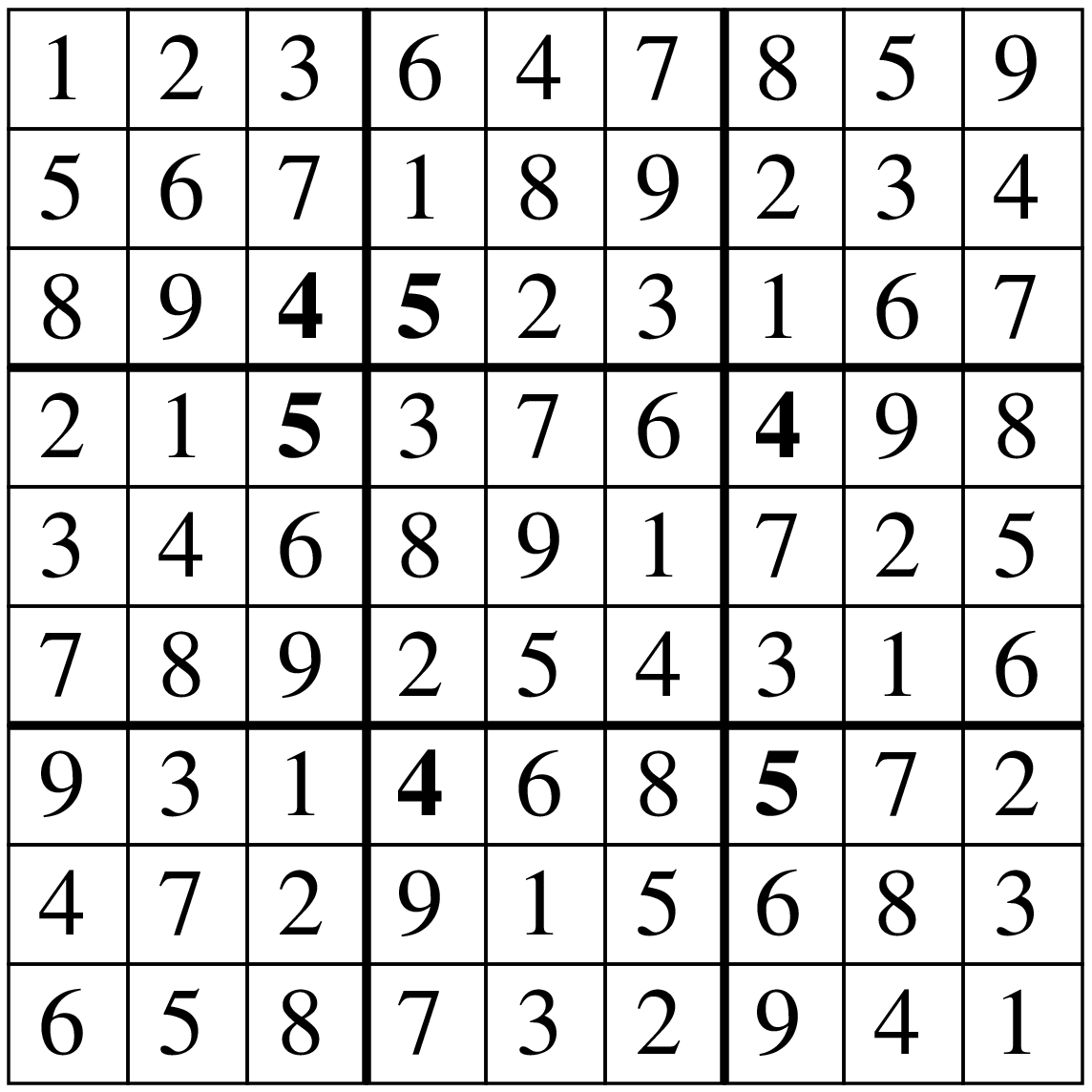}
~~~
\includegraphics[height=0.13\textheight,keepaspectratio]{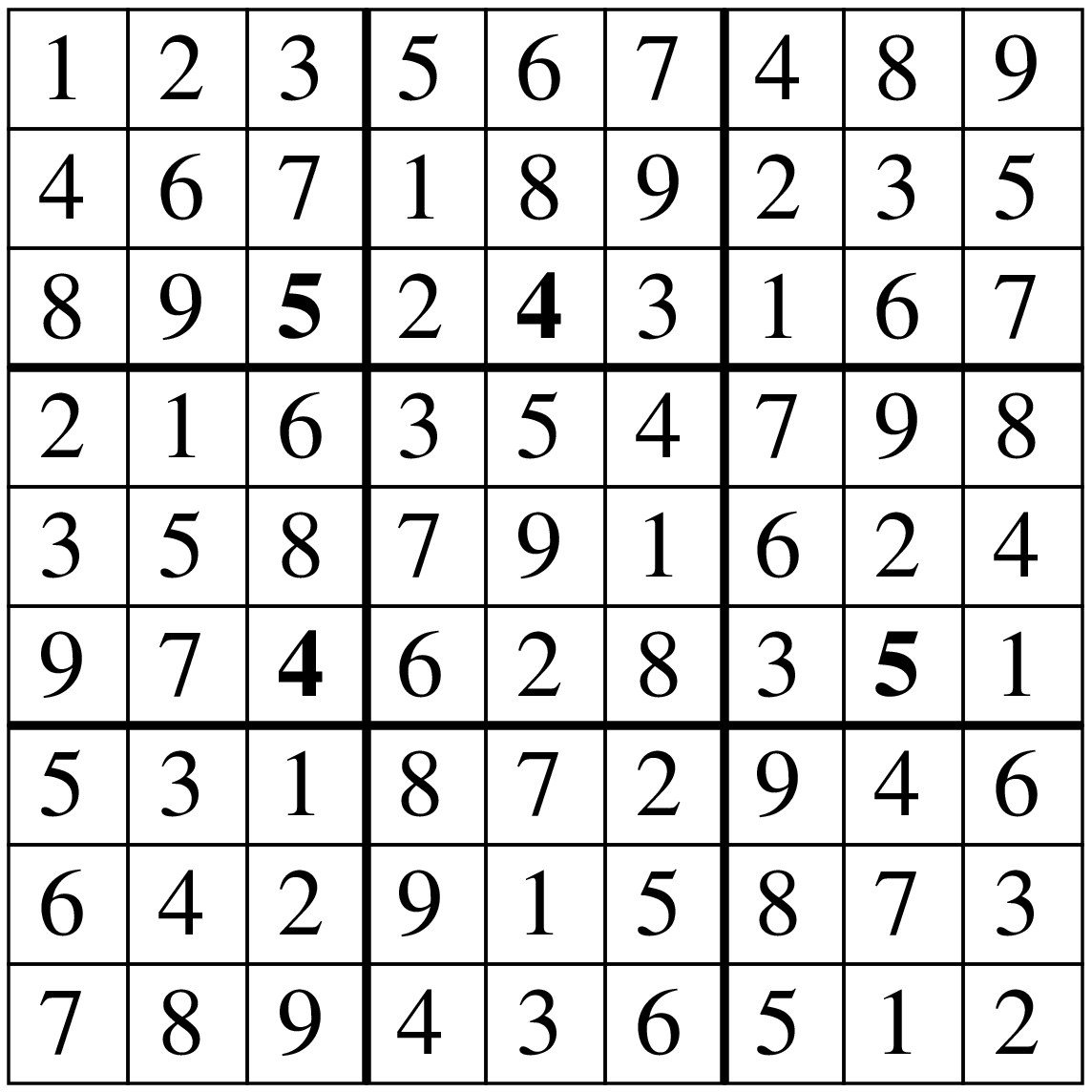}
~~~
\includegraphics[height=0.13\textheight,keepaspectratio]{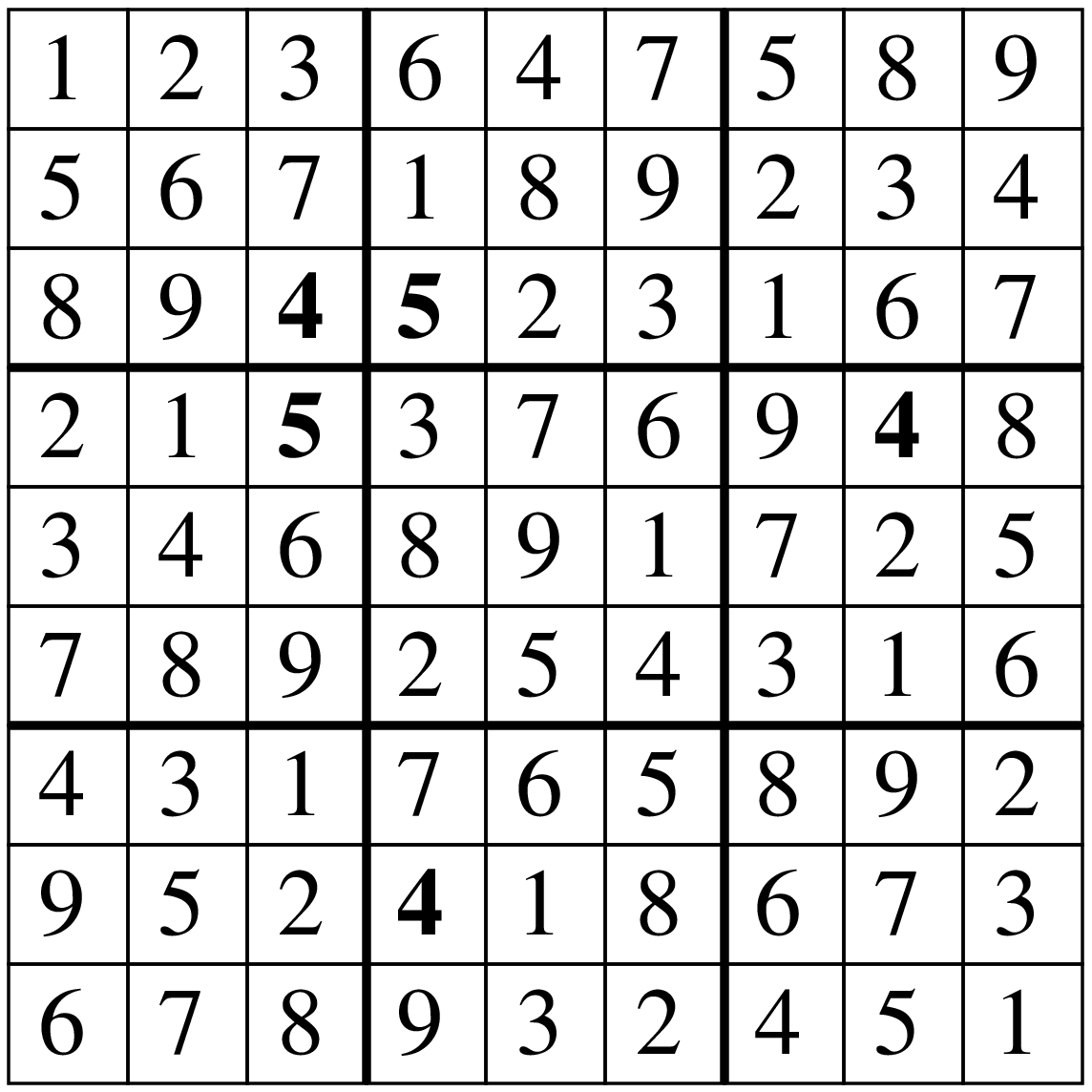}
\end{center}

\end{document}